%% file: main.tex
\documentclass[10pt,twocolumn,letterpaper]{article}

\usepackage[pagenumbers]{cvpr} 

\usepackage{multirow}
\usepackage{comment}
\usepackage{siunitx}
\usepackage{xcolor}
\usepackage{booktabs}
\usepackage{subcaption}
\usepackage{tabularx}
\usepackage{subcaption}
\usepackage{titletoc}
\usepackage[export]{adjustbox}

\newcolumntype{N}{S[table-format=3.0]}   
\newcolumntype{I}{S[table-format=-2.0]}  
\newcolumntype{D}{S[table-format=2.2]}   
\newcolumntype{F}{S[table-format=1.2]}   

\input{preamble}

\definecolor{cvprblue}{rgb}{0.21,0.49,0.74}
\usepackage[pagebackref,breaklinks,colorlinks,allcolors=cvprblue]{hyperref}

\def\confName{CVPR}
\def\confYear{2027}

\title{How Audio-Visual LLMs Build Their Words: Tracing Generated-Token Logits across Modalities}

\author{
Lucia Cascone, Valeria Fraenza, Michele Nappi, 
Fabio Narducci, Benedetto Simone\thanks{Corresponding author.}\\
Department of Computer Science, University of Salerno\\
Via Giovanni Paolo II, 132, 84084, Fisciano, SA, Italy\\
{\tt\small \{lcascone,vfraenza,mnappi,fnarducci,bsimone\}@unisa.it}
}

\begin{document}
\maketitle
\input{0_abstract}    
\input{1_intro}
\input{2_related_work}

\input{3_method}
\input{4_exp_protocol}

\input{5_results}

\input{6_conclusions}
{
    \small
    \bibliographystyle{unsrtnat}
    \bibliography{main}
}


\input{X_suppl}

\end{document}

%% file: preamble.tex
\usepackage{booktabs, multirow, siunitx}
\usepackage[table]{xcolor}
\definecolor{famgray}{gray}{0.94}
\definecolor{luciacolor}{rgb}{0.90,0.10,0.50}

\definecolor{Vcol}{HTML}{FA8D7C}
\definecolor{Acol}{HTML}{FFE368}
\definecolor{Qcol}{HTML}{9CE197}
\definecolor{Gcol}{HTML}{91C1F4}
\definecolor{Kcol}{HTML}{BFA6E8}



%% file: 0_abstract.tex
\begin{abstract}
Audio-visual large language models (AVLLMs) jointly process video, audio, and text to describe multimodal scenes, yet how each source contributes to individual generated tokens remains underexplored. We introduce
MALT (Multimodal Attribution via Logit Tracing), to the best of our
knowledge the first framework for source-resolved decomposition of
generated-token logits in AVLLMs. Conditioning on the observed forward pass, MALT traces each logit backward through the decoder and assigns signed contributions to the positions from which attention reads values.  Grouping these contributions by source (video, audio, prompt, or previously generated text) and adding explicit non-source terms
reconstructs the logit exactly.  Across nine AVLLMs, video positions
dominate aggregate contributions, while audio positions contribute
relatively more to words describing sound events than to words naming
their sources. When audio and video depict different events, words
describing the heard event receive larger contributions from audio than
from video positions in most models. Modality ablations largely agree with these attributions while revealing compensatory shifts across sources. Code will be released.
\end{abstract}

%% file: 1_intro.tex
\section{Introduction}
\label{sec:intro}

\begin{figure}[t!]
    \centering
    \includegraphics[width=1\linewidth]{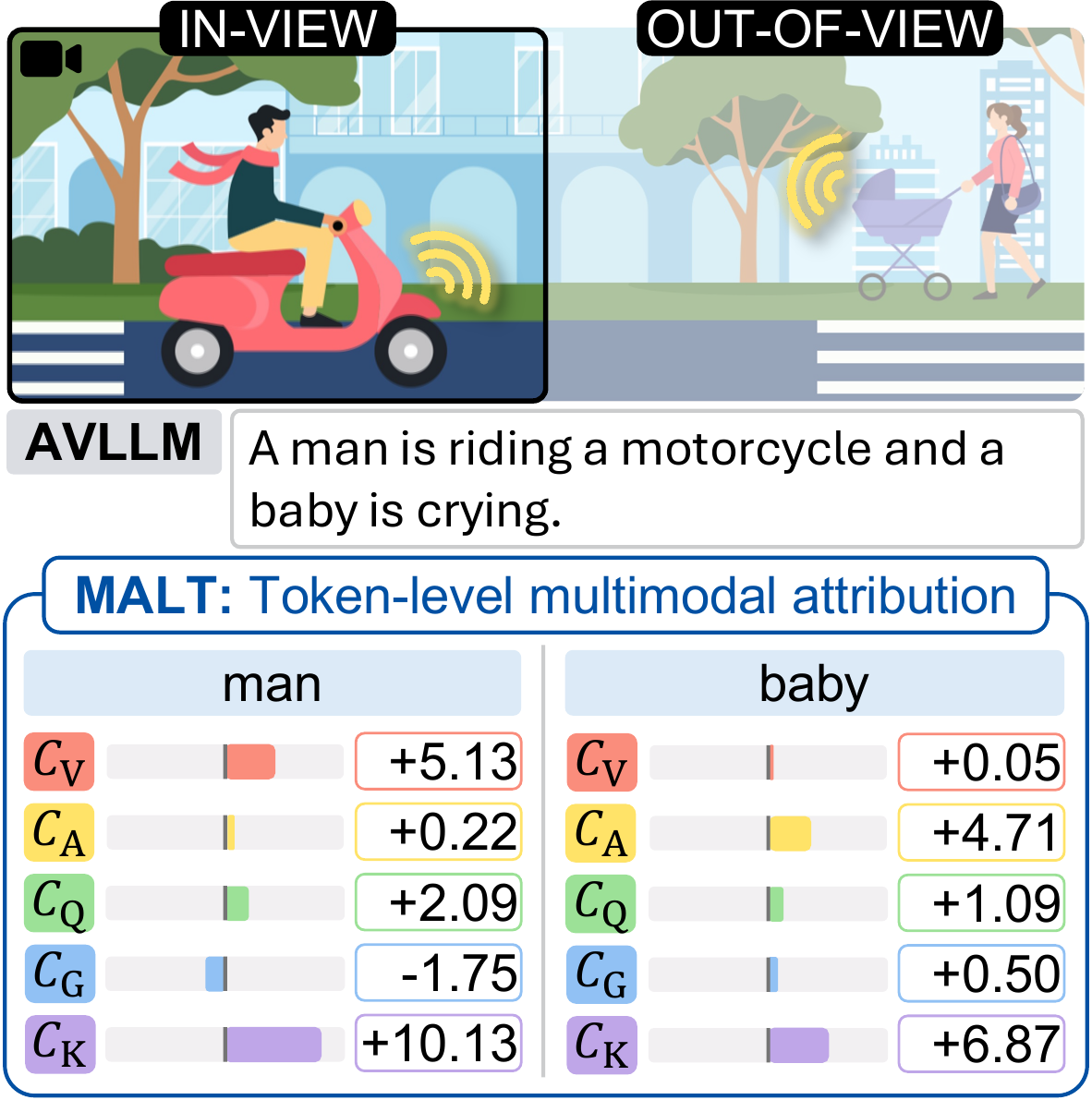}\caption{MALT decomposes each generated-token logit into signed contributions from video ($C_{\mathrm V}$), audio ($C_{\mathrm A}$), prompt ($C_{\mathrm Q}$), and previously generated text ($C_{\mathrm G}$) positions, together with a non-source component ($C_{\mathrm K}$). The full decomposition (\cref{eq:source_decomposition}) additionally includes the direct decoder-input term $E$ and numerical reconstruction residual $\epsilon$, which are not shown. In this example captioned by Qwen2.5-Omni 3B, video contributes more than audio to \emph{man}, while audio contributes more than video to \emph{baby}, whose cry is heard off-screen.}
    \label{fig:ga}
\end{figure}

Audio-visual large language models (AVLLMs) integrate visual, acoustic, and textual information to describe dynamic scenes, yet they may over-rely on visual cues, neglect acoustic evidence, or exhibit cross-modal biases~\cite{yao2026survey,baid2026don,chen2026some,sung2025avhbench}. Understanding such behavior requires knowing which inputs support each generated word. In \Cref{fig:ga}, the man is visible on screen, while a cry signals a baby outside the camera's view:  when the model generates \textit{man} or \textit{baby}, what are the contributions of video, audio, and textual positions to the corresponding token logit?
Existing analyses address related but distinct questions.
Modality interventions~\cite{chen2026some} reveal prediction changes but cannot separate source contributions from compensation, information-theoretic decompositions~\cite{fang2026towards} focus on dataset-level multiple-choice decisions, and Transformer attribution methods~\cite{ali2022xai,jantsch2026dpa} were developed for text-only models. None of them provides an exact, source-resolved accounting of generated-token logits in AVLLMs that separates audio, video, and text value reads from explicit non-source terms.
We introduce \textbf{MALT} (\textbf{M}ultimodal \textbf{A}ttribution via \textbf{L}ogit \textbf{T}racing), which decomposes each generated-token logit into signed contributions from video, audio, prompt, and previously generated text positions. Conditioning on the observed forward pass, MALT traces the logit backward through the decoder, assigning contributions to the positions from which attention reads values. Together with explicit non-source terms accounting for prefix, bias, and decoder-input terms, these contributions exactly reconstruct the logit. Since values are contextualized, contributions reflect the positions read rather than the original provenance of their information: a generated-text position, for instance, may carry information about an earlier sound. Our contributions are threefold:

\begin{itemize}
\item MALT, to our knowledge the first framework to provide an exact, source-resolved decomposition of generated-token logits in AVLLMs with explicit non-source terms, so that logit changes under interventions also decompose exactly into per-component changes.
\item A word-level study of nine AVLLMs, showing that, although video dominates in aggregate,  audio positions contribute consistently more to words describing sound events than
    to words naming their sources.  
\item Evidence of attribution consistency with model behavior: MALT attributions largely agree with counterfactual modality obscuration while revealing compensatory shifts across sources, and under conflicting audio-video evidence, words describing heard events receive greater contributions from audio than from video.

\end{itemize}

%% file: 2_related_work.tex
\section{Related work}
\label{sec:related_work}

Measuring how source positions support a generated token requires (i) separating each source's contribution from the effect of removing it, (ii) targeting the token's logit, and (iii) decomposing that logit exactly into source and non-source terms. Existing lines of work address different subsets of these requirements; we review them accordingly.

\noindent\textbf{(i) Source contributions.} AVLLM studies alter or remove inputs~\cite{ye2026eyes,chen2026some,ishikawa2026listening}, while perturbation-based attribution such as
MM-SHAP~\cite{parcalabescu2023mmshap} masks each modality to estimate its contribution to individual predictions. Removal effects, however, mix direct evidence with compensation and depend on the substitute~\cite{sturmfels2020baselines,chang2018counterfactual}, as we
observe for silence in \cref{subsec:faithfulness}.

\noindent\textbf{(ii) Token-level attribution.}
Probing and causal tracing~\cite{Selvakumar2026CVPR,jung2026probing}
characterize internal representations rather than generated-token
logits. Attention weights and rollout~\cite{abnar2020rollout} are
computed before the token is emitted, so they are shared by all
candidate tokens, and they ignore the content of the values
read~\cite{kobayashi2020attention}. Gradient-based
methods~\cite{Ancona2017TowardsBU} target the token logit, but
their scores do not generally sum to it and mix the evidence read with
changes in routing.

\noindent\textbf{(iii) Exact accounting.}
Residual-stream decompositions such as ALTI-Logit~\cite{ferrando2023explaining} retain MLP updates as separate terms and need estimated mixing coefficients to reach input positions.
Conservative propagation~\cite{ali2022xai,3692070.3692076} lets bias terms absorb unreported relevance, and the AttnLRP softmax rule does not conserve it~\cite{arras2025closelook}.
Closest to our approach, DPA~\cite{jantsch2026dpa} propagates the target unembedding backward through SwiGLU Transformers to the input embeddings, splitting attention between value and query--key paths; with its default weights, its rules coincide with those of AttnLRP~\cite{3692070.3692076},
including a softmax linearization that does not conserve the logit. Our gated-MLP rule is the uniform rule shared by both methods, for which we prove conservation and a Shapley interpretation (Supp.~\cref{sec:supp_mlp}). For attention, MALT instead assigns the full readout to the values read under fixed routing~\cite{ali2022xai} and stops at the first value read
rather than continuing to the input embeddings, reporting value- and output-projection biases and prefix reads as explicit terms.

%% file: 3_method.tex
\begin{figure*}[t!]
    \centering
    \includegraphics[width=1\linewidth]{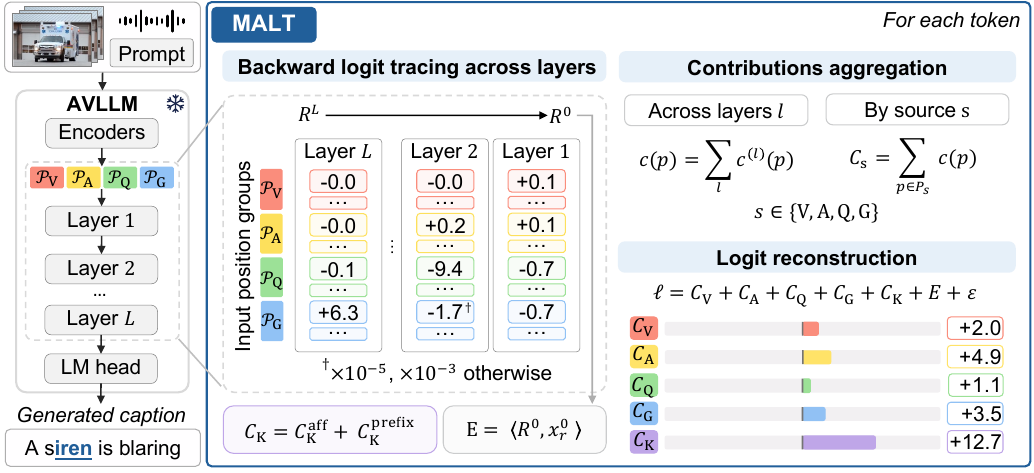}\caption{
Overview of MALT on an LU-AVS~\cite{liu2024benchmarking} example captioned by Qwen2.5-Omni 3B. For a target token $y_t$ (here, \emph{iren}), MALT propagates the logit readout backward from $R^L$ to $R^0$, assigning signed contributions $c^{(l)}(p)$ to the positions from which attention reads values (center; one illustrative position per source). Contributions are summed across layers and then over positions belonging to each source: video ($\mathrm V$), audio ($\mathrm A$), prompt ($\mathrm Q$), and previously generated text ($\mathrm G$). Together with the non-source component $C_{\mathrm K}$ (prefix reads and affine terms), the direct decoder-input term $E$, and numerical reconstruction residual $\epsilon$, they reconstruct the observed logit (right). 
}
    \label{fig:method}
\end{figure*}

\section{Method}
\label{sec:method}

Given a video $v$, an audio track $a$, and a prompt $q$, let $y=(y_1,\ldots,y_T)$ denote the generated caption.
At generation step $t$, let $r$ be the decoder position predicting $y_t$, and $\ell(y_t)$ its pre-softmax logit conditioned on $(v,a,q,y_{<t})$.
MALT traces this logit backward through the decoder, assigning signed attention-value contributions to the positions from which values are read (\Cref{fig:method}). 
Among the positions $\mathcal{P}_r$ accessible to $r$ under causal attention, those preceding the first multimodal token form the prefix $\mathcal{P}_{\mathrm{prefix}}$, while the remaining positions are partitioned into sources $\mathcal{S}=\{\mathrm{V},\mathrm{A},\mathrm{Q},\mathrm{G}\}$ (video, audio, prompt/template, and previously generated text), with $\mathcal{P}_s$ denoting the positions belonging to source $s$.


\subsection{Source-resolved logit decomposition}
\label{sec:source}

Let $c_{y_t}^{l}(p)$ denote the signed contribution to
$\ell(y_t)$ carried by the bias-free values read from $p$
at $r$ in layer $l$, summed over attention heads
(\cref{eq:attention_contribution}). Aggregating over layers and over the positions of each group gives
\begin{equation}
    c_{y_t}(p) = \sum_{l=1}^{L} c_{y_t}^{l}(p),
    \quad
    C_s(y_t) = \sum_{p\in\mathcal{P}_s} c_{y_t}(p),
    \quad s\in\mathcal{S}.
    \label{eq:source_contribution}
\end{equation}
The resulting logit decomposition is
\begin{equation}
    \ell(y_t)
    = \sum_{s\in\mathcal{S}} C_s(y_t)
    + C_{\mathrm{K}}(y_t) + E(y_t) + \epsilon(y_t),
    \label{eq:source_decomposition}
\end{equation}
where $C_{\mathrm{K}}$ collects prefix reads and attention
bias terms (\cref{sec:constant}), $E$ is the direct
decoder-input contribution at $r$
(\cref{sec:tracing}), and $\epsilon$ is the
numerical reconstruction residual, which vanishes
in exact arithmetic. 
Since the trace is linear in the initial readout direction (\cref{sec:tracing}), the margin $\ell(y_t)-\ell(y_t^{\mathrm{alt}})$ against an alternative token $y_t^{\mathrm{alt}}$ at the same position decomposes term by term, with group $s$ contributing
$C_s(y_t)-C_s(y_t^{\mathrm{alt}})$; the difference in $E$ is generally
nonzero, since $R^0$ depends on the target token.
These differences are invariant to common logit shifts.

\subsection{Logit tracing through the decoder}
\label{sec:tracing}

Let $x_p^0\in\mathbb{R}^D$ denote the decoder input at position $p$
and $x_p^l$ the residual-stream state after layer $l$. We derive the
trace for the pre-norm Qwen-family decoder blocks used by the
evaluated AVLLMs~\cite{Yang2024Qwen2TR,
Yang2024Qwen25TR,yang2025qwen3technicalreport,
Bai2025Qwen3VLTR}. At the prediction position $r$, layer $l$ updates the residual stream as
\begin{equation}
    x_r^{l,\mathrm{mid}} = x_r^{l-1} + a_r^l,
    \qquad
    x_r^{l} = x_r^{l,\mathrm{mid}} + m^l(\hat{x}_r^l),
    \label{eq:block}
\end{equation}
where $a_r^l$ is the attention output and
$\hat{x}_r^l=\operatorname{Norm}_m^l(x_r^{l,\mathrm{mid}})$ is the
MLP input, normalized by the pre-MLP RMSNorm.

\paragraph{Design principle.}
MALT conditions on the attention weights, RMSNorm denominators, and SwiGLU sigmoid factors observed during the forward pass.
With attention weights fixed, each head output is a weighted sum of value vectors, so its readout decomposes by value position; with sigmoid factors fixed, the SwiGLU readout becomes quadratic in the normalized MLP input.
For a fixed forward pass and interaction-allocation rule, the trace is linear in the initial readout direction and conserves the observed logit.
The derivation assumes bias-free SwiGLU projections and LM heads; attention value- and output-projection biases are accounted for separately (Supp.~\cref{sec:supp_attention}).

\paragraph{Output readout and normalization.}
For a bias-free LM head with unembedding matrix $W_U$ and final RMSNorm scale $\gamma_f$, the logit of $y_t$ reads out the final residual state $x_r^L$ along $R^L$:
\begin{equation}
    \ell(y_t)=\langle R^L,x_r^L\rangle,
    \qquad
    R^L=\frac{W_U[y_t]\odot\gamma_f}{\operatorname{rms}(x_r^L)},
    \label{eq:logit}
\end{equation}
where $W_U[y_t]$ is the row of $W_U$ for $y_t$ and $\operatorname{rms}$ is defined in Supp.~\cref{sec:supp_rms}.
With the observed denominator fixed~\cite{3692070.3692076}, an RMSNorm $\hat{x}=\gamma\odot x/\operatorname{rms}(x)$ maps a readout direction $R$ at its output to $R_{\mathrm{in}}=\gamma\odot R/\operatorname{rms}(x)$ at its input.
For a residual update $x'=x+F(x)$, $R$ propagates unchanged along the skip path, while the branch contributes $\langle R,F(x)\rangle$.
We denote by $R^l$ and $R^{l,\mathrm{mid}}$ the readout directions at $x_r^l$ and $x_r^{l,\mathrm{mid}}$.
The pre-attention RMSNorm is not traversed separately, as its effect is already incorporated into the value vectors $\tilde v_p^{l,\kappa(h)}=W_V^{l,\kappa(h)}\,\mathrm{Norm}_a^l(x_p^{l-1})$.

\paragraph{Gated MLP.}
For the SwiGLU branch
$m^l(\hat{x})=W_d^l[\operatorname{SiLU}(z)\odot u]$,
with $z=W_g^l\hat{x}$ and $u=W_u^l\hat{x}$,
let $w=(W_d^l)^\top R^l$ and
$\bar{\sigma}=\sigma(z)$.
Fixing $\bar{\sigma}$ yields the quadratic form
$f(\xi)=\langle w\odot\bar{\sigma},
(W_g^l\xi)\odot(W_u^l\xi)\rangle$,
which equals the observed MLP branch readout at
$\xi=\hat{x}$. We propagate

\begin{equation}
\begin{aligned}
R_{\mathrm{mlp}}
=\tfrac12\nabla f(\hat{x})
=\tfrac12\bigl[&
(W_g^l)^\top(w\odot\bar{\sigma}\odot u)\\
&+(W_u^l)^\top(w\odot\bar{\sigma}\odot z)
\bigr].
\end{aligned}
\label{eq:mlp_direction}
\end{equation}
This preserves the observed branch readout and,
for any additive decomposition of $\hat{x}$,
allocates pairwise interactions symmetrically,
corresponding to the Shapley allocation of the
local quadratic game with fixed sigmoid factors
(Supp.~\cref{sec:supp_mlp}).
Weighting the two terms by $\lambda$ and
$1-\lambda$, with $\lambda\in[0,1]$, yields
a family of conservative rules; we use
$\lambda=1/2$ and assess its sensitivity
in \cref{sec:sensitivity}.
Propagating $R_{\mathrm{mlp}}$ through the
pre-MLP RMSNorm and adding the residual path
gives $R^{l,\mathrm{mid}}$.

\paragraph{Attention routing.}
Let $\alpha_{r,p}^{l,h}$ be the observed weight with which query
head $h$ at $r$ reads $p\in\mathcal{P}_r$, $\kappa(h)$ its key-value
head under grouped-query attention, $\tilde v_p^{l,\kappa(h)}$ the
value without its projection bias, and $W_O^{l,h}$ the block of the
output projection acting on head $h$. Holding the attention weights
fixed~\cite{ali2022xai} gives
\begin{equation}
    c_{y_t}^{l}(p)
    =\sum_h\alpha_{r,p}^{l,h}
      \bigl\langle(W_O^{l,h})^\top R^{l,\mathrm{mid}},
                   \tilde v_p^{l,\kappa(h)}\bigr\rangle.
    \label{eq:attention_contribution}
\end{equation}
This yields an exact decomposition of the bias-free attention readout across value positions under fixed routing, without introducing additional attribution parameters. Query--key interactions,
including rotary position embeddings, determine the observed
routing but are not separately attributed. Value and
output-projection biases, when present, contribute a term
$C_{\mathrm K}^{l,\mathrm{aff}}(y_t)$ independent of the
attention weights for a fixed readout direction, since each
head's weights sum to one
(Supp.~\cref{sec:supp_attention}).

\paragraph{Conservation.}
Conditioned on the observed forward pass, MALT preserves the readout algebraically: skip paths carry it unchanged, attention decomposes into value reads and affine terms
under fixed routing (Supp.~\cref{sec:supp_attention}),
and the MLP rule conserves its branch readout for any $\lambda$ (Supp.~\cref{sec:supp_mlp}).
Summing the layer-wise conservation identity
(Supp.~\cref{eq:layer_conservation}) across decoder layers cancels the intermediate readout terms and yields
\cref{eq:source_decomposition}, with
$E(y_t)=\langle R^0,x_r^0\rangle$ and
$\epsilon(y_t)=0$ in exact arithmetic.
Value reads from $r$ itself are counted once, in $c_{y_t}(r)$, separately from $E$, and assigned to $\mathrm Q$ at the first generation step and to $\mathrm G$ thereafter. Across $74{,}771$ generated tokens, the mean absolute reconstruction residual is $0.041$ logit units (mean relative error $0.11\%$ overall, with per-model means no greater than $0.14\%$; Supp.~\cref{tab:supp_reconstruction}).
Reconstruction accuracy does not by itself establish attribution faithfulness; we therefore examine consistency with counterfactual interventions in
\Cref{subsec:faithfulness}.

\subsection{Non-source terms and interpretation}
\label{sec:constant}

The non-source contribution is
\begin{equation}
    C_{\mathrm K}(y_t)
    =C_{\mathrm K}^{\mathrm{aff}}(y_t)
     +C_{\mathrm K}^{\mathrm{prefix}}(y_t),
    \label{eq:non_source}
\end{equation}
where
$C_{\mathrm K}^{\mathrm{aff}}(y_t)
=\sum_{l=1}^{L}C_{\mathrm K}^{l,\mathrm{aff}}(y_t)$
collects attention-projection bias contributions at $r$
(Supp.~\cref{sec:supp_attention}), and
$C_{\mathrm K}^{\mathrm{prefix}}(y_t)
=\sum_{p\in\mathcal P_{\mathrm{prefix}}}c_{y_t}(p)$
collects reads from positions preceding the first
multimodal token, whose extent depends on the model's
input template (Supp.~\cref{subsec:supp_models}).
The affine contribution depends on fixed projection biases
and input-dependent readout directions $R^{l,\mathrm{mid}}$,
which also vary with the MLP propagation rule;
biases absorbed into contextualized values in earlier layers
remain assigned to the positions read.
Prefix states are unaffected by later multimodal tokens,
but their contributions vary through routing and readout directions. $C_{\mathrm K}$ thus collects contributions outside the defined source groups and is reported as a single component.

\paragraph{Interpretation.}
$C_s(y_t)$ measures the signed contribution associated with value reads from group $s$ under the observed attention routing.
Since values are contextualized, generated-text positions may relay information from earlier audio or video tokens, while
interleaved audio and video positions may carry information
transferred from the other modality. MALT assigns each
contribution to the position from which the value is read,
rather than to the original provenance of its information. The decomposition preserves the observed token logit under
the specified propagation rules, but does not represent
the full differential of the model. In particular, changes
in attention weights, RMSNorm denominators, and SwiGLU
sigmoid factors are not separately attributed.
Consequently, $C_s(y_t)$ represents position-associated
logit support rather than the total causal effect of
removing source $s$, which may also alter attention routing
and contextualized representations. We therefore assess
agreement with counterfactual interventions in sign and
ordering, without expecting equality in magnitude (\cref{subsec:faithfulness}).

%% file: 4_exp_protocol.tex
\begin{table*}[t]
\centering
\caption{Attribution of generated event words by linguistic role
(\textit{Subj.}: subjects; \textit{Pred.}: predicates).
Models are grouped by the presence of value-projection biases
(Supp.~\cref{tab:supp_affine}).
Cov.: event coverage (\%); $n$: number of matched words.
Component share, A-V preference, and deciding component are reported in \%;
A-V density is per $10^3$ modality positions.
Bold indicates the most frequent deciding component per row.
$^\dagger$Ola-7B processes 249/300 clips due to its context limit.
$^\ddagger$Shares exclude $E$ and are rounded, so rows may not sum to 100\%.}
\label{tab:source_attribution}

\setlength{\tabcolsep}{2.7pt}
\sisetup{table-number-alignment=center}
\newcommand{\decidingwidth}{1.3em}
\newcommand{\componenthead}[2]{%
  \multicolumn{1}{%
    >{\columncolor{#1!45}%
      [\dimexpr\tabcolsep-1pt\relax]%
      [\dimexpr\tabcolsep-1pt\relax]}c%
  }{#2}%
}

\begin{tabular}{@{}c l c c
  S[table-format=3.0]
  *{5}{S[table-format=-2.0]}
  c
  S[table-format=1.2] S[table-format=2.2]
  c
  S[table-format=2.0]
  c
  *{5}{S[
    table-format=2.0,
    table-fixed-width=true,
    table-column-width=\decidingwidth
  ]}}
\toprule
 & & & &
 & \multicolumn{5}{c}{\textbf{Component share$^\ddagger$}}
 & \hspace{3pt}
 & \multicolumn{2}{c}{\textbf{A-V density}}
 & \hspace{3pt}
 & {\textbf{A-V pref.}}
 & \hspace{8pt}
 & \multicolumn{5}{c}{%
     \makebox[0pt][c]{\textbf{Deciding component}}%
   } \\
\cmidrule(lr){6-10}
\cmidrule(lr){12-13}
\cmidrule(lr){15-15}
\cmidrule(l){17-21}
 & Model & Cov. & Role & {$n$}
 & \componenthead{Vcol}{V}
 & \componenthead{Acol}{A}
 & \componenthead{Qcol}{Q}
 & \componenthead{Gcol}{G}
 & \componenthead{Kcol}{K}
 & & {V/tok} & {A/tok}
 & & {A$>$V}
 & & \componenthead{Vcol}{\makebox[\decidingwidth][c]{V}}
 & \componenthead{Acol}{\makebox[\decidingwidth][c]{A}}
 & \componenthead{Qcol}{\makebox[\decidingwidth][c]{Q}}
 & \componenthead{Gcol}{\makebox[\decidingwidth][c]{G}}
 & \componenthead{Kcol}{\makebox[\decidingwidth][c]{K}} \\
\midrule

\multirow{12}{*}{%
  \rotatebox[origin=c]{90}{%
    \shortstack[c]{\textit{With}\\
                   \textit{value-projection}\\
                   \textit{bias}}%
  }%
}
 & \multirow{2}{*}{Omni 3B} & \multirow{2}{*}{76} & \textit{Subj.} & 333
 & 41 & 5 & 8 & 7 & 38
 & & 0.82 & 1.03 & & 5
 & & {\textbf{51}} & 3 & 3 & 12 & 30 \\
 & & & \textit{Pred.} & 210
 & 26 & 9 & 9 & 13 & 43
 & & 0.51 & 1.83 & & 19
 & & 26 & 19 & 0 & 14 & {\textbf{41}} \\
\addlinespace[2pt]
 & \multirow{2}{*}{Omni 7B} & \multirow{2}{*}{76} & \textit{Subj.} & 391
 & 33 & 5 & 10 & -5 & 58
 & & 0.64 & 0.94 & & 8
 & & 40 & 3 & 2 & 14 & {\textbf{41}} \\
 & & & \textit{Pred.} & 260
 & 21 & 8 & 7 & 5 & 59
 & & 0.43 & 1.60 & & 22
 & & 21 & 15 & 0 & 17 & {\textbf{47}} \\
\addlinespace[2pt]
 & \multirow{2}{*}{MCP 2.6} & \multirow{2}{*}{67} & \textit{Subj.} & 326
 & 23 & 3 & 12 & -26 & 88
 & & 0.44 & 2.28 & & 2
 & & 38 & 3 & 3 & 13 & {\textbf{43}} \\
 & & & \textit{Pred.} & 172
 & 15 & 5 & 12 & -16 & 84
 & & 0.31 & 4.90 & & 15
 & & 25 & 12 & 2 & 14 & {\textbf{48}} \\
\addlinespace[2pt]
 & \multirow{2}{*}{VL-AV} & \multirow{2}{*}{76} & \textit{Subj.} & 293
 & 29 & 10 & 8 & -12 & 65
 & & 8.32 & 1.24 & & 14
 & & 38 & 8 & 1 & 10 & {\textbf{43}} \\
 & & & \textit{Pred.} & 225
 & 18 & 12 & 11 & -12 & 71
 & & 5.04 & 1.52 & & 32
 & & 25 & 12 & 0 & 21 & {\textbf{41}} \\
\addlinespace[2pt]
 & \multirow{2}{*}{Ola 7B} & \multirow{2}{*}{75\rlap{$^\dagger$}}
 & \textit{Subj.} & 245
 & 22 & 8 & 26 & -19 & 62
 & & 0.39 & 5.48 & & 16
 & & {\textbf{40}} & 8 & 11 & 10 & 31 \\
 & & & \textit{Pred.} & 174
 & 14 & 10 & 23 & 0 & 52
 & & 0.31 & 8.02 & & 40
 & & {\textbf{32}} & 28 & 6 & 17 & 17 \\
\addlinespace[2pt]
 & \multirow{2}{*}{AVF} & \multirow{2}{*}{82} & \textit{Subj.} & 403
 & 27 & 8 & 6 & -4 & 64
 & & 1.11 & 1.48 & & 16
 & & 26 & 6 & 1 & 20 & {\textbf{47}} \\
 & & & \textit{Pred.} & 232
 & 15 & 12 & 7 & 6 & 60
 & & 0.65 & 2.54 & & 37
 & & 13 & 16 & 0 & 24 & {\textbf{47}} \\
\midrule

\multirow{6}{*}{%
  \rotatebox[origin=c]{90}{%
    \shortstack[c]{\textit{Without}\\
                   \textit{value-projection}\\
                   \textit{bias}}%
  }%
}
 & \multirow{2}{*}{MCP 4.5} & \multirow{2}{*}{74} & \textit{Subj.} & 308
 & 25 & 2 & 30 & 27 & 15
 & & 0.72 & 6.84 & & 0
 & & {\textbf{54}} & 1 & 18 & 24 & 2 \\
 & & & \textit{Pred.} & 239
 & 20 & 4 & 27 & 35 & 13
 & & 0.62 & 15.91 & & 6
 & & {\textbf{55}} & 17 & 8 & 18 & 3 \\
\addlinespace[2pt]
 & \multirow{2}{*}{vSAL 4B} & \multirow{2}{*}{77} & \textit{Subj.} & 349
 & 39 & 1 & 13 & 24 & 22
 & & 0.32 & 3.17 & & 0
 & & {\textbf{58}} & 0 & 4 & 24 & 15 \\
 & & & \textit{Pred.} & 216
 & 30 & 2 & 16 & 30 & 21
 & & 0.24 & 3.98 & & 0
 & & {\textbf{51}} & 0 & 4 & 19 & 25 \\
\addlinespace[2pt]
 & \multirow{2}{*}{vSAL 8B} & \multirow{2}{*}{75} & \textit{Subj.} & 333
 & 39 & 0 & 12 & 28 & 18
 & & 0.33 & 1.23 & & 0
 & & {\textbf{60}} & 0 & 7 & 23 & 10 \\
 & & & \textit{Pred.} & 210
 & 29 & 1 & 15 & 37 & 17
 & & 0.26 & 2.98 & & 0
 & & {\textbf{54}} & 1 & 8 & 17 & 20 \\
\bottomrule
\end{tabular}
\end{table*}

\begin{table*}[t]
\centering
\caption{Counterfactual source intervention.
Mean changes $\Delta C_s = C_s - C'_s$ and $\Delta\ell = \ell - \ell'$ in logit units after obscuring audio with silence (A) or video with black frames (V), reported separately for event subjects (\textit{Subj.}) and predicates (\textit{Pred.}); positive values denote a decrease after obscuration.
Bold marks the contribution of the obscured modality. $\Delta E$ and $\Delta\epsilon$ are omitted, so the component changes may not sum exactly to $\Delta\ell$.
}
\label{tab:counterfactual_intervention}

\setlength{\tabcolsep}{3pt}
\robustify\bfseries
\sisetup{
  table-number-alignment=center,
  text-series-to-math=true
}

\resizebox{\linewidth}{!}{%
\begin{tabular}{@{}c l c
  S[table-format=-2.2,table-model-setup=\bfseries]
  S[table-format=-1.2,table-model-setup=\bfseries]
  S[table-format=-1.2,table-model-setup=\bfseries]
  S[table-format=-1.2,table-model-setup=\bfseries]
  S[table-format=-1.2] S[table-format=-1.2]
  S[table-format=-2.2] S[table-format=-1.2]
  S[table-format=-2.2] S[table-format=-1.2]
  S[table-format=-2.2] S[table-format=-1.2]@{}}
\toprule
 & &
 & \multicolumn{2}{>{\columncolor{Vcol!45}[0pt][0pt]}c}{$\mathbf{\Delta C_{\mathrm V}}$}
 & \multicolumn{2}{>{\columncolor{Acol!45}[0pt][0pt]}c}{$\mathbf{\Delta C_{\mathrm A}}$}
 & \multicolumn{2}{>{\columncolor{Qcol!45}[0pt][0pt]}c}{$\mathbf{\Delta C_{\mathrm Q}}$}
 & \multicolumn{2}{>{\columncolor{Gcol!45}[0pt][0pt]}c}{$\mathbf{\Delta C_{\mathrm G}}$}
 & \multicolumn{2}{>{\columncolor{Kcol!45}[0pt][0pt]}c}{$\mathbf{\Delta C_{\mathrm K}}$}
 & \multicolumn{2}{c}{$\boldsymbol{\Delta\ell}$} \\
\cmidrule(l{\tabcolsep}r{\tabcolsep}){4-5}
\cmidrule(l{\tabcolsep}r{\tabcolsep}){6-7}
\cmidrule(l{\tabcolsep}r{\tabcolsep}){8-9}
\cmidrule(l{\tabcolsep}r{\tabcolsep}){10-11}
\cmidrule(l{\tabcolsep}r{\tabcolsep}){12-13}
\cmidrule(l{\tabcolsep}){14-15}
 & Model & Obs.
 & {\textit{Subj.}} & {\textit{Pred.}}
 & {\textit{Subj.}} & {\textit{Pred.}}
 & {\textit{Subj.}} & {\textit{Pred.}}
 & {\textit{Subj.}} & {\textit{Pred.}}
 & {\textit{Subj.}} & {\textit{Pred.}}
 & {\textit{Subj.}} & {\textit{Pred.}} \\
\midrule
\multirow{12}{*}{%
  \rotatebox[origin=c]{90}{%
    \shortstack[c]{\textit{With}\\
                   \textit{value-projection}\\
                   \textit{bias}}%
  }%
}
 & \multirow{2}{*}{Omni 3B}
 & A & -0.69 & -0.55 & \bfseries 0.66 & \bfseries 1.19
 & 0.02 & 0.02 & -0.12 & -0.13 & 0.08 & 0.15 & -0.06 & 0.69 \\
 & & V & \bfseries 7.02 & \bfseries 3.99 & -3.17 & -1.83
 & 0.18 & 0.29 & -1.36 & -1.11 & 0.64 & 0.60 & 3.30 & 1.94 \\
\addlinespace[2pt]
 & \multirow{2}{*}{Omni 7B}
 & A & -0.13 & -0.27 & \bfseries 0.67 & \bfseries 1.08
 & 0.16 & 0.04 & -0.38 & -0.46 & 0.68 & 0.61 & 1.00 & 1.00 \\
 & & V & \bfseries 5.48 & \bfseries 3.25 & -2.13 & -1.03
 & 0.17 & 0.06 & -2.00 & -1.46 & 1.70 & 1.22 & 3.22 & 2.05 \\
\addlinespace[2pt]
 & \multirow{2}{*}{MCP 2.6}
 & A & 0.12 & -0.33 & \bfseries 0.27 & \bfseries 1.72
 & -0.29 & -0.57 & -0.24 & -0.46 & -0.18 & -0.96 & -0.32 & -0.59 \\
 & & V & \bfseries 13.23 & \bfseries 8.58 & -2.05 & -0.64
 & 1.83 & 1.58 & -10.56 & -9.03 & 12.51 & 8.46 & 14.97 & 8.98 \\
\addlinespace[2pt]
 & \multirow{2}{*}{VL-AV}
 & A & -0.77 & -0.46 & \bfseries 1.62 & \bfseries 1.99
 & -0.04 & -0.01 & -0.17 & -0.54 & 0.58 & 1.06 & 1.22 & 2.04 \\
 & & V & \bfseries 5.96 & \bfseries 3.65 & -3.38 & -1.89
 & -0.36 & 0.12 & -1.70 & -2.38 & 3.19 & 3.69 & 3.73 & 3.19 \\
\addlinespace[2pt]
 & \multirow{2}{*}{Ola 7B}
 & A & -0.94 & -0.95 & \bfseries 1.24 & \bfseries 1.87
 & 0.52 & 0.27 & -0.27 & -0.35 & 0.33 & 0.36 & 0.89 & 1.20 \\
 & & V & \bfseries 3.51 & \bfseries 2.56 & -1.87 & -1.15
 & -0.69 & -0.08 & -1.28 & -1.70 & 0.96 & 1.61 & 0.62 & 1.25 \\
\addlinespace[2pt]
 & \multirow{2}{*}{AVF}
 & A & -0.51 & -0.51 & \bfseries 0.94 & \bfseries 1.52
 & 0.04 & 0.11 & -0.27 & -0.50 & 0.05 & 0.57 & 0.25 & 1.19 \\
 & & V & \bfseries 4.26 & \bfseries 2.50 & -0.78 & 0.03
 & -0.03 & 0.23 & -1.89 & -1.50 & 0.10 & -0.01 & 1.63 & 1.23 \\
\midrule
\multirow{6}{*}{%
  \rotatebox[origin=c]{90}{%
    \shortstack[c]{\textit{Without}\\
                   \textit{value-projection}\\
                   \textit{bias}}%
  }%
}
 & \multirow{2}{*}{MCP 4.5}
 & A & -0.47 & -0.72 & \bfseries 0.79 & \bfseries 2.01
 & -0.06 & -1.59 & -0.76 & -1.31 & -0.67 & -1.33 & -1.20 & -2.98 \\
 & & V & \bfseries 14.71 & \bfseries 9.70 & -4.62 & -0.98
 & 5.26 & 3.46 & -4.62 & -8.32 & 1.04 & -0.63 & 11.62 & 3.03 \\
\addlinespace[2pt]
 & \multirow{2}{*}{vSAL 4B}
 & A & 0.16 & 0.40 & \bfseries 0.14 & \bfseries 0.22
 & 0.07 & 0.19 & -0.23 & -0.41 & 0.01 & -0.01 & 0.14 & 0.38 \\
 & & V & \bfseries 7.20 & \bfseries 4.65 & -0.33 & -0.18
 & 0.24 & 0.76 & -3.86 & -3.49 & 0.04 & -0.01 & 3.22 & 1.71 \\
\addlinespace[2pt]
 & \multirow{2}{*}{vSAL 8B}
 & A & 0.12 & 0.43 & \bfseries 0.10 & \bfseries 0.24
 & 0.04 & 0.15 & -0.12 & -0.34 & -0.01 & -0.01 & 0.12 & 0.47 \\
 & & V & \bfseries 7.53 & \bfseries 4.80 & -0.34 & -0.22
 & -0.13 & 0.25 & -3.50 & -3.41 & -0.23 & -0.25 & 3.22 & 1.12 \\
\bottomrule
\end{tabular}%
}
\end{table*}

\section{Experimental evaluation}
\label{sec:experiments}

\subsection{Experimental setup}
\label{sec:setup}

\paragraph{Dataset and sampling.}
We use the test split of LU-AVS~\cite{liu2024benchmarking}, which contains 1,333 untrimmed videos annotated with sound events from 78 classes and their temporal extent.
We select 300 videos by stratified sampling on the class with the largest annotated duration in each video, excluding clips whose multimodal sequence would exceed 30,000 tokens; the subset contains 385 event instances covering all classes (Supp.~\cref{subsec:supp_dataset}).

\paragraph{Models and generation.}
We consider nine open-weight AVLLMs: Audio-Visual Flamingo (AVF)~\cite{Ghosh2026AudioVisualFO}, VideoLLaMA2.1-7B-AV (VL-AV)~\cite{damonlpsg2024videollama2}, Qwen2.5-Omni (Omni) 3B/7B~\cite{Qwen2.5-Omni}, MiniCPM-o (MCP) 2.6/4.5~\cite{yao2024minicpm,Cui2026MiniCPMo4T}, Ola-7B~\cite{Liu2025OlaPT}, and video-SALMONN 2 Pro (vSAL) 4B/8B~\cite{tang2025videosalmonn2captionenhancedaudiovisual}.
Each model uses its official checkpoint, native audio-visual preprocessing, chat template, and decoding configuration (greedy for six models, sampling with seed~0 for MiniCPM-o and Ola-7B); VL-AV processes eight 2-s audio segments rather than the full track. All models receive the same prompt: \emph{``Write a one-sentence caption that describes the scene and its sounds together, as a plain description.''}
After generation, MALT is applied in a teacher-forced forward pass over the generated caption, without altering the model output. Model-specific settings and implementation details are provided in Supp.~\cref{subsec:supp_models,subsec:supp_implementation}.

\subsection{Evaluation protocol}
\label{sec:protocol}

\paragraph{Generated words and event matching.}
\label{sec:event_matching}
Each generated word is represented by its first model token, which marks the onset of its generation and avoids overweighting multi-token words.
Captions are parsed with spaCy~\cite{honnibal2020spacy}, and content words are matched to the events annotated for each video using lexical, WordNet~\cite{miller-1994-wordnet}, and dependency rules.
Matched words are labeled \textit{predicates} if they denote the sound event (\eg, \emph{crying}, \emph{siren}) and \textit{subjects} if they denote its source (\eg, \emph{baby}, \emph{ambulance}); these roles are semantic rather than strictly grammatical.
All word-level attribution analyses are restricted to matched words (Supp.~\cref{subsec:supp_word_matching}).

\paragraph{Attribution metrics.}
We report the components $\{\mathrm{V},\mathrm{A},\mathrm{Q},\mathrm{G},\mathrm{K}\}$ separately for subjects and predicates; $E$ is retained only in the full decomposition.
\emph{Component Share} is a component's summed contribution divided by the summed target logit over matched words; shares are signed, may be negative or exceed $100\%$, and sum to $100\%$ minus the share of $E$, up to the residual.
\emph{Audio-Video Density} divides each word's video or audio contribution by the number of positions of that modality in its sample, then averages across words and scales to $10^3$ modality positions; as a measure of support per decoder position, it is compared only within models.
\emph{Audio-Video Preference} is the percentage of words whose audio contribution exceeds their video contribution.
\emph{Deciding Component} assigns each word to the component maximizing $C_s(y_t)-C_s(y_t^{\mathrm{alt}})$ (\cref{sec:source}), where $y_t^{\mathrm{alt}}$ is the highest-logit alternative token.
Formal definitions are given in Supp.~\cref{subsec:supp_metrics}.

\paragraph{Counterfactual intervention.}
We obscure audio with silence or video with black frames and recompute MALT on the same teacher-forced sequence, keeping the target token, preceding generated tokens, and prediction position fixed.
Denoting post-obscuration quantities with a prime, $\Delta C_s=C_s-C'_s$ and $\Delta\ell=\ell-\ell'$ satisfy
\begin{equation}
    \Delta\ell(y_t)=\sum_{s\in\mathcal S} \Delta C_s(y_t)+\Delta C_{\mathrm K}(y_t)+\Delta E(y_t)+\Delta\epsilon(y_t),
    \label{eq:counterfactual_decomposition}
\end{equation}
decomposing the net logit change into changes in individual contributions; positive values denote a decrease after obscuration.
To assess consistency, we compare $C_m(y_t)$ with the logit change $\Delta\ell_m(y_t)$ from obscuring $m\in\{\mathrm{V},\mathrm{A}\}$, reporting \emph{Sign Agreement} (percentage of words with matching signs among those with $|C_m|,|\Delta\ell_m|\geq0.1$), Pearson correlation, and \emph{Dominant Modality Agreement} (percentage of words for which the modality with larger $C_m$ also yields the larger signed logit change).

\paragraph{Controlled audio-video swaps.}
To test modality selectivity under conflicting evidence,
we use 25 pairs of distinct single-event 10-s clips
from the evaluation set, spanning 31 event classes
(Supp.~\cref{subsec:supp_swap}). For each pair, with video and audio streams $(V_1,A_1)$ and $(V_2,A_2)$, we generate captions for the two original combinations and the two cross-modal swaps $(V_1,A_2)$ and $(V_2,A_1)$, yielding 100 conditions, with the same prompt and decoding protocol as above.
Under swaps, words referring to the video event are called \emph{seen} and those referring to the audio event \emph{heard}.
Groups with fewer than five words are not interpreted.

\paragraph{Sensitivity analysis.}
We recompute MALT for $\lambda\in\{0,0.25,0.5,0.75,1\}$, keeping inputs, captions, and matched words fixed so that only the MLP rule changes, and report Component Share and Audio-Video Preference.

%% file: 5_results.tex
\begin{table*}[t]
\centering
\caption{Controlled audio-video swaps.
For each word group, $n$ is the number of matched generated words, and
$C_{\mathrm V}$ and $C_{\mathrm A}$ are the mean visual and audio
contributions in logit units.
Under cross-modal swaps, \textit{Seen words} and \textit{Heard words}
refer to the events carried by the video and audio streams, respectively,
and \textit{Caption counts} report how many of the 50 swapped conditions
yield captions mentioning only the seen event (Seen), only the heard
event (Heard), both, or neither.
Gray entries are based on fewer than five words and are not interpreted.
$^\dagger$Ola-7B processed 92/100 conditions (46/50 swaps) due to its context limit.
$^\ddagger$Audio-Visual Flamingo processed 96/100 conditions (48/50 swaps)
because four clips had too few frames.}
\label{tab:swap}

\setlength{\tabcolsep}{4pt}
\sisetup{table-number-alignment=center}
\newcommand{\swapscorewidth}{3.1em}
\newcommand{\swapscorehead}[2]{%
  \multicolumn{1}{%
    >{\columncolor{#1!45}%
      [\dimexpr\tabcolsep-1pt\relax]%
      [\dimexpr\tabcolsep-1pt\relax]}c%
  }{\makebox[\swapscorewidth][c]{$C_{\mathrm{#2}}$}}%
}

\newcommand{\swapgraycount}[1]{%
  \multicolumn{1}{S[table-format=2.0,color=gray]}{#1}%
}

\newcommand{\swapgrayscore}[1]{%
  \multicolumn{1}{S[
    table-format=-2.2,
    table-fixed-width=true,
    table-column-width=\swapscorewidth,
    color=gray
  ]}{#1}%
}

\begin{tabular}{@{}c l
  *{3}{
    S[table-format=2.0]
    *{2}{S[
      table-format=-2.2,
      table-fixed-width=true,
      table-column-width=\swapscorewidth
    ]}
  }
  *{4}{S[table-format=2.0]}@{}}
\toprule
 & & \multicolumn{3}{c}{\textbf{Original A--V}}
 & \multicolumn{10}{c}{\textbf{Cross-modal swaps}} \\
\cmidrule(lr){3-5} \cmidrule(l){6-15}
 & & \multicolumn{3}{c}{\textit{Matched words}}
 & \multicolumn{3}{c}{\textit{Seen words}}
 & \multicolumn{3}{c}{\textit{Heard words}}
 & \multicolumn{4}{c}{\textit{Caption counts}} \\
\cmidrule(lr){3-5} \cmidrule(lr){6-8}
\cmidrule(lr){9-11} \cmidrule(l){12-15}
 & Model
 & {$n$} & \swapscorehead{Vcol}{V} & \swapscorehead{Acol}{A}
 & {$n$} & \swapscorehead{Vcol}{V} & \swapscorehead{Acol}{A}
 & {$n$} & \swapscorehead{Vcol}{V} & \swapscorehead{Acol}{A}
 & {Seen} & {Heard} & {Both} & {Neither} \\
\midrule

\multirow{6}{*}{%
  \rotatebox[origin=c]{90}{%
    \shortstack[c]{\textit{With}\\
                   \textit{value-}\\
                   \textit{projection}\\
                   \textit{bias}}%
  }%
}
 & Omni 3B
 & 84 & 6.06 & 1.98
 & 63 & 8.32 & 0.19
 & 43 & 0.00 & 5.90
 & 19 & 2 & 26 & 3 \\
\addlinespace[3pt]
 & Omni 7B
 & 94 & 5.12 & 1.54
 & 64 & 7.50 & 0.14
 & 35 & 0.09 & 5.04
 & 21 & 3 & 23 & 3 \\
\addlinespace[3pt]
 & MCP 2.6
 & 64 & 12.52 & 3.18
 & 49 & 16.72 & 1.10
 & \swapgraycount{3} 
 & \swapgrayscore{1.10} 
 & \swapgrayscore{10.61}
 & 35 & 0 & 2 & 13 \\
\addlinespace[3pt]
 & VL-AV
 & 72 & 5.51 & 2.50
 & 53 & 7.76 & 0.22
 & 26 & -0.49 & 7.65
 & 28 & 5 & 11 & 6 \\
\addlinespace[3pt]
 & Ola 7B$^\dagger$
 & 62 & 3.83 & 1.95
 & 32 & 4.91 & 0.49
 & 39 & -0.04 & 4.46
 & 13 & 17 & 10 & 6 \\
\addlinespace[3pt]
 & AVF$^\ddagger$
 & 74 & 5.02 & 1.62
 & 54 & 6.97 & 0.18
 & 42 & -0.17 & 5.94
 & 14 & 3 & 27 & 4 \\
\midrule

\multirow{3}{*}{%
  \rotatebox[origin=c]{90}{%
    \shortstack[c]{\textit{Without}\\
                   \textit{value-}\\
                   \textit{projection}\\
                   \textit{bias}}%
  }%
}
 & MCP 4.5
 & 78 & 24.94 & 5.11
 & 60 & 28.75 & 2.19
 & 20 & 10.40 & 16.94
 & 31 & 2 & 12 & 5 \\
\addlinespace[3pt]
 & vSAL 4B
 & 72 & 9.58 & 0.47
 & 57 & 10.55 & 0.37
 & \swapgraycount{2} 
 & \swapgrayscore{3.35} 
 & \swapgrayscore{0.83}
 & 39 & 0 & 2 & 9 \\
\addlinespace[3pt]
 & vSAL 8B
 & 74 & 9.14 & 0.23
 & 62 & 9.88 & 0.13
 & \swapgraycount{3} 
 & \swapgrayscore{3.93} 
 & \swapgrayscore{1.74}
 & 39 & 0 & 3 & 8 \\
\bottomrule
\end{tabular}
\end{table*}

\section{Results}
\label{sec:experiments}

\subsection{Component attribution patterns}
\label{subsec:Component_attribution}

\Cref{tab:source_attribution} reports attribution results for subject and predicate words.
Video contributes more than audio in aggregate across all models and both roles.
In all nine models, visual shares are larger for subjects, whereas audio shares are larger for predicates.
The non-modal components differ substantially across architectures: $\mathrm{K}$ accounts for a much larger share in models with value-projection biases than in those without them (38--88\% vs.\ 13--22\%), consistent with the affine contributions included in $\mathrm{K}$ (\cref{eq:non_source}; Supp.~\cref{tab:supp_affine}). In models without value-projection biases, $C_{\mathrm K}^{\mathrm{aff}}=0$, so $C_{\mathrm K}$ consists entirely of causal-prefix reads. Normalizing by the number of modality positions reverses the video-audio ordering in eight of nine models for both roles, the exception being VideoLLaMA2.1-7B-AV. In these eight models, lower aggregate audio shares thus coexist with higher contributions per audio position than per video position.
Because densities are expressed in logit units per modality position, they depend on both the number of positions representing each modality and the model's logit scale; they therefore measure position-normalized support rather than modality importance.
At the word level, video contributions exceed audio for most event words in every model.
Audio contributions exceed video contributions more often for predicates than for subjects in seven models, and for almost no words in video-SALMONN 2 Pro, indicating a role-dependent attribution pattern. 
The contrastive analysis measures support for the generated token relative to its highest-logit alternative rather than aggregate support.
On predicates, the most frequent deciding component is video in the three decoders without value-projection biases and Ola-7B, and $\mathrm{K}$ in the remaining five models.
Audio is never the most frequent deciding component, consistent with semantically similar alternatives receiving comparable sensory contributions, which narrows audio margins.
Event coverage and 95\% confidence intervals are reported in Supp.~\cref{sec:supp_attribution_patterns}.

\subsection{Counterfactual intervention}
\label{subsec:faithfulness}

As shown in \Cref{tab:counterfactual_intervention}, obscuring either modality reduces its attributed contribution in every model and for both roles.
Audio obscuration reduces $C_{\mathrm A}$ more for predicates than for subjects, whereas video obscuration reduces $C_{\mathrm V}$ more for subjects than for predicates, in all nine models.
These role-dependent responses mirror the attribution patterns in \Cref{tab:source_attribution}. On average, video obscuration increases the audio contribution in all models except for Audio-Visual Flamingo predicates, where it is essentially unchanged, while audio obscuration increases the visual contribution in six of nine models.
The contribution of previously generated text increases under both interventions in every model ($\Delta C_{\mathrm G}<0$), consistent with the teacher-forced prefix retaining information from the original input.
In both MiniCPM-o variants, audio obscuration decreases $C_{\mathrm A}$ but increases the target logit on average ($\Delta\ell<0$), indicating that silence is not a neutral audio-removal intervention for these models.
At the word level, we test whether the original contribution $C_m$ predicts the logit change $\Delta\ell_m$ caused by obscuring modality $m$ (Supp.~\cref{subsec:supp_intervention}).
For video, Sign Agreement is 88--94\% in six models and 59--77\% in the remaining three, with positive correlations between $C_{\mathrm V}$ and $\Delta\ell_{\mathrm V}$ in all models.
For audio, Sign Agreement reaches 81--91\% for Qwen2.5-Omni 7B, VideoLLaMA2.1-7B-AV, and Ola-7B, and 61--65\% for video-SALMONN 2 Pro, whereas both MiniCPM-o reach 43--44\% with near-zero correlations.
It is also lower for Qwen2.5-Omni 3B and Audio-Visual Flamingo (48\% and 54\%) despite strong correlations ($r=0.82$ and $0.79$), consistent with visual compensation offsetting decreases in audio contribution: for Qwen2.5-Omni 3B subjects, the mean decrease in $C_{\mathrm A}$ is matched by an increase in $C_{\mathrm V}$, leaving $\Delta\ell$ near zero.
The modality with the larger contribution also produces the larger logit decrease for 77--91\% of words in seven models, compared with 68\% and 57\% in Ola-7B and MiniCPM-o~4.5.
A two-player Shapley reference, computed for three models, averages the marginal effects of removing each modality across both removal orders, thus also accounting for audio removal when video is absent (Supp.~\cref{subsec:supp_counterfactual_shapley}).
Against this reference, audio Sign Agreement reaches 95\% for Qwen2.5-Omni~3B and 97\% for VideoLLaMA2.1-7B-AV, but only 31\% for MiniCPM-o~4.5, where obscuring both modalities raises the median target logit from 106.75 to 129.38.
Overall, MALT attributions agree with intervention effects when obscuration suppresses modality-specific evidence, while the exact decomposition reveals compensatory shifts that can attenuate or reverse target-logit changes.
Agreement is weakest where the intervention itself is not neutral, most clearly for silence in MiniCPM-o.

\subsection{Controlled audio-video swaps}
\label{subsec:swaps}

\Cref{tab:swap} reports attributions under matched and conflicting audio-video evidence.
For the original combinations, video contributions exceed audio contributions in all models, consistent with \cref{tab:source_attribution}.
Under cross-modal swaps, words referring to the seen event remain video-dominant in all nine models.
Words referring to the heard event receive larger audio than video contributions in all six models with enough heard words to be interpreted, although MiniCPM-o~4.5 retains a substantial visual contribution; MiniCPM-o~2.6 and video-SALMONN 2 Pro produce only two or three heard words and are not interpreted.
Qwen2.5-Omni and Audio-Visual Flamingo frequently mention both events, whereas MiniCPM-o, VideoLLaMA2.1-7B-AV, and video-SALMONN 2 Pro more often mention only the seen event, and Ola-7B mentions only the heard event in 17 of 46 processed swaps.
Whether a model mentions the heard event thus depends on its architecture, but when it does, MALT attributes the corresponding words mainly to audio.
Complete results with 95\% confidence intervals are reported in Supp.~\cref{subsec:supp_swap_res}.

\subsection{Comparison with adapted attribution methods}
\label{subsec:baselines}

Since existing methods were not designed for source-level logit decomposition in AVLLMs, we adapt attention-based scores, Gradient$\times$Input, AttnLRP~\cite{arras2025closelook}, and ALTI-Logit~\cite{ferrando2023explaining} by aggregating them over modality positions (Supp.~\cref{subsec:supp_attr_met}).
Under its default pathway weights, DPA~\cite{jantsch2026dpa}
uses propagation rules equivalent to those of AttnLRP
(Supp.~\cref{sec:supp_dpa_equiv}).
We therefore treat AttnLRP as representative of this
equivalent configuration rather than reporting DPA as
a separate baseline.
We evaluate them on three models spanning high agreement (Qwen2.5-Omni~3B), the weakest agreement (MiniCPM-o~4.5), and a distinct audio pipeline (VideoLLaMA2.1-7B-AV), using single-modality interventions, a two-player Shapley reference, and controlled swaps (Supp.~\cref{tab:supp_baselines_consistency,tab:supp_baselines_shapley,tab:supp_baselines_swap}).
Attention-based scores agree less with intervention effects, and Gradient$\times$Input remains near or below chance in sign agreement. AttnLRP and ALTI-Logit perform comparably to MALT under interventions and the Shapley reference, as AttnLRP does under controlled swaps.
However, their position-level scores account for a median of only 18--33\% and 12--54\% of the target logit, respectively, leaving the remainder unreported.
MALT's terms other than $C_{\mathrm K}^{\mathrm{aff}}$, \ie, source, prefix, and decoder-input contributions, account for a median of 100\%, 63\%, and 24\% for MiniCPM-o~4.5, Qwen2.5-Omni~3B, and VideoLLaMA2.1-7B-AV; the remainder is $C_{\mathrm K}^{\mathrm{aff}}$, zero for the bias-free MiniCPM-o~4.5 and 37\% and 76\% for the other two.
Exact reconstruction therefore does not imply greater positional coverage (ALTI-Logit reaches 35\% on VideoLLaMA2.1-7B-AV), but MALT reports the non-positional term explicitly, so that the logit and its intervention-induced changes decompose exhaustively into component changes (\cref{eq:counterfactual_decomposition}).

\subsection{Sensitivity to the MLP rule}
\label{sec:sensitivity}

Varying $\lambda$ changes the absolute component shares, most strongly for generated context and the non-source component, whose balance across architectures therefore depends on this convention.
The main modality-level conclusions remain stable: video contributions exceed audio contributions for every model and both linguistic roles at all values of $\lambda$, and visual shares remain larger for subjects and audio shares larger for predicates, up to rounding.
Audio-video preference varies by at most 8 percentage points for predicates and 5 for subjects across the full sweep.
The main audio-video findings are therefore robust to this attribution convention, while absolute shares are reported under the symmetric choice $\lambda=0.5$.
Full results in Supp.~\cref{subsec:supp_lambda}.

%% file: 6_conclusions.tex
\section{Conclusions}
\label{sec:conclusions}

We introduced MALT, a framework that decomposes each generated-token logit of an AVLLM into signed contributions from video, audio, prompt, and generated-text positions, together with explicit non-source terms that complete an exact reconstruction.
Applied to nine AVLLMs, MALT shows that video positions dominate aggregate support, while audio positions contribute relatively more to predicates, which typically describe sound events, than to subjects.
Under conflicting evidence, when models mention the heard event, its words are attributed mainly to audio.
The share assigned to non-source terms is also tied to decoder architecture, being much larger in decoders with value-projection biases.
These attributions agree with modality interventions comparably to the strongest adapted decomposition methods, while, unlike them, MALT accounts for the full logit by reporting the part not attributable to input positions, mainly value-projection biases, as an explicit term rather than leaving it unreported; this exactness reveals how sources compensate for one another when evidence is removed.
Analyses of modality reliance in AVLLMs should therefore distinguish direct evidence from compensation rather than rely on ablations alone.
MALT measures support from the values read rather than the original provenance of information; extending it to multi-hop tracing and to decoder families beyond those evaluated here are natural next steps.

\section*{LLM Usage Statement}
OpenAI’s ChatGPT was used for language and consistency edits and Anthropic’s Claude to refine code. All AI-assisted content was reviewed by the authors, who take full responsibility for the submission.

%% file: X_suppl.tex
\clearpage
\setcounter{page}{1}
\maketitlesupplementary
\appendix

\startcontents[supp]
\startlist[supp]{lot}

\section*{Supplementary Contents}
\printcontents[supp]{}{1}{\setcounter{tocdepth}{2}}

\section{Additional method details}
\label{sec:supp_method}

This section collects the derivations supporting the propagation
rules in \cref{sec:tracing}. Throughout, $y_t$ is fixed, and the
RMS denominators, attention weights, and sigmoid factors are those
observed in the forward pass.

\subsection{RMSNorm and residual updates}
\label{sec:supp_rms}

Each RMSNorm computes $\hat{x}=\gamma\odot x/\operatorname{rms}(x)$,
with $\operatorname{rms}(x)=\sqrt{\lVert x\rVert_2^2/D+\eta}$, where
$\eta$ is the stability constant of that module. Holding the
observed denominator fixed~\cite{3692070.3692076}, the readout of a
direction $R$ at the output can be rewritten as
   \begin{equation}
       \langle R,\hat{x}\rangle=\langle R_{\mathrm{in}},x\rangle,
       \qquad
       R_{\mathrm{in}}=\frac{\gamma\odot R}{\operatorname{rms}(x)}.
       \label{eq:rms_propagation}
   \end{equation}
With the denominator fixed, $\hat{x}$ is linear in $x$, so any
additive decomposition of $x$ induces an additive decomposition of
$\hat{x}$. For a residual update $x'=x+b(x)$, linearity of the readout in its argument yields
\begin{equation}
    \langle R,x'\rangle=\langle R,x\rangle+\langle R,b(x)\rangle,
    \label{eq:residual_identity}
\end{equation}
so the skip path carries $R$ unchanged, while the branch term is
propagated through the corresponding operation.

\subsection{Attention routing and affine terms}
\label{sec:supp_attention}

At the prediction position $r$, the attention output of layer $l$ is
\begin{equation}
    a_r^l=b_O^l+\sum_h W_O^{l,h}\sum_{p\in\mathcal{P}_r}
          \alpha_{r,p}^{l,h}\,v_p^{l,\kappa(h)},
\end{equation}
where $b_O^l$ is the output-projection bias. We write each value as
$v_p^{l,\kappa(h)}=\tilde v_p^{l,\kappa(h)}+b_V^{l,\kappa(h)}$, and
take any absent bias to be zero. Holding the attention weights
fixed~\cite{ali2022xai}, the readout of $a_r^l$ along
$R^{l,\mathrm{mid}}$ is
   \begin{equation}
   \begin{aligned}
       \langle R^{l,\mathrm{mid}},a_r^l\rangle
       ={}&\sum_{p\in\mathcal{P}_r}\sum_h\alpha_{r,p}^{l,h}
           \bigl\langle(W_O^{l,h})^\top R^{l,\mathrm{mid}},
                        \tilde v_p^{l,\kappa(h)}\bigr\rangle\\
         &+\sum_h\Bigl(\sum_{p\in\mathcal{P}_r}\alpha_{r,p}^{l,h}\Bigr)
           \bigl\langle(W_O^{l,h})^\top R^{l,\mathrm{mid}},
                        b_V^{l,\kappa(h)}\bigr\rangle\\
         &+\langle R^{l,\mathrm{mid}},b_O^l\rangle.
   \end{aligned}
   \end{equation}
The first term is $\sum_{p}c_{y_t}^{l}(p)$
(\cref{eq:attention_contribution}). Since the weights of each head
sum to one, the remaining terms are independent of the attention
weights for a fixed readout direction and form the affine contribution
\begin{equation}
    C_{\mathrm K}^{l,\mathrm{aff}}(y_t)
    =\langle R^{l,\mathrm{mid}},b_O^l\rangle
     +\sum_h\bigl\langle(W_O^{l,h})^\top R^{l,\mathrm{mid}},
                         b_V^{l,\kappa(h)}\bigr\rangle,
    \label{eq:affine_contribution}
\end{equation}
so that
\begin{equation}
    \langle R^{l,\mathrm{mid}},a_r^l\rangle
    =\sum_{p\in\mathcal P_r}c_{y_t}^{l}(p)
     +C_{\mathrm K}^{l,\mathrm{aff}}(y_t).
    \label{eq:attention_readout}
\end{equation}
Because it is introduced by the projection biases rather than by
the content of any position, this term is not assigned to a source.
It is accumulated as
$C_{\mathrm K}^{\mathrm{aff}}(y_t)
=\sum_{l=1}^{L}C_{\mathrm K}^{l,\mathrm{aff}}(y_t)$ and reported in
the non-source component. Affine effects already absorbed into the
state of a position in earlier layers remain part of the value read
from that position and are assigned to its source.
\cref{tab:supp_affine} lists the bias terms present in each
evaluated decoder.

Query and key projections, including their biases and rotary
position embeddings, determine the observed weights
$\alpha_{r,p}^{l,h}$ but receive no additive term. Attributing these routing effects would require an additional convention for handling the softmax. Instead, fixing the observed attention weights yields an exact decomposition of the position-dependent value-read term, with projection biases accounted for separately.
No evaluated decoder contains additional affine attention terms
assigned to $C_{\mathrm K}^{\mathrm{aff}}$.

\begin{table}[t]
    \centering
    \caption{Presence of value-projection biases in the
    evaluated decoders. All models have $b_O^l=0$;
    thus, only value-projection biases can contribute
    to $C_{\mathrm{K}}^{\mathrm{aff}}(y_t)$.}
    \label{tab:supp_affine}
    \begin{tabular}{ll}
        \toprule
        Model & Value-projection bias \\
        \midrule
        Omni 3B/7B  & Present \\
        MCP 2.6     & Present \\
        MCP 4.5     & Absent \\
        VL-AV       & Present \\
        Ola-7B      & Present \\
        AVF         & Present \\
        vSAL 4B/8B  & Absent \\
        \bottomrule
    \end{tabular}
\end{table}

\subsection{Gated MLP}
\label{sec:supp_mlp}

\paragraph{Quadratic form.}
Let $w=(W_d^l)^\top R^l$ and $\bar{\sigma}=\sigma(z)$. Since
$\operatorname{SiLU}(z)=z\odot\sigma(z)$, the branch readout is
\begin{equation}
    \langle R^l,m^l(\hat{x})\rangle
    =\langle w,\operatorname{SiLU}(z)\odot u\rangle
    =\langle w\odot\bar{\sigma},z\odot u\rangle
    =f(\hat{x}).
\end{equation}
Writing the elementwise product as a diagonal matrix,
\begin{equation}
    f(\xi)=\xi^\top M\xi,
    \qquad
    M=(W_g^l)^\top\operatorname{diag}(w\odot\bar{\sigma})\,W_u^l.
    \label{eq:supp_mlp_quadratic}
\end{equation}

\paragraph{General propagation rule.}
For $\lambda\in[0,1]$, we define
   \begin{equation}
   \begin{aligned}
       R_{\mathrm{mlp}}^{\lambda}
       ={}&\lambda(W_g^l)^\top(w\odot\bar{\sigma}\odot u)\\
          &+(1-\lambda)(W_u^l)^\top(w\odot\bar{\sigma}\odot z)\\
       ={}&\bigl(\lambda M+(1-\lambda)M^\top\bigr)\hat{x},
   \end{aligned}
   \label{eq:supp_mlp_lambda}
   \end{equation}
where the second equality uses $z=W_g^l\hat{x}$ and
$u=W_u^l\hat{x}$. Since $\hat{x}^\top M\hat{x}=\hat{x}^\top
M^\top\hat{x}$, we have
$\langle R_{\mathrm{mlp}}^{\lambda},\hat{x}\rangle=f(\hat{x})$ for
every $\lambda$, so all rules in the family preserve the observed
readout. For $\lambda=1/2$,
$R_{\mathrm{mlp}}^{1/2}=\tfrac12(M+M^\top)\hat{x}
=\tfrac12\nabla f(\hat{x})$, which is the rule in
\cref{eq:mlp_direction}. For $\lambda=0$,
$R_{\mathrm{mlp}}^{0}=(W_u^l)^\top(w\odot\operatorname{SiLU}(z))$,
which is equivalent to holding the full gate activation fixed.

\paragraph{Allocation of self and cross terms.}
With the RMS denominator fixed, any additive decomposition of the
residual stream at $r$ induces an additive decomposition
$\hat{x}=\sum_k\hat{x}_k$ (\cref{sec:supp_rms}). By
\cref{eq:supp_mlp_lambda}, component $k$ receives
\begin{equation}
   \begin{aligned}
       \langle R_{\mathrm{mlp}}^{\lambda},\hat{x}_k\rangle
       ={}&\hat{x}_k^\top M\hat{x}_k\\
          &+\sum_{k'\neq k}\bigl(\lambda\,\hat{x}_k^\top M\hat{x}_{k'}
           +(1-\lambda)\,\hat{x}_{k'}^\top M\hat{x}_k\bigr).
   \end{aligned}
   \label{eq:supp_mlp_allocation}
   \end{equation}
Each self term $\hat{x}_k^\top M\hat{x}_k$ is assigned to its
component for every $\lambda$. Each cross term
$\hat{x}_k^\top M\hat{x}_{k'}+\hat{x}_{k'}^\top M\hat{x}_k$ is
shared between $k$ and $k'$; for $\lambda\neq1/2$, the division
depends on which component enters through the gate branch and which
through the up branch. Summing \cref{eq:supp_mlp_allocation} over $k$
recovers $f(\hat{x})$.

\paragraph{Shapley interpretation.}
Consider the cooperative game
$v(S)=f\bigl(\sum_{k\in S}\hat{x}_k\bigr)
=\sum_{i,j\in S}\hat{x}_i^\top M\hat{x}_j$ over the components.
The game is a sum of single-player terms $\hat{x}_k^\top
M\hat{x}_k$ and pairwise unanimity terms $\hat{x}_k^\top
M\hat{x}_{k'}+\hat{x}_{k'}^\top M\hat{x}_k$. By linearity and
symmetry of the Shapley value, each single-player term is assigned
to its player and each pairwise term is split equally between its
two players, which coincides with \cref{eq:supp_mlp_allocation} at
$\lambda=1/2$. This interpretation is local to each MLP: it holds
for the readout of a single block with $\bar{\sigma}$, the RMS
denominators, and $R^l$ fixed, and does not make the full
decomposition a Shapley value over the network.

\paragraph{Propagation to the residual stream.}
$R_{\mathrm{mlp}}$ is propagated through the pre-MLP RMSNorm by
\cref{eq:rms_propagation}, giving $R_{\mathrm{in}}$, and combined
with the skip path by \cref{eq:residual_identity}, giving
$R^{l,\mathrm{mid}}=R^l+R_{\mathrm{in}}$.


\subsection{Layer-wise conservation and position assignment}
\label{sec:supp_conservation}

\paragraph{Layer-wise identity.}
We show that the propagation rules in \cref{sec:tracing}
yield the complete logit decomposition in
\cref{eq:source_decomposition}.
By the residual updates in \cref{eq:block},    conservation of the MLP readout (\cref{eq:mlp_direction};
   \cref{eq:supp_mlp_lambda} for general $\lambda$), and RMSNorm propagation
(\cref{eq:rms_propagation}),
\begin{equation}
\begin{aligned}
    \langle R^l,x_r^l\rangle
    &=\langle R^l,x_r^{l,\mathrm{mid}}\rangle
      +\langle R_{\mathrm{mlp}},\hat{x}_r^l\rangle\\
    &=\langle R^{l,\mathrm{mid}},x_r^{l,\mathrm{mid}}\rangle\\
    &=\langle R^{l,\mathrm{mid}},x_r^{l-1}\rangle
      +\langle R^{l,\mathrm{mid}},a_r^l\rangle.
\end{aligned}
\label{eq:supp_layer_readout}
\end{equation}
Here $R^{l,\mathrm{mid}}$ combines the skip-path direction $R^l$
with the MLP direction propagated through its input RMSNorm.
The attention skip path carries this direction unchanged,
$R^{l-1}=R^{l,\mathrm{mid}}$.
Substituting the attention readout decomposition
(\cref{eq:attention_readout}) gives
\begin{equation}
\begin{aligned}
    \langle R^l,x_r^l\rangle
    ={}&\langle R^{l-1},x_r^{l-1}\rangle\\
       &+\sum_{p\in\mathcal{P}_r}c_{y_t}^{l}(p)
        +C_{\mathrm K}^{l,\mathrm{aff}}(y_t).
\end{aligned}
\label{eq:layer_conservation}
\end{equation}
The attention output also depends on $x_r^{l-1}$ through the
query and, when $r$ is among the readable positions, through its
value. With observed attention weights held fixed, query--key
routing receives no separate attribution. The value read from
$r$ is instead counted in $c_{y_t}^{l}(r)$, not propagated
again through the residual direction $R^{l-1}$.

\paragraph{Telescoping and reconstruction.}
Summing \cref{eq:layer_conservation} over all $L$ layers and
using the final readout identity (\cref{eq:logit}) cancels the
intermediate readouts:
\begin{equation}
\begin{aligned}
    \ell(y_t)
    ={}&\langle R^0,x_r^0\rangle\\
       &+\sum_{l=1}^{L}\sum_{p\in\mathcal{P}_r}
          c_{y_t}^{l}(p)\\
       &+\sum_{l=1}^{L}C_{\mathrm K}^{l,\mathrm{aff}}(y_t).
\end{aligned}
\label{eq:supp_telescoping}
\end{equation}
The first term is the direct decoder-input readout,
$E(y_t)=\langle R^0,x_r^0\rangle$.
It includes the original input at the prediction position,
measured along the propagated direction $R^0$, but excludes
the self-attention value reads from that position. Partitioning readable positions into the source groups and the
   causal prefix, with $C_{\mathrm K}$ as in \cref{eq:non_source},
   recovers \cref{eq:source_decomposition} with $\epsilon(y_t)=0$
   in exact arithmetic.
For finite-precision computation, we define the signed
reconstruction residual as
\begin{equation}
    \epsilon(y_t)=\ell(y_t)
    -\sum_{s\in\mathcal{S}}C_s(y_t)
    -C_{\mathrm K}(y_t)-E(y_t).
    \label{eq:supp_reconstruction_residual}
\end{equation}
This residual measures numerical reconstruction accuracy,
not attribution faithfulness.

\paragraph{Position assignment.}
For the evaluated AVLLMs, video and audio positions are identified
from the modality masks provided by each model. Positions before
the first multimodal token form the causal prefix and contribute
to $C_{\mathrm K}^{\mathrm{prefix}}$. Remaining prompt/template
positions, including model-specific structural tokens that follow
the first multimodal token (e.g., delimiters, timestamps, and
separators), belong to $\mathrm Q$; previously generated positions
belong to $\mathrm G$. A value read from the prediction position
$r$ belongs to $\mathrm Q$ at the first generation step and to
$\mathrm G$ at subsequent steps. The prefix boundary is defined
uniformly, but its resulting span is model-dependent because
input templates differ (\cref{subsec:supp_models}).

\section{Experimental setup details}
\label{sec:exp_setup}

\subsection{Dataset selection}
\label{subsec:supp_dataset}

The number of videos selected from each event-class stratum is proportional to its frequency in the full test split, with at least one video retained per class. Videos are then sampled uniformly without
replacement within each stratum. The sequence-length threshold  is used as a computational filter, since
multimodal sequence length depends on both video duration and spatial resolution. We estimate this cost using Qwen2.5-Omni 3B, whose native visual representation exhibits the largest sequence growth among the evaluated models. The filter therefore mainly affects long, high-resolution clips while retaining long videos at lower resolution. The sampled videos contain 616 annotated temporal segments. Multiple
segments corresponding to the same event class within a video are
merged into a single event instance for evaluation, yielding 385 event
instances in total.

\subsection{Models and generation}
\label{subsec:supp_models}

We follow the released audiovisual preprocessing and generation
configuration of each model.
\Cref{tab:supp_models} summarizes the main differences in video sampling
and decoding across the evaluated AVLLMs. All models process audio at
16\,kHz. VideoLLaMA2.1-7B-AV processes eight 2-s audio segments distributed across the video, while the remaining models process the full audio track through their native audiovisual pipelines. For source aggregation, multimodal content positions are identified from each model's decoder sequence.
Model-specific delimiters, timestamps, separators, and other structural
positions following the first multimodal token are assigned to
$\mathrm{Q}$, while positions preceding this boundary are retained in
$\mathrm{K}$.

\begin{table}[t]
\centering
\caption{Model-specific video sampling and decoding.
Frame counts indicate a fixed number of frames sampled uniformly across the clip, whereas fps indicates a fixed sampling rate.}
\label{tab:supp_models}
\setlength{\tabcolsep}{5pt}
\begin{tabular}{@{}lcc@{}}
\toprule
Model & Video sampling & Decoding \\
\midrule
Omni 3B/7B            & 2~fps          & Greedy   \\
MCP 2.6/4.5              & 1~fps          & Sampling \\
VL-AV           & 8~frames       & Greedy   \\
Ola-7B                         & 64~frames      & Sampling \\
AVF          & 128~frames     & Greedy   \\
vSAL 4B/8B      & 10~fps         & Greedy   \\
\bottomrule
\end{tabular}
\end{table}

\subsection{Implementation and numerical details}
\label{subsec:supp_implementation}

Experiments are conducted on a workstation equipped with an Intel Xeon
w9-3495X CPU, 512\,GB of RAM, and two NVIDIA RTX 5000 Ada Generation GPUs with 32\,GB of memory each. All environments use Python 3.11.16. We use a separate software environment for each model family to accommodate differences among the released codebases. The main software versions are reported in \Cref{tab:supp_environment}. We use the released
audiovisual preprocessing pipelines, which rely on libraries such as \texttt{qwen-omni-utils}, \texttt{decord}, MoviePy, \texttt{librosa}, PyAV, and OpenAI Whisper, depending on the model. Model forward passes use \texttt{bfloat16}, except for
VideoLLaMA2.1-7B-AV, which follows its native \texttt{float16} configuration. Residual states, value vectors, and embeddings are stored in native precision and cast to \texttt{float32} for the decomposition. MALT uses \texttt{float32} for readout propagation, MLP recomputation, value and output projections, dot products, and
contribution accumulation, with TF32 enabled for matrix
multiplications. Attention rows are recomputed in \texttt{float32} from the query and key representations after native query/key
normalization and positional encoding, where applicable. The
recomputation preserves the decoder's attention scaling, masking,
and query-to-key/value head mapping. Only rows at positions
predicting generated tokens are retained for attribution. The evaluated
MiniCPM-o 2.6 and 4.5 models use Qwen2 and Qwen3 decoders,
respectively, without the \texttt{scale\_depth} residual scaling of
the original MiniCPM LLM.
We assess numerical consistency from the residual
$\epsilon(y_t)$ between the MALT reconstruction and the corresponding
target-token logit. \Cref{tab:supp_reconstruction} summarizes the
absolute residual $|\epsilon(y_t)|$ over 74,771 generated tokens across
all models. The mean absolute residual is 0.041 logit units, and the
mean relative error is 0.11\%. MiniCPM-o exhibits larger absolute
residuals on substantially larger residual-stream and target-logit
scales, with relative reconstruction errors comparable to those of
the other \texttt{bfloat16} models.

\begin{table}[t]
\centering
\caption{Software environments for the evaluated model families.}
\label{tab:supp_environment}

\setlength{\tabcolsep}{5pt}
\begin{tabular}{@{}lccc@{}}
\toprule
Model              & PyTorch & CUDA & Transformers \\
\midrule
Omni 3B/7B        & 2.6.0   & 12.4 & 5.16.1       \\
MCP 2.6/4.5         & 2.6.0   & 12.4 & 4.51.0       \\
VL-AV       & 2.6.0   & 12.4 & 4.42.3       \\
Ola-7B                    & 2.6.0   & 12.4 & 4.43.4       \\
AVF     & 2.8.0   & 12.6 & 5.7.0.dev0   \\
vSAL 4B/8B & 2.8.0   & 12.6 & 4.57.6       \\
\bottomrule
\end{tabular}
\end{table}

\begin{table*}[t]
\centering
\caption{{Logit reconstruction error.}
Absolute errors are computed from the magnitude of the numerical
residual $|\epsilon(y_t)|$ for each generated token and reported in logit units. Relative errors are computed for each generated token and reported as mean percentages. The final row summarizes all generated tokens across models.}
\label{tab:supp_reconstruction}

\setlength{\tabcolsep}{5pt}
\sisetup{table-number-alignment=center}

\newcommand{\reconbiaslabel}[1]{%
  \rotatebox[origin=c]{90}{%
    \scriptsize
    \shortstack[c]{\textit{#1 value-}\\\textit{projection bias}}%
  }%
}
\newcommand{\reconbiasmid}[1]{%
  \noalign{%
    \vskip 2pt
    \vbox to 0pt{\vss\hbox{\reconbiaslabel{#1}}\vss}%
    \vskip 2pt
  }%
}
\newcommand{\reconbiascell}[1]{%
  \raisebox{\dimexpr
    (\ht\strutbox-\dp\strutbox-\height+\depth)/2\relax
  }[0pt][0pt]{\reconbiaslabel{#1}}%
}

\begin{tabular}{@{}c l
  S[table-format=1.3]
  S[table-format=1.3]
  S[table-format=1.3]
  S[table-format=1.2]@{}}
\toprule
 & & \multicolumn{3}{c}{\textbf{Absolute error}}
 & \multicolumn{1}{c}{\textbf{Relative error}} \\
\cmidrule(lr){3-5}
\cmidrule(l){6-6}
\smash{\phantom{\reconbiaslabel{Without}}}
 & Model & {Mean} & {95th percentile} & {Maximum} & {Mean} \\
\midrule
 & Qwen2.5-Omni 3B       & 0.023 & 0.057 & 0.153 & 0.13 \\
\addlinespace[4pt]
 & Qwen2.5-Omni 7B       & 0.025 & 0.066 & 0.174 & 0.12 \\
\addlinespace[4pt]
 & MiniCPM-o 2.6         & 0.109 & 0.289 & 0.619 & 0.14 \\
\reconbiasmid{With}
 & VideoLLaMA2.1-7B-AV   & 0.003 & 0.008 & 0.015 & 0.01 \\
\addlinespace[4pt]
 & Ola-7B                & 0.024 & 0.064 & 0.153 & 0.12 \\
\addlinespace[4pt]
 & Audio-Visual Flamingo & 0.022 & 0.057 & 0.152 & 0.11 \\
\midrule
 & MiniCPM-o 4.5         & 0.105 & 0.309 & 0.762 & 0.13 \\
\addlinespace[4pt]
\reconbiascell{Without}
 & video-SALMONN 2 Pro 4B & 0.028 & 0.071 & 0.173 & 0.11 \\
\addlinespace[4pt]
 & video-SALMONN 2 Pro 8B & 0.023 & 0.057 & 0.128 & 0.09 \\
 \addlinespace[2pt]
\toprule
 & All models            & 0.041 & 0.165 & 0.762 & 0.11 \\
\bottomrule
\end{tabular}
\end{table*}

\section{Evaluation protocol details}
\subsection{Generated-word matching and semantic roles}
\label{subsec:supp_word_matching}

Generated token strings are concatenated and parsed with spaCy \texttt{en\_core\_web\_sm} (v3.8.0). Each tokenizer token is assigned to the spaCy word associated with the largest character overlap, and the earliest assigned token defines the word-onset attribution used in our analyses. We consider nouns, proper nouns, verbs, and adjectives as content words; punctuation, special tokens, and all remaining part-of-speech categories are excluded from event matching. For each video, generated content words are matched only against the LU-AVS event classes annotated for that video. Matching is case-insensitive and considers the spaCy lemma and surface form of each
word. Label forms are additionally expanded using WordNet 3.0 synonyms from the first corresponding synset and derivationally related forms. Multi-word labels are matched at the level of their content words, and no manual aliases are used. A generated word may therefore match more than one annotated event, whereas unmatched content words are excluded from the event-conditioned analyses. Matched words are then assigned to a \emph{predicate} or \emph{subject} role. Verbal or \textit{-ing} forms in the event label are treated as predicates; for multi-word labels without such a form, the syntactic head is used instead, while the remaining content words are treated as subjects. Dependency rules additionally recover nominal subjects of matched verbal predicates and generic event verbs governing matched subjects, including passive constructions.

\subsection{{Attribution metrics formulation}}
\label{subsec:supp_metrics}
Let $\mathcal W$ denote the set of matched words for a given linguistic role.
For $s\in\{\mathrm{V},\mathrm{A},\mathrm{Q},\mathrm{G},\mathrm{K}\}$,
Component Share is
\begin{equation}
\mathrm{Share}_s
=
100\,
\frac{\sum_{w\in\mathcal W} C_s(w)}
     {\sum_{w\in\mathcal W} \ell(w)}.
\end{equation}

Because contributions are signed, Component Share is a ratio of signed sums and may be negative or exceed
$100\%$. The direct decoder-input term $E$ is not included among the reported
components.

For modality $m\in\{\mathrm{V},\mathrm{A}\}$, Audio-Video Density is
\begin{equation}
D_m
=
\frac{10^3}{|\mathcal W|}
\sum_{w\in\mathcal W}
\frac{C_m(w)}{|\mathcal P_m(w)|},
\end{equation}
where $\mathcal P_m(w)$ is the set of decoder positions associated with
modality $m$ in the sample containing $w$.

Audio-Video Preference is
\begin{equation}
\mathrm{Pref}_{A>V}
=
\frac{100}{|\mathcal W|}
\sum_{w\in\mathcal W}
\mathbf{1}\!\left[C_A(w)>C_V(w)\right].
\end{equation}

For each word $w$, let $y_w$ denote its first model token and
$y_w^{\mathrm{alt}}$ the highest-logit vocabulary token other than $y_w$.
We define
\begin{equation}
d_s(w)=C_s(y_w)-C_s(y_w^{\mathrm{alt}}),
\end{equation}
and assign $w$ to the component with the largest signed $d_s(w)$.
Deciding Component reports the percentage of words assigned to each component.

\subsection{Controlled audio-video swaps}
\label{subsec:supp_swap}

The 25 clip pairs were manually curated from the evaluation subset. Candidate clips contain a single annotated LU-AVS event extending for at least the first 10\,s of the video, and speech-related classes are excluded. Pairs were
selected to provide clear conflicting audio-video evidence: the two clips in each pair correspond to distinct event classes, belong to different event domains, and have non-overlapping lexical forms under the event-matching vocabulary. For swapped inputs, generated words are matched against the event labels of both clips in the pair, allowing words referring to the video and audio events to be identified separately as seen and heard, respectively.
The resulting pairs are listed in
\Cref{tab:supp_swap_pairs}.

\begin{table*}[t]
\centering
\caption{Clip pairs for the controlled audio-video swaps.
Clip identifiers refer to the original LU-AVS videos.}
\label{tab:supp_swap_pairs}

\setlength{\tabcolsep}{5pt}
\begin{tabular}{@{}c l l @{\hspace{1.25em}} l l@{}}
\toprule
 & \multicolumn{2}{c}{\textbf{Clip A}}
 & \multicolumn{2}{c}{\textbf{Clip B}} \\
\cmidrule(lr){2-3}
\cmidrule(rr){4-5}
Pair & ID & Event & ID & Event \\
\midrule
 1 & \texttt{6UYP\_v7eipM} & frog croaking
   & \texttt{5NqpmNqPDeQ} & chainsawing trees \\
 2 & \texttt{6YS3ewiOBkU} & lions roaring
   & \texttt{5o-s547x2e8} & helicopter \\
 3 & \texttt{28liZ3kli0s} & baby crying
   & \texttt{BxnLAGsByCI} & driving motorcycle \\
 4 & \texttt{2kOOgrG6LH0} & cat meowing
   & \texttt{2tFn61fIwQ0} & ambulance siren \\
 5 & \texttt{2BYBDRLHpzo} & frog croaking
   & \texttt{DhVTM1lxnYI} & typing on computer keyboard \\
 6 & \texttt{-z--VHgH4ho} & horse clip-clop
   & \texttt{21SK6lAVKrw} & playing banjo \\
 7 & \texttt{1YqmTvXOwKs} & church bell ringing
   & \texttt{4\_4D9gYSfeg} & playing harmonica \\
 8 & \texttt{2WogzOS\_D9Q} & chainsawing trees
   & \texttt{-Sz4z0QwEuM} & people burping \\
 9 & \texttt{1Bn0IuUuMI4} & train wheels squealing
   & \texttt{3CFm\_4LkBck} & playing flute \\
10 & \texttt{EXrUYmn6SXc} & hammering nails
   & \texttt{2Jt4iqSqNTg} & bird chirping \\
11 & \texttt{0TQQn\_S8ZDM} & helicopter
   & \texttt{ALdjQ9kbwd8} & playing accordion \\
12 & \texttt{5koOuSwqY1s} & vehicle honking
   & \texttt{3mebfy8rMKs} & frog croaking \\
13 & \texttt{1zmHC3OitT4} & playing snare drum
   & \texttt{-aMHwbGIJzg} & chainsawing trees \\
14 & \texttt{1TdCttdPZck} & tap dancing
   & \texttt{7DH5fqj8j6Q} & ambulance siren \\
15 & \texttt{BOvv1l0H\_to} & typing on computer keyboard
   & \texttt{5f7z2-05sBk} & chainsawing trees \\
16 & \texttt{IJXrQCDz-wc} & people eating
   & \texttt{2qBbIZsK0nQ} & playing flute \\
17 & \texttt{16DCCYYdgmg} & playing violin
   & \texttt{0ZGt4GjihSY} & helicopter \\
18 & \texttt{6hB1E-ge0UU} & playing accordion
   & \texttt{CZgx\_6XaEkg} & driving motorcycle \\
19 & \texttt{R1RGIgC-xKw} & playing banjo
   & \texttt{0lq5Oey7ChY} & baby crying \\
20 & \texttt{69qOqfMjBOg} & playing trombone
   & \texttt{0XcOI-i6DI0} & police car siren \\
21 & \texttt{-3kt0GobDIM} & playing accordion
   & \texttt{hfewjg0CTQM} & church bell ringing \\
22 & \texttt{8KWjhWRGBYY} & playing clarinet
   & \texttt{-GHPfOXLP2o} & car passing by \\
23 & \texttt{0zc-1aX4tJI} & playing trumpet
   & \texttt{xanOeYcQpcM} & lawn mowing \\
24 & \texttt{5S38QZLUtuU} & playing violin
   & \texttt{4-hRP3tS4Zw} & police car siren \\
25 & \texttt{5vPhf41ITLw} & playing cornet
   & \texttt{53aXTceSfrY} & helicopter \\
\bottomrule
\end{tabular}
\end{table*}

\section{Additional Results}
\subsection{Component attribution patterns}
\label{sec:supp_attribution_patterns}

\Cref{tab:supp_coverage} reports event coverage for each model. An annotated event is considered covered when at least one generated subject or predicate word is matched to that event. Coverage ranges from 67\% to 82\% across models. Ola-7B is evaluated on the 249 clips that fit within its native context limit, while all the others processed the entire evaluation subset. \Cref{tab:supp_aggregation_ic_95} reports the attribution metrics presented in \Cref{subsec:Component_attribution} along with their 95\% confidence intervals. We recomputed each metric on 1,000 bootstrap samples drawn with replacement at the clip level, keeping all words from the same caption together. The intervals support the main attribution patterns under clip-level resampling.
Because predicates may occur later than subjects, we additionally control for generation position by matching their relative caption-position distributions; the role-dependent attribution pattern persists after matching (\Cref{sec:supp_position_control}).

\begin{table*}[t]
\centering
\caption{Event coverage. {Total} counts the annotated event instances in the videos
processed by each model. An instance is {Matched} if at least
one generated subject or predicate word refers to it.
Coverage is the percentage of annotated instances matched.
The last three columns count instances matched by subjects only,
predicates only, or both.}
\label{tab:supp_coverage}

\setlength{\tabcolsep}{5pt}
\sisetup{table-number-alignment=center}

\newcommand{\covbiaslabel}[1]{%
  \rotatebox[origin=c]{90}{%
    \scriptsize
    \shortstack[c]{\textit{#1 value-}\\\textit{projection bias}}%
  }%
}
\newcommand{\covbiasmid}[1]{%
  \noalign{%
    \vskip 2pt
    \vbox to 0pt{\vss\hbox{\covbiaslabel{#1}}\vss}%
    \vskip 2pt
  }%
}
\newcommand{\covbiascell}[1]{%
  \raisebox{\dimexpr
    (\ht\strutbox-\dp\strutbox-\height+\depth)/2\relax
  }[0pt][0pt]{\covbiaslabel{#1}}%
}

\begin{tabular}{@{}c l
  S[table-format=3.0]
  S[table-format=3.0]
  S[table-format=2.0]
  S[table-format=3.0]
  S[table-format=2.0]
  S[table-format=3.0]@{}}
\toprule
 & & \multicolumn{3}{c}{\textbf{Event coverage}}
 & \multicolumn{3}{c}{\textbf{Match type}} \\
\cmidrule(lr){3-5}
\cmidrule(l){6-8}
\smash{\phantom{\covbiaslabel{Without}}}
 & Model
 & {Total} & {Matched} & {Coverage}
 & {Subj. only} & {Pred. only} & {Both} \\
\midrule
 & Qwen2.5-Omni 3B       & 385 & 291 & 76 &  89 & 23 & 179 \\
\addlinespace[4pt]
 & Qwen2.5-Omni 7B       & 385 & 292 & 76 &  84 & 21 & 187 \\
\addlinespace[4pt]
 & MiniCPM-o 2.6         & 385 & 259 & 67 &  99 & 17 & 143 \\
\covbiasmid{With}
 & VideoLLaMA2.1-7B-AV   & 385 & 294 & 76 &  77 & 33 & 184 \\
\addlinespace[4pt]
 & Ola-7B                & 310 & 234 & 75 &  63 & 22 & 149 \\
\addlinespace[4pt]
 & Audio-Visual Flamingo & 385 & 315 & 82 & 121 & 38 & 156 \\
\midrule
 & MiniCPM-o 4.5         & 385 & 286 & 74 &  62 & 36 & 188 \\
\addlinespace[4pt]
\covbiascell{Without}
 & video-SALMONN 2 Pro 4B & 385 & 295 & 77 & 103 & 39 & 153 \\
\addlinespace[4pt]
 & video-SALMONN 2 Pro 8B & 385 & 289 & 75 & 105 & 39 & 145 \\
\bottomrule
\end{tabular}
\end{table*}

\begin{table*}[t]
\centering
\caption{Attribution of generated event words by linguistic role with
95\% confidence intervals.
Results are reported separately for subjects (\textit{Subj.}) and
predicates (\textit{Pred.}), with $n$ denoting the number of matched
generated words.
Models are grouped by the presence of value-projection biases
(\cref{tab:supp_affine}).
Component share, A-V preference, and deciding component are reported
as percentages. A-V density is reported per $10^3$ modality positions.
Bold indicates the most frequent deciding component per row.
Brackets indicate 95\% confidence intervals.}
\label{tab:supp_aggregation_ic_95}

\scriptsize
\setlength{\tabcolsep}{2pt}

\newcommand{\aggci}[3]{%
  \shortstack[c]{%
    $#1$\\[1pt]
    {\setlength{\fboxsep}{1pt}%
     \colorbox{black!8}{$[#2,\,#3]$}}%
  }%
}
\newcommand{\aggboldci}[3]{%
  \aggci{\mathbf{#1}}{#2}{#3}%
}
\newcommand{\aggcomponenthead}[2]{%
  \multicolumn{1}{%
    >{\columncolor{#1!45}%
      [\dimexpr\tabcolsep-1pt\relax]%
      [\dimexpr\tabcolsep-1pt\relax]}c%
  }{#2}%
}
\newcommand{\aggbiasmid}[1]{%
  \noalign{%
    \vskip\dimexpr 5pt/2\relax
    \vbox to 0pt{%
      \vss
      \hbox{\rotatebox[origin=c]{90}{\textit{#1 value-projection bias}}}%
      \vss
    }%
    \vskip\dimexpr 5pt/2\relax
  }%
}

\resizebox{\linewidth}{!}{%
\begin{tabular}{@{}c l c c *{5}{c} *{2}{c} c *{5}{c}@{}}
\toprule
 & & &
 & \multicolumn{5}{c}{\textbf{Component share}}
 & \multicolumn{2}{c}{\textbf{A-V density}}
 & \textbf{A-V pref.}
 & \multicolumn{5}{c}{\textbf{Deciding component}} \\
\cmidrule(lr){5-9}
\cmidrule(lr){10-11}
\cmidrule(lr){12-12}
\cmidrule(l){13-17}
\smash{\phantom{\rotatebox[origin=c]{90}{%
  \textit{Without value-projection bias}}}}
 & Model & Role & $n$
 & \aggcomponenthead{Vcol}{V}
 & \aggcomponenthead{Acol}{A}
 & \aggcomponenthead{Qcol}{Q}
 & \aggcomponenthead{Gcol}{G}
 & \aggcomponenthead{Kcol}{K}
 & V/tok & A/tok & A$>$V
 & \aggcomponenthead{Vcol}{V}
 & \aggcomponenthead{Acol}{A}
 & \aggcomponenthead{Qcol}{Q}
 & \aggcomponenthead{Gcol}{G}
 & \aggcomponenthead{Kcol}{K} \\
\midrule

 & \multirow{2}{*}{Omni 3B} & \textit{Subj.} & 333
 & \aggci{41}{40}{43} & \aggci{5}{4}{5} & \aggci{8}{8}{9}
 & \aggci{7}{5}{9} & \aggci{38}{36}{40}
 & \aggci{0.82}{0.75}{0.89} & \aggci{1.03}{0.84}{1.23} & \aggci{5}{2}{7}
 & \aggboldci{51}{46}{56} & \aggci{3}{2}{5} & \aggci{3}{2}{5}
 & \aggci{12}{9}{15} & \aggci{30}{25}{35} \\
\addlinespace[5pt]
 & & \textit{Pred.} & 210
 & \aggci{26}{25}{28} & \aggci{9}{7}{10} & \aggci{9}{8}{10}
 & \aggci{13}{11}{16} & \aggci{43}{40}{45}
 & \aggci{0.51}{0.44}{0.59} & \aggci{1.83}{1.42}{2.22} & \aggci{19}{14}{24}
 & \aggci{26}{20}{32} & \aggci{19}{14}{24} & \aggci{0}{0}{1}
 & \aggci{14}{9}{19} & \aggboldci{41}{34}{49} \\
\addlinespace[7pt]

 & \multirow{2}{*}{Omni 7B} & \textit{Subj.} & 391
 & \aggci{33}{31}{34} & \aggci{5}{4}{5} & \aggci{10}{9}{11}
 & \aggci{-5}{-7}{-3} & \aggci{58}{56}{59}
 & \aggci{0.64}{0.57}{0.71} & \aggci{0.94}{0.81}{1.10} & \aggci{8}{6}{11}
 & \aggci{40}{35}{46} & \aggci{3}{1}{4} & \aggci{2}{1}{3}
 & \aggci{14}{10}{17} & \aggboldci{41}{36}{47} \\
\addlinespace[5pt]
 & & \textit{Pred.} & 260
 & \aggci{21}{20}{23} & \aggci{8}{6}{9} & \aggci{7}{6}{7}
 & \aggci{5}{2}{7} & \aggci{59}{57}{62}
 & \aggci{0.43}{0.36}{0.49} & \aggci{1.60}{1.30}{1.90} & \aggci{22}{17}{28}
 & \aggci{21}{15}{27} & \aggci{15}{11}{19} & \aggci{0}{0}{0}
 & \aggci{17}{12}{22} & \aggboldci{47}{41}{54} \\
\addlinespace[7pt]

 & \multirow{2}{*}{MCP 2.6} & \textit{Subj.} & 326
 & \aggci{23}{22}{24} & \aggci{3}{2}{3} & \aggci{12}{11}{13}
 & \aggci{-26}{-28}{-24} & \aggci{88}{87}{90}
 & \aggci{0.44}{0.40}{0.47} & \aggci{2.28}{1.92}{2.70} & \aggci{2}{1}{4}
 & \aggci{38}{32}{43} & \aggci{3}{1}{5} & \aggci{3}{2}{5}
 & \aggci{13}{9}{17} & \aggboldci{43}{38}{49} \\
\addlinespace[5pt]
 & & \textit{Pred.} & 172
 & \aggci{15}{14}{16} & \aggci{5}{4}{6} & \aggci{12}{11}{13}
 & \aggci{-16}{-18}{-14} & \aggci{84}{83}{86}
 & \aggci{0.31}{0.28}{0.34} & \aggci{4.90}{4.04}{5.96} & \aggci{15}{9}{20}
 & \aggci{25}{19}{31} & \aggci{12}{7}{17} & \aggci{2}{0}{5}
 & \aggci{14}{9}{20} & \aggboldci{48}{41}{55} \\
 \addlinespace[2pt]
\aggbiasmid{With}

 & \multirow{2}{*}{VL-AV} & \textit{Subj.} & 293
 & \aggci{29}{28}{31} & \aggci{10}{9}{11} & \aggci{8}{7}{9}
 & \aggci{-12}{-15}{-10} & \aggci{65}{63}{68}
 & \aggci{8.32}{7.81}{8.85} & \aggci{1.24}{1.11}{1.38} & \aggci{14}{10}{18}
 & \aggci{38}{32}{43} & \aggci{8}{5}{11} & \aggci{1}{0}{3}
 & \aggci{10}{7}{13} & \aggboldci{43}{37}{48} \\
\addlinespace[5pt]
 & & \textit{Pred.} & 225
 & \aggci{18}{16}{19} & \aggci{12}{10}{13} & \aggci{11}{10}{12}
 & \aggci{-12}{-15}{-9} & \aggci{71}{68}{74}
 & \aggci{5.04}{4.59}{5.52} & \aggci{1.52}{1.33}{1.71} & \aggci{32}{25}{38}
 & \aggci{25}{19}{31} & \aggci{12}{8}{17} & \aggci{0}{0}{0}
 & \aggci{21}{16}{26} & \aggboldci{41}{35}{48} \\
\addlinespace[7pt]

 & \multirow{2}{*}{Ola-7B} & \textit{Subj.} & 245
 & \aggci{22}{20}{24} & \aggci{8}{7}{9} & \aggci{26}{25}{28}
 & \aggci{-19}{-22}{-17} & \aggci{62}{60}{64}
 & \aggci{0.39}{0.35}{0.44} & \aggci{5.48}{4.80}{6.29} & \aggci{16}{11}{21}
 & \aggboldci{40}{34}{46} & \aggci{8}{5}{12} & \aggci{11}{7}{15}
 & \aggci{10}{6}{14} & \aggci{31}{25}{37} \\
\addlinespace[5pt]
 & & \textit{Pred.} & 174
 & \aggci{14}{13}{15} & \aggci{10}{9}{12} & \aggci{23}{22}{25}
 & \aggci{0}{-2}{2} & \aggci{52}{51}{54}
 & \aggci{0.31}{0.27}{0.35} & \aggci{8.02}{6.74}{9.35} & \aggci{40}{32}{48}
 & \aggboldci{32}{25}{39} & \aggci{28}{22}{35} & \aggci{6}{3}{10}
 & \aggci{17}{11}{23} & \aggci{17}{12}{23} \\
\addlinespace[7pt]

 & \multirow{2}{*}{AVF} & \textit{Subj.} & 403
 & \aggci{27}{25}{28} & \aggci{8}{7}{8} & \aggci{6}{5}{6}
 & \aggci{-4}{-7}{-2} & \aggci{64}{62}{67}
 & \aggci{1.11}{1.05}{1.17} & \aggci{1.48}{1.28}{1.71} & \aggci{16}{12}{20}
 & \aggci{26}{21}{30} & \aggci{6}{4}{9} & \aggci{1}{0}{2}
 & \aggci{20}{16}{24} & \aggboldci{47}{42}{52} \\
\addlinespace[5pt]
 & & \textit{Pred.} & 232
 & \aggci{15}{14}{17} & \aggci{12}{10}{13} & \aggci{7}{7}{8}
 & \aggci{6}{4}{7} & \aggci{60}{58}{62}
 & \aggci{0.65}{0.57}{0.72} & \aggci{2.54}{2.13}{3.05} & \aggci{37}{31}{43}
 & \aggci{13}{8}{18} & \aggci{16}{12}{22} & \aggci{0}{0}{0}
 & \aggci{24}{18}{29} & \aggboldci{47}{40}{54} \\
 \addlinespace[2pt]
\midrule

 & \multirow{2}{*}{MCP 4.5} & \textit{Subj.} & 308
 & \aggci{25}{24}{25} & \aggci{2}{2}{2} & \aggci{30}{28}{32}
 & \aggci{27}{25}{29} & \aggci{15}{14}{15}
 & \aggci{0.72}{0.66}{0.79} & \aggci{6.84}{5.79}{8.09} & \aggci{0}{0}{0}
 & \aggboldci{54}{48}{60} & \aggci{1}{0}{3} & \aggci{18}{14}{22}
 & \aggci{24}{19}{28} & \aggci{2}{1}{4} \\
\addlinespace[5pt]
 & & \textit{Pred.} & 239
 & \aggci{20}{19}{21} & \aggci{4}{4}{5} & \aggci{27}{26}{28}
 & \aggci{35}{33}{36} & \aggci{13}{13}{13}
 & \aggci{0.62}{0.56}{0.68} & \aggci{15.91}{13.39}{18.60} & \aggci{6}{4}{10}
 & \aggboldci{55}{49}{62} & \aggci{17}{12}{22} & \aggci{8}{5}{11}
 & \aggci{18}{13}{22} & \aggci{3}{1}{5} \\
\addlinespace[7pt]

 & \multirow{2}{*}{vSAL 4B} & \textit{Subj.} & 349
 & \aggci{39}{38}{41} & \aggci{1}{1}{1} & \aggci{13}{12}{13}
 & \aggci{24}{22}{25} & \aggci{22}{22}{23}
 & \aggci{0.32}{0.29}{0.36} & \aggci{3.17}{2.79}{3.67} & \aggci{0}{0}{0}
 & \aggboldci{58}{52}{63} & \aggci{0}{0}{0} & \aggci{4}{2}{6}
 & \aggci{24}{19}{28} & \aggci{15}{11}{19} \\
\aggbiasmid{Without}
 & & \textit{Pred.} & 216
 & \aggci{30}{29}{32} & \aggci{2}{2}{2} & \aggci{16}{15}{17}
 & \aggci{30}{29}{32} & \aggci{21}{21}{22}
 & \aggci{0.24}{0.21}{0.28} & \aggci{3.98}{3.44}{4.63} & \aggci{0}{0}{0}
 & \aggboldci{51}{44}{58} & \aggci{0}{0}{0} & \aggci{4}{2}{7}
 & \aggci{19}{14}{25} & \aggci{25}{19}{31} \\
\addlinespace[7pt]

 & \multirow{2}{*}{vSAL 8B} & \textit{Subj.} & 333
 & \aggci{39}{38}{41} & \aggci{0}{0}{1} & \aggci{12}{11}{13}
 & \aggci{28}{27}{29} & \aggci{18}{18}{19}
 & \aggci{0.33}{0.30}{0.37} & \aggci{1.23}{0.95}{1.74} & \aggci{0}{0}{0}
 & \aggboldci{60}{54}{65} & \aggci{0}{0}{0} & \aggci{7}{4}{10}
 & \aggci{23}{19}{28} & \aggci{10}{7}{14} \\
\addlinespace[5pt]
 & & \textit{Pred.} & 210
 & \aggci{29}{28}{30} & \aggci{1}{1}{1} & \aggci{15}{14}{16}
 & \aggci{37}{36}{38} & \aggci{17}{17}{18}
 & \aggci{0.26}{0.23}{0.31} & \aggci{2.98}{2.25}{3.75} & \aggci{0}{0}{0}
 & \aggboldci{54}{47}{61} & \aggci{1}{0}{3} & \aggci{8}{5}{13}
 & \aggci{17}{12}{22} & \aggci{20}{14}{25} \\
\bottomrule
\end{tabular}%
}
\end{table*}

\subsection{Counterfactual intervention}
\label{subsec:supp_intervention}
\Cref{tab:supp_counterfactual} reports the counterfactual changes presented
in \Cref{subsec:faithfulness}, while \Cref{tab:supp_consistency} reports the
corresponding word-level attribution-intervention consistency in terms of
Sign Agreement, Pearson correlation, and Dominant Modality Agreement.
Both tables include 95\% confidence intervals obtained from 1,000 bootstrap
samples drawn with replacement at the clip level, keeping all words from the
same caption together. Sensitivity to the choice of audio replacement is
examined in \Cref{subsec:supp_audio_replacement}.

\newcommand{\maltcounterci}[4]{%
  \shortstack[c]{%
    $#1$\if\relax\detokenize{#4}\relax\else\,#4\fi\\[1pt]
    {\scriptsize\setlength{\fboxsep}{1pt}%
     \colorbox{black!8}{$[#2,\,#3]$}}%
  }%
}

\begin{table*}[t]
\centering
\caption{Counterfactual intervention with confidence intervals.
Results are reported
as mean changes in logit units, with separate columns for event subjects
(\textit{Subj.}) and predicates (\textit{Pred.}). Each row identifies the
obscured modality (Obs.) as audio (A) or video (V). Brackets report 95\%
confidence intervals from 1,000 clip-level bootstrap resamples.}
\label{tab:supp_counterfactual}

\setlength{\tabcolsep}{2pt}
\scriptsize

\newcommand{\suppcfhead}[2]{%
  \multicolumn{2}{>{\columncolor{#1!45}[0pt][0pt]}c}%
    {$\mathbf{\Delta C_{\mathrm{#2}}}$}%
}
\newcommand{\suppcfbiaslabel}[1]{%
  \rotatebox[origin=c]{90}{\textit{#1 value-projection bias}}%
}
\newcommand{\suppcfbiasmid}[1]{%
  \noalign{%
    \vskip\dimexpr 5pt/2\relax
    \vbox to 0pt{\vss\hbox{\suppcfbiaslabel{#1}}\vss}%
    \vskip\dimexpr 5pt/2\relax
  }%
}

\resizebox{\linewidth}{!}{%
\begin{tabular}{@{}c l c *{6}{cc}@{}}
\toprule
 & &
 & \suppcfhead{Vcol}{V}
 & \suppcfhead{Acol}{A}
 & \suppcfhead{Qcol}{Q}
 & \suppcfhead{Gcol}{G}
 & \suppcfhead{Kcol}{K}
 & \multicolumn{2}{c}{$\boldsymbol{\Delta\ell}$} \\
\cmidrule(l{\tabcolsep}r{\tabcolsep}){4-5}
\cmidrule(l{\tabcolsep}r{\tabcolsep}){6-7}
\cmidrule(l{\tabcolsep}r{\tabcolsep}){8-9}
\cmidrule(l{\tabcolsep}r{\tabcolsep}){10-11}
\cmidrule(l{\tabcolsep}r{\tabcolsep}){12-13}
\cmidrule(l{\tabcolsep}r{\tabcolsep}){14-15}
\smash{\phantom{\suppcfbiaslabel{Without}}}
 & \textbf{Model} & \textbf{Obs.}
 & \textit{Subj.} & \textit{Pred.}
 & \textit{Subj.} & \textit{Pred.}
 & \textit{Subj.} & \textit{Pred.}
 & \textit{Subj.} & \textit{Pred.}
 & \textit{Subj.} & \textit{Pred.}
 & \textit{Subj.} & \textit{Pred.} \\
\midrule
 & \multirow{2}{*}{Omni 3B}
& A
& \maltcounterci{-0.69}{-0.75}{-0.64}{}
& \maltcounterci{-0.55}{-0.63}{-0.47}{}
& \maltcounterci{0.66}{0.56}{0.76}{}
& \maltcounterci{1.19}{0.96}{1.43}{}
& \maltcounterci{0.02}{0.01}{0.04}{}
& \maltcounterci{0.02}{-0.01}{0.05}{}
& \maltcounterci{-0.12}{-0.15}{-0.09}{}
& \maltcounterci{-0.13}{-0.20}{-0.07}{}
& \maltcounterci{0.08}{0.05}{0.11}{}
& \maltcounterci{0.15}{0.10}{0.21}{}
& \maltcounterci{-0.06}{-0.14}{0.04}{}
& \maltcounterci{0.69}{0.47}{0.95}{} \\
\addlinespace[5pt]
 & & V
& \maltcounterci{7.02}{6.73}{7.33}{}
& \maltcounterci{3.99}{3.73}{4.27}{}
& \maltcounterci{-3.17}{-3.41}{-2.94}{}
& \maltcounterci{-1.83}{-2.02}{-1.64}{}
& \maltcounterci{0.18}{0.13}{0.22}{}
& \maltcounterci{0.29}{0.25}{0.33}{}
& \maltcounterci{-1.36}{-1.55}{-1.20}{}
& \maltcounterci{-1.11}{-1.25}{-0.98}{}
& \maltcounterci{0.64}{0.52}{0.76}{}
& \maltcounterci{0.60}{0.48}{0.70}{}
& \maltcounterci{3.30}{3.04}{3.55}{}
& \maltcounterci{1.94}{1.65}{2.21}{} \\
\addlinespace[7pt]
 & \multirow{2}{*}{Omni 7B}
& A
& \maltcounterci{-0.13}{-0.19}{-0.07}{}
& \maltcounterci{-0.27}{-0.33}{-0.20}{}
& \maltcounterci{0.67}{0.58}{0.77}{}
& \maltcounterci{1.08}{0.90}{1.27}{}
& \maltcounterci{0.16}{0.12}{0.19}{}
& \maltcounterci{0.04}{0.01}{0.06}{}
& \maltcounterci{-0.38}{-0.44}{-0.33}{}
& \maltcounterci{-0.46}{-0.55}{-0.38}{}
& \maltcounterci{0.68}{0.59}{0.78}{}
& \maltcounterci{0.61}{0.51}{0.73}{}
& \maltcounterci{1.00}{0.88}{1.13}{}
& \maltcounterci{1.00}{0.82}{1.20}{} \\
\addlinespace[5pt]
 & & V
& \maltcounterci{5.48}{5.19}{5.75}{}
& \maltcounterci{3.25}{2.99}{3.53}{}
& \maltcounterci{-2.13}{-2.30}{-1.96}{}
& \maltcounterci{-1.03}{-1.18}{-0.88}{}
& \maltcounterci{0.17}{0.10}{0.25}{}
& \maltcounterci{0.06}{0.03}{0.10}{}
& \maltcounterci{-2.00}{-2.14}{-1.87}{}
& \maltcounterci{-1.46}{-1.60}{-1.31}{}
& \maltcounterci{1.70}{1.50}{1.92}{}
& \maltcounterci{1.22}{1.00}{1.44}{}
& \maltcounterci{3.22}{2.94}{3.52}{}
& \maltcounterci{2.05}{1.75}{2.35}{} \\
\addlinespace[5pt]
 & \multirow{2}{*}{MCP 2.6}
& A
& \maltcounterci{0.12}{-0.02}{0.27}{}
& \maltcounterci{-0.33}{-0.58}{-0.10}{}
& \maltcounterci{0.27}{0.07}{0.47}{}
& \maltcounterci{1.72}{1.22}{2.27}{}
& \maltcounterci{-0.29}{-0.38}{-0.22}{}
& \maltcounterci{-0.57}{-0.69}{-0.45}{}
& \maltcounterci{-0.24}{-0.40}{-0.07}{}
& \maltcounterci{-0.46}{-0.69}{-0.24}{}
& \maltcounterci{-0.18}{-0.47}{0.13}{}
& \maltcounterci{-0.96}{-1.51}{-0.47}{}
& \maltcounterci{-0.32}{-0.57}{-0.06}{}
& \maltcounterci{-0.59}{-1.09}{-0.11}{} \\
\addlinespace[7pt]
 & & V
& \maltcounterci{13.23}{12.58}{13.90}{}
& \maltcounterci{8.58}{7.78}{9.38}{}
& \maltcounterci{-2.05}{-2.34}{-1.76}{}
& \maltcounterci{-0.64}{-1.07}{-0.18}{}
& \maltcounterci{1.83}{1.56}{2.11}{}
& \maltcounterci{1.58}{1.33}{1.82}{}
& \maltcounterci{-10.56}{-11.15}{-9.96}{}
& \maltcounterci{-9.03}{-9.66}{-8.38}{}
& \maltcounterci{12.51}{11.44}{13.67}{}
& \maltcounterci{8.46}{7.23}{9.72}{}
& \maltcounterci{14.97}{13.57}{16.45}{}
& \maltcounterci{8.98}{7.44}{10.48}{} \\
\suppcfbiasmid{With}
 & \multirow{2}{*}{VL-AV}
& A
& \maltcounterci{-0.77}{-0.86}{-0.68}{}
& \maltcounterci{-0.46}{-0.55}{-0.38}{}
& \maltcounterci{1.62}{1.45}{1.80}{}
& \maltcounterci{1.99}{1.72}{2.27}{}
& \maltcounterci{-0.04}{-0.10}{0.01}{}
& \maltcounterci{-0.01}{-0.06}{0.05}{}
& \maltcounterci{-0.17}{-0.24}{-0.11}{}
& \maltcounterci{-0.54}{-0.66}{-0.41}{}
& \maltcounterci{0.58}{0.45}{0.72}{}
& \maltcounterci{1.06}{0.83}{1.29}{}
& \maltcounterci{1.22}{1.04}{1.42}{}
& \maltcounterci{2.04}{1.70}{2.40}{} \\
\addlinespace[5pt]
 & & V
& \maltcounterci{5.96}{5.62}{6.33}{}
& \maltcounterci{3.65}{3.34}{3.99}{}
& \maltcounterci{-3.38}{-3.61}{-3.14}{}
& \maltcounterci{-1.89}{-2.14}{-1.63}{}
& \maltcounterci{-0.36}{-0.47}{-0.25}{}
& \maltcounterci{0.12}{0.03}{0.20}{}
& \maltcounterci{-1.70}{-1.89}{-1.50}{}
& \maltcounterci{-2.38}{-2.61}{-2.15}{}
& \maltcounterci{3.19}{2.89}{3.50}{}
& \maltcounterci{3.69}{3.34}{4.01}{}
& \maltcounterci{3.73}{3.43}{4.05}{}
& \maltcounterci{3.19}{2.87}{3.57}{} \\
\addlinespace[7pt]
 & \multirow{2}{*}{Ola-7B}
& A
& \maltcounterci{-0.94}{-1.05}{-0.84}{}
& \maltcounterci{-0.95}{-1.06}{-0.84}{}
& \maltcounterci{1.24}{1.10}{1.38}{}
& \maltcounterci{1.87}{1.59}{2.13}{}
& \maltcounterci{0.52}{0.44}{0.59}{}
& \maltcounterci{0.27}{0.19}{0.36}{}
& \maltcounterci{-0.27}{-0.37}{-0.18}{}
& \maltcounterci{-0.35}{-0.48}{-0.23}{}
& \maltcounterci{0.33}{0.21}{0.49}{}
& \maltcounterci{0.36}{0.17}{0.54}{}
& \maltcounterci{0.89}{0.73}{1.07}{}
& \maltcounterci{1.20}{0.86}{1.52}{} \\
\addlinespace[5pt]
 & & V
& \maltcounterci{3.51}{3.24}{3.78}{}
& \maltcounterci{2.56}{2.32}{2.81}{}
& \maltcounterci{-1.87}{-2.01}{-1.74}{}
& \maltcounterci{-1.15}{-1.30}{-1.00}{}
& \maltcounterci{-0.69}{-0.76}{-0.62}{}
& \maltcounterci{-0.08}{-0.14}{-0.01}{}
& \maltcounterci{-1.28}{-1.41}{-1.13}{}
& \maltcounterci{-1.70}{-1.85}{-1.57}{}
& \maltcounterci{0.96}{0.83}{1.10}{}
& \maltcounterci{1.61}{1.41}{1.83}{}
& \maltcounterci{0.62}{0.41}{0.85}{}
& \maltcounterci{1.25}{0.95}{1.59}{} \\
\addlinespace[7pt]
 & \multirow{2}{*}{AVF}
& A
& \maltcounterci{-0.51}{-0.57}{-0.46}{}
& \maltcounterci{-0.51}{-0.58}{-0.44}{}
& \maltcounterci{0.94}{0.83}{1.07}{}
& \maltcounterci{1.52}{1.29}{1.78}{}
& \maltcounterci{0.04}{0.02}{0.06}{}
& \maltcounterci{0.11}{0.08}{0.14}{}
& \maltcounterci{-0.27}{-0.35}{-0.19}{}
& \maltcounterci{-0.50}{-0.62}{-0.40}{}
& \maltcounterci{0.05}{-0.04}{0.14}{}
& \maltcounterci{0.57}{0.40}{0.75}{}
& \maltcounterci{0.25}{0.13}{0.37}{}
& \maltcounterci{1.19}{0.90}{1.51}{} \\
\addlinespace[5pt]
 & & V
& \maltcounterci{4.26}{4.03}{4.49}{}
& \maltcounterci{2.50}{2.19}{2.78}{}
& \maltcounterci{-0.78}{-0.91}{-0.67}{}
& \maltcounterci{0.03}{-0.15}{0.19}{}
& \maltcounterci{-0.03}{-0.08}{0.01}{}
& \maltcounterci{0.23}{0.18}{0.29}{}
& \maltcounterci{-1.89}{-2.07}{-1.72}{}
& \maltcounterci{-1.50}{-1.71}{-1.26}{}
& \maltcounterci{0.10}{-0.12}{0.32}{}
& \maltcounterci{-0.01}{-0.22}{0.21}{}
& \maltcounterci{1.63}{1.38}{1.91}{}
& \maltcounterci{1.23}{1.00}{1.48}{} \\
\addlinespace[2pt]
\midrule
 & \multirow{2}{*}{MCP 4.5}
& A
& \maltcounterci{-0.47}{-0.85}{-0.08}{}
& \maltcounterci{-0.72}{-1.59}{0.09}{}
& \maltcounterci{0.79}{0.62}{0.96}{}
& \maltcounterci{2.01}{1.61}{2.43}{}
& \maltcounterci{-0.06}{-0.57}{0.43}{}
& \maltcounterci{-1.59}{-2.29}{-0.92}{}
& \maltcounterci{-0.76}{-1.34}{-0.25}{}
& \maltcounterci{-1.31}{-2.28}{-0.43}{}
& \maltcounterci{-0.67}{-1.05}{-0.31}{}
& \maltcounterci{-1.33}{-1.87}{-0.80}{}
& \maltcounterci{-1.20}{-2.78}{0.19}{}
& \maltcounterci{-2.98}{-5.71}{-0.37}{} \\
\addlinespace[5pt]
 & & V
& \maltcounterci{14.71}{13.50}{15.87}{}
& \maltcounterci{9.70}{8.65}{10.79}{}
& \maltcounterci{-4.62}{-5.25}{-4.02}{}
& \maltcounterci{-0.98}{-1.42}{-0.50}{}
& \maltcounterci{5.26}{3.73}{6.80}{}
& \maltcounterci{3.46}{2.14}{4.72}{}
& \maltcounterci{-4.62}{-6.19}{-3.15}{}
& \maltcounterci{-8.32}{-10.40}{-6.31}{}
& \maltcounterci{1.04}{-0.04}{2.19}{}
& \maltcounterci{-0.63}{-1.62}{0.35}{}
& \maltcounterci{11.62}{6.97}{16.01}{}
& \maltcounterci{3.03}{-1.74}{7.98}{} \\
\addlinespace[7pt]
 & \multirow{2}{*}{vSAL 4B}
& A
& \maltcounterci{0.16}{0.09}{0.24}{}
& \maltcounterci{0.40}{0.30}{0.52}{}
& \maltcounterci{0.14}{0.12}{0.17}{}
& \maltcounterci{0.22}{0.19}{0.26}{}
& \maltcounterci{0.07}{0.05}{0.10}{}
& \maltcounterci{0.19}{0.13}{0.25}{}
& \maltcounterci{-0.23}{-0.29}{-0.18}{}
& \maltcounterci{-0.41}{-0.52}{-0.33}{}
& \maltcounterci{0.01}{-0.01}{0.03}{}
& \maltcounterci{-0.01}{-0.04}{0.02}{}
& \maltcounterci{0.14}{0.05}{0.26}{}
& \maltcounterci{0.38}{0.26}{0.50}{} \\
\suppcfbiasmid{Without}
 & & V
& \maltcounterci{7.20}{6.83}{7.58}{}
& \maltcounterci{4.65}{4.30}{4.98}{}
& \maltcounterci{-0.33}{-0.36}{-0.30}{}
& \maltcounterci{-0.18}{-0.21}{-0.15}{}
& \maltcounterci{0.24}{0.18}{0.30}{}
& \maltcounterci{0.76}{0.67}{0.86}{}
& \maltcounterci{-3.86}{-4.12}{-3.62}{}
& \maltcounterci{-3.49}{-3.75}{-3.21}{}
& \maltcounterci{0.04}{-0.04}{0.11}{}
& \maltcounterci{-0.01}{-0.07}{0.04}{}
& \maltcounterci{3.22}{2.95}{3.54}{}
& \maltcounterci{1.71}{1.41}{1.99}{} \\
\addlinespace[7pt]
 & \multirow{2}{*}{vSAL 8B}
& A
& \maltcounterci{0.12}{0.08}{0.16}{}
& \maltcounterci{0.43}{0.32}{0.54}{}
& \maltcounterci{0.10}{0.08}{0.13}{}
& \maltcounterci{0.24}{0.19}{0.30}{}
& \maltcounterci{0.04}{0.01}{0.07}{}
& \maltcounterci{0.15}{0.09}{0.21}{}
& \maltcounterci{-0.12}{-0.17}{-0.08}{}
& \maltcounterci{-0.34}{-0.42}{-0.25}{}
& \maltcounterci{-0.01}{-0.03}{0.01}{}
& \maltcounterci{-0.01}{-0.04}{0.01}{}
& \maltcounterci{0.12}{0.06}{0.19}{}
& \maltcounterci{0.47}{0.33}{0.61}{} \\
\addlinespace[5pt]
 & & V
& \maltcounterci{7.53}{7.17}{7.90}{}
& \maltcounterci{4.80}{4.49}{5.12}{}
& \maltcounterci{-0.34}{-0.37}{-0.31}{}
& \maltcounterci{-0.22}{-0.24}{-0.19}{}
& \maltcounterci{-0.13}{-0.19}{-0.06}{}
& \maltcounterci{0.25}{0.16}{0.35}{}
& \maltcounterci{-3.50}{-3.77}{-3.21}{}
& \maltcounterci{-3.41}{-3.67}{-3.17}{}
& \maltcounterci{-0.23}{-0.30}{-0.16}{}
& \maltcounterci{-0.25}{-0.30}{-0.20}{}
& \maltcounterci{3.22}{2.95}{3.50}{}
& \maltcounterci{1.12}{0.91}{1.36}{} \\
\bottomrule
\end{tabular}%
}
\end{table*}

\begin{table*}[t]
\centering
\caption{Attribution-intervention consistency. For each matched subject
or predicate word ($n$), the original modality contribution $C_m$ is
compared with the target-logit change $\Delta\ell_m$ induced by obscuring
the same modality.
Sign Agreement (Sign Agr.) is reported for
$|C_m|,|\Delta\ell_m|\geq0.1$ and, separately, for
$C_m\geq1$ logit while retaining
$|\Delta\ell_m|\geq0.1$, with subset sizes in parentheses.
$r$ denotes the Pearson correlation between $C_m$ and $\Delta\ell_m$.
Dominant agreement (Dominant Agr.) denotes agreement on the modality
with the higher score.
Agreement metrics are percentages. Brackets report 95\% confidence
intervals from 1,000 clip-level bootstrap resamples.}
\label{tab:supp_consistency}

\setlength{\tabcolsep}{3.5pt}
\footnotesize

\newcommand{\maltconsbiaslabel}[1]{%
  \rotatebox[origin=c]{90}{%
    \scriptsize
    \shortstack[c]{\textit{#1 value-}\\\textit{projection bias}}%
  }%
}
\newcommand{\maltconsbiasmid}[1]{%
  \noalign{%
    \vskip\dimexpr 5pt/2\relax
    \vbox to 0pt{\vss\hbox{\maltconsbiaslabel{#1}}\vss}%
    \vskip\dimexpr 5pt/2\relax
  }%
}
\newsavebox{\maltconsrowref}
\sbox{\maltconsrowref}{\maltcounterci{0.00}{-0.00}{0.00}{}}
\newcommand{\maltconsbiascell}[1]{%
  \raisebox{\dimexpr
    (\ht\maltconsrowref-\dp\maltconsrowref-\height+\depth)/2\relax
  }[0pt][0pt]{\maltconsbiaslabel{#1}}%
}

\begin{tabular}{@{}c l c ccc ccc c@{}}
\toprule
 & & & \multicolumn{3}{c}{\textbf{Video obscured}}
     & \multicolumn{3}{c}{\textbf{Audio obscured}} & \\
\cmidrule(lr){4-6}
\cmidrule(lr){7-9}
\smash{\phantom{\maltconsbiaslabel{Without}}}
 & Model & $n$
 & Sign Agr. & $C_{\mathrm V}\!\geq\!1$ & $r$
 & Sign Agr. & $C_{\mathrm A}\!\geq\!1$ & $r$
 & Dominant Agr. \\
\midrule

 & Omni 3B & 543
 & \maltcounterci{93}{91}{96}{}
 & \maltcounterci{94}{91}{96}{(512)}
 & \maltcounterci{0.46}{0.40}{0.51}{}
 & \maltcounterci{48}{42}{54}{}
 & \maltcounterci{76}{68}{84}{(134)}
 & \maltcounterci{0.82}{0.75}{0.86}{}
 & \maltcounterci{91}{88}{94}{} \\
\addlinespace[5pt]
 & Omni 7B & 651
 & \maltcounterci{91}{89}{94}{}
 & \maltcounterci{92}{89}{94}{(577)}
 & \maltcounterci{0.47}{0.42}{0.52}{}
 & \maltcounterci{91}{89}{94}{}
 & \maltcounterci{93}{90}{97}{(184)}
 & \maltcounterci{0.57}{0.48}{0.65}{}
 & \maltcounterci{82}{78}{85}{} \\
\addlinespace[5pt]
 & MCP 2.6 & 498
 & \maltcounterci{92}{89}{95}{}
 & \maltcounterci{92}{89}{95}{(497)}
 & \maltcounterci{0.58}{0.52}{0.63}{}
 & \maltcounterci{43}{37}{48}{}
 & \maltcounterci{41}{36}{47}{(361)}
 & \maltcounterci{0.13}{-0.10}{0.35}{}
 & \maltcounterci{90}{86}{93}{} \\
\maltconsbiasmid{With}
 & VL-AV & 518
 & \maltcounterci{94}{92}{97}{}
 & \maltcounterci{96}{94}{97}{(463)}
 & \maltcounterci{0.30}{0.20}{0.39}{}
 & \maltcounterci{91}{88}{93}{}
 & \maltcounterci{94}{91}{97}{(341)}
 & \maltcounterci{0.78}{0.72}{0.83}{}
 & \maltcounterci{82}{79}{86}{} \\
\addlinespace[5pt]
 & Ola-7B & 419
 & \maltcounterci{65}{59}{70}{}
 & \maltcounterci{69}{63}{74}{(340)}
 & \maltcounterci{0.50}{0.43}{0.57}{}
 & \maltcounterci{81}{76}{85}{}
 & \maltcounterci{93}{89}{96}{(213)}
 & \maltcounterci{0.76}{0.70}{0.81}{}
 & \maltcounterci{68}{63}{73}{} \\
\addlinespace[5pt]
 & AVF & 635
 & \maltcounterci{77}{73}{81}{}
 & \maltcounterci{78}{74}{82}{(516)}
 & \maltcounterci{0.35}{0.27}{0.42}{}
 & \maltcounterci{54}{49}{58}{}
 & \maltcounterci{76}{71}{80}{(277)}
 & \maltcounterci{0.79}{0.74}{0.83}{}
 & \maltcounterci{77}{74}{80}{} \\
\midrule

 & MCP 4.5 & 547
 & \maltcounterci{59}{55}{63}{}
 & \maltcounterci{59}{55}{63}{(544)}
 & \maltcounterci{0.21}{0.14}{0.27}{}
 & \maltcounterci{44}{40}{49}{}
 & \maltcounterci{45}{40}{49}{(416)}
 & \maltcounterci{0.03}{-0.09}{0.14}{}
 & \maltcounterci{57}{53}{62}{} \\
\addlinespace[5pt]
\maltconsbiascell{Without}
 & vSAL 4B & 565
 & \maltcounterci{91}{88}{94}{}
 & \maltcounterci{91}{88}{94}{(549)}
 & \maltcounterci{0.69}{0.64}{0.73}{}
 & \maltcounterci{61}{56}{65}{}
 & \maltcounterci{80}{57}{100}{(15)}
 & \maltcounterci{0.29}{0.18}{0.39}{}
 & \maltcounterci{82}{79}{85}{} \\
\addlinespace[5pt]
 & vSAL 8B & 543
 & \maltcounterci{88}{85}{91}{}
 & \maltcounterci{88}{85}{91}{(523)}
 & \maltcounterci{0.67}{0.62}{0.72}{}
 & \maltcounterci{65}{58}{73}{}
 & \maltcounterci{100}{100}{100}{(11)}
 & \maltcounterci{0.42}{0.28}{0.55}{}
 & \maltcounterci{79}{76}{83}{} \\
\bottomrule
\end{tabular}
\end{table*}

\subsubsection{Two-player Shapley}
\label{subsec:supp_counterfactual_shapley}

The single-modality interventions in \Cref{tab:supp_consistency} measure the
marginal effect of removing one modality while the other remains present.
To additionally account for the order in which video and audio are removed,
we construct a two-player Shapley reference for the three models used in our
baseline comparison. Let $f(VA)$ denote the target-token logit on the original
input, $f(V)$ the logit with audio replaced by silence, $f(A)$ the logit with
video replaced by black frames, and $f(\emptyset)$ the logit with both
modalities obscured. All four conditions use the same teacher-forced generated
sequence. The Shapley values are
\begin{align}
\phi_V
&=
\frac{1}{2}\left[f(VA)-f(A)\right]
+
\frac{1}{2}\left[f(V)-f(\emptyset)\right],
\\
\phi_A
&=
\frac{1}{2}\left[f(VA)-f(V)\right]
+
\frac{1}{2}\left[f(A)-f(\emptyset)\right].
\end{align}
They satisfy
\begin{equation}
\phi_V+\phi_A=f(VA)-f(\emptyset).
\end{equation}
Thus, unlike the single-removal reference, each modality is evaluated both
with and without the other modality present.

We additionally quantify non-additivity between the two interventions as
\begin{equation}
I
=
f(VA)-f(V)-f(A)+f(\emptyset).
\end{equation}
Here, $I=0$ corresponds to additive video and audio effects under these
perturbations, while nonzero values capture differences between the two
removal orders.

\Cref{tab:supp_counterfactual_shapley} compares the original MALT contributions $C_V$ and $C_A$
with $\phi_V$ and $\phi_A$ on the same matched words used in
\Cref{tab:supp_consistency}. We report Sign Agreement, Pearson correlation,
and Dominant Modality Agreement metrics, together with 95\% confidence
intervals from 1,000 clip-level bootstrap resamples.

\begin{table*}[t]
\centering
\caption{Agreement between MALT and the two-player Shapley reference.
$n$ denotes the number of matched subject and predicate words.
Original MALT modality contributions ($C_{\mathrm V}$, $C_{\mathrm A}$)
are evaluated against the corresponding Shapley values
($\phi_{\mathrm V}$, $\phi_{\mathrm A}$).
Sign agreement (Sign Agr.) is reported for
$|C_m|,|\phi_m|\geq0.1$. The $C_m\geq1$ columns further restrict Sign Agreement to positive
contributions of at least one logit unit, while retaining the
$|\phi_m|\geq0.1$ threshold; subset sizes are shown in parentheses.
Pearson's $r$ denotes correlation between $C_m$ and $\phi_m$.
Dominant agreement (Dominant Agr.) denotes agreement on the modality with
the higher score. Agreement metrics and $|I|\geq1$ are reported as
percentages. The medians of $I$ and $|I|$ are in logit units.
Brackets report 95\% confidence intervals from 1,000 clip-level
bootstrap resamples.}

\label{tab:supp_counterfactual_shapley}

\footnotesize
\setlength{\tabcolsep}{2.5pt}

\newcommand{\shapleyci}[4][]{%
  \shortstack[c]{%
    $#2$#1\\[1pt]
    {\scriptsize
     \setlength{\fboxsep}{1pt}%
     \colorbox{black!8}{$[#3,\,#4]$}}%
  }%
}
\newcommand{\shapleymaltbiaslabel}[1]{%
  \rotatebox[origin=c]{90}{%
    \scriptsize
    \shortstack[c]{\textit{#1 value-}\\\textit{projection bias}}%
  }%
}
\newcommand{\shapleymaltbiasmid}[1]{%
  \noalign{%
    \vskip 2.5pt
    \vbox to 0pt{\vss\hbox{\shapleymaltbiaslabel{#1}}\vss}%
    \vskip 2.5pt
  }%
}
\newsavebox{\shapleymaltrowref}
\sbox{\shapleymaltrowref}{\shapleyci{0.00}{-0.00}{0.00}}
\newcommand{\shapleymaltbiascell}[1]{%
  \raisebox{\dimexpr
    (\ht\shapleymaltrowref-\dp\shapleymaltrowref-\height+\depth)/2\relax
  }{\shapleymaltbiaslabel{#1}}%
}

\begin{tabular}{@{}c l c ccc ccc c ccc@{}}
\toprule
 & & &
 \multicolumn{3}{c}{\textbf{Video} ($C_{\mathrm V}$ vs.\ $\phi_{\mathrm V}$)}
 & \multicolumn{3}{c}{\textbf{Audio} ($C_{\mathrm A}$ vs.\ $\phi_{\mathrm A}$)}
 & & \multicolumn{3}{c}{\textbf{Interaction} ($I$)} \\
\cmidrule(lr){4-6}
\cmidrule(lr){7-9}
\cmidrule(l){11-13}
 & Model & $n$
 & Sign Agr. & $C_{\mathrm V}\!\geq\!1$ & $r$
 & Sign Agr. & $C_{\mathrm A}\!\geq\!1$ & $r$
 & \shortstack{Dominant\\Agr.}
 & Median $I$ & Median $|I|$ & $|I|\!\geq\!1$ \\
\midrule

 & Omni 3B & 543
 & \shapleyci{97}{95}{98}
 & \shapleyci[\, (521)]{97}{96}{98}
 & \shapleyci{0.65}{0.60}{0.69}
 & \shapleyci{95}{93}{97}
 & \shapleyci[\, (151)]{97}{94}{99}
 & \shapleyci{0.48}{0.36}{0.57}
 & \shapleyci{91}{88}{94}
 & \shapleyci{-3.43}{-3.76}{-3.06}
 & \shapleyci{3.46}{3.19}{3.80}
 & \shapleyci{86}{83}{89} \\
\shapleymaltbiasmid{With}
 & VL-AV & 518
 & \shapleyci{96}{94}{98}
 & \shapleyci[\, (472)]{99}{97}{100}
 & \shapleyci{0.46}{0.39}{0.53}
 & \shapleyci{97}{96}{99}
 & \shapleyci[\, (350)]{98}{97}{99}
 & \shapleyci{0.62}{0.54}{0.69}
 & \shapleyci{82}{79}{86}
 & \shapleyci{-3.46}{-3.80}{-3.05}
 & \shapleyci{3.67}{3.29}{4.08}
 & \shapleyci{83}{80}{86} \\
\midrule

\shapleymaltbiascell{Without}
 & MCP 4.5 & 547
 & \shapleyci{48}{44}{52}
 & \shapleyci[\, (545)]{48}{44}{52}
 & \shapleyci{0.25}{0.18}{0.31}
 & \shapleyci{31}{26}{35}
 & \shapleyci[\, (422)]{29}{24}{33}
 & \shapleyci{0.08}{-0.02}{0.17}
 & \shapleyci{57}{53}{62}
 & \shapleyci{8.73}{6.28}{13.85}
 & \shapleyci{19.32}{16.96}{22.53}
 & \shapleyci{96}{95}{98} \\
\bottomrule
\end{tabular}
\end{table*}

MALT remains strongly consistent with the order-averaged reference for
Qwen2.5-Omni 3B and VideoLLaMA2.1-7B-AV, with video/audio Sign Agreement of
97/95\% and 96/97\%, respectively, and Dominant Modality Agreement of 91\%
and 82\%. MiniCPM-o~4.5 shows substantially lower agreement and stronger
non-additivity: its median $|I|$ is 19.32 logits, compared with 3.46 and 3.67
for the other two models, and 96\% of its matched words have $|I|\geq1$.
Moreover, the median target logit increases from 106.75 on the original input
to 129.38 when both modalities are obscured. These results indicate strong
redistribution under the joint perturbation for MiniCPM-o~4.5 and show that
averaging over removal orders does not eliminate the sensitivity of
perturbation-based references to the obscured inputs. Accordingly, we treat
the Shapley values as an order-symmetrized counterfactual reference rather
than as ground-truth modality contributions.

\subsection{Controlled audio-video swaps}
\label{subsec:supp_swap_res}

\Cref{tab:swap_ci} reports the modality contributions presented in
\Cref{tab:swap} together with their 95\% confidence intervals. Intervals are
obtained from 1,000 bootstrap samples drawn with replacement at the clip level, keeping all words from the same clip together.

\begin{table*}[t]
\centering
\caption{Controlled audio-video swaps with 95\% confidence intervals.
For each word group, $n$ is the number of matched generated words, and
$C_{\mathrm V}$ and $C_{\mathrm A}$ are the mean visual and audio
contributions in logit units. Under cross-modal swaps, \textit{Seen}
and \textit{Heard words} refer to the events carried by the video and audio
streams, respectively. \textit{Caption counts} distinguish captions that mention
only the seen event, only the heard event, both, or neither.
Brackets show 95\% percentile confidence intervals from 1,000 clip-level
bootstrap resamples that keep words from the same clip together.
$^\dagger$Ola-7B processed 92/100 conditions due to its context limit.
$^\ddagger$AV-Flamingo processed 96/100 conditions because four had too few frames.}
\label{tab:swap_ci}

\footnotesize
\setlength{\tabcolsep}{2.2pt}

\newcommand{\swapci}[3]{%
  \shortstack[c]{%
    $#1$\\[1pt]
    {\scriptsize
     \setlength{\fboxsep}{1pt}%
     \colorbox{black!8}{$[#2,\,#3]$}}%
  }%
}
\newcommand{\swapcihead}[2]{%
  \multicolumn{1}{%
    >{\columncolor{#1!45}%
      [\dimexpr\tabcolsep-1pt\relax]%
      [\dimexpr\tabcolsep-1pt\relax]}c%
  }{$C_{\mathrm{#2}}$}%
}
\newcommand{\swapcibiaslabel}[1]{%
  \rotatebox[origin=c]{90}{%
    \scriptsize
    \shortstack[c]{\textit{#1 value-}\\\textit{projection bias}}%
  }%
}
\newcommand{\swapcibiasmid}[1]{%
  \noalign{%
    \vskip\dimexpr 5pt/2\relax
    \vbox to 0pt{\vss\hbox{\swapcibiaslabel{#1}}\vss}%
    \vskip\dimexpr 5pt/2\relax
  }%
}
\newsavebox{\swapcirowref}
\sbox{\swapcirowref}{\swapci{0.00}{-0.00}{0.00}}
\newcommand{\swapcibiascell}[1]{%
  \raisebox{\dimexpr
    (\ht\swapcirowref-\dp\swapcirowref-\height+\depth)/2\relax
  }[0pt][0pt]{\swapcibiaslabel{#1}}%
}

\begin{tabular*}{\textwidth}{@{\extracolsep{\fill}}c l ccc ccc ccc cccc@{}}
\toprule
  & & \multicolumn{3}{c}{\textbf{Original A--V}}
  & \multicolumn{10}{c}{\textbf{Cross-modal swaps}} \\
\cmidrule(lr){3-5}
\cmidrule(l){6-15}
  & & \multicolumn{3}{c}{\textit{Matched words}}
  & \multicolumn{3}{c}{\textit{Seen words}}
  & \multicolumn{3}{c}{\textit{Heard words}}
  & \multicolumn{4}{c}{\textit{Caption counts}} \\
\cmidrule(lr){3-5}
\cmidrule(lr){6-8}
\cmidrule(lr){9-11}
\cmidrule(l){12-15}
\smash{\phantom{\swapcibiaslabel{Without}}}
  & Model
  & $n$ & \swapcihead{Vcol}{V} & \swapcihead{Acol}{A}
  & $n$ & \swapcihead{Vcol}{V} & \swapcihead{Acol}{A}
  & $n$ & \swapcihead{Vcol}{V} & \swapcihead{Acol}{A}
  & {Seen} & {Heard} & Both & Neither \\
\midrule

  & Omni 3B
  & 84 & \swapci{6.06}{5.34}{6.86}   & \swapci{1.98}{1.62}{2.41}
  & 63 & \swapci{8.32}{7.37}{9.33}   & \swapci{0.19}{0.10}{0.29}
  & 43 & \swapci{0.00}{-0.06}{0.06}  & \swapci{5.90}{5.42}{6.39}
  & 19 & 2 & 26 & 3 \\
\addlinespace[5pt]
  & Omni 7B
  & 94 & \swapci{5.12}{4.60}{5.70}   & \swapci{1.54}{1.30}{1.84}
  & 64 & \swapci{7.50}{6.78}{8.30}   & \swapci{0.14}{0.06}{0.23}
  & 35 & \swapci{0.09}{-0.04}{0.26}  & \swapci{5.04}{4.66}{5.45}
  & 21 & 3 & 23 & 3 \\
\addlinespace[5pt]
  & MCP 2.6
  & 64 & \swapci{12.52}{10.64}{14.25} & \swapci{3.18}{2.58}{3.80}
  & 49 & \swapci{16.72}{14.78}{18.68} & \swapci{1.10}{0.79}{1.48}
  &  3 & \swapci{1.10}{0.28}{2.72}   & \swapci{10.61}{9.88}{12.06}
  & 35 & 0 & 2 & 13 \\
\swapcibiasmid{With}
  & VL-AV
  & 72 & \swapci{5.51}{4.66}{6.42}     & \swapci{2.50}{1.97}{3.04}
  & 53 & \swapci{7.76}{7.07}{8.62}     & \swapci{0.22}{0.10}{0.35}
  & 26 & \swapci{-0.49}{-0.80}{-0.11} & \swapci{7.65}{6.91}{8.52}
  & 28 & 5 & 11 & 6 \\
\addlinespace[5pt]
  & Ola-7B$^\dagger$
  & 62 & \swapci{3.83}{3.22}{4.49}    & \swapci{1.95}{1.62}{2.31}
  & 32 & \swapci{4.91}{4.25}{5.58}    & \swapci{0.49}{0.35}{0.67}
  & 39 & \swapci{-0.04}{-0.15}{0.09} & \swapci{4.46}{3.99}{4.97}
  & 13 & 17 & 10 & 6 \\
\addlinespace[5pt]
  & AVF$^\ddagger$
  & 74 & \swapci{5.02}{4.35}{5.75}     & \swapci{1.62}{1.19}{2.05}
  & 54 & \swapci{6.97}{6.41}{7.61}     & \swapci{0.18}{0.14}{0.23}
  & 42 & \swapci{-0.17}{-0.28}{-0.07} & \swapci{5.94}{5.57}{6.34}
  & 14 & 3 & 27 & 4 \\
\midrule

  & MCP 4.5
  & 78 & \swapci{24.94}{22.59}{27.46} & \swapci{5.11}{4.12}{6.20}
  & 60 & \swapci{28.75}{25.91}{32.19} & \swapci{2.19}{1.50}{2.91}
  & 20 & \swapci{10.40}{6.74}{14.77}  & \swapci{16.94}{13.99}{20.67}
  & 31 & 2 & 12 & 5 \\
\addlinespace[5pt]
\swapcibiascell{Without}
  & vSAL 4B
  & 72 & \swapci{9.58}{8.86}{10.32}   & \swapci{0.47}{0.41}{0.54}
  & 57 & \swapci{10.55}{9.88}{11.26}  & \swapci{0.37}{0.33}{0.42}
  &  2 & \swapci{3.35}{3.08}{3.63}    & \swapci{0.83}{0.69}{0.96}
  & 39 & 0 & 2 & 9 \\
\addlinespace[5pt]
  & vSAL 8B
  & 74 & \swapci{9.14}{8.40}{9.95}   & \swapci{0.23}{0.19}{0.28}
  & 62 & \swapci{9.88}{9.13}{10.64}  & \swapci{0.13}{0.10}{0.17}
  &  3 & \swapci{3.93}{3.05}{4.57}   & \swapci{1.74}{1.70}{1.82}
  & 39 & 0 & 3 & 8 \\
\bottomrule
\end{tabular*}
\end{table*}

\subsection{Comparison with attribution methods}
\label{subsec:supp_attr_met}

We compare attribution methods along two distinct dimensions:
behavioral consistency, assessing agreement with counterfactual
interventions and whether modality scores favor the evidence
associated with mentioned events under controlled swaps;
and logit accounting, assessing how much of the observed
target-token logit is represented by the reported scores.
These properties are distinct: agreement with interventions
does not imply complete logit accounting.

For a controlled comparison, all methods are evaluated on
the same teacher-forced sequences, matched target words,
and audio and video decoder-input positions.
For each baseline, modality scores are obtained by
aggregating the scores assigned to the corresponding positions.\\
\textbf{Attention-based scores.}
Attention mass sums the attention weights from the prediction position
to the positions of a modality. We evaluate two variants, one that
averages the resulting mass across heads and decoder layers and one
that averages across heads in the final layer only.
Attention$\times\|v\|$ multiplies each attention weight by the norm
of the corresponding value vector before summing over modality positions
and averaging over all heads and decoder layers~\cite{kobayashi2020attention}.\\
\textbf{Gradient$\times$Input.}
For a target token $y_t$, we compute the gradient of its logit
$\ell(y_t)$ with respect to the decoder-input vector $x_p^0$ at
position $p$. The position score is
$\langle \partial\ell(y_t)/\partial x_p^0, x_p^0\rangle$.
We sum these signed scores over the positions of each modality.\\
\textbf{AttnLRP.}
We implement AttnLRP on the evaluated decoders using the modified
Gradient$\times$Input formulation of~\cite{arras2025closelook}.
The modality score is the sum of the relevance assigned to its
positions at the decoder input. Under its default pathway weights,
DPA uses the same propagation rules as AttnLRP, including the softmax
linearization. The reported AttnLRP scores therefore represent this
default DPA configuration.\\
\textbf{ALTI-Logit.}
ALTI-Logit~\cite{ferrando2023explaining} uses layerwise mixing
coefficients to trace contributions to decoder-input positions.
We sum the resulting position scores within each modality.
MLP updates are retained as separate terms and excluded from
both the modality sums and the numerator of the summed-score
ratio.\\
\textbf{Results.} \Cref{tab:supp_baselines_consistency} compares the audio and video
scores on the original input with the target-logit changes induced by
obscuring the corresponding modality.
For Qwen2.5-Omni 3B and VideoLLaMA2.1-7B-AV, MALT, AttnLRP, and
ALTI-Logit reach 91--96\% video sign agreement, but their audio
results differ across models. On Qwen2.5-Omni 3B, MALT and AttnLRP
have similar audio Pearson correlations ($r=0.82$ and $0.80$),
despite sign agreement of 48\% and 58\%, respectively.
On VideoLLaMA2.1-7B-AV, all three methods reach
91\% audio Sign Agreement, whereas the attention-based
variants show weaker audio correlations on both models.
On MiniCPM-o 4.5, no signed method exceeds 51\%
audio Sign Agreement, and Dominant Agreement ranges
from 52\% to 59\% across methods.
Thus, consistency with single-modality obscuration
varies substantially across architectures.

\Cref{tab:supp_baselines_shapley} compares modality scores on the
same matched words with two-player Shapley values, which average
the marginal logit effects over both modality removal orders.
Compared with single-modality obscuration, video Pearson correlations for MALT, AttnLRP, and ALTI-Logit increase from $0.38$--$0.47$ to $0.59$--$0.65$ on Qwen2.5-Omni 3B and from $0.30$--$0.38$ to $0.46$--$0.53$ on VideoLLaMA2.1-7B-AV.
On Qwen2.5-Omni 3B, audio sign agreement also rises from
44--58\% to 95--96\%, but audio Pearson correlations fall from
$0.71$--$0.82$ to $0.48$--$0.52$. The scores thus agree more
often with the sign of the audio reference without tracking
its variation across words more closely. On VideoLLaMA2.1-7B-AV,
audio sign agreement reaches 97--98\% for all three methods.
In contrast, on MiniCPM-o 4.5, their sign agreement falls
from 59\% to 48\% for video and from 44--46\% to
30--32\% for audio. Gradient$\times$Input shows a different pattern. Its audio sign agreement reaches 75\% on Qwen2.5-Omni 3B but is only 26\% on
VideoLLaMA2.1-7B-AV. Even on Qwen2.5-Omni 3B, the relatively high
audio sign agreement is accompanied by a weak correlation
($r=0.14$) and 51\% dominant agreement. Dominant agreement is
unchanged for all methods because the two-player Shapley values
preserve the ordering of the single-modality effects,
$\phi_{\mathrm V}-\phi_{\mathrm A}
= \Delta\ell_{\mathrm V}-\Delta\ell_{\mathrm A}$.\\

\Cref{tab:supp_baselines_swap} compares MALT, AttnLRP, and ALTI-Logit on the same matched words to assess whether the higher score is assigned to the modality carrying evidence for the mentioned event.
Under swaps, all three methods assign higher video scores to
98--100\% of seen words across the three models. For heard words,
all methods favor audio in 100\% of cases on Qwen2.5-Omni~3B.
On VideoLLaMA2.1-7B-AV, MALT and AttnLRP favor audio in 96\% of
cases, compared with 81\% for ALTI-Logit. Differences also emerge
on MiniCPM-o~4.5, where AttnLRP favors audio for all 20 heard
words, compared with 90\% for MALT and 80\% for ALTI-Logit.
Despite these differences, all three methods assign higher mean
scores to audio than to video for heard words. ALTI-Logit retains
relatively substantial video scores on VideoLLaMA2.1-7B-AV
(1.85 vs.\ 2.89 for audio) and MiniCPM-o~4.5
(2.63 vs.\ 3.46), consistent with its weaker word-level audio
preference. MALT also retains substantial visual contributions
on MiniCPM-o~4.5 (10.40 vs.\ 16.94 logit units for audio).
Overall, the three methods generally follow the modality carrying
the mentioned event, although ALTI-Logit shows less consistent
audio preference under conflicting evidence.

Beyond behavioral agreement, we assess how much
of the target logit is accounted for by the
reported attribution scores
(\cref{tab:supp_baselines_consistency}).
On VideoLLaMA2.1-7B-AV, ALTI-Logit has a higher
median summed-score ratio than MALT's non-affine
terms (0.35 versus 0.24), while MALT additionally
reports the affine contribution, whose median
ratio is 0.76.
On MiniCPM-o 4.5, which has no value-projection bias,
MALT's non-affine terms already account for the full
logit up to numerical precision, whereas the
reported AttnLRP and ALTI-Logit scores account for
median ratios of 0.33 and 0.12, respectively.
MALT propagates MLP readouts through the residual stream
and explicitly separates source, prefix, affine, and direct
decoder-input contributions. Together, these terms reconstruct
the logit and account for its intervention-induced changes
up to numerical precision.

\subsubsection{Equivalence of DPA and AttnLRP}
\label{sec:supp_dpa_equiv}

DPA~\cite{jantsch2026dpa} and AttnLRP~\cite{3692070.3692076} both propagate the
target unembedding backward to the input embeddings and score each position by
the dot product between its embedding and the propagated vector, replacing the
gradient with a modified backward pass at each nonlinearity.
With DPA's default path weights
($\mu_V=0.5$, $\mu_Q=\mu_K=0.25$, $\mu_{\mathrm{gate}}=\mu_{\mathrm{up}}=0.5$),
the modifications coincide with the AttnLRP rules (Tab.~\ref{tab:supp_dpa_equiv}).
All remaining operations are linear, where both reduce to the ordinary gradient,
and both start from the same readout direction (\cref{eq:logit}). We verified the equivalence numerically by running the released implementations
of both methods on the same inputs: input attributions agree up to $10^{-6}$, a
difference due only to the RMSNorm stability constant
(DPA uses a fixed $10^{-7}$, AttnLRP the model's own value).
The equivalence holds for these default weights; other DPA configurations,
such as placing all weight on the content paths, yield different attributions.
We therefore report AttnLRP as representative of both methods.

\begin{table}[t]
\centering
\caption{Correspondence between AttnLRP and DPA propagation rules under DPA's default path weights.}
\label{tab:supp_dpa_equiv}
\small
\begin{tabular}{@{}p{0.22\linewidth}p{0.34\linewidth}p{0.34\linewidth}@{}}
\toprule
Operation & AttnLRP & DPA (default weights) \\
\midrule
RMSNorm & denominator held fixed & denominator held fixed \\
SiLU in gate & identity rule: $\mathrm{SiLU}'(z)\to\sigma(z)$ & gate path scaled by $\sigma(z)$ \\
Gate $\times$ up & uniform rule ($\tfrac12$, $\tfrac12$) & $\mu_{\mathrm{gate}}=\mu_{\mathrm{up}}=\tfrac12$ \\
Attention & uniform on $A\cdot V$ and $QK^\top$; softmax by Gradient$\times$Input: $V\,\tfrac12$, $Q\,\tfrac14$, $K\,\tfrac14$ & $\mu_V=\tfrac12$ (fixed routing), $\mu_Q=\mu_K=\tfrac14$ (softmax Jacobian) \\
\bottomrule
\end{tabular}
\end{table}

\begin{table*}[t]
\centering
\caption{Attribution-intervention consistency of MALT and the
baseline methods under modality obscuration.
$n$ denotes the number of matched subject and predicate words.
Original-input modality scores are compared with the target-logit
change $\Delta\ell_m$ induced by obscuring modality $m$.
Sign agreement (Sign Agr.) is evaluated on words with
$|\mathrm{score}_m|,|\Delta\ell_m|\geq0.1$; the retained subset
can differ across methods. It is not reported for non-negative
attention scores.
$r$ and $\rho$ denote Pearson and Spearman correlations,
respectively, between the attribution score and $\Delta\ell_m$.
Dominant agreement (Dom. Agr.) measures how often the modality
with the larger attribution score also has the larger signed
$\Delta\ell_m$. Agreement metrics are reported as percentages.
Accounting columns report median per-target ratios to the
target logit. For MALT,
$\Sigma=C_V+C_A+C_Q+C_G+C_K^{\mathrm{prefix}}+E$,
with $C_K^{\mathrm{aff}}$ reported separately.
For each signed baseline, $\Sigma$ sums its reported decoder-input
position scores; ALTI-Logit's separate MLP terms are excluded.
Brackets report 95\% confidence intervals from 1,000 clip-level
bootstrap resamples.}
\label{tab:supp_baselines_consistency}

\footnotesize
\setlength{\tabcolsep}{3pt}

\newcommand{\baselineci}[3]{%
  \shortstack[c]{%
    $#1$\\[1pt]
    {\scriptsize
     \setlength{\fboxsep}{1pt}%
     \colorbox{black!8}{$[#2,\,#3]$}}%
  }%
}
\newcommand{\consistencybiaslabel}[1]{%
  \rotatebox[origin=c]{90}{\textit{#1 value-projection bias}}%
}
\newcommand{\consistencybiasmid}[1]{%
  \noalign{%
    \vskip\dimexpr 5pt/2\relax
    \vbox to 0pt{\vss\hbox{\consistencybiaslabel{#1}}\vss}%
    \vskip\dimexpr 5pt/2\relax
  }%
}
\newsavebox{\consistencyrowref}
\sbox{\consistencyrowref}{\baselineci{0.00}{-0.00}{0.00}}
\newcommand{\consistencybiascell}[1]{%
  \raisebox{\dimexpr
    (\ht\consistencyrowref-\dp\consistencyrowref-\height+\depth)/2\relax
  }[0pt][0pt]{\consistencybiaslabel{#1}}%
}

\resizebox{\linewidth}{!}{%
\begin{tabular}{@{}c l ccc ccc c cc@{}}
\toprule
 & & \multicolumn{3}{c}{\textbf{Video obscuration}}
 & \multicolumn{3}{c}{\textbf{Audio obscuration}}
 &
 & \multicolumn{2}{c}{\textbf{Accounting}} \\
\cmidrule(lr){3-5}
\cmidrule(lr){6-8}
\cmidrule(l){10-11}
\smash{\phantom{\consistencybiaslabel{Without}}}
 & Method
 & Sign Agr. & $r$ & $\rho$
 & Sign Agr. & $r$ & $\rho$
 & Dom. Agr. & $\Sigma/\ell$ & $C_K^{\mathrm{aff}}/\ell$ \\
\midrule

 & \multicolumn{10}{l@{}}{\textit{Qwen2.5-Omni 3B} ($n=543$ words)} \\
\addlinespace[2pt]
 & \textbf{MALT}
 & \baselineci{93}{91}{96}
 & \baselineci{0.46}{0.40}{0.51}
 & \baselineci{0.48}{0.42}{0.54}
 & \baselineci{48}{42}{54}
 & \baselineci{0.82}{0.75}{0.86}
 & \baselineci{0.50}{0.41}{0.58}
 & \baselineci{91}{88}{94}
 & \baselineci{0.63}{0.59}{0.65}
 & 0.37 \\
\addlinespace[5pt]
 & Attention mass (mean)
 & --
 & \baselineci{0.39}{0.33}{0.45}
 & \baselineci{0.40}{0.33}{0.47}
 & --
 & \baselineci{0.39}{0.30}{0.47}
 & \baselineci{0.25}{0.15}{0.35}
 & \baselineci{86}{82}{89}
 & -- & -- \\
\addlinespace[5pt]
 & Attention mass (last layer)
 & --
 & \baselineci{0.30}{0.23}{0.38}
 & \baselineci{0.35}{0.28}{0.42}
 & --
 & \baselineci{0.02}{-0.07}{0.12}
 & \baselineci{-0.01}{-0.12}{0.09}
 & \baselineci{85}{82}{89}
 & -- & -- \\
\addlinespace[5pt]
 & Attention$\times\|v\|$ (mean)
 & --
 & \baselineci{0.37}{0.31}{0.43}
 & \baselineci{0.38}{0.31}{0.44}
 & --
 & \baselineci{0.54}{0.45}{0.61}
 & \baselineci{0.31}{0.21}{0.41}
 & \baselineci{86}{82}{89}
 & -- & -- \\
\addlinespace[5pt]
 & Gradient$\times$Input
 & \baselineci{58}{53}{63}
 & \baselineci{0.08}{-0.02}{0.17}
 & \baselineci{0.08}{-0.00}{0.17}
 & \baselineci{47}{41}{53}
 & \baselineci{0.16}{-0.03}{0.34}
 & \baselineci{0.07}{-0.02}{0.17}
 & \baselineci{51}{47}{56}
 & \baselineci{0.02}{0.01}{0.02}
 & -- \\
\addlinespace[5pt]
 & AttnLRP
 & \baselineci{93}{91}{95}
 & \baselineci{0.47}{0.41}{0.52}
 & \baselineci{0.50}{0.44}{0.56}
 & \baselineci{58}{51}{65}
 & \baselineci{0.80}{0.74}{0.85}
 & \baselineci{0.51}{0.43}{0.58}
 & \baselineci{91}{88}{93}
 & \baselineci{0.19}{0.19}{0.20}
 & -- \\
\addlinespace[5pt]
 & ALTI-Logit
 & \baselineci{94}{91}{96}
 & \baselineci{0.38}{0.31}{0.44}
 & \baselineci{0.40}{0.33}{0.46}
 & \baselineci{44}{39}{50}
 & \baselineci{0.71}{0.62}{0.79}
 & \baselineci{0.41}{0.32}{0.50}
 & \baselineci{90}{88}{93}
 & \baselineci{0.54}{0.52}{0.55}
 & -- \\
\consistencybiasmid{With}
\addlinespace[5pt]
 & \multicolumn{10}{l@{}}{\textit{VideoLLaMA2.1-7B-AV} ($n=518$ words)} \\
\addlinespace[2pt]
 & \textbf{MALT}
 & \baselineci{94}{92}{97}
 & \baselineci{0.30}{0.20}{0.39}
 & \baselineci{0.27}{0.18}{0.35}
 & \baselineci{91}{88}{93}
 & \baselineci{0.78}{0.72}{0.83}
 & \baselineci{0.50}{0.42}{0.58}
 & \baselineci{82}{79}{86}
 & \baselineci{0.24}{0.21}{0.29}
 & 0.76 \\
\addlinespace[5pt]
 & Attention mass (mean)
 & --
 & \baselineci{0.13}{0.03}{0.22}
 & \baselineci{0.13}{0.03}{0.21}
 & --
 & \baselineci{0.32}{0.23}{0.40}
 & \baselineci{0.17}{0.07}{0.25}
 & \baselineci{79}{75}{82}
 & -- & -- \\
\addlinespace[5pt]
 & Attention mass (last layer)
 & --
 & \baselineci{0.15}{0.06}{0.23}
 & \baselineci{0.16}{0.07}{0.24}
 & --
 & \baselineci{0.45}{0.36}{0.54}
 & \baselineci{0.36}{0.26}{0.44}
 & \baselineci{78}{74}{82}
 & -- & -- \\
\addlinespace[5pt]
 & Attention$\times\|v\|$ (mean)
 & --
 & \baselineci{0.14}{0.04}{0.22}
 & \baselineci{0.14}{0.05}{0.22}
 & --
 & \baselineci{0.52}{0.45}{0.59}
 & \baselineci{0.32}{0.23}{0.40}
 & \baselineci{78}{75}{82}
 & -- & -- \\
\addlinespace[5pt]
 & Gradient$\times$Input
 & \baselineci{50}{46}{55}
 & \baselineci{0.15}{0.05}{0.26}
 & \baselineci{0.12}{0.03}{0.22}
 & \baselineci{33}{28}{38}
 & \baselineci{-0.00}{-0.13}{0.12}
 & \baselineci{0.05}{-0.04}{0.14}
 & \baselineci{59}{55}{63}
 & \baselineci{-0.02}{-0.03}{-0.02}
 & -- \\
\addlinespace[5pt]
 & AttnLRP
 & \baselineci{91}{88}{94}
 & \baselineci{0.38}{0.29}{0.47}
 & \baselineci{0.36}{0.28}{0.44}
 & \baselineci{91}{89}{94}
 & \baselineci{0.73}{0.67}{0.78}
 & \baselineci{0.51}{0.43}{0.59}
 & \baselineci{79}{75}{83}
 & \baselineci{0.18}{0.17}{0.18}
 & -- \\
\addlinespace[5pt]
 & ALTI-Logit
 & \baselineci{96}{94}{97}
 & \baselineci{0.33}{0.25}{0.41}
 & \baselineci{0.32}{0.25}{0.40}
 & \baselineci{91}{89}{94}
 & \baselineci{0.68}{0.60}{0.76}
 & \baselineci{0.45}{0.37}{0.53}
 & \baselineci{83}{80}{86}
 & \baselineci{0.35}{0.34}{0.36}
 & -- \\
\midrule

 & \multicolumn{10}{l@{}}{\textit{MiniCPM-o 4.5} ($n=547$ words)} \\
\addlinespace[2pt]
 & \textbf{MALT}
 & \baselineci{59}{55}{63}
 & \baselineci{0.21}{0.14}{0.27}
 & \baselineci{0.20}{0.13}{0.27}
 & \baselineci{44}{40}{49}
 & \baselineci{0.03}{-0.09}{0.14}
 & \baselineci{0.00}{-0.10}{0.08}
 & \baselineci{57}{53}{62}
 & \baselineci{1.00}{1.00}{1.00}
 & 0.00 \\
\addlinespace[5pt]
 & Attention mass (mean)
 & --
 & \baselineci{0.07}{0.01}{0.13}
 & \baselineci{0.05}{-0.02}{0.12}
 & --
 & \baselineci{-0.04}{-0.14}{0.05}
 & \baselineci{-0.11}{-0.20}{-0.03}
 & \baselineci{59}{55}{64}
 & -- & -- \\
\addlinespace[5pt]
 & Attention mass (last layer)
 & --
 & \baselineci{-0.09}{-0.17}{-0.01}
 & \baselineci{-0.07}{-0.15}{0.00}
 & --
 & \baselineci{-0.03}{-0.14}{0.06}
 & \baselineci{-0.06}{-0.15}{0.02}
 & \baselineci{59}{55}{64}
 & -- & -- \\
\addlinespace[5pt]
\consistencybiascell{Without}
 & Attention$\times\|v\|$ (mean)
 & --
 & \baselineci{-0.01}{-0.09}{0.06}
 & \baselineci{-0.03}{-0.11}{0.05}
 & --
 & \baselineci{0.01}{-0.11}{0.14}
 & \baselineci{-0.04}{-0.13}{0.04}
 & \baselineci{59}{55}{64}
 & -- & -- \\
\addlinespace[5pt]
 & Gradient$\times$Input
 & \baselineci{52}{48}{56}
 & \baselineci{0.02}{-0.02}{0.15}
 & \baselineci{0.09}{-0.00}{0.18}
 & \baselineci{51}{46}{55}
 & \baselineci{0.24}{0.11}{0.34}
 & \baselineci{0.09}{-0.00}{0.19}
 & \baselineci{52}{48}{57}
 & \baselineci{0.04}{0.03}{0.06}
 & -- \\
\addlinespace[5pt]
 & AttnLRP
 & \baselineci{59}{55}{63}
 & \baselineci{0.20}{0.13}{0.27}
 & \baselineci{0.17}{0.09}{0.24}
 & \baselineci{46}{41}{50}
 & \baselineci{0.14}{-0.01}{0.28}
 & \baselineci{0.06}{-0.03}{0.15}
 & \baselineci{58}{54}{62}
 & \baselineci{0.33}{0.33}{0.34}
 & -- \\
\addlinespace[5pt]
 & ALTI-Logit
 & \baselineci{59}{55}{63}
 & \baselineci{-0.01}{-0.08}{0.07}
 & \baselineci{-0.03}{-0.11}{0.05}
 & \baselineci{44}{39}{48}
 & \baselineci{0.12}{-0.07}{0.28}
 & \baselineci{0.00}{-0.08}{0.09}
 & \baselineci{59}{56}{64}
 & \baselineci{0.12}{0.11}{0.12}
 & -- \\
\bottomrule
\end{tabular}%
}
\end{table*}

\begin{table*}[t]
\centering
\caption{Agreement with the two-player Shapley reference for MALT and
the attribution baselines.
For the same $n$ matched subject and predicate words, modality scores
assigned on the original input are compared with the corresponding
Shapley values $\phi_{\mathrm V}$ and $\phi_{\mathrm A}$.
Sign agreement (Sign Agr.) measures how often the attribution score
and the corresponding Shapley value have the same sign.
Dashes indicate that this metric is not reported for non-negative
attention scores.
$r$ and $\rho$ denote Pearson and Spearman correlations, respectively,
between the paired scores and Shapley values.
Dominant agreement (Dom. Agr.) measures how often the modality with
the higher attribution score also has the larger Shapley value.
Sign and dominant-modality agreement are reported as percentages.
Brackets report 95\% confidence intervals from 1,000 clip-level
bootstrap resamples.}
\label{tab:supp_baselines_shapley}

\footnotesize
\renewcommand{\arraystretch}{0.98}
\setlength{\tabcolsep}{3pt}

\newcommand{\shapleybaselineci}[3]{%
  \shortstack[c]{%
    $#1$\\[1pt]
    {\scriptsize
     \setlength{\fboxsep}{1pt}%
     \colorbox{black!8}{$[#2,\,#3]$}}%
  }%
}
\newcommand{\shapleybiaslabel}[1]{%
  \rotatebox[origin=c]{90}{\textit{#1 value-projection bias}}%
}
\newcommand{\shapleybiasmid}[1]{%
  \noalign{%
    \vskip\dimexpr 5pt/2\relax
    \vbox to 0pt{\vss\hbox{\shapleybiaslabel{#1}}\vss}%
    \vskip\dimexpr 5pt/2\relax
  }%
}
\newsavebox{\shapleyrowref}
\sbox{\shapleyrowref}{\shapleybaselineci{0.00}{-0.00}{0.00}}
\newcommand{\shapleybiascell}[1]{%
  \raisebox{\dimexpr
    (\ht\shapleyrowref-\dp\shapleyrowref-\height+\depth)/2\relax
  }[0pt][0pt]{\shapleybiaslabel{#1}}%
}

\begin{tabular}{@{}c l ccc ccc c@{}}
\toprule
 & & \multicolumn{3}{c}{\textbf{Video} (score vs.\ $\phi_{\mathrm V}$)}
   & \multicolumn{3}{c}{\textbf{Audio} (score vs.\ $\phi_{\mathrm A}$)}
   & \\
\cmidrule(lr){3-5}
\cmidrule(lr){6-8}
\smash{\phantom{\shapleybiaslabel{Without}}}
 & Method
 & Sign Agr. & $r$ & $\rho$
 & Sign Agr. & $r$ & $\rho$
 & Dom. Agr. \\
\midrule

 & \multicolumn{8}{l@{}}{\textit{Qwen2.5-Omni 3B} ($n=543$ words)} \\
\addlinespace[2pt]
 & \textbf{MALT}
 & \shapleybaselineci{97}{95}{98}
 & \shapleybaselineci{0.65}{0.60}{0.69}
 & \shapleybaselineci{0.65}{0.59}{0.70}
 & \shapleybaselineci{95}{93}{97}
 & \shapleybaselineci{0.48}{0.36}{0.57}
 & \shapleybaselineci{0.45}{0.37}{0.53}
 & \shapleybaselineci{91}{88}{94} \\
\addlinespace[5pt]
 & Attention mass (mean)
 & --
 & \shapleybaselineci{0.57}{0.51}{0.62}
 & \shapleybaselineci{0.54}{0.48}{0.60}
 & --
 & \shapleybaselineci{0.31}{0.21}{0.39}
 & \shapleybaselineci{0.29}{0.19}{0.37}
 & \shapleybaselineci{86}{82}{89} \\
\addlinespace[5pt]
 & Attention mass (last layer)
 & --
 & \shapleybaselineci{0.46}{0.40}{0.52}
 & \shapleybaselineci{0.51}{0.44}{0.57}
 & --
 & \shapleybaselineci{0.24}{0.15}{0.33}
 & \shapleybaselineci{0.29}{0.21}{0.38}
 & \shapleybaselineci{85}{82}{89} \\
\addlinespace[5pt]
 & Attention$\times\|v\|$ (mean)
 & --
 & \shapleybaselineci{0.54}{0.48}{0.60}
 & \shapleybaselineci{0.52}{0.45}{0.58}
 & --
 & \shapleybaselineci{0.33}{0.22}{0.42}
 & \shapleybaselineci{0.29}{0.20}{0.37}
 & \shapleybaselineci{86}{82}{89} \\
\addlinespace[5pt]
 & Gradient$\times$Input
 & \shapleybaselineci{57}{52}{62}
 & \shapleybaselineci{0.05}{-0.05}{0.14}
 & \shapleybaselineci{0.07}{-0.03}{0.15}
 & \shapleybaselineci{75}{70}{80}
 & \shapleybaselineci{0.14}{0.03}{0.26}
 & \shapleybaselineci{0.14}{0.05}{0.22}
 & \shapleybaselineci{51}{47}{56} \\
\addlinespace[5pt]
 & AttnLRP
 & \shapleybaselineci{97}{95}{98}
 & \shapleybaselineci{0.64}{0.59}{0.69}
 & \shapleybaselineci{0.64}{0.59}{0.69}
 & \shapleybaselineci{96}{94}{98}
 & \shapleybaselineci{0.52}{0.42}{0.60}
 & \shapleybaselineci{0.48}{0.40}{0.55}
 & \shapleybaselineci{91}{88}{93} \\
\addlinespace[5pt]
 & ALTI-Logit
 & \shapleybaselineci{96}{95}{98}
 & \shapleybaselineci{0.59}{0.53}{0.64}
 & \shapleybaselineci{0.57}{0.50}{0.62}
 & \shapleybaselineci{95}{93}{97}
 & \shapleybaselineci{0.48}{0.40}{0.55}
 & \shapleybaselineci{0.43}{0.35}{0.50}
 & \shapleybaselineci{90}{88}{93} \\
\shapleybiasmid{With}

 & \multicolumn{8}{l@{}}{\textit{VideoLLaMA2.1-7B-AV} ($n=518$ words)} \\
\addlinespace[2pt]
 & \textbf{MALT}
 & \shapleybaselineci{96}{94}{98}
 & \shapleybaselineci{0.46}{0.39}{0.53}
 & \shapleybaselineci{0.43}{0.36}{0.50}
 & \shapleybaselineci{97}{96}{99}
 & \shapleybaselineci{0.62}{0.54}{0.69}
 & \shapleybaselineci{0.47}{0.40}{0.54}
 & \shapleybaselineci{82}{79}{86} \\
\addlinespace[5pt]
 & Attention mass (mean)
 & --
 & \shapleybaselineci{0.41}{0.32}{0.50}
 & \shapleybaselineci{0.43}{0.34}{0.51}
 & --
 & \shapleybaselineci{0.39}{0.32}{0.46}
 & \shapleybaselineci{0.36}{0.29}{0.43}
 & \shapleybaselineci{79}{75}{82} \\
\addlinespace[5pt]
 & Attention mass (last layer)
 & --
 & \shapleybaselineci{0.42}{0.34}{0.50}
 & \shapleybaselineci{0.44}{0.36}{0.51}
 & --
 & \shapleybaselineci{0.42}{0.32}{0.50}
 & \shapleybaselineci{0.28}{0.19}{0.36}
 & \shapleybaselineci{78}{74}{82} \\
\addlinespace[5pt]
 & Attention$\times\|v\|$ (mean)
 & --
 & \shapleybaselineci{0.44}{0.35}{0.51}
 & \shapleybaselineci{0.45}{0.37}{0.52}
 & --
 & \shapleybaselineci{0.47}{0.39}{0.54}
 & \shapleybaselineci{0.36}{0.29}{0.44}
 & \shapleybaselineci{78}{75}{82} \\
\addlinespace[5pt]
 & Gradient$\times$Input
 & \shapleybaselineci{50}{45}{55}
 & \shapleybaselineci{0.07}{-0.03}{0.17}
 & \shapleybaselineci{0.06}{-0.03}{0.15}
 & \shapleybaselineci{26}{21}{31}
 & \shapleybaselineci{0.02}{-0.08}{0.12}
 & \shapleybaselineci{0.02}{-0.06}{0.11}
 & \shapleybaselineci{59}{55}{63} \\
\addlinespace[5pt]
 & AttnLRP
 & \shapleybaselineci{94}{92}{97}
 & \shapleybaselineci{0.51}{0.44}{0.57}
 & \shapleybaselineci{0.48}{0.42}{0.55}
 & \shapleybaselineci{97}{96}{98}
 & \shapleybaselineci{0.58}{0.51}{0.65}
 & \shapleybaselineci{0.45}{0.37}{0.52}
 & \shapleybaselineci{79}{75}{83} \\
\addlinespace[5pt]
 & ALTI-Logit
 & \shapleybaselineci{97}{95}{98}
 & \shapleybaselineci{0.53}{0.47}{0.59}
 & \shapleybaselineci{0.52}{0.45}{0.58}
 & \shapleybaselineci{98}{96}{99}
 & \shapleybaselineci{0.52}{0.44}{0.60}
 & \shapleybaselineci{0.37}{0.29}{0.44}
 & \shapleybaselineci{83}{80}{86} \\
\midrule

 & \multicolumn{8}{l@{}}{\textit{MiniCPM-o 4.5} ($n=547$ words)} \\
\addlinespace[2pt]
 & \textbf{MALT}
 & \shapleybaselineci{48}{44}{52}
 & \shapleybaselineci{0.25}{0.18}{0.31}
 & \shapleybaselineci{0.24}{0.18}{0.31}
 & \shapleybaselineci{31}{26}{35}
 & \shapleybaselineci{0.08}{-0.02}{0.17}
 & \shapleybaselineci{0.05}{-0.05}{0.14}
 & \shapleybaselineci{57}{53}{62} \\
\addlinespace[5pt]
 & Attention mass (mean)
 & --
 & \shapleybaselineci{0.05}{-0.02}{0.12}
 & \shapleybaselineci{0.03}{-0.04}{0.11}
 & --
 & \shapleybaselineci{-0.03}{-0.12}{0.04}
 & \shapleybaselineci{-0.12}{-0.22}{-0.04}
 & \shapleybaselineci{59}{55}{64} \\
\addlinespace[5pt]
 & Attention mass (last layer)
 & --
 & \shapleybaselineci{-0.11}{-0.19}{-0.02}
 & \shapleybaselineci{-0.09}{-0.16}{-0.00}
 & --
 & \shapleybaselineci{-0.03}{-0.12}{0.05}
 & \shapleybaselineci{-0.07}{-0.15}{0.01}
 & \shapleybaselineci{59}{55}{64} \\
\addlinespace[5pt]
\shapleybiascell{Without}
 & Attention$\times\|v\|$ (mean)
 & --
 & \shapleybaselineci{-0.04}{-0.12}{0.04}
 & \shapleybaselineci{-0.06}{-0.14}{0.02}
 & --
 & \shapleybaselineci{0.02}{-0.07}{0.10}
 & \shapleybaselineci{-0.06}{-0.15}{0.03}
 & \shapleybaselineci{59}{55}{64} \\
\addlinespace[5pt]
 & Gradient$\times$Input
 & \shapleybaselineci{52}{48}{56}
 & \shapleybaselineci{0.03}{-0.02}{0.12}
 & \shapleybaselineci{0.05}{-0.04}{0.15}
 & \shapleybaselineci{54}{50}{58}
 & \shapleybaselineci{0.16}{0.06}{0.25}
 & \shapleybaselineci{0.12}{0.03}{0.20}
 & \shapleybaselineci{52}{48}{57} \\
\addlinespace[5pt]
 & AttnLRP
 & \shapleybaselineci{48}{44}{52}
 & \shapleybaselineci{0.26}{0.19}{0.32}
 & \shapleybaselineci{0.21}{0.14}{0.29}
 & \shapleybaselineci{32}{28}{36}
 & \shapleybaselineci{0.13}{0.03}{0.23}
 & \shapleybaselineci{0.07}{-0.03}{0.16}
 & \shapleybaselineci{58}{54}{62} \\
\addlinespace[5pt]
 & ALTI-Logit
 & \shapleybaselineci{48}{44}{52}
 & \shapleybaselineci{0.06}{-0.02}{0.14}
 & \shapleybaselineci{0.03}{-0.05}{0.11}
 & \shapleybaselineci{30}{26}{34}
 & \shapleybaselineci{0.05}{-0.06}{0.13}
 & \shapleybaselineci{-0.05}{-0.14}{0.03}
 & \shapleybaselineci{59}{56}{64} \\
\bottomrule
\end{tabular}
\end{table*}

\begin{table*}[t]
\centering
\caption{MALT, AttnLRP, and ALTI-Logit under controlled audio-video swaps.
Models are grouped by the presence of value-projection biases.
\textit{Seen} and \textit{Heard words} refer to events carried
by the video and audio streams, respectively. For each word group, $n$
denotes the number of matched words, and $C_{\mathrm V}$ and
$C_{\mathrm A}$ denote the mean scores for video and audio positions.
MALT reports contributions in logit units, while AttnLRP and ALTI-Logit
report decoder-input attribution scores.
The $V>A$ and $A>V$ columns give the percentages of words with higher
video and audio scores, respectively. The corresponding margins
$C_V-C_A$ and $C_A-C_V$ are the mean signed differences in the
indicated direction. Brackets report 95\% confidence intervals from 1,000 clip-level bootstrap resamples.}
\label{tab:supp_baselines_swap}

\footnotesize
\setlength{\tabcolsep}{2.5pt}

\newcommand{\swapbaselineci}[3]{%
  \shortstack[c]{%
    $#1$\\[1pt]
    {\scriptsize
     \setlength{\fboxsep}{1pt}%
     \colorbox{black!8}{$[#2,\,#3]$}}%
  }%
}
\newcommand{\swapbaselinehead}[2]{%
  \multicolumn{1}{%
    >{\columncolor{#1!45}%
      [\dimexpr\tabcolsep-1pt\relax]%
      [\dimexpr\tabcolsep-1pt\relax]}c%
  }{$C_{\mathrm{#2}}$}%
}
\newcommand{\swapbiaslabel}[1]{%
  \rotatebox[origin=c]{90}{%
    \scriptsize
    \shortstack[c]{\textit{#1 value-}\\\textit{projection bias}}%
  }%
}
\newcommand{\swapbiasmid}[1]{%
  \noalign{%
    \vskip\dimexpr 5pt/2\relax
    \vbox to 0pt{%
      \vss
      \hbox{\swapbiaslabel{#1}}%
      \vss
    }%
    \vskip\dimexpr 5pt/2\relax
  }%
}

\resizebox{\linewidth}{!}{%
\begin{tabular}{@{}c l ccc ccccc ccccc@{}}
\toprule
 & & \multicolumn{3}{c}{\textbf{Original A-V}}
 & \multicolumn{10}{c}{\textbf{Cross-modal swaps}} \\
\cmidrule(lr){3-5}
\cmidrule(l){6-15}
 & & \multicolumn{3}{c}{\textit{Matched words}}
 & \multicolumn{5}{c}{\textit{Seen words}}
 & \multicolumn{5}{c}{\textit{Heard words}} \\
\cmidrule(lr){3-5}
\cmidrule(lr){6-10}
\cmidrule(l){11-15}
\smash{\phantom{\swapbiaslabel{Without bias}}}
 & Method
 & $n$ & \swapbaselinehead{Vcol}{V} & \swapbaselinehead{Acol}{A}
 & $n$ & \swapbaselinehead{Vcol}{V} & \swapbaselinehead{Acol}{A}
 & $V>A$ & $C_{\mathrm V}-C_{\mathrm A}$
 & $n$ & \swapbaselinehead{Vcol}{V} & \swapbaselinehead{Acol}{A}
 & $A>V$ & $C_{\mathrm A}-C_{\mathrm V}$ \\
\midrule

 & \multicolumn{14}{l@{}}{\textit{Qwen2.5-Omni 3B}} \\
\addlinespace[2pt]
 & MALT
 & 84 & \swapbaselineci{6.06}{5.34}{6.85}
      & \swapbaselineci{1.98}{1.62}{2.41}
 & 63 & \swapbaselineci{8.32}{7.39}{9.37}
      & \swapbaselineci{0.19}{0.10}{0.30}
      & \swapbaselineci{100}{100}{100}
      & \swapbaselineci{8.13}{7.19}{9.20}
 & 43 & \swapbaselineci{-0.00}{-0.06}{0.06}
      & \swapbaselineci{5.90}{5.45}{6.45}
      & \swapbaselineci{100}{100}{100}
      & \swapbaselineci{5.91}{5.45}{6.44} \\
\addlinespace[5pt]
 & AttnLRP
 & 84 & \swapbaselineci{2.16}{1.86}{2.50}
      & \swapbaselineci{0.63}{0.52}{0.76}
 & 63 & \swapbaselineci{3.10}{2.72}{3.54}
      & \swapbaselineci{0.02}{-0.02}{0.05}
      & \swapbaselineci{100}{100}{100}
      & \swapbaselineci{3.09}{2.70}{3.51}
 & 43 & \swapbaselineci{0.11}{0.07}{0.16}
      & \swapbaselineci{1.63}{1.48}{1.81}
      & \swapbaselineci{100}{100}{100}
      & \swapbaselineci{1.52}{1.37}{1.68} \\
\addlinespace[5pt]
 & ALTI-Logit
 & 84 & \swapbaselineci{6.11}{5.40}{6.84}
      & \swapbaselineci{1.23}{1.01}{1.48}
 & 63 & \swapbaselineci{7.77}{7.13}{8.49}
      & \swapbaselineci{0.20}{0.11}{0.26}
      & \swapbaselineci{100}{100}{100}
      & \swapbaselineci{7.58}{6.91}{8.30}
 & 43 & \swapbaselineci{0.85}{0.72}{0.99}
      & \swapbaselineci{3.33}{2.90}{3.80}
      & \swapbaselineci{100}{100}{100}
      & \swapbaselineci{2.48}{2.12}{2.86} \\

\swapbiasmid{With}

\addlinespace[5pt]
 & \multicolumn{14}{l@{}}{\textit{VideoLLaMA2.1-7B-AV}} \\
\addlinespace[2pt]
 & MALT
 & 72 & \swapbaselineci{5.51}{4.67}{6.39}
      & \swapbaselineci{2.50}{1.96}{3.05}
 & 53 & \swapbaselineci{7.76}{7.06}{8.55}
      & \swapbaselineci{0.22}{0.12}{0.34}
      & \swapbaselineci{100}{100}{100}
      & \swapbaselineci{7.53}{6.81}{8.34}
 & 26 & \swapbaselineci{-0.49}{-0.81}{-0.14}
      & \swapbaselineci{7.65}{6.90}{8.55}
      & \swapbaselineci{96}{88}{100}
      & \swapbaselineci{8.13}{7.24}{8.99} \\
\addlinespace[5pt]
 & AttnLRP
 & 72 & \swapbaselineci{1.90}{1.53}{2.28}
      & \swapbaselineci{1.00}{0.80}{1.21}
 & 53 & \swapbaselineci{3.05}{2.66}{3.44}
      & \swapbaselineci{-0.02}{-0.08}{0.04}
      & \swapbaselineci{98}{95}{100}
      & \swapbaselineci{3.08}{2.67}{3.49}
 & 26 & \swapbaselineci{-0.29}{-0.61}{-0.01}
      & \swapbaselineci{3.10}{2.81}{3.49}
      & \swapbaselineci{96}{88}{100}
      & \swapbaselineci{3.39}{2.95}{3.92} \\
\addlinespace[5pt]
 & ALTI-Logit
 & 72 & \swapbaselineci{6.19}{5.56}{6.90}
      & \swapbaselineci{0.82}{0.57}{1.11}
 & 53 & \swapbaselineci{7.46}{6.82}{8.15}
      & \swapbaselineci{0.10}{0.04}{0.18}
      & \swapbaselineci{98}{94}{100}
      & \swapbaselineci{7.36}{6.73}{8.03}
 & 26 & \swapbaselineci{1.85}{1.46}{2.25}
      & \swapbaselineci{2.89}{2.21}{3.78}
      & \swapbaselineci{81}{68}{93}
      & \swapbaselineci{1.04}{0.41}{1.79} \\

\midrule

 & \multicolumn{14}{l@{}}{\textit{MiniCPM-o 4.5}} \\
\addlinespace[2pt]
 & MALT
 & 78 & \swapbaselineci{24.94}{22.49}{27.51}
      & \swapbaselineci{5.11}{4.20}{6.15}
 & 60 & \swapbaselineci{28.75}{25.48}{32.06}
      & \swapbaselineci{2.19}{1.46}{3.02}
      & \swapbaselineci{100}{100}{100}
      & \swapbaselineci{26.56}{23.37}{29.98}
 & 20 & \swapbaselineci{10.40}{6.67}{14.72}
      & \swapbaselineci{16.94}{13.89}{20.67}
      & \swapbaselineci{90}{76}{100}
      & \swapbaselineci{6.54}{2.20}{11.25} \\

\swapbiasmid{Without}

 & AttnLRP
 & 78 & \swapbaselineci{13.50}{11.99}{14.99}
      & \swapbaselineci{2.41}{1.97}{2.93}
 & 60 & \swapbaselineci{16.07}{14.49}{17.64}
      & \swapbaselineci{0.94}{0.63}{1.27}
      & \swapbaselineci{100}{100}{100}
      & \swapbaselineci{15.13}{13.42}{16.83}
 & 20 & \swapbaselineci{1.94}{0.95}{3.13}
      & \swapbaselineci{9.47}{8.08}{10.77}
      & \swapbaselineci{100}{100}{100}
      & \swapbaselineci{7.54}{5.86}{8.97} \\
\addlinespace[5pt]
 & ALTI-Logit
 & 78 & \swapbaselineci{8.34}{7.65}{8.99}
      & \swapbaselineci{1.53}{1.29}{1.80}
 & 60 & \swapbaselineci{9.55}{9.00}{10.02}
      & \swapbaselineci{0.79}{0.65}{0.94}
      & \swapbaselineci{100}{100}{100}
      & \swapbaselineci{8.76}{8.19}{9.30}
 & 20 & \swapbaselineci{2.63}{2.23}{3.04}
      & \swapbaselineci{3.46}{2.95}{4.07}
      & \swapbaselineci{80}{58}{100}
      & \swapbaselineci{0.83}{0.25}{1.40} \\

\bottomrule
\end{tabular}%
}
\end{table*}

\subsection{Sensitivity to the MLP Rule}
\label{subsec:supp_lambda}

\Cref{tab:supp_lambda} complements the sensitivity analysis presented in \Cref{sec:sensitivity} by reporting the full results for the MLP propagation parameter $\lambda$. Changing $\lambda$ can redistribute Component Share, particularly between generated context $G$ and the non-source component $K$. For example, for Omni 3B predicates, $G$ increases from $5\%$ to $28\%$ while $K$ decreases from $59\%$ to $23\%$ between $\lambda=0$ and $\lambda=1$; for MCP 2.6 subjects, $G$ changes from $-16\%$ to $-37\%$ while $K$ increases from $79\%$ to $107\%$. Negative or above-$100\%$ values can arise because Component Share is defined as a signed ratio of summed contributions to summed target logits. These variations are expected because $\lambda$ directly controls how the multiplicative MLP interaction is allocated between its two branches.

\begin{table*}[t]
\centering
\caption{Sensitivity to the MLP propagation rule.
Results are reported separately for subjects and predicates.
Models are grouped by the presence of value-projection biases
(\cref{tab:supp_affine}).
Component share and audio-video preference (A-V pref.) are expressed
as percentages. The dagger marks the main setting, $\lambda=0.5$.}
\label{tab:supp_lambda}

\setlength{\tabcolsep}{4pt}
\renewcommand{\arraystretch}{0.86}
\setlength{\aboverulesep}{0.8pt}
\setlength{\belowrulesep}{0.8pt}
\sisetup{table-number-alignment=center}

\newcommand{\lambdacomponenthead}[2]{%
  \multicolumn{1}{%
    >{\columncolor{#1!45}%
      [\dimexpr\tabcolsep-1pt\relax]%
      [\dimexpr\tabcolsep-1pt\relax]}c%
  }{#2}%
}

\begin{tabular}{@{}c l >{\vphantom{$0.5^\dagger$}}c
  *{2}{
    S[table-format=2.0]
    S[table-format=2.0]
    S[table-format=2.0]
    S[table-format=-2.0]
    S[table-format=3.0]
    S[table-format=2.0]
  }@{}}
\toprule
 & & & \multicolumn{6}{c}{\textbf{Subjects}}
     & \multicolumn{6}{c}{\textbf{Predicates}} \\
\cmidrule(lr){4-9}
\cmidrule(l){10-15}
 & & & \multicolumn{5}{c}{\textit{Component share}}
     & \multicolumn{1}{c}{\textit{A-V pref.}}
     & \multicolumn{5}{c}{\textit{Component share}}
     & \multicolumn{1}{c}{\textit{A-V pref.}} \\
\cmidrule(lr){4-8}
\cmidrule(lr){9-9}
\cmidrule(lr){10-14}
\cmidrule(l){15-15}
 & Model & $\lambda$
 & \lambdacomponenthead{Vcol}{V}
 & \lambdacomponenthead{Acol}{A}
 & \lambdacomponenthead{Qcol}{Q}
 & \lambdacomponenthead{Gcol}{G}
 & \lambdacomponenthead{Kcol}{K}
 & {$A>V$}
 & \lambdacomponenthead{Vcol}{V}
 & \lambdacomponenthead{Acol}{A}
 & \lambdacomponenthead{Qcol}{Q}
 & \lambdacomponenthead{Gcol}{G}
 & \lambdacomponenthead{Kcol}{K}
 & {$A>V$} \\
\midrule

\multirow[c]{30}{*}{%
  \rotatebox[origin=c]{90}{\textit{With value-projection bias}}%
}
 & \multirow{5}{*}{Qwen2.5-Omni 3B}
 & 0             & 34 & 4 & 6 &  2 & 53 & 5 & 21 &  8 & 7 &  5 & 59 & 19 \\
 & & 0.25        & 37 & 4 & 7 &  4 & 47 & 5 & 23 &  8 & 8 &  9 & 51 & 20 \\
 & & $0.5^\dagger$ & 41 & 5 & 8 &  7 & 38 & 5 & 26 &  9 & 9 & 13 & 43 & 19 \\
 & & 0.75        & 47 & 5 & 8 & 11 & 28 & 3 & 31 & 10 & 8 & 19 & 33 & 19 \\
 & & 1           & 48 & 5 & 3 & 16 & 28 & 4 & 37 & 12 & 0 & 28 & 23 & 20 \\
\addlinespace[3pt]
 & \multirow{5}{*}{Qwen2.5-Omni 7B}
 & 0             & 29 & 4 & 11 & -17 & 73 & 9 & 19 &  6 & 9 & -9 & 74 & 24 \\
 & & 0.25        & 30 & 4 & 10 & -11 & 66 & 9 & 20 &  7 & 8 & -3 & 68 & 23 \\
 & & $0.5^\dagger$ & 33 & 5 & 10 &  -5 & 58 & 8 & 21 &  8 & 7 &  5 & 59 & 22 \\
 & & 0.75        & 37 & 6 &  8 &   3 & 46 & 8 & 25 &  9 & 5 & 14 & 47 & 22 \\
 & & 1           & 45 & 7 &  7 &  14 & 26 & 9 & 34 & 12 & 0 & 24 & 30 & 22 \\
\addlinespace[3pt]
 & \multirow{5}{*}{MiniCPM-o 2.6}
 & 0             & 22 & 3 & 11 & -16 &  79 & 3 & 16 & 6 & 13 &  -8 &  73 & 17 \\
 & & 0.25        & 23 & 3 & 12 & -20 &  82 & 3 & 15 & 5 & 13 & -12 &  78 & 16 \\
 & & $0.5^\dagger$ & 23 & 3 & 12 & -26 &  88 & 2 & 15 & 5 & 12 & -16 &  84 & 15 \\
 & & 0.75        & 21 & 2 & 12 & -31 &  96 & 2 & 13 & 4 & 10 & -20 &  92 & 14 \\
 & & 1           & 17 & 1 & 11 & -37 & 107 & 3 & 10 & 3 &  8 & -26 & 105 & 16 \\
\addlinespace[3pt]
 & \multirow{5}{*}{VideoLLaMA2.1-7B-AV}
 & 0             & 28 &  9 &  7 & -24 & 79 & 13 & 19 & 11 & 10 & -26 & 85 & 31 \\
 & & 0.25        & 28 &  9 &  7 & -18 & 73 & 13 & 18 & 11 & 11 & -19 & 79 & 33 \\
 & & $0.5^\dagger$ & 29 & 10 &  8 & -12 & 65 & 14 & 18 & 12 & 11 & -12 & 71 & 32 \\
 & & 0.75        & 32 & 10 &  9 &  -6 & 54 & 14 & 19 & 13 & 11 &  -5 & 61 & 32 \\
 & & 1           & 39 & 12 & 12 &   1 & 36 & 13 & 24 & 15 &  9 &   5 & 47 & 32 \\
\addlinespace[3pt]
 & \multirow{5}{*}{Ola-7B}
 & 0             & 18 &  7 & 33 & -28 & 69 & 17 & 13 &  9 & 28 & -8 & 59 & 36 \\
 & & 0.25        & 20 &  7 & 30 & -23 & 66 & 17 & 13 &  9 & 26 & -5 & 56 & 36 \\
 & & $0.5^\dagger$ & 22 &  8 & 26 & -19 & 62 & 16 & 14 & 10 & 23 &  0 & 52 & 40 \\
 & & 0.75        & 25 &  9 & 23 & -14 & 57 & 14 & 16 & 12 & 20 &  6 & 46 & 41 \\
 & & 1           & 30 & 10 & 18 &  -7 & 49 & 12 & 19 & 15 & 13 & 14 & 38 & 44 \\
\addlinespace[3pt]
 & \multirow{5}{*}{Audio-Visual Flamingo}
 & 0             & 24 &  7 & 6 & -18 & 80 & 15 & 14 &  9 & 7 & -5 & 75 & 35 \\
 & & 0.25        & 25 &  7 & 6 & -11 & 73 & 16 & 14 & 10 & 7 &  0 & 68 & 37 \\
 & & $0.5^\dagger$ & 27 &  8 & 6 &  -4 & 64 & 16 & 15 & 12 & 7 &  6 & 60 & 37 \\
 & & 0.75        & 29 &  9 & 6 &   3 & 53 & 16 & 18 & 14 & 7 & 13 & 48 & 37 \\
 & & 1           & 34 & 10 & 6 &  11 & 38 & 15 & 21 & 18 & 7 & 22 & 32 & 38 \\
\midrule
\addlinespace[3pt]
\multirow[c]{15}{*}{%
  \rotatebox[origin=c]{90}{\textit{Without value-projection bias}}%
}
 & \multirow{5}{*}{MiniCPM-o 4.5}
 & 0             & 25 & 2 & 32 & 28 & 12 & 0 & 20 & 4 & 29 & 34 & 11 &  5 \\
 & & 0.25        & 25 & 2 & 32 & 28 & 13 & 0 & 20 & 4 & 28 & 35 & 12 &  5 \\
 & & $0.5^\dagger$ & 25 & 2 & 30 & 27 & 15 & 0 & 20 & 4 & 27 & 35 & 13 &  6 \\
 & & 0.75        & 25 & 2 & 27 & 27 & 17 & 0 & 21 & 5 & 24 & 35 & 14 &  8 \\
 & & 1           & 26 & 1 & 19 & 28 & 23 & 2 & 22 & 5 & 16 & 38 & 18 & 10 \\
\addlinespace[3pt]
 & \multirow{5}{*}{video-SALMONN~2~Pro~4B}
 & 0             & 37 & 2 & 15 & 26 & 20 & 0 & 28 & 2 & 17 & 31 & 21 & 0 \\
 & & 0.25        & 38 & 1 & 14 & 25 & 21 & 0 & 29 & 2 & 17 & 31 & 21 & 0 \\
 & & $0.5^\dagger$ & 39 & 1 & 13 & 24 & 22 & 0 & 30 & 2 & 16 & 30 & 21 & 0 \\
 & & 0.75        & 40 & 1 & 11 & 21 & 26 & 0 & 32 & 2 & 13 & 29 & 23 & 0 \\
 & & 1           & 38 & 0 &  9 & 16 & 35 & 0 & 31 & 2 & 10 & 28 & 29 & 1 \\
\addlinespace[3pt]
 & \multirow{5}{*}{video-SALMONN~2~Pro~8B}
 & 0             & 38 & 0 & 15 & 30 & 16 & 0 & 28 & 1 & 16 & 38 & 16 & 0 \\
 & & 0.25        & 39 & 0 & 14 & 29 & 17 & 0 & 28 & 1 & 16 & 37 & 17 & 0 \\
 & & $0.5^\dagger$ & 39 & 0 & 12 & 28 & 18 & 0 & 29 & 1 & 15 & 37 & 17 & 0 \\
 & & 0.75        & 40 & 0 & 11 & 27 & 20 & 0 & 30 & 1 & 13 & 37 & 18 & 0 \\
 & & 1           & 39 & 0 & 10 & 25 & 24 & 0 & 31 & 1 & 11 & 38 & 19 & 0 \\
\bottomrule
\end{tabular}
\end{table*}

\section{Additional robustness analyses}
\label{sec:supp_robustness}

\subsection{Controlling for caption position}
\label{sec:supp_position_control}
A potential confound in the role-dependent patterns of
\Cref{tab:source_attribution} is generation position. Later words are
generated after a longer textual prefix has accumulated, and the
balance between audio and video contributions may vary over the course
of generation. Because predicates can occur later than subjects, their
larger relative audio contribution could therefore partly reflect
caption position rather than semantic role.
We test whether the observed differences persist when the two roles
are compared under matched distributions of relative caption position.
For each word, we define relative position as $k/L$, where $k$ is the
position of its first model token and $L$ is the caption length in model
tokens. We divide relative position into five equal-width bins and use
direct standardization in both directions: predicates are reweighted to
match the position distribution of subjects, and subjects are
reweighted to match that of predicates. Reweighting changes only the
weight assigned to each word; word identities, role assignments, and
attribution values remain unchanged. Position support overlaps almost
completely between the two roles, with unmatched reference mass only
for Ola-7B (1.6\%). This unmatched mass is excluded, equivalently renormalizing the reference distribution over the remaining bins. The column reporting mean $k$ in
\Cref{tab:position_matched} is descriptive only; matching is performed
on relative position $k/L$. Confidence intervals use the same
1,000-resample clip-level bootstrap as
\Cref{tab:supp_aggregation_ic_95}.
Position balancing preserves the role-dependent attribution patterns.
When predicates are standardized to the subject position distribution,
their Audio Share remains larger than that of subjects in all nine
models. The reverse comparison is consistent with this pattern: moving
subjects to the predicate position distribution changes their Audio
Share only marginally, with predicates remaining larger in eight
models and tied in video-SALMONN 2 Pro 8B at the reported precision.
The complementary visual pattern is also preserved, with subjects
retaining larger Video Share than predicates.
Audio Contribution Density shows the same robustness after accounting
for the number of audio positions. Predicates retain higher audio
density than subjects in all nine models under both standardizations.
Matching changes the magnitude of some gaps, but never reverses their
direction.
The word-level comparisons are similarly stable. In the models with
nonzero Audio-Video Preference, audio exceeds video more often for
predicates than for subjects after position matching, while both
video-SALMONN~2 Pro variants remain near zero. Reweighting predicates
to subject positions also changes the frequency with which audio is the
Deciding Component only modestly; wherever audio is a deciding
component for predicates, it remains more frequent for predicates than for subjects.
Overall, caption position contributes to the magnitude of some
attribution differences but does not account for the role-dependent
pattern in \Cref{tab:source_attribution}. The larger relative audio
support for predicates and the larger visual support for subjects
persist after matching the two roles by generation position.

\newcommand{\posmatchci}[3]{%
  $#1$\,%
  {\setlength{\fboxsep}{1pt}%
   \colorbox{black!8}{$[#2,\,#3]$}}%
}
\newcommand{\posmatchhead}[2]{%
  \multicolumn{1}{%
    >{\columncolor{#1!45}%
       [\dimexpr\tabcolsep-1pt\relax]%
       [\dimexpr\tabcolsep-1pt\relax]}c%
  }{#2}%
}
\newcommand{\posmatchbiaslabel}[1]{%
  \rotatebox[origin=c]{90}{\textit{#1 value-projection bias}}%
}
\newsavebox{\posmatchrowref}
\newcommand{\posmatchbiascell}[1]{%
  \raisebox{\dimexpr
    (\ht\posmatchrowref-\dp\posmatchrowref-\height+\depth)/2\relax
  }[0pt][0pt]{\posmatchbiaslabel{#1}}%
}

\newsavebox{\posmatchtablebox}
\newsavebox{\posmatchscaledbox}
\newcommand{\posmatchbiasmid}[1]{%
  \noalign{%
    \vskip 1pt
    \vbox to 0pt{\vss\hbox{\posmatchbiaslabel{#1}}\vss}%
    \vskip 1pt
  }%
}

\begin{table*}[t]
\centering
\caption{Attribution at matched caption positions.
Models are grouped by the presence of value-projection biases. $n$ is the number of words, with the effective sample size in parentheses
for reweighted rows. Mean $k$ is the mean position index within the caption.
Component shares, audio-video preference ($A>V$), and deciding-component frequencies are reported as percentages. Audio contribution density is reported per $10^3$ audio input positions. Brackets denote 95\% confidence intervals from 1,000 clip-level bootstrap resamples.}

\label{tab:position_matched}

\scriptsize
\setlength{\tabcolsep}{2.2pt}
\renewcommand{\arraystretch}{0.94}
\sbox{\posmatchrowref}{\posmatchci{0.00}{-0.00}{0.00}}

\sbox{\posmatchtablebox}{%
\begin{tabular}{@{}c l r r cc c c cc@{}}
\toprule
 & & &
 & \multicolumn{2}{c}{\textbf{Component share}}
 & \textbf{A density} &
 & \multicolumn{2}{c}{\textbf{Deciding component}} \\
\cmidrule(lr){5-6}
\cmidrule(lr){7-7}
\cmidrule(lr){9-10}
\smash{\phantom{\posmatchbiaslabel{Without}}}
 & Model / word group & $n$ & Mean $k$
 & \posmatchhead{Vcol}{V} & \posmatchhead{Acol}{A}
 & A/tok & $A>V$
 & \posmatchhead{Acol}{A} & \posmatchhead{Vcol}{V} \\
\midrule
 & \multicolumn{9}{l@{}}{\emph{Qwen2.5-Omni 3B}} \\
  & \quad predicate & 210 & 7.8
 & \posmatchci{26}{25}{28} & \posmatchci{9}{7}{10} & \posmatchci{1.83}{1.42}{2.22}
 & \posmatchci{19}{14}{24} & \posmatchci{19}{14}{24} & \posmatchci{26}{20}{32} \\
  & \quad predicate, at subject positions & 210 (188) & 6.2
 & \posmatchci{28}{26}{30} & \posmatchci{7}{6}{9} & \posmatchci{1.56}{1.21}{1.89}
 & \posmatchci{14}{10}{19} & \posmatchci{15}{11}{20} & \posmatchci{26}{20}{34} \\
  & \quad subject & 333 & 5.7
 & \posmatchci{41}{40}{43} & \posmatchci{5}{4}{5} & \posmatchci{1.03}{0.84}{1.23}
 & \posmatchci{5}{2}{7} & \posmatchci{3}{2}{5} & \posmatchci{51}{46}{56} \\
  & \quad subject, at predicate positions & 333 (294) & 7.6
 & \posmatchci{41}{39}{43} & \posmatchci{5}{4}{6} & \posmatchci{1.17}{0.91}{1.46}
 & \posmatchci{7}{4}{11} & \posmatchci{4}{2}{7} & \posmatchci{52}{47}{58} \\
\addlinespace[4pt]
 & \multicolumn{9}{l@{}}{\emph{Qwen2.5-Omni 7B}} \\
  & \quad predicate & 260 & 13.1
 & \posmatchci{21}{20}{23} & \posmatchci{8}{6}{9} & \posmatchci{1.60}{1.30}{1.90}
 & \posmatchci{22}{17}{28} & \posmatchci{15}{11}{19} & \posmatchci{21}{15}{27} \\
  & \quad predicate, at subject positions & 260 (227) & 10.6
 & \posmatchci{24}{22}{26} & \posmatchci{7}{6}{8} & \posmatchci{1.47}{1.20}{1.75}
 & \posmatchci{19}{14}{24} & \posmatchci{15}{10}{19} & \posmatchci{23}{17}{29} \\
 & \quad subject & 391 & 10.2
 & \posmatchci{33}{31}{34} & \posmatchci{5}{4}{5} & \posmatchci{0.94}{0.81}{1.10}
 & \posmatchci{8}{6}{11} & \posmatchci{3}{1}{4} & \posmatchci{40}{35}{46} \\
  & \quad subject, at predicate positions & 391 (346) & 12.9
 & \posmatchci{30}{28}{33} & \posmatchci{5}{4}{6} & \posmatchci{1.02}{0.86}{1.20}
 & \posmatchci{11}{7}{15} & \posmatchci{4}{2}{6} & \posmatchci{40}{34}{45} \\
\addlinespace[4pt]
 & \multicolumn{9}{l@{}}{\emph{MiniCPM-o 2.6}} \\
  & \quad predicate & 172 & 14.4
 & \posmatchci{15}{14}{16} & \posmatchci{5}{4}{6} & \posmatchci{4.90}{4.04}{5.96}
 & \posmatchci{15}{9}{20} & \posmatchci{12}{7}{17} & \posmatchci{25}{19}{31} \\
  & \quad predicate, at subject positions & 172 (158) & 11.7
 & \posmatchci{15}{15}{16} & \posmatchci{5}{4}{5} & \posmatchci{4.34}{3.73}{5.06}
 & \posmatchci{10}{7}{15} & \posmatchci{10}{6}{14} & \posmatchci{24}{18}{31} \\
  & \quad subject & 326 & 11.2
 & \posmatchci{23}{22}{24} & \posmatchci{3}{2}{3} & \posmatchci{2.28}{1.92}{2.70}
 & \posmatchci{2}{1}{4} & \posmatchci{3}{1}{5} & \posmatchci{38}{32}{43} \\
  & \quad subject, at predicate positions & 326 (292) & 13.8
 & \posmatchci{21}{20}{22} & \posmatchci{3}{3}{3} & \posmatchci{2.41}{1.94}{3.00}
 & \posmatchci{4}{1}{7} & \posmatchci{3}{1}{5} & \posmatchci{35}{30}{40} \\
\posmatchbiasmid{With}
\addlinespace[2pt]
 & \multicolumn{9}{l@{}}{\emph{VideoLLaMA2.1-7B-AV}} \\
  & \quad predicate & 225 & 4.8
 & \posmatchci{18}{16}{19} & \posmatchci{12}{10}{13} & \posmatchci{1.52}{1.33}{1.71}
 & \posmatchci{32}{25}{38} & \posmatchci{12}{8}{17} & \posmatchci{25}{19}{31} \\
  & \quad predicate, at subject positions & 225 (156) & 3.9
 & \posmatchci{18}{16}{20} & \posmatchci{13}{11}{14} & \posmatchci{1.62}{1.39}{1.86}
 & \posmatchci{32}{24}{40} & \posmatchci{14}{8}{20} & \posmatchci{32}{24}{40} \\
  & \quad subject & 293 & 3.4
 & \posmatchci{29}{28}{31} & \posmatchci{10}{9}{11} & \posmatchci{1.24}{1.11}{1.38}
 & \posmatchci{14}{10}{18} & \posmatchci{8}{5}{11} & \posmatchci{38}{32}{43} \\
  & \quad subject, at predicate positions & 293 (212) & 4.8
 & \posmatchci{33}{32}{35} & \posmatchci{9}{8}{9} & \posmatchci{1.12}{1.02}{1.23}
 & \posmatchci{8}{5}{11} & \posmatchci{5}{3}{8} & \posmatchci{44}{37}{51} \\
\addlinespace[4pt]
 & \multicolumn{9}{l@{}}{\emph{Ola-7B}} \\
  & \quad predicate & 174 & 4.3
 & \posmatchci{14}{13}{15} & \posmatchci{10}{9}{12} & \posmatchci{8.02}{6.74}{9.35}
 & \posmatchci{40}{32}{48} & \posmatchci{28}{22}{35} & \posmatchci{32}{25}{39} \\
  & \quad predicate, at subject positions & 174 (154) & 3.4
 & \posmatchci{15}{14}{16} & \posmatchci{10}{8}{11} & \posmatchci{7.44}{6.19}{8.81}
 & \posmatchci{34}{26}{42} & \posmatchci{26}{20}{33} & \posmatchci{39}{31}{47} \\
 & \quad subject & 245 & 3.2
 & \posmatchci{22}{20}{24} & \posmatchci{8}{7}{9} & \posmatchci{5.48}{4.80}{6.29}
 & \posmatchci{16}{11}{21} & \posmatchci{8}{5}{12} & \posmatchci{40}{34}{46} \\
  & \quad subject, at predicate positions & 245 (203) & 4.2
 & \posmatchci{22}{20}{24} & \posmatchci{8}{7}{9} & \posmatchci{5.38}{4.51}{6.56}
 & \posmatchci{18}{12}{25} & \posmatchci{8}{4}{11} & \posmatchci{39}{32}{46} \\
\addlinespace[4pt]
 & \multicolumn{9}{l@{}}{\emph{AV-Flamingo}} \\
  & \quad predicate & 232 & 21.0
 & \posmatchci{15}{14}{17} & \posmatchci{12}{10}{13} & \posmatchci{2.54}{2.13}{3.05}
 & \posmatchci{37}{31}{43} & \posmatchci{16}{12}{22} & \posmatchci{13}{8}{18} \\
  & \quad predicate, at subject positions & 232 (201) & 18.6
 & \posmatchci{17}{14}{19} & \posmatchci{11}{9}{12} & \posmatchci{2.31}{1.91}{2.82}
 & \posmatchci{32}{26}{39} & \posmatchci{14}{9}{19} & \posmatchci{15}{9}{23} \\
  & \quad subject & 403 & 23.4
 & \posmatchci{27}{25}{28} & \posmatchci{8}{7}{8} & \posmatchci{1.48}{1.28}{1.71}
 & \posmatchci{16}{12}{20} & \posmatchci{6}{4}{9} & \posmatchci{26}{21}{30} \\
  & \quad subject, at predicate positions & 403 (357) & 28.3
 & \posmatchci{25}{24}{27} & \posmatchci{8}{7}{9} & \posmatchci{1.59}{1.37}{1.86}
 & \posmatchci{19}{15}{24} & \posmatchci{7}{4}{10} & \posmatchci{25}{20}{30} \\
\midrule
\addlinespace[2pt]
 & \multicolumn{9}{l@{}}{\emph{MiniCPM-o 4.5}} \\
  & \quad predicate & 239 & 8.0
 & \posmatchci{20}{19}{21} & \posmatchci{4}{4}{5} & \posmatchci{15.91}{13.39}{18.60}
 & \posmatchci{6}{4}{10} & \posmatchci{17}{12}{22} & \posmatchci{55}{49}{62} \\
  & \quad predicate, at subject positions & 239 (193) & 6.4
 & \posmatchci{21}{20}{22} & \posmatchci{4}{4}{4} & \posmatchci{13.89}{11.68}{16.22}
 & \posmatchci{4}{3}{7} & \posmatchci{14}{10}{19} & \posmatchci{60}{53}{66} \\
  & \quad subject & 308 & 6.1
 & \posmatchci{25}{24}{25} & \posmatchci{2}{2}{2} & \posmatchci{6.84}{5.79}{8.09}
 & \posmatchci{0}{0}{0} & \posmatchci{1}{0}{3} & \posmatchci{54}{48}{60} \\
  & \quad subject, at predicate positions & 308 (256) & 7.9
 & \posmatchci{24}{23}{25} & \posmatchci{2}{2}{3} & \posmatchci{7.84}{6.46}{9.37}
 & \posmatchci{0}{0}{0} & \posmatchci{2}{0}{4} & \posmatchci{55}{49}{62} \\
\addlinespace[4pt]
 & \multicolumn{9}{l@{}}{\emph{video-SALMONN 2 Pro 4B}} \\
  & \quad predicate & 216 & 13.2
 & \posmatchci{30}{29}{32} & \posmatchci{2}{2}{2} & \posmatchci{3.98}{3.44}{4.63}
 & \posmatchci{0}{0}{0} & \posmatchci{0}{0}{0} & \posmatchci{51}{44}{58} \\
  & \quad predicate, at subject positions & 216 (195) & 11.5
 & \posmatchci{31}{29}{32} & \posmatchci{2}{1}{2} & \posmatchci{3.88}{3.38}{4.48}
 & \posmatchci{0}{0}{0} & \posmatchci{0}{0}{0} & \posmatchci{55}{48}{62} \\
\posmatchbiascell{Without} & \quad subject & 349 & 11.4
 & \posmatchci{39}{38}{41} & \posmatchci{1}{1}{1} & \posmatchci{3.17}{2.79}{3.67}
 & \posmatchci{0}{0}{0} & \posmatchci{0}{0}{0} & \posmatchci{58}{52}{63} \\
  & \quad subject, at predicate positions & 349 (313) & 13.2
 & \posmatchci{39}{38}{40} & \posmatchci{1}{1}{1} & \posmatchci{3.15}{2.78}{3.66}
 & \posmatchci{0}{0}{0} & \posmatchci{0}{0}{0} & \posmatchci{58}{53}{64} \\
\addlinespace[4pt]
 & \multicolumn{9}{l@{}}{\emph{video-SALMONN 2 Pro 8B}} \\
  & \quad predicate & 210 & 13.3
 & \posmatchci{29}{28}{30} & \posmatchci{1}{1}{1} & \posmatchci{2.98}{2.25}{3.75}
 & \posmatchci{0}{0}{0} & \posmatchci{1}{0}{3} & \posmatchci{54}{47}{61} \\
  & \quad predicate, at subject positions & 210 (170) & 10.3
 & \posmatchci{30}{29}{32} & \posmatchci{1}{1}{1} & \posmatchci{2.45}{1.85}{3.10}
 & \posmatchci{0}{0}{0} & \posmatchci{1}{0}{2} & \posmatchci{55}{48}{63} \\
  & \quad subject & 333 & 10.3
 & \posmatchci{39}{38}{41} & \posmatchci{0}{0}{1} & \posmatchci{1.23}{0.95}{1.74}
 & \posmatchci{0}{0}{0} & \posmatchci{0}{0}{0} & \posmatchci{60}{54}{65} \\
  & \quad subject, at predicate positions & 333 (282) & 13.7
 & \posmatchci{39}{37}{40} & \posmatchci{1}{0}{1} & \posmatchci{1.37}{1.00}{2.03}
 & \posmatchci{0}{0}{0} & \posmatchci{0}{0}{0} & \posmatchci{59}{53}{65} \\
\bottomrule
\end{tabular}%
}
\ifdim\dimexpr\ht\posmatchtablebox+\dp\posmatchtablebox\relax>.82\textheight
  \sbox{\posmatchscaledbox}{%
    \resizebox*{!}{.82\textheight}{\usebox{\posmatchtablebox}}%
  }%
\else
  \sbox{\posmatchscaledbox}{\usebox{\posmatchtablebox}}%
\fi
\ifdim\wd\posmatchscaledbox>\linewidth
  \resizebox{\linewidth}{!}{\usebox{\posmatchscaledbox}}%
\else
  \usebox{\posmatchscaledbox}%
\fi
\end{table*}

\subsection{Agreement assessment}
\label{subsec:supp_agreement_baselines}
High agreement with modality obscuration can arise from class imbalance when most interventions produce logit changes of the same sign or favor the same modality.
\Cref{tab:trivial_baselines} therefore compares attribution
methods with two constant baselines. For Sign Agreement, \emph{always
positive} predicts a positive logit decrease after obscuration.
For Dominant Agreement, \emph{always video} predicts that obscuring
video produces the larger signed logit decrease. We also report
agreement separately for words on which the audio or video intervention
produces the larger decrease. Balanced Agreement gives the two groups
equal weight, so either constant modality prediction scores 50\%.
The always-positive baseline reaches 94\% for video on
Qwen2.5-Omni~3B and 96\% on VideoLLaMA2.1-7B-AV, close to the
values for MALT, AttnLRP, and ALTI-Logit. The same holds for
audio on VideoLLaMA2.1-7B-AV, where the baseline reaches 90\%
and all three methods reach 91\%. These high values therefore
largely reflect the prevalence of positive logit decreases.
Sign Agreement alone cannot show whether the attribution scores
track differences in intervention effects across words.
On Qwen2.5-Omni~3B, video obscuration produces the larger logit
decrease for 464 of 543 words. Always predicting video thus reaches
85\% overall agreement, close to the 85--86\% obtained by the
attention variants. These variants rank video above audio for
97--100\% of words and recover only 0--11\% of audio-dominant
cases. MALT identifies 54\% of audio-dominant cases and 97\% of
video-dominant cases, reaching 76\% balanced agreement. AttnLRP
and ALTI-Logit reach 75\% and 71\%, respectively, compared with
50--55\% for the attention variants. Gradient$\times$Input
recovers more audio-dominant cases than MALT (67\% vs.\ 54\%),
but identifies fewer video-dominant cases (48\% vs.\ 97\%),
resulting in 58\% balanced agreement.
The distinction also matters on VideoLLaMA2.1-7B-AV:
ALTI-Logit achieves similar overall agreement to MALT
(83\% vs.\ 82\%), but identifies fewer audio-dominant words
(23\% vs.\ 59\%). Once the two groups receive equal weight,
MALT and AttnLRP both reach 74\%, compared with 61\% for
ALTI-Logit and 50--52\% for the attention variants.
MiniCPM-o~4.5 shows a different pattern. Although 223 of 547
words are audio-dominant under obscuration, MALT, AttnLRP, and
ALTI-Logit identify only 1--6\% of these cases, and their
balanced agreement remains close to the constant baseline
(48--51\%). Thus, attribution rankings identify the modality whose obscuration has the larger effect more reliably than an always-video prediction on Qwen2.5-Omni~3B and VideoLLaMA2.1-7B-AV, but not on MiniCPM-o~4.5.

\begin{table*}[t]
\caption{Agreement under modality obscuration relative to trivial baselines.
\emph{Always $+$} assigns a positive sign to every modality and therefore
measures the marginal prevalence of positive logit changes under obscuration.
\emph{Always video} ranks video above audio for every word.
Dominant-modality agreement is reported over all words and separately
for audio-dominant (A-dom. $\Delta\ell_A>\Delta\ell_V$) and
video-dominant (V-dom. $\Delta\ell_V>\Delta\ell_A$) words.
Balanced agreement is the mean of the two class-conditional agreements
and equals 50\% for either constant prediction.
Video-Audio Preference (V-A Pref.) is the fraction of words for which
a method assigns the larger score to video.
All agreement and preference metrics are reported as percentages.
Brackets denote 95\% confidence intervals from 1,000 clip-level
bootstrap resamples.}
\label{tab:trivial_baselines}
\centering
\scriptsize
\setlength{\tabcolsep}{3pt}

\newcommand{\trivialci}[3]{%
  \shortstack[c]{%
    $#1$\\[1pt]
    {\setlength{\fboxsep}{1pt}%
     \colorbox{black!8}{$[#2,\,#3]$}}%
  }%
}
\newcommand{\trivialbiaslabel}[1]{%
  \rotatebox[origin=c]{90}{\textit{#1 value-projection bias}}%
}
\newcommand{\trivialbiasmid}[2]{%
  \noalign{%
    \vskip\dimexpr#2/2\relax
    \vbox to 0pt{\vss\hbox{\trivialbiaslabel{#1}}\vss}%
    \vskip\dimexpr#2/2\relax
  }%
}

\begin{tabular}{@{}c l cc cccc c@{}}
\toprule
 & & \multicolumn{2}{c}{\textbf{Sign agreement}}
 & \multicolumn{4}{c}{\textbf{Dominant agreement}}
 & \textbf{V-A pref.}\\
\cmidrule(lr){3-4}
\cmidrule(lr){5-8}
\cmidrule(lr){9-9}
\smash{\phantom{\trivialbiaslabel{Without}}}
 & Method & Video & Audio
 & All & A-dom. & V-dom. & Balanced & $V>A$ \\
\midrule

 & \multicolumn{8}{l@{}}{\emph{Qwen2.5-Omni 3B}
   ($n=543$: 79 A-dom., 464 V-dom.; sign test: 530 V / 421 A)} \\
\addlinespace[2pt]
 & Always $+$ / always video
 & \trivialci{94}{91}{96} & \trivialci{44}{38}{50}
 & \trivialci{85}{82}{89} & \trivialci{0}{0}{0}
 & \trivialci{100}{100}{100} & \trivialci{50}{50}{50}
 & 100 \\
\addlinespace[2pt]
 & MALT
 & \trivialci{93}{91}{96} & \trivialci{48}{42}{54}
 & \trivialci{91}{88}{94} & \trivialci{54}{42}{66}
 & \trivialci{97}{96}{99} & \trivialci{76}{70}{82}
 & 90 \\
\addlinespace[2pt]
 & Attention mass, mean
 & -- & --
 & \trivialci{86}{82}{89} & \trivialci{3}{0}{7}
 & \trivialci{100}{99}{100} & \trivialci{51}{50}{53}
 & 99 \\
\addlinespace[2pt]
 & Attention mass, last layer
 & -- & --
 & \trivialci{85}{82}{89} & \trivialci{0}{0}{0}
 & \trivialci{100}{100}{100} & \trivialci{50}{50}{50}
 & 100 \\
\addlinespace[2pt]
 & Attention $\times \|v\|$, mean
 & -- & --
 & \trivialci{86}{82}{89} & \trivialci{11}{5}{19}
 & \trivialci{99}{98}{100} & \trivialci{55}{52}{59}
 & 97 \\
\addlinespace[2pt]
 & Gradient$\times$Input
 & \trivialci{58}{53}{63} & \trivialci{47}{41}{53}
 & \trivialci{51}{47}{56} & \trivialci{67}{57}{77}
 & \trivialci{48}{44}{53} & \trivialci{58}{52}{63}
 & 46 \\
\addlinespace[2pt]
 & AttnLRP
 & \trivialci{93}{91}{95} & \trivialci{58}{51}{65}
 & \trivialci{91}{88}{93} & \trivialci{53}{41}{65}
 & \trivialci{97}{96}{99} & \trivialci{75}{69}{81}
 & 90 \\
\addlinespace[2pt]
 & ALTI-Logit
 & \trivialci{94}{91}{96} & \trivialci{44}{39}{50}
 & \trivialci{90}{88}{93} & \trivialci{43}{31}{54}
 & \trivialci{98}{97}{100} & \trivialci{71}{65}{76}
 & 92 \\

\trivialbiasmid{With}{4pt}
\addlinespace[4pt]

 & \multicolumn{8}{l@{}}{\emph{VideoLLaMA2.1-7B-AV}
   ($n=518$: 115 A-dom., 403 V-dom.; sign test: 507 V / 500 A)} \\
\addlinespace[2pt]
 & Always $+$ / always video
 & \trivialci{96}{94}{97} & \trivialci{90}{88}{93}
 & \trivialci{78}{74}{82} & \trivialci{0}{0}{0}
 & \trivialci{100}{100}{100} & \trivialci{50}{50}{50}
 & 100 \\
\addlinespace[2pt]
 & MALT
 & \trivialci{94}{92}{97} & \trivialci{91}{88}{93}
 & \trivialci{82}{79}{86} & \trivialci{59}{50}{69}
 & \trivialci{89}{86}{92} & \trivialci{74}{69}{79}
 & 78 \\
\addlinespace[2pt]
 & Attention mass, mean
 & -- & --
 & \trivialci{79}{75}{82} & \trivialci{5}{1}{10}
 & \trivialci{100}{99}{100} & \trivialci{52}{50}{55}
 & 99 \\
\addlinespace[2pt]
 & Attention mass, last layer
 & -- & --
 & \trivialci{78}{74}{82} & \trivialci{0}{0}{0}
 & \trivialci{100}{100}{100} & \trivialci{50}{50}{50}
 & 100 \\
\addlinespace[2pt]
 & Attention $\times \|v\|$, mean
 & -- & --
 & \trivialci{78}{75}{82} & \trivialci{1}{0}{3}
 & \trivialci{100}{100}{100} & \trivialci{50}{50}{51}
 & 100 \\
\addlinespace[2pt]
 & Gradient$\times$Input
 & \trivialci{50}{46}{55} & \trivialci{33}{28}{38}
 & \trivialci{59}{55}{63} & \trivialci{45}{36}{54}
 & \trivialci{63}{59}{68} & \trivialci{54}{49}{59}
 & 61 \\
\addlinespace[2pt]
 & AttnLRP
 & \trivialci{91}{88}{94} & \trivialci{91}{89}{94}
 & \trivialci{79}{75}{83} & \trivialci{65}{56}{74}
 & \trivialci{83}{79}{87} & \trivialci{74}{70}{79}
 & 72 \\
\addlinespace[2pt]
 & ALTI-Logit
 & \trivialci{96}{94}{97} & \trivialci{91}{89}{94}
 & \trivialci{83}{80}{86} & \trivialci{23}{16}{31}
 & \trivialci{100}{99}{100} & \trivialci{61}{58}{65}
 & 94 \\
\midrule

 & \multicolumn{8}{l@{}}{\emph{MiniCPM-o 4.5}
   ($n=547$: 223 A-dom., 324 V-dom.; sign test: 544 V / 540 A)} \\
\addlinespace[2pt]
 & Always $+$ / always video
 & \trivialci{59}{55}{63} & \trivialci{44}{39}{48}
 & \trivialci{59}{55}{64} & \trivialci{0}{0}{0}
 & \trivialci{100}{100}{100} & \trivialci{50}{50}{50}
 & 100 \\
\addlinespace[2pt]
 & MALT
 & \trivialci{59}{55}{63} & \trivialci{44}{40}{49}
 & \trivialci{57}{53}{62} & \trivialci{1}{0}{2}
 & \trivialci{96}{94}{98} & \trivialci{48}{47}{50}
 & 97 \\
\addlinespace[2pt]
 & Attention mass, mean
 & -- & --
 & \trivialci{59}{55}{64} & \trivialci{0}{0}{0}
 & \trivialci{100}{100}{100} & \trivialci{50}{50}{50}
 & 100 \\
\addlinespace[2pt]
 & Attention mass, last layer
 & -- & --
 & \trivialci{59}{55}{64} & \trivialci{0}{0}{0}
 & \trivialci{100}{100}{100} & \trivialci{50}{50}{50}
 & 100 \\
\trivialbiasmid{Without}{2pt}
 & Attention $\times \|v\|$, mean
 & -- & --
 & \trivialci{59}{55}{64} & \trivialci{0}{0}{1}
 & \trivialci{100}{99}{100} & \trivialci{50}{50}{51}
 & 100 \\
\addlinespace[2pt]
 & Gradient$\times$Input
 & \trivialci{52}{48}{56} & \trivialci{51}{46}{55}
 & \trivialci{52}{48}{57} & \trivialci{52}{45}{58}
 & \trivialci{53}{47}{59} & \trivialci{52}{48}{57}
 & 51 \\
\addlinespace[2pt]
 & AttnLRP
 & \trivialci{59}{55}{63} & \trivialci{46}{41}{50}
 & \trivialci{58}{54}{62} & \trivialci{6}{3}{9}
 & \trivialci{93}{90}{96} & \trivialci{50}{47}{52}
 & 94 \\
\addlinespace[2pt]
 & ALTI-Logit
 & \trivialci{59}{55}{63} & \trivialci{44}{39}{48}
 & \trivialci{59}{56}{64} & \trivialci{4}{1}{7}
 & \trivialci{97}{95}{99} & \trivialci{51}{49}{52}
 & 97 \\
\bottomrule
\end{tabular}
\end{table*}

\subsection{Sensitivity to audio-removal baseline}
\label{subsec:supp_audio_replacement}
Counterfactual experiments in \Cref{tab:counterfactual_intervention} replace the original audio waveform with digital silence, which removes the acoustic evidence but may also induce model-specific responses. This is particularly relevant for MiniCPM-o, where audio obscuration can increase the target logit despite reducing the attributed audio contribution, and attribution-intervention consistency is weak (\Cref{tab:supp_consistency}). To examine whether these effects are specific to digital silence, we repeat the audio intervention on MiniCPM-o 2.6 and 4.5 using low-amplitude white noise at $-40$\,dBFS RMS. The generated captions, target words, teacher-forced sequences, and clip-level bootstrap procedure remain unchanged. Video obscuration is also kept fixed.
\Cref{tab:noise_baseline} compares the component-level responses
and attribution-intervention consistency under the two audio
replacements.
\paragraph{Counterfactual response.}
For MiniCPM-o 4.5, both interventions reduce the audio contribution,
but their effects on the target logit differ across linguistic roles.
For predicates, the reduction in audio contribution is larger under
silence than under white noise
($2.01$ and $1.44$). Nevertheless, both interventions increase the
target logit ($-2.98$ and $-3.54$, respectively),
accompanied by negative changes in the remaining MALT components.
For subjects, the audio contribution also decreases under both
interventions ($0.79$ and $0.32$), but the mean
target-logit change reverses sign, from $-1.20$
under silence to $+2.15$ under white noise. This difference
is accompanied by changes in the redistribution of contributions:
$\Delta C_Q$, $\Delta C_G$, and $\Delta C_K$ are negative under
silence but positive under white noise. Thus, the anomalous
logit increase persists for predicates under both interventions,
whereas the response for subjects depends on the replacement signal.
MiniCPM-o 2.6 exhibits smaller differences between the two
interventions. For predicates, $\Delta C_A$ changes from
$1.72$ under silence to $1.66$ under white noise, with
corresponding target-logit changes of $-0.59$ and $-0.15$.
For subjects, $\Delta C_A$ remains positive ($0.27$ and $0.23$),
and the target-logit changes are also negative under both
interventions ($-0.32$ and $-0.57$, respectively).
\paragraph{Attribution-intervention consistency.}
Changing the audio replacement does not resolve the weak consistency
observed for MiniCPM-o 4.5 (\Cref{tab:supp_consistency}). Audio Sign Agreement increases from
44\% under silence to 52\% under white noise, but the
always-positive reference increases by the same amount.
Pearson and Spearman correlations remain near zero
($r=0.03$ and $\rho=0.00$ under silence; $r=0.01$ and
$\rho=0.03$ under white noise), indicating that the higher
Sign Agreement does not correspond to improved association
between attribution scores and intervention effects.
Dominant-modality agreement leads to the same conclusion.
Balanced agreement remains close to the 50\% constant-reference
level (48\% under silence and 49\% under white noise).
MALT ranks video above audio for 97\% of words under both
interventions, identifying only 1\% and 2\% of audio-dominant
words, respectively.
For MiniCPM-o 2.6, audio Sign Agreement changes from 43\% to
41\%, and balanced dominant agreement from 61\% to 60\%.
Agreement on audio-dominant words remains at 26\%.
Pearson correlation increases from $0.13$ to $0.20$, although
both confidence intervals include zero. Under low-amplitude white noise, MiniCPM-o 4.5 retains near-zero correlations and near-constant balanced agreement, despite the increase in raw Sign Agreement.
Overall, the two replacements produce different counterfactual
responses, showing that the choice of replacement affects
the redistribution of contributions and, in some cases, even
the direction of the target-logit change. Nevertheless, the
near-zero correlations and near-constant balanced agreement
of MiniCPM-o 4.5 persist under white noise. Its weak
attribution-intervention consistency therefore cannot be
explained solely by the use of digital silence, although
neither replacement can be considered a neutral removal of audio.

\begin{table*}[t]
\centering
\scriptsize
\caption{Sensitivity to audio replacement in MiniCPM-o.
Digital silence and low-amplitude white noise are compared
using the same generated captions and matched target words.
Results are reported separately for MiniCPM-o 4.5
(without value-projection bias) and MiniCPM-o 2.6
(with value-projection bias).
Brackets denote 95\% confidence intervals from
1,000 clip-level bootstrap resamples.}
\label{tab:noise_baseline}

\setlength{\tabcolsep}{3pt}
\captionsetup[subtable]{position=top}

\newcolumntype{Y}{>{\centering\arraybackslash}X}

\newcommand{\noiseci}[4][]{%
  \shortstack[c]{%
    $#2$#1\\[1pt]
    {\setlength{\fboxsep}{1pt}%
     \colorbox{black!8}{$[#3,\,#4]$}}%
  }%
}
\newcommand{\noisedeltahead}[2]{%
  \colorbox{#1!45}{%
    \makebox[\dimexpr\linewidth-2\fboxsep\relax][c]{%
      $\mathbf{\Delta C_{\mathrm{#2}}}$%
    }%
  }%
}


\subcaptionbox{Counterfactual component changes.
Results are reported separately for event subjects
(\textit{Subj.}) and predicates (\textit{Pred.}), with
$n$ indicating the number of target words.
$\Delta C_s$ denotes the change in each MALT component
($s\in\{V,A,Q,G,K\}$), and $\Delta\ell$ the corresponding
target-logit change, computed as original minus
post-intervention values. All changes are expressed
in logit units.
\label{tab:noise_components}}[\linewidth]{%
\centering
\begin{tabularx}{\linewidth}{@{}llc*{6}{Y}@{}}
\toprule
Role & Replacement & $n$
 & \noisedeltahead{Vcol}{V}
 & \noisedeltahead{Acol}{A}
 & \noisedeltahead{Qcol}{Q}
 & \noisedeltahead{Gcol}{G}
 & \noisedeltahead{Kcol}{K}
 & $\mathbf{\Delta\ell}$ \\
\midrule
\multicolumn{9}{@{}l@{}}{\textit{MiniCPM-o 2.6 (With value-projection bias)}} \\
\addlinespace[2pt]
\multirow[c]{2}{*}{\textit{Pred.}}
 & silence & 172
 & \noiseci{-0.33}{-0.58}{-0.10}
 & \noiseci{1.72}{1.22}{2.27}
 & \noiseci{-0.57}{-0.69}{-0.45}
 & \noiseci{-0.46}{-0.69}{-0.24}
 & \noiseci{-0.96}{-1.51}{-0.47}
 & \noiseci{-0.59}{-1.09}{-0.11} \\
 & white noise & 172
 & \noiseci{-0.20}{-0.44}{0.04}
 & \noiseci{1.66}{1.18}{2.19}
 & \noiseci{-0.33}{-0.44}{-0.21}
 & \noiseci{-0.08}{-0.31}{0.12}
 & \noiseci{-1.22}{-1.73}{-0.76}
 & \noiseci{-0.15}{-0.62}{0.36} \\

\addlinespace[2pt]
\multirow[c]{2}{*}{\textit{Subj.}}
 & silence & 326
 & \noiseci{0.12}{-0.02}{0.27}
 & \noiseci{0.27}{0.07}{0.47}
 & \noiseci{-0.29}{-0.38}{-0.22}
 & \noiseci{-0.24}{-0.40}{-0.07}
 & \noiseci{-0.18}{-0.47}{0.13}
 & \noiseci{-0.32}{-0.57}{-0.06} \\
 & white noise & 326
 & \noiseci{0.04}{-0.10}{0.20}
 & \noiseci{0.23}{0.04}{0.43}
 & \noiseci{-0.27}{-0.35}{-0.18}
 & \noiseci{0.16}{-0.01}{0.34}
 & \noiseci{-0.75}{-1.04}{-0.45}
 & \noiseci{-0.57}{-0.84}{-0.31} \\

 \midrule
\multicolumn{9}{@{}l@{}}{\textit{MiniCPM-o 4.5 (Without value-projection bias)}} \\
\addlinespace[2pt]
\multirow[c]{2}{*}{\textit{Pred.}}
 & silence & 239
 & \noiseci{-0.72}{-1.59}{0.09}
 & \noiseci{2.01}{1.61}{2.43}
 & \noiseci{-1.59}{-2.29}{-0.92}
 & \noiseci{-1.31}{-2.28}{-0.43}
 & \noiseci{-1.33}{-1.87}{-0.80}
 & \noiseci{-2.98}{-5.71}{-0.37} \\
 & white noise & 239
 & \noiseci{-1.90}{-2.86}{-1.04}
 & \noiseci{1.44}{1.04}{1.86}
 & \noiseci{-1.27}{-2.00}{-0.59}
 & \noiseci{-0.92}{-1.93}{0.05}
 & \noiseci{-0.86}{-1.42}{-0.36}
 & \noiseci{-3.54}{-6.23}{-0.97} \\

\addlinespace[2pt]
\multirow[c]{2}{*}{\textit{Subj.}}
 & silence & 308
 & \noiseci{-0.47}{-0.85}{-0.08}
 & \noiseci{0.79}{0.62}{0.96}
 & \noiseci{-0.06}{-0.57}{0.43}
 & \noiseci{-0.76}{-1.34}{-0.25}
 & \noiseci{-0.67}{-1.05}{-0.31}
 & \noiseci{-1.20}{-2.78}{0.19} \\
 & white noise & 308
 & \noiseci{-0.70}{-1.14}{-0.22}
 & \noiseci{0.32}{0.14}{0.51}
 & \noiseci{1.70}{1.15}{2.32}
 & \noiseci{0.40}{-0.25}{1.04}
 & \noiseci{0.41}{0.01}{0.82}
 & \noiseci{2.15}{0.49}{3.89} \\

\bottomrule
\end{tabularx}%
}

\par\vspace{8pt}


\subcaptionbox{Attribution-intervention consistency.
$n$ denotes the number of target words, with counts
in parentheses indicating the words evaluated for
the corresponding metric.
Same sign reports Sign Agreement for MALT and an
always-positive reference; $r$ and $\rho$ denote
Pearson and Spearman correlations, respectively.
Dominant-modality agreement is reported over all words,
for an always-video reference, over audio-dominant words,
and as the balanced average across the two modality classes.
$V>A$ denotes the percentage of words for which MALT
assigns greater contribution to video than audio.
Agreement and preference metrics are expressed as percentages.
\label{tab:noise_consistency}}[\linewidth]{%
\resizebox{\linewidth}{!}{%
\begin{tabular}{@{}l l *{9}{c}@{}}
\toprule
 & & \multicolumn{2}{c}{\textbf{Same sign}}
 & & &
 \multicolumn{4}{c}{\textbf{Dominant modality}}
 & \textbf{V-A pref.} \\
\cmidrule(lr){3-4}
\cmidrule(lr){7-10}
\cmidrule(lr){11-11}
Replacement & $n$
 & MALT & always $+$
 & $r$ & $\rho$
 & all words & always video
 & audio-dominant & balanced
 & $V>A$ \\
\midrule

\multicolumn{11}{@{}l@{}}{\textit{MiniCPM-o 2.6 (With value-projection bias)}} \\
\addlinespace[2pt]
silence & 498
 & \noiseci[\, (454)]{43}{37}{48}
 & \noiseci[\, (458)]{41}{36}{46}
 & \noiseci{0.13}{-0.10}{0.35}
 & \noiseci{0.03}{-0.07}{0.14}
 & \noiseci{90}{86}{93}
 & \noiseci{92}{88}{94}
 & \noiseci[\, (42)]{26}{12}{44}
 & \noiseci{61}{54}{70}
 & 94 \\
\addlinespace[2pt]
white noise & 498
 & \noiseci[\, (466)]{41}{36}{46}
 & \noiseci[\, (469)]{40}{35}{44}
 & \noiseci{0.20}{-0.04}{0.41}
 & \noiseci{0.09}{-0.01}{0.19}
 & \noiseci{89}{86}{93}
 & \noiseci{91}{88}{94}
 & \noiseci[\, (43)]{26}{12}{45}
 & \noiseci{60}{54}{70}
 & 94 \\

 \midrule

 \multicolumn{11}{@{}l@{}}{\textit{MiniCPM-o 4.5 (Without value-projection bias)}} \\
\addlinespace[2pt]
silence & 547
 & \noiseci[\, (536)]{44}{40}{49}
 & \noiseci[\, (540)]{44}{39}{48}
 & \noiseci{0.03}{-0.09}{0.14}
 & \noiseci{0.00}{-0.10}{0.08}
 & \noiseci{57}{53}{62}
 & \noiseci{59}{55}{64}
 & \noiseci[\, (223)]{1}{0}{2}
 & \noiseci{48}{47}{50}
 & 97 \\
\addlinespace[2pt]
white noise & 547
 & \noiseci[\, (537)]{52}{48}{56}
 & \noiseci[\, (541)]{52}{48}{56}
 & \noiseci{0.01}{-0.11}{0.11}
 & \noiseci{0.03}{-0.06}{0.12}
 & \noiseci{58}{54}{62}
 & \noiseci{59}{55}{63}
 & \noiseci[\, (224)]{2}{0}{4}
 & \noiseci{49}{48}{51}
 & 97 \\
\bottomrule
\end{tabular}%
}%
}

\end{table*}

\section{Qualitative examples}

This section presents qualitative examples across the evaluated models, counterfactual interventions, and attribution methods. Each panel visualizes the attribution scores for a selected generated event token, projected onto sampled video frames, the audio waveform, prompt tokens, and preceding generated text. Red, white and blue indicate positive, near-zero and negative scores, respectively, with opacity proportional to magnitude and separate color scales for each input modality. Green annotations indicate the temporal extent of the corresponding sound event and its spatial extent in the video frames. Only in the modality-swap examples, audio annotations are shown in orange, while video annotations remain green to distinguish the conflicting events. The colored boxes at the bottom summarize the scores aggregated by component.

\subsection{MALT attribution}
\begin{figure*}[t]
    \centering
    \setlength{\fboxsep}{0pt}
    \setlength{\fboxrule}{0.15pt}
    \colorlet{attrallborder}{black!35}

    \newcommand{\attrpanel}[2]{%
        \begin{minipage}[t]{0.326\textwidth}
            \centering
            \textbf{#1}\par\vspace{2pt}
            {\color{attrallborder}%
            \fbox{%
                \includegraphics[
                    width=\dimexpr\linewidth-2\fboxrule\relax
                ]{#2}%
            }}%
        \end{minipage}%
    }

    \noindent
    \attrpanel{Omni 3B}
        {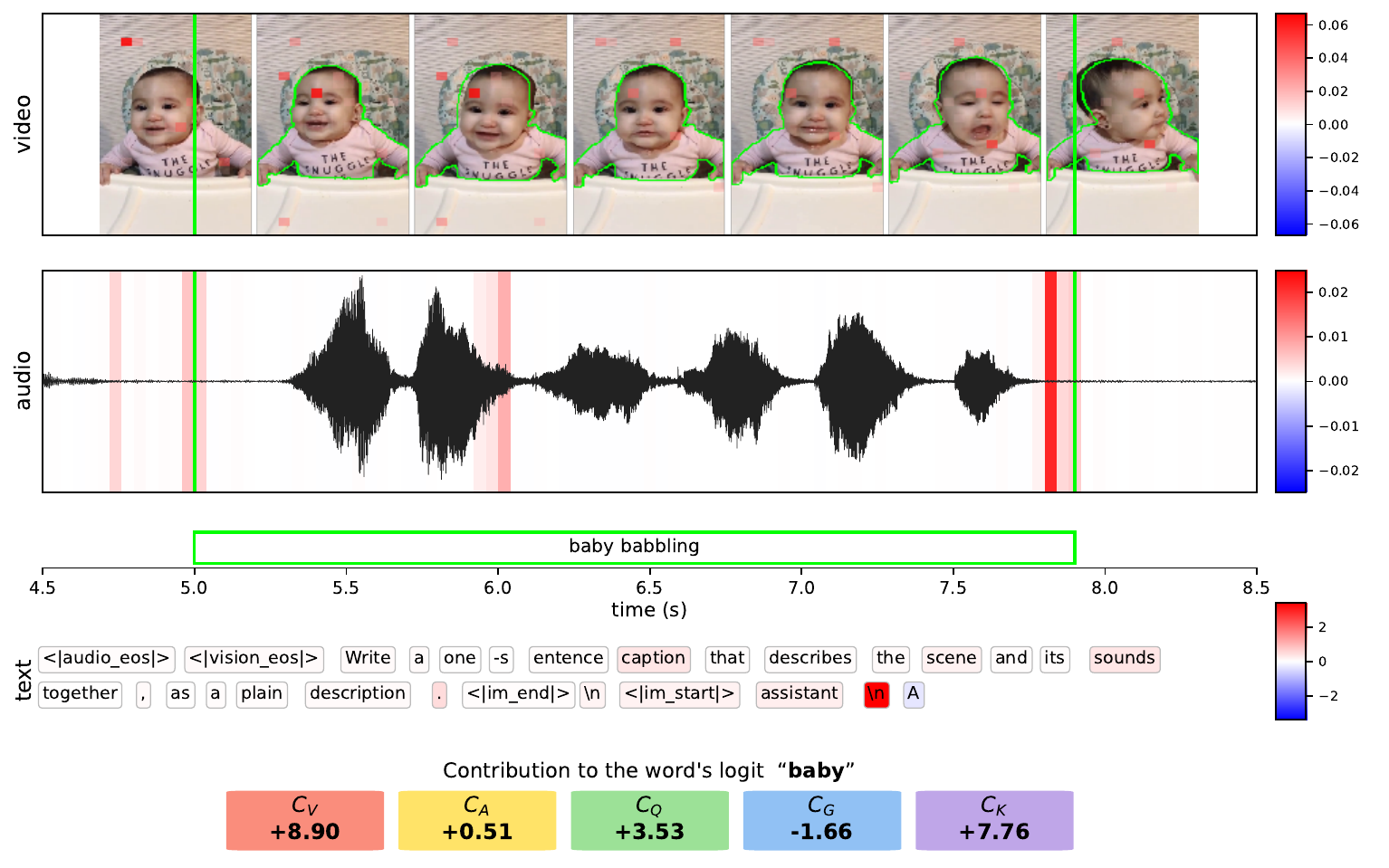}\hfill
    \attrpanel{Omni 7B}
        {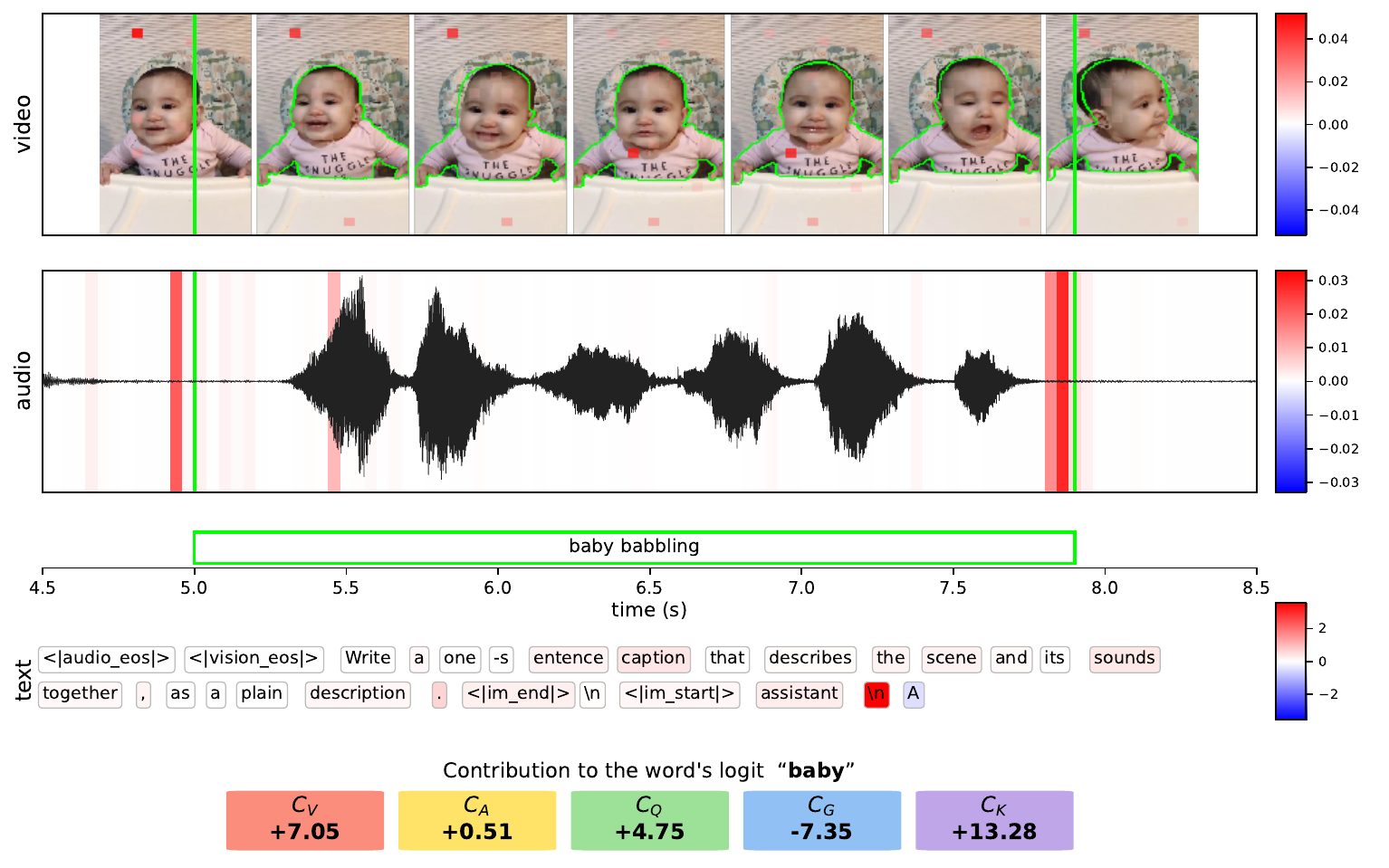}\hfill
    \attrpanel{MCP 2.6}
        {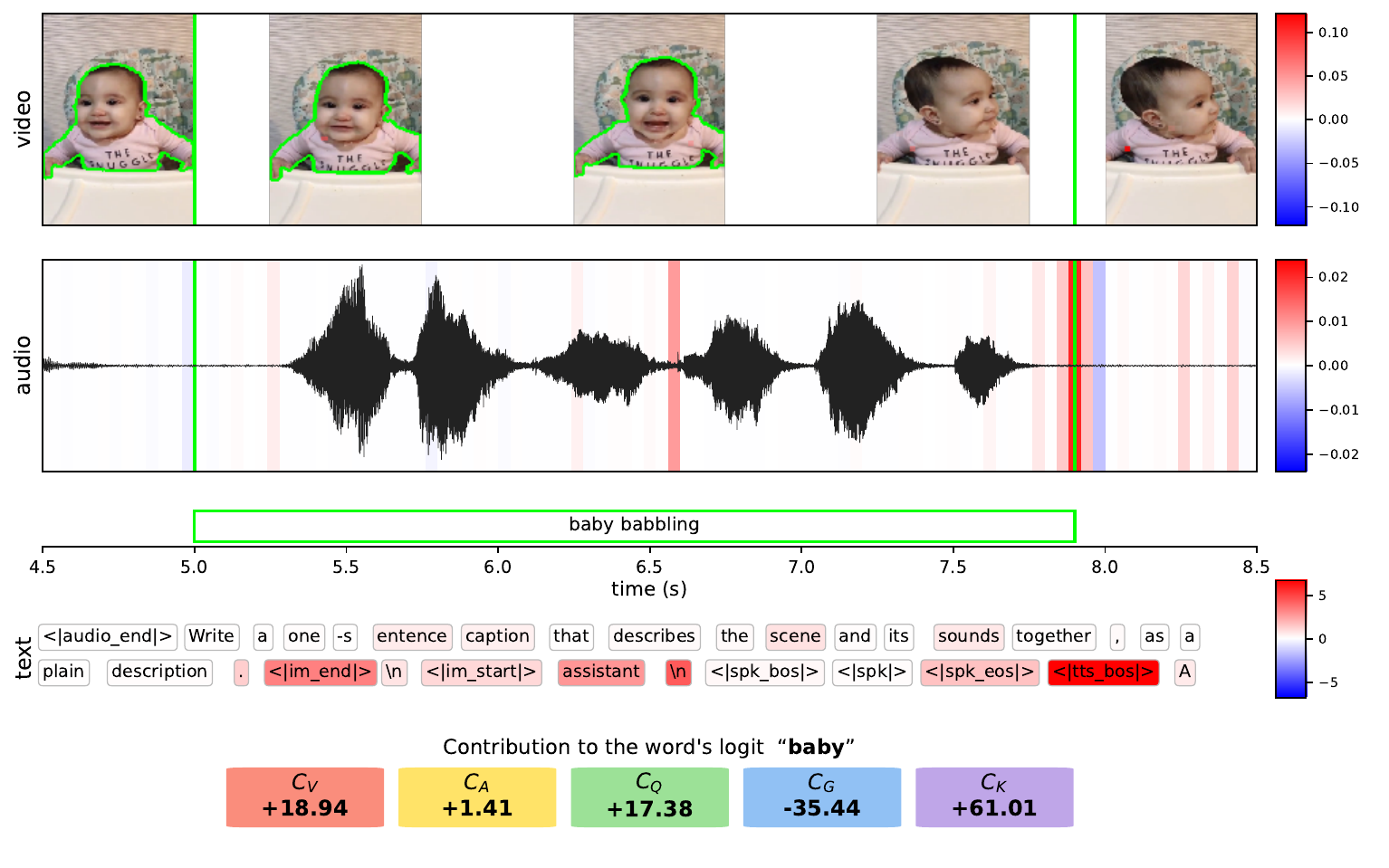}

    \par\vspace{7pt}

    \noindent
    \attrpanel{VL-AV}
        {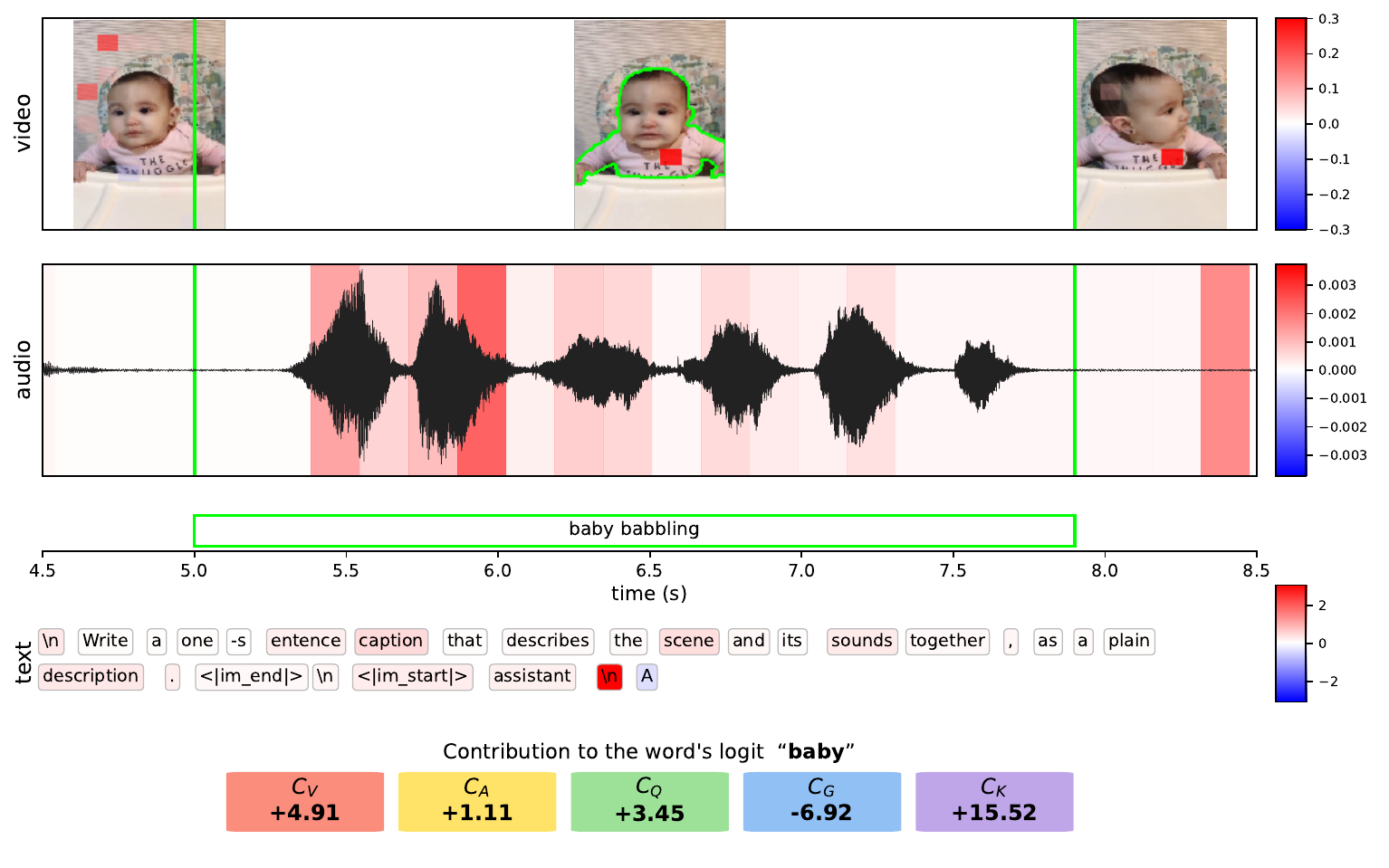}\hfill
    \attrpanel{Ola-7B}
        {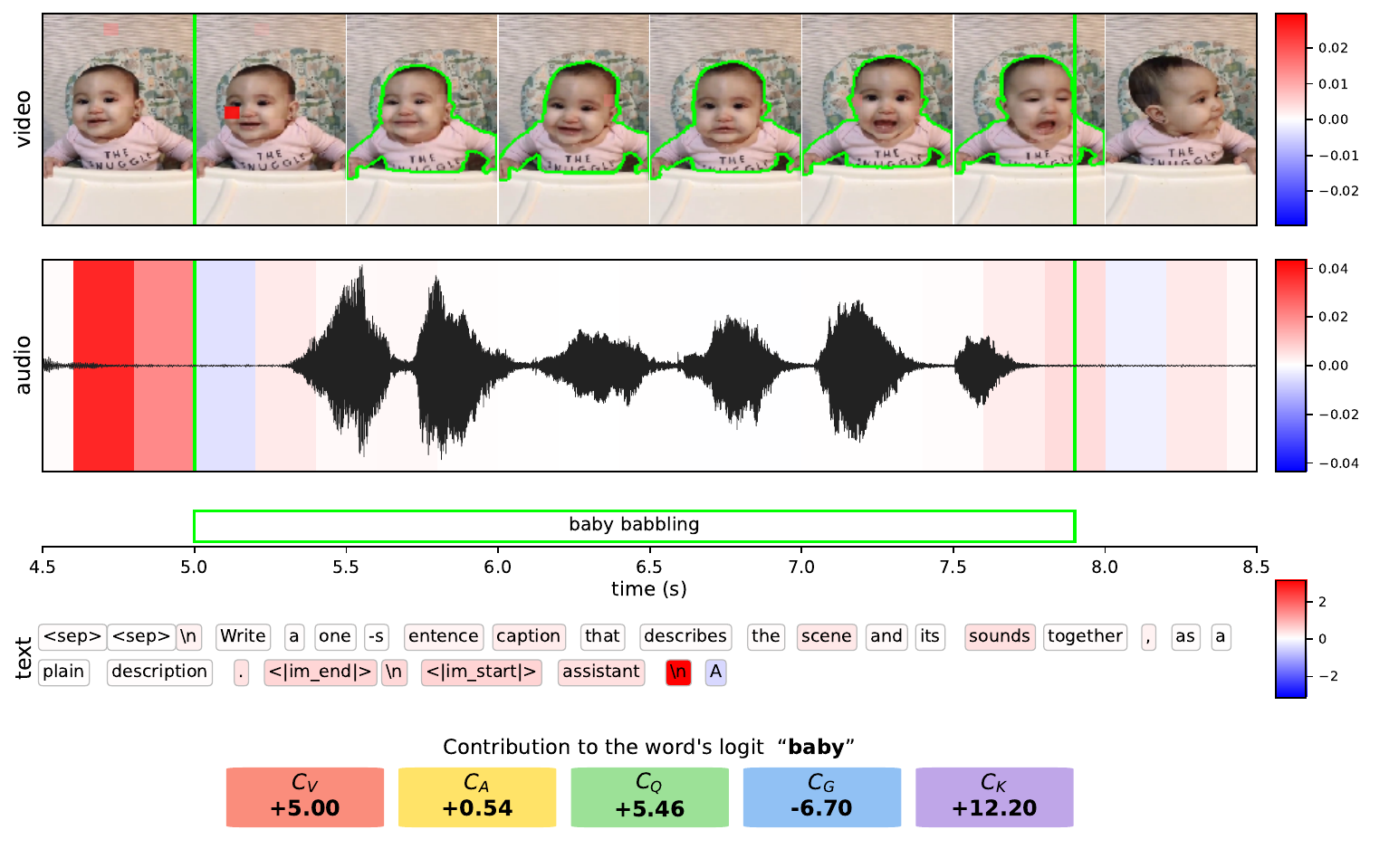}\hfill
    \attrpanel{AVF}
        {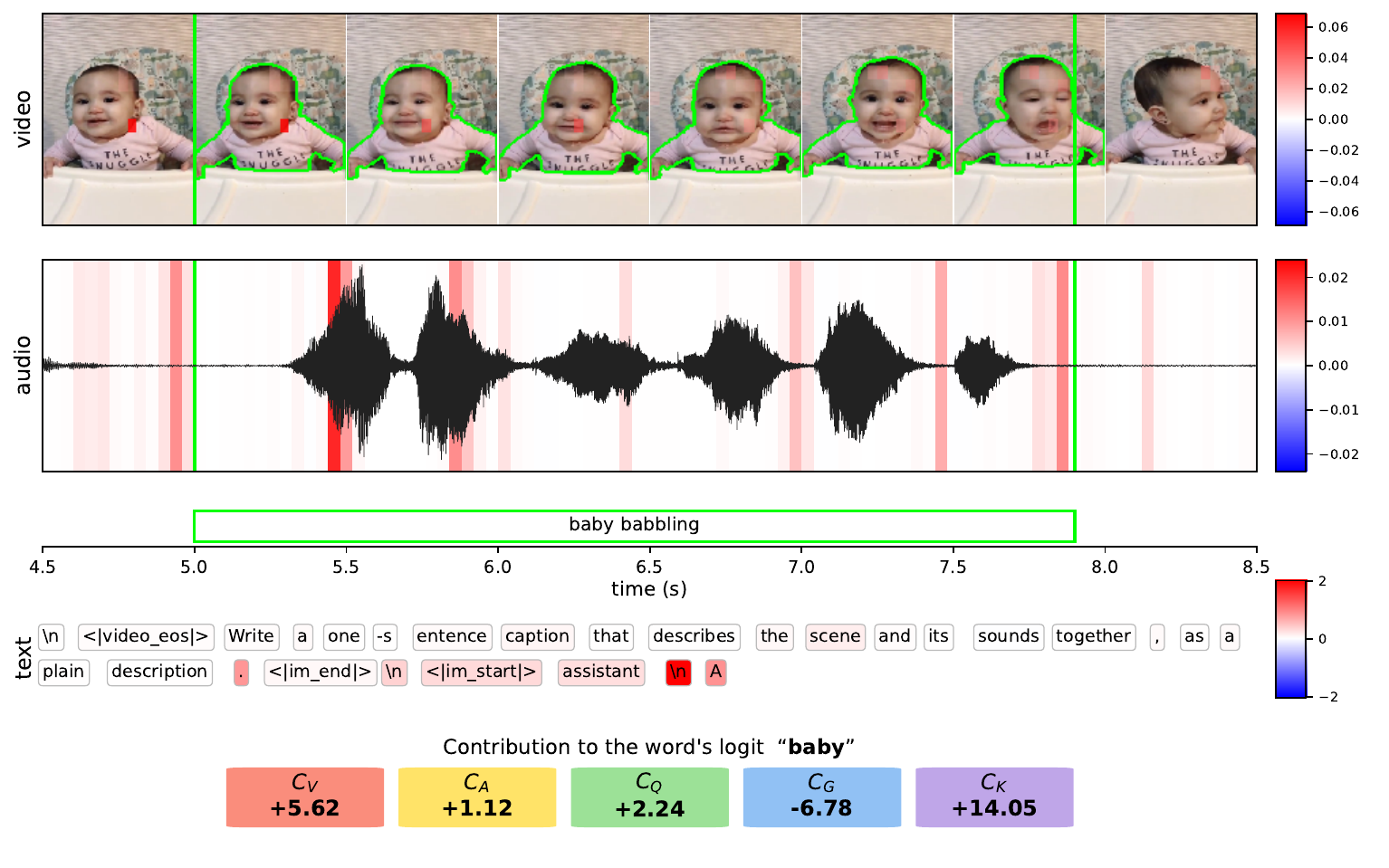}

    \par\vspace{7pt}

    \noindent
    \attrpanel{MCP 4.5}
        {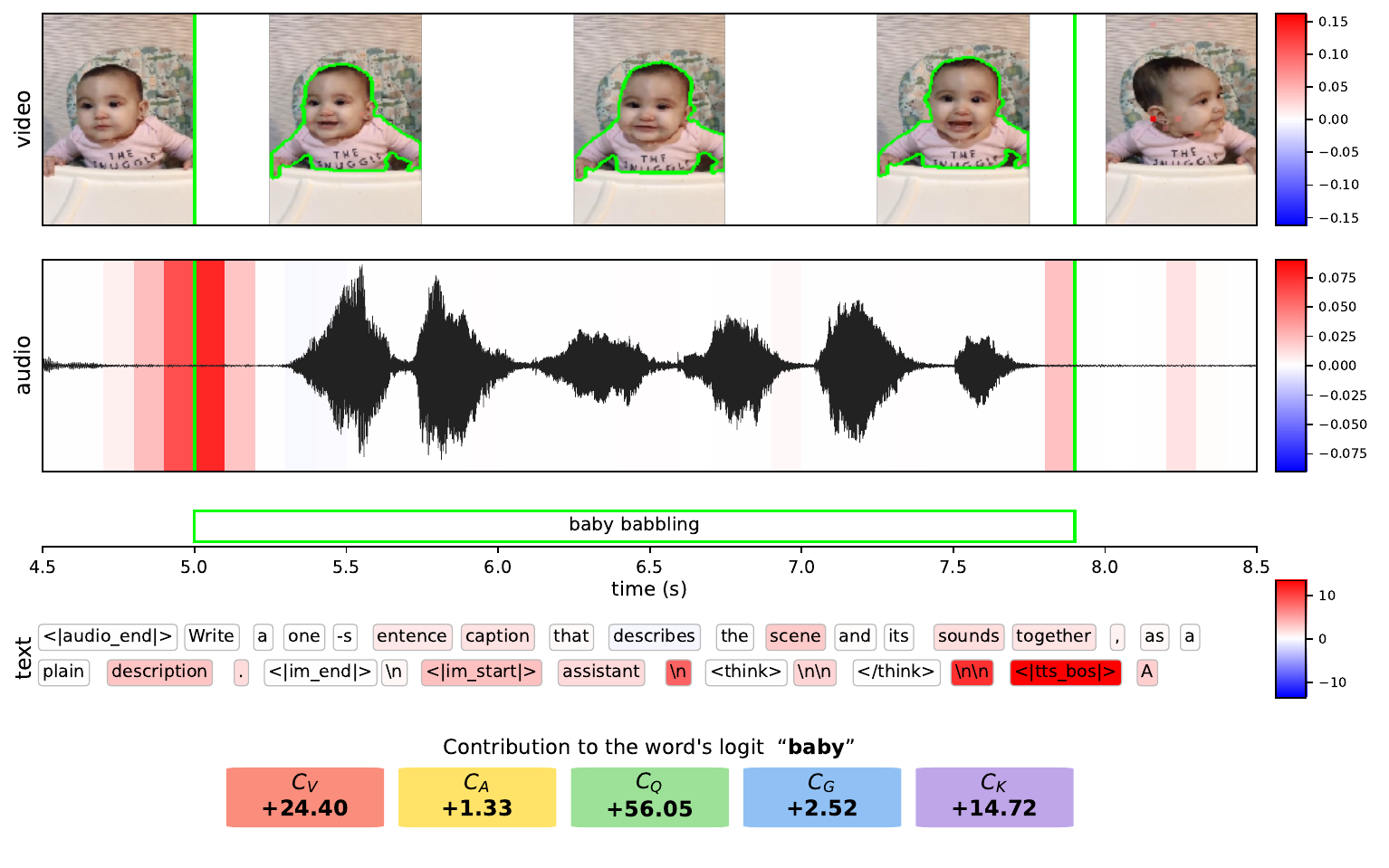}\hfill
    \attrpanel{vSAL 4B}
        {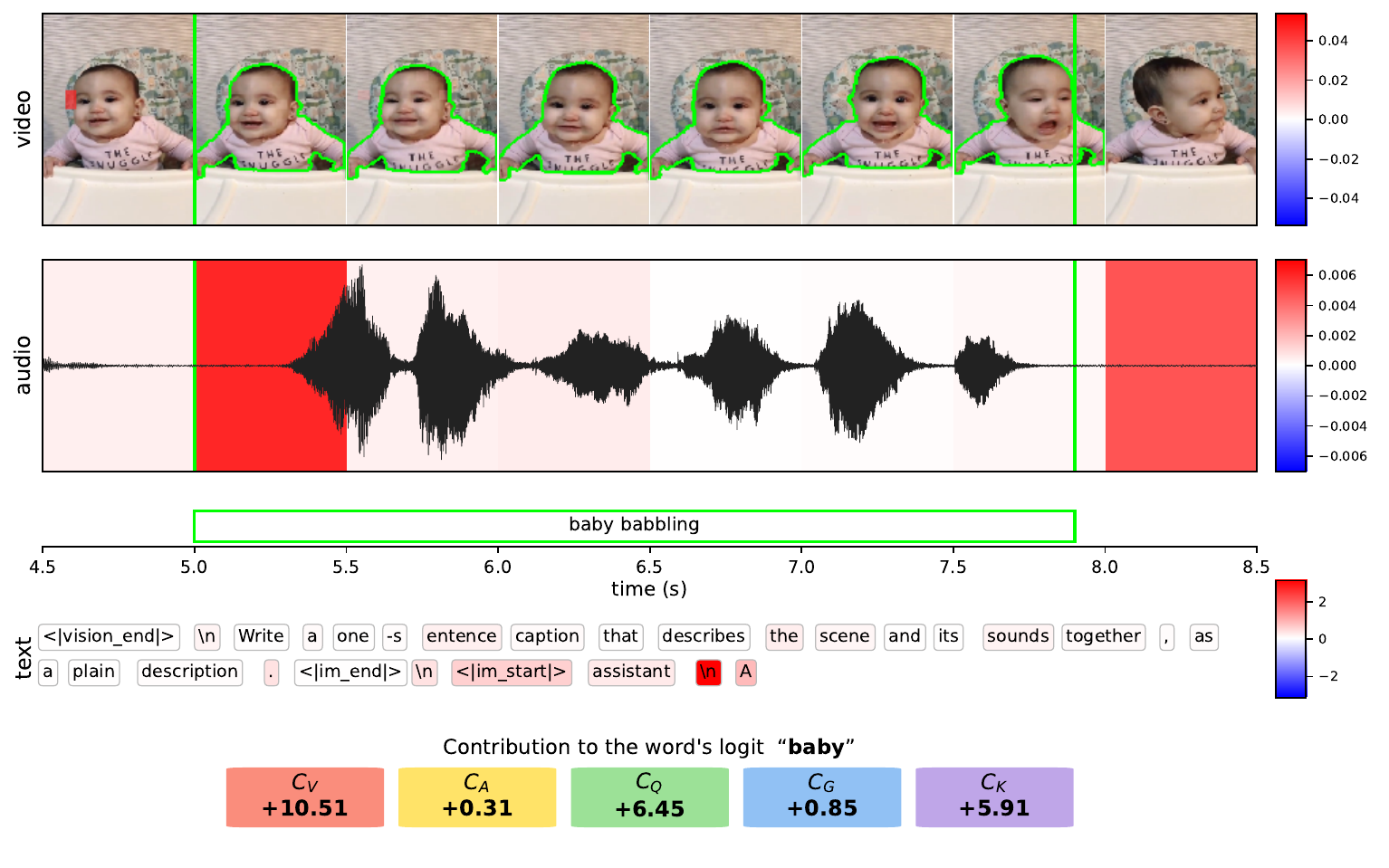}\hfill
    \attrpanel{vSAL 8B}
        {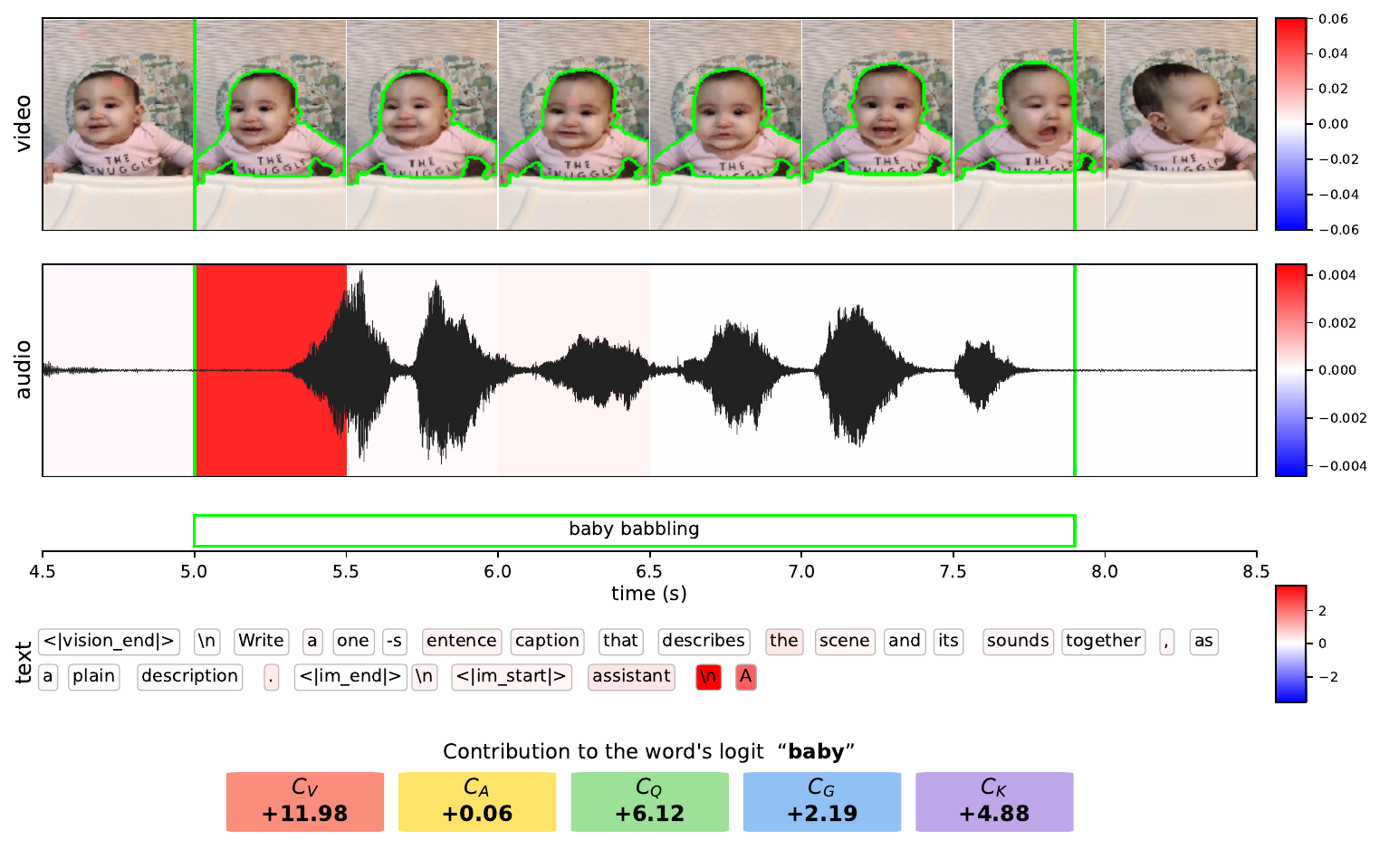}

    \caption{Qualitative MALT attributions across the nine
    evaluated AVLLMs. Each panel shows signed contributions
    from individual video, audio, and text positions to the
    logit of the selected generated token. Red indicates
    positive contributions, blue negative contributions,
    and white near-zero values. Colorbars report contributions
    in logit units, with independent scales for each modality
    and model; color intensities are therefore not directly
    comparable across heatmaps. Video frames differ across
    models due to native temporal sampling (Sec.~\ref{subsec:supp_models}).
    Green annotations indicate the spatial and temporal extent
    of the annotated sound event, where available. Bottom boxes
    report aggregated contributions from video ($C_V$),
    audio ($C_A$), prompt ($C_Q$), generated text ($C_G$),
    and non-source terms ($C_K$). The direct decoder-input
    term $E$ and numerical residual $\epsilon$ are omitted.}

    \label{fig:qualitative_all_models}
\end{figure*}

Figure~\ref{fig:qualitative_all_models} presents qualitative examples of MALT attributions across all nine AVLLMs, illustrating how individual generated event words receive contributions from video, audio, prompt, and previously generated text positions.

\subsection{Counterfactual intervention}
\Cref{fig:qualitative_counterfactual,fig:qualitative_counterfactual_b,fig:qualitative_counterfactual_c} illustrate how the attribution patterns change under modality obscuration, comparing the original input with audio-silenced and video-obscured conditions across the nine models.

\colorlet{attrborder}{black!35}

\colorlet{attrborder}{black!35}

\newcommand{\cfwideimg}[2]{%
    \begin{minipage}[t]{0.41\textwidth}
        \centering
        {\footnotesize\textbf{#1}}\par\vspace{1pt}
        {%
            \setlength{\fboxsep}{0pt}%
            \setlength{\fboxrule}{0.15pt}%
            \color{attrborder}%
            \fbox{%
                \includegraphics[
                    width=\dimexpr\linewidth-2\fboxrule\relax
                ]{#2}%
            }%
        }%
    \end{minipage}%
}

\newcommand{\cfwidegroup}[4]{%
    \par\noindent\centering
    \makebox[\textwidth][c]{%
        \begin{tabular}{@{}c@{\hspace{6pt}}c@{}}

            \adjustbox{valign=c}{%
                \rotatebox[origin=c]{90}{\textbf{#1}}%
            }
            &
            \adjustbox{valign=c}{%
                \begin{tabular}{@{}c@{}}

                    \makebox[\dimexpr0.82\textwidth+6pt\relax][c]{%
                        \cfwideimg{Original}{#2}%
                    }
                    \\[1pt]

                    \makebox[\dimexpr0.82\textwidth+6pt\relax][c]{%
                        \cfwideimg{Video obscuration}{#3}%
                        \hspace{6pt}%
                        \cfwideimg{Audio silence}{#4}%
                    }
                \end{tabular}%
            }
        \end{tabular}%
    }%
    \par\vspace{3pt}
}

\begin{figure*}[p]
    \centering

    \cfwidegroup{Omni 3B}
        {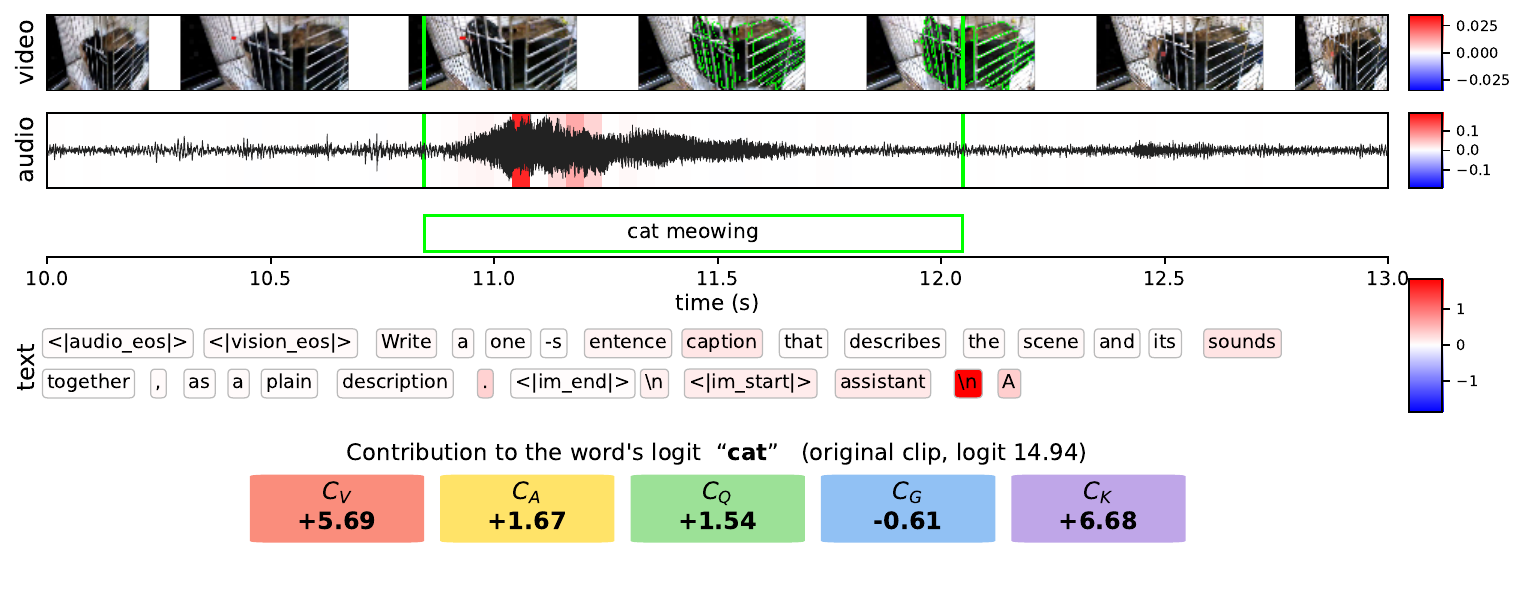}
        {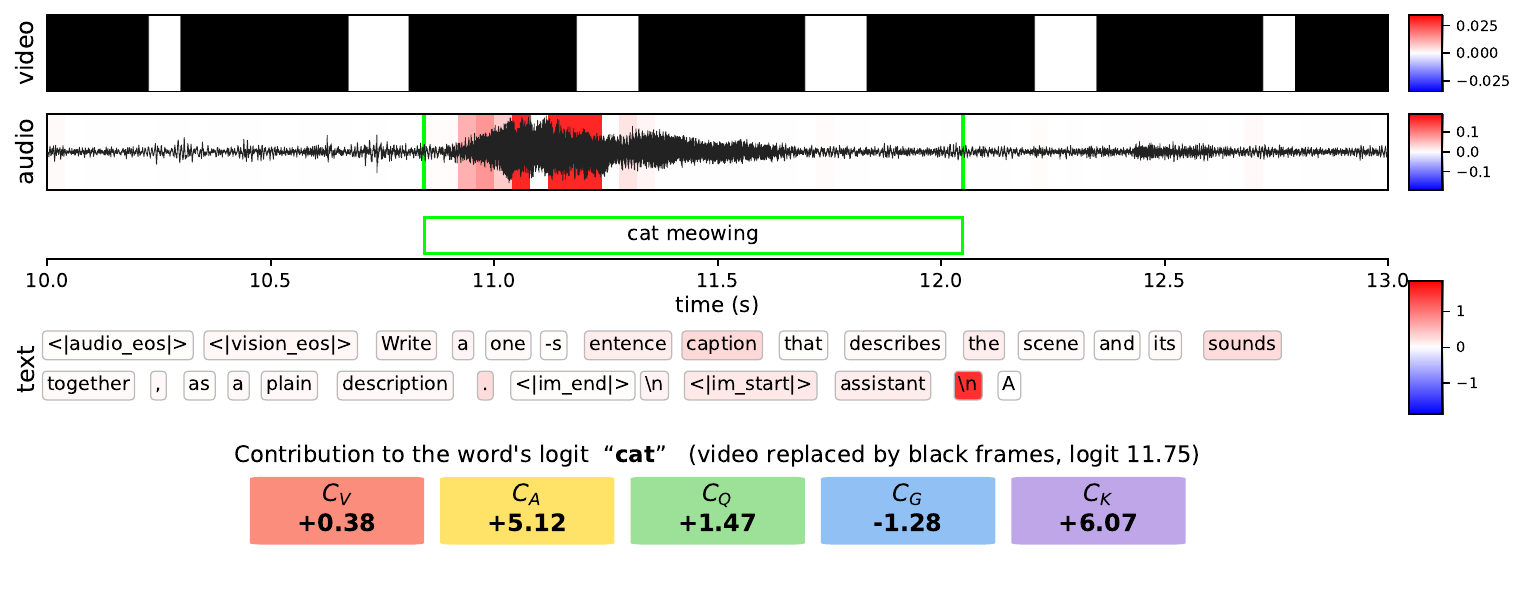}
        {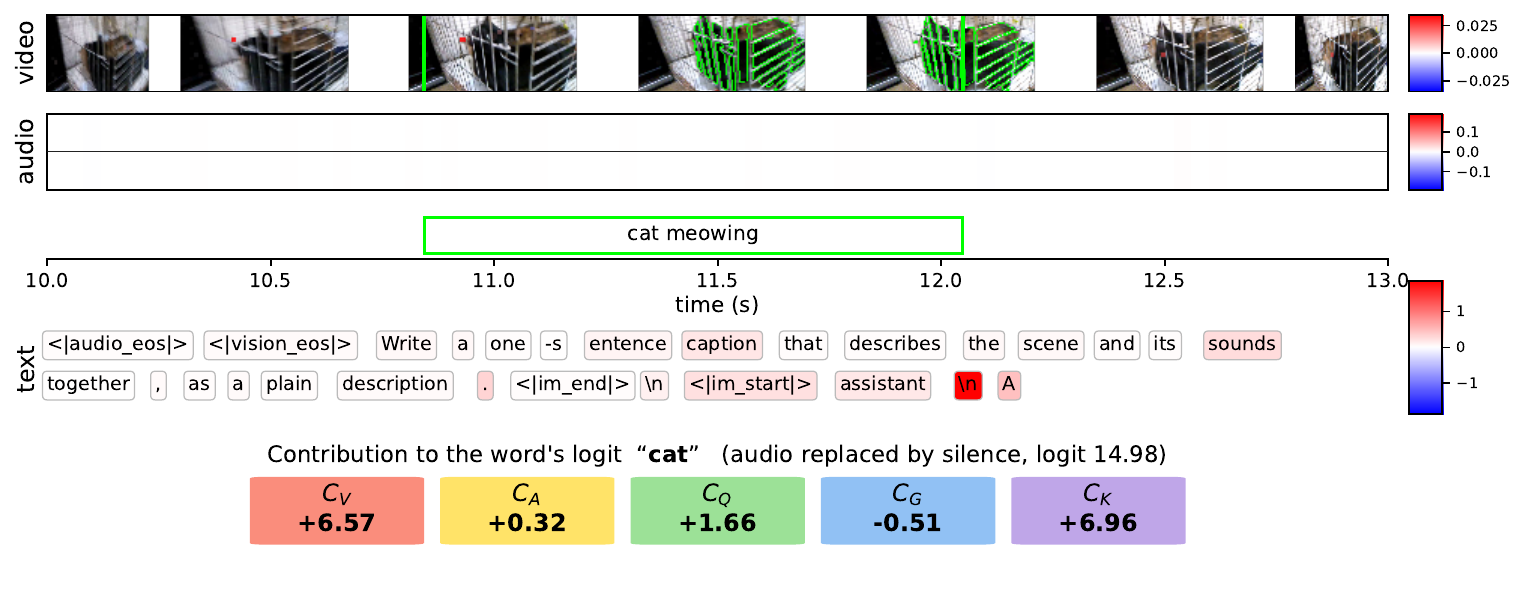}

    \cfwidegroup{Omni 7B}
        {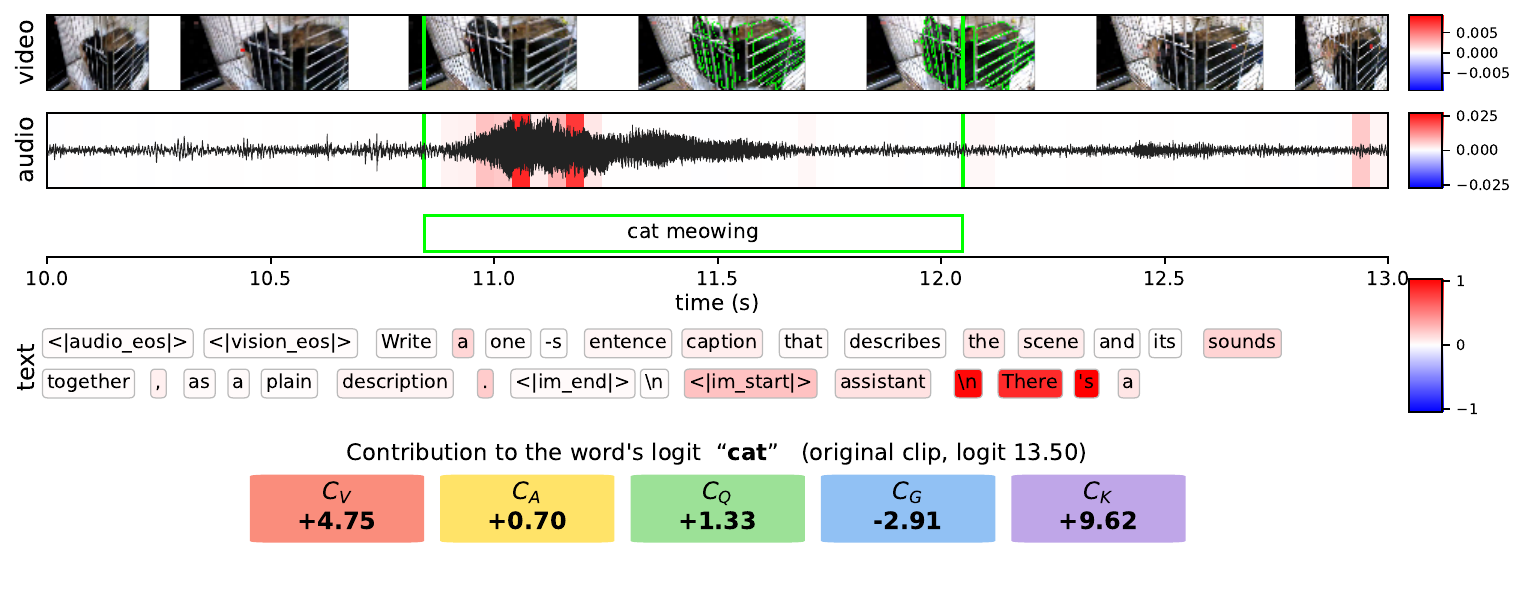}
        {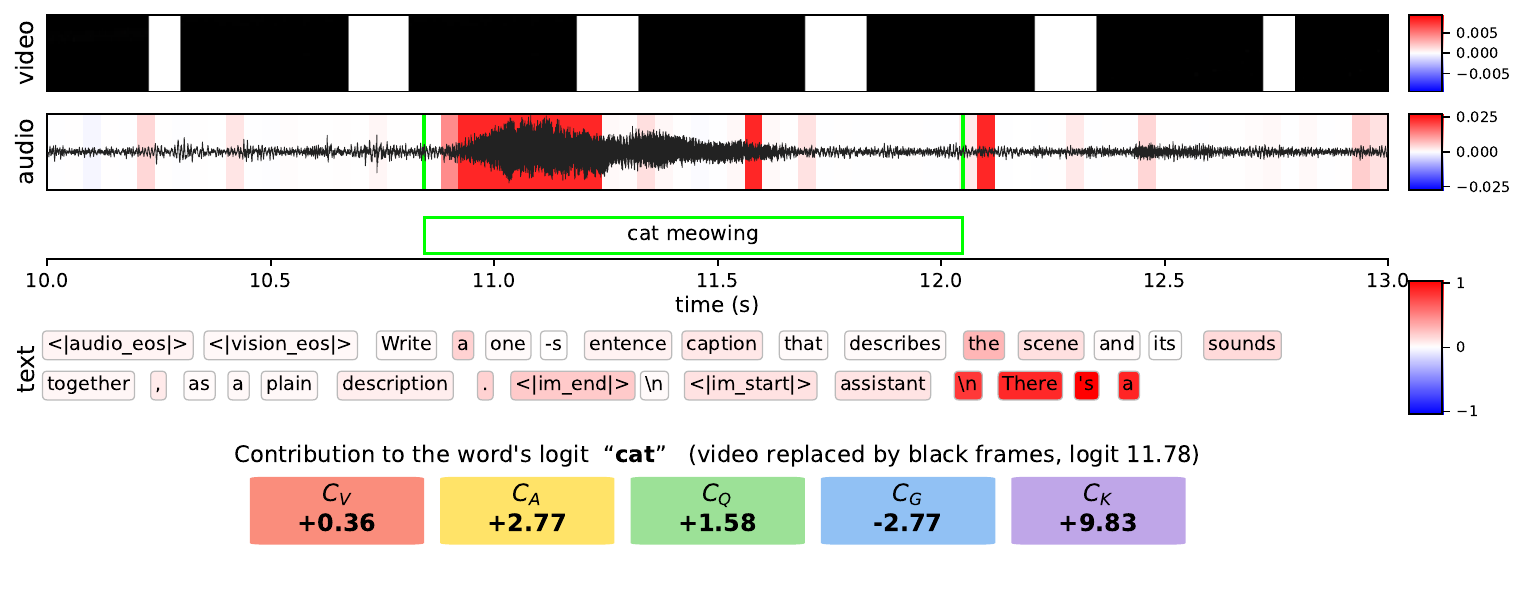}
        {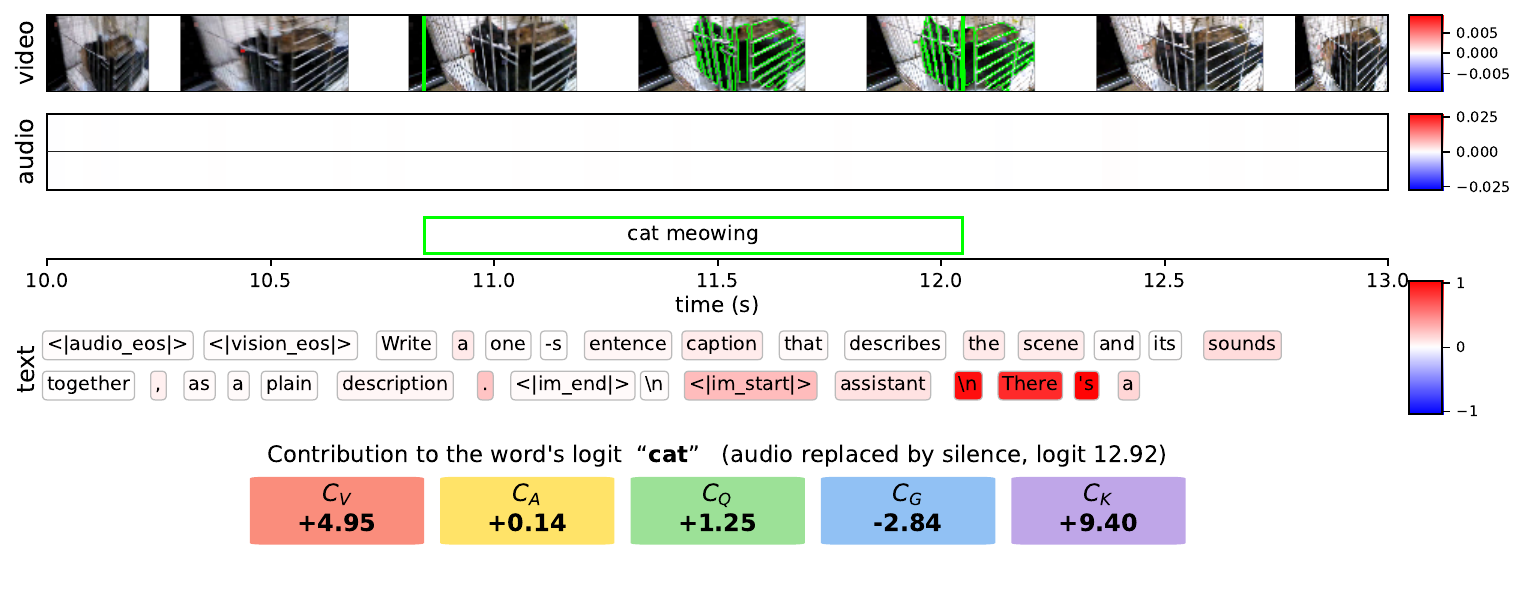}

    \cfwidegroup{MCP 2.6}
        {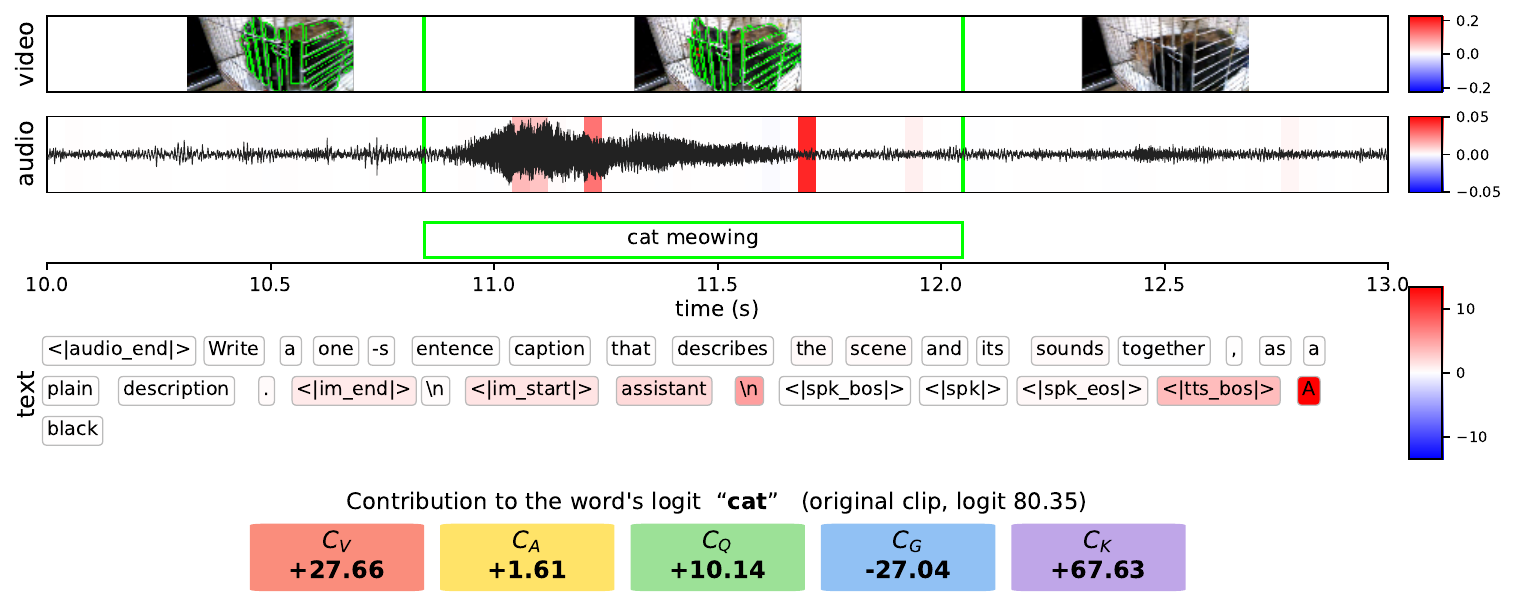}
        {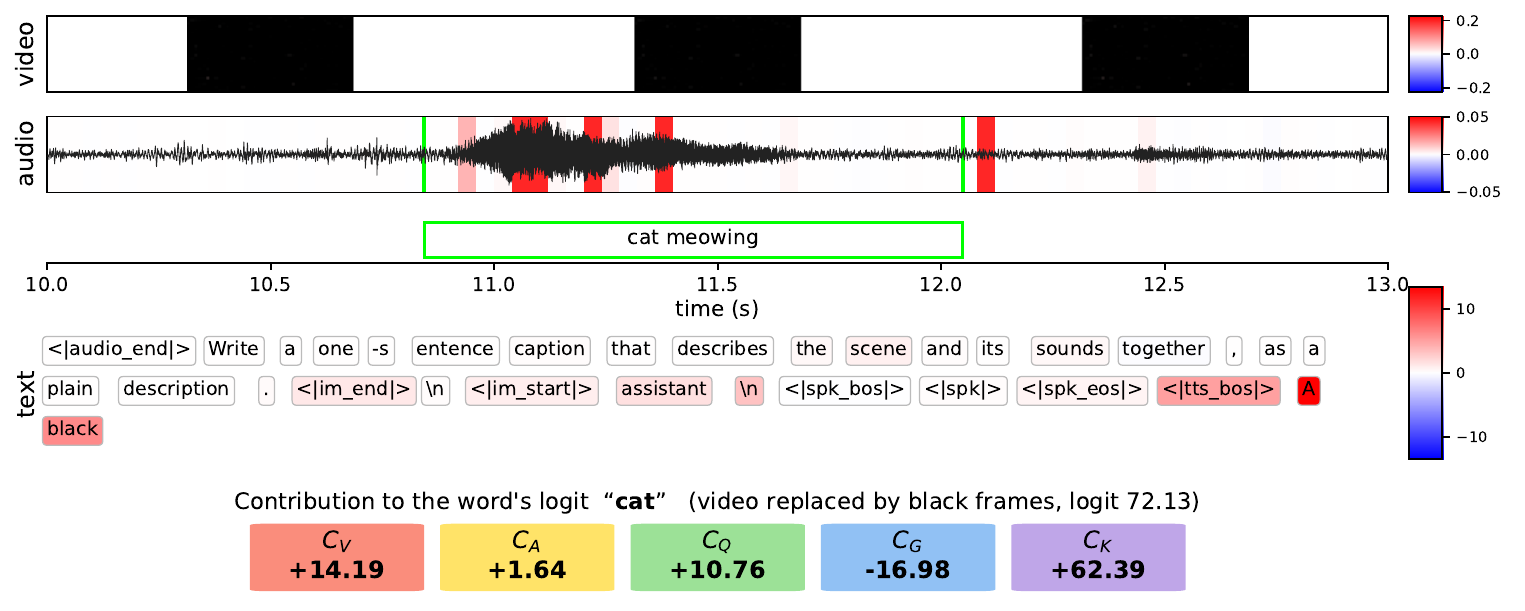}
        {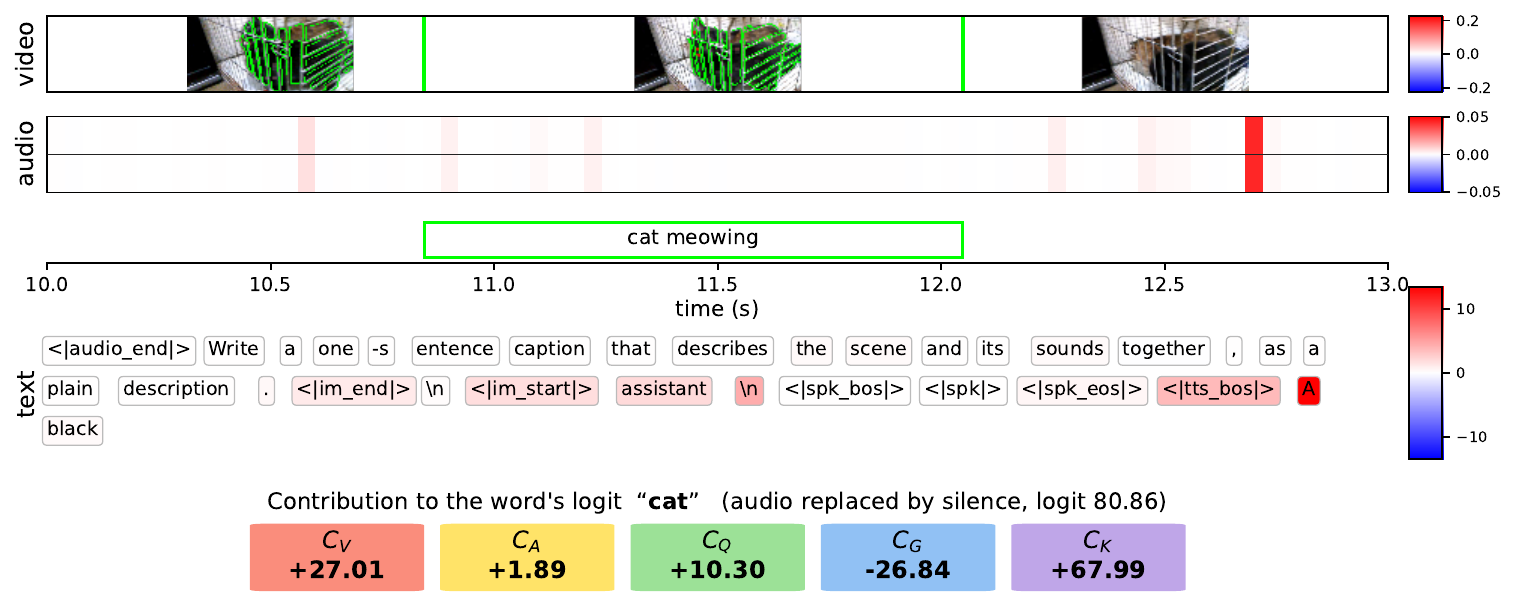}

    \caption{Qualitative MALT attributions under counterfactual modality obscuration (Omni 3B, Omni 7B, and MCP 2.6). Each model is shown under three conditions: original input, video replaced by black frames, and audio replaced by digital silence. The generated sequence and target token are held fixed under teacher forcing. Heatmaps show signed per-position contributions in logit units (red: positive; blue: negative; white: near zero). Color scales are independent across models and modalities. Green annotations mark the annotated sound event. Bottom boxes report aggregated contributions $C_V$, $C_A$, $C_Q$, $C_G$, and $C_K$. The direct term $E$ and numerical residual $\epsilon$ are omitted.}
    \label{fig:qualitative_counterfactual}
\end{figure*}

\begin{figure*}[p]
    \centering

    \cfwidegroup{VL-AV}
        {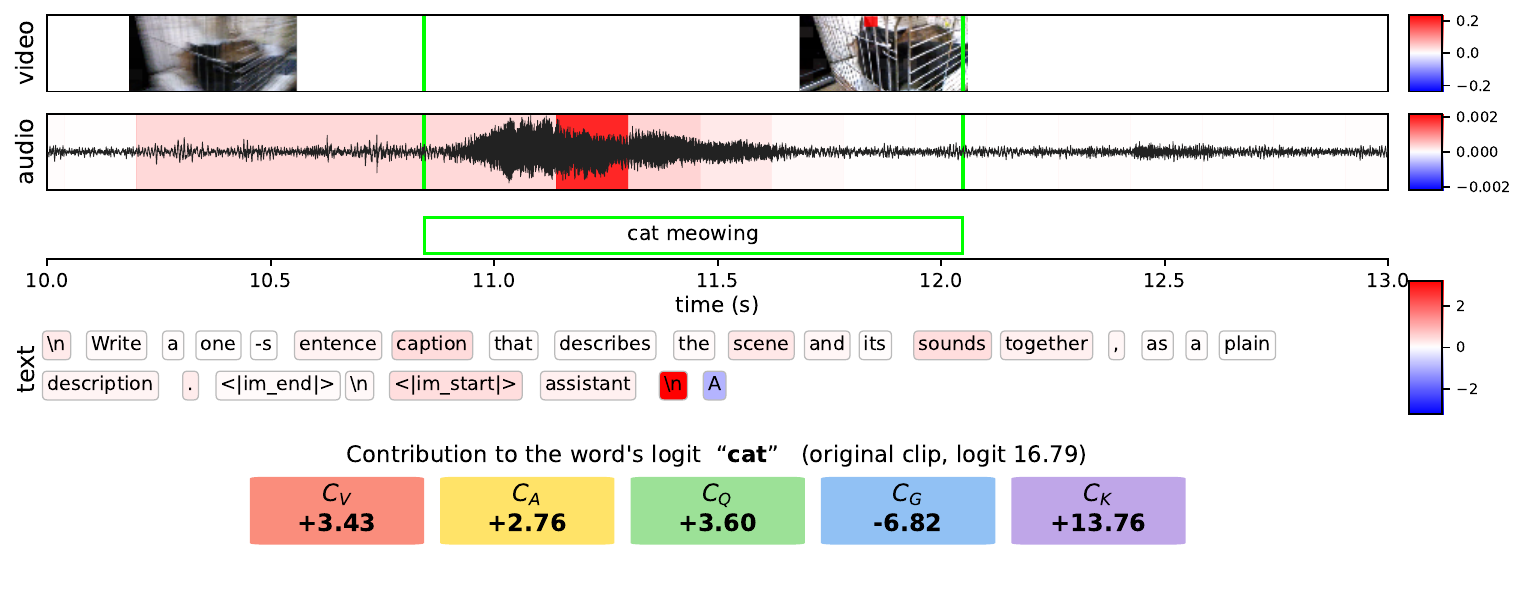}
        {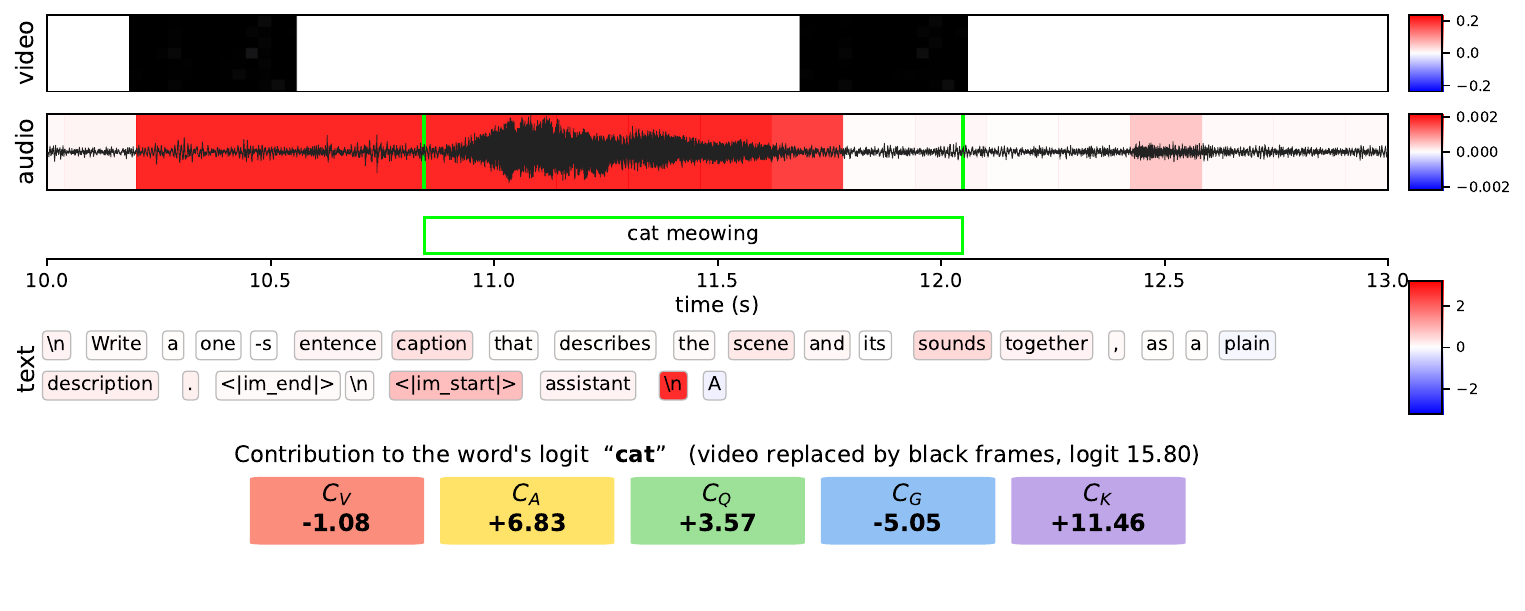}
        {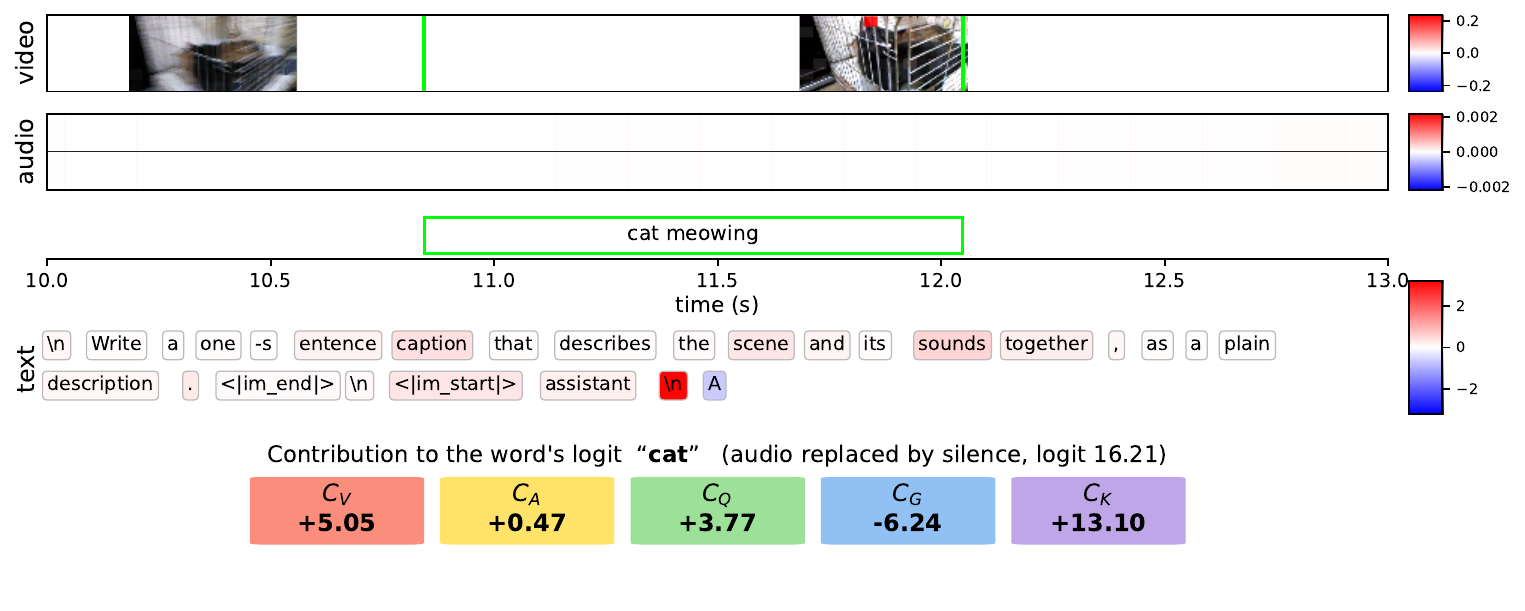}

    \cfwidegroup{Ola-7B}
        {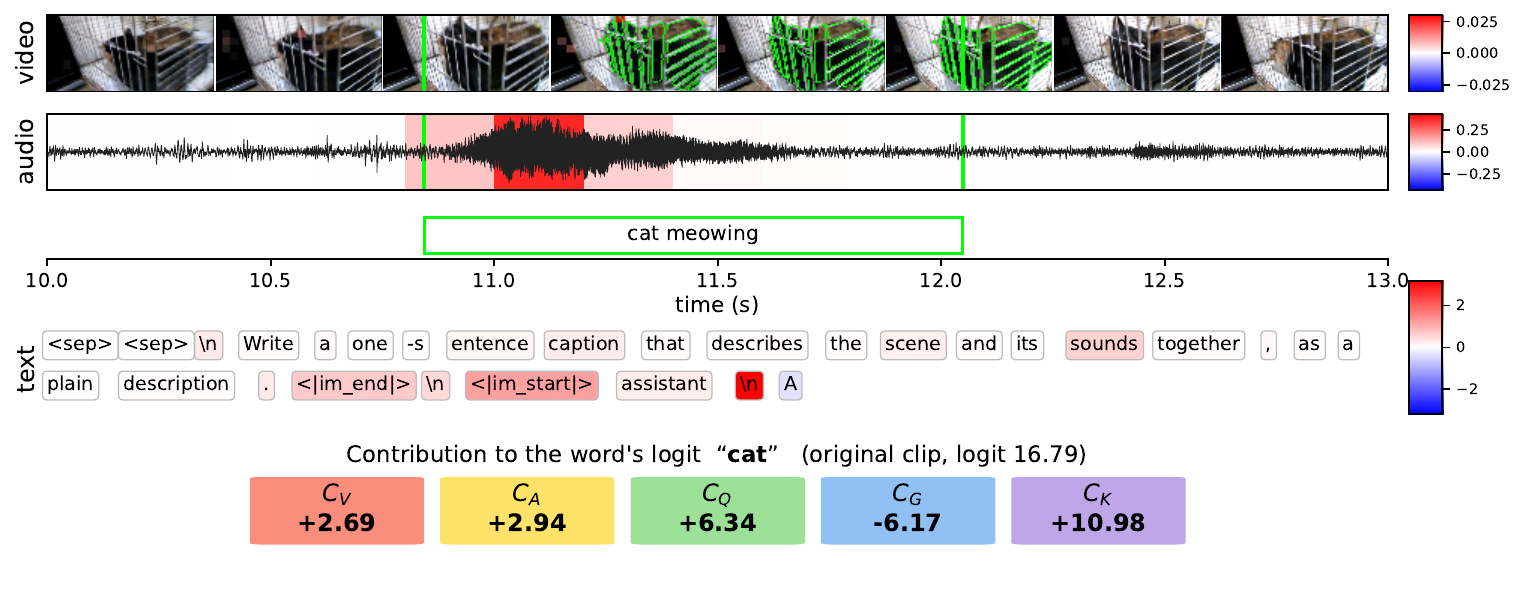}
        {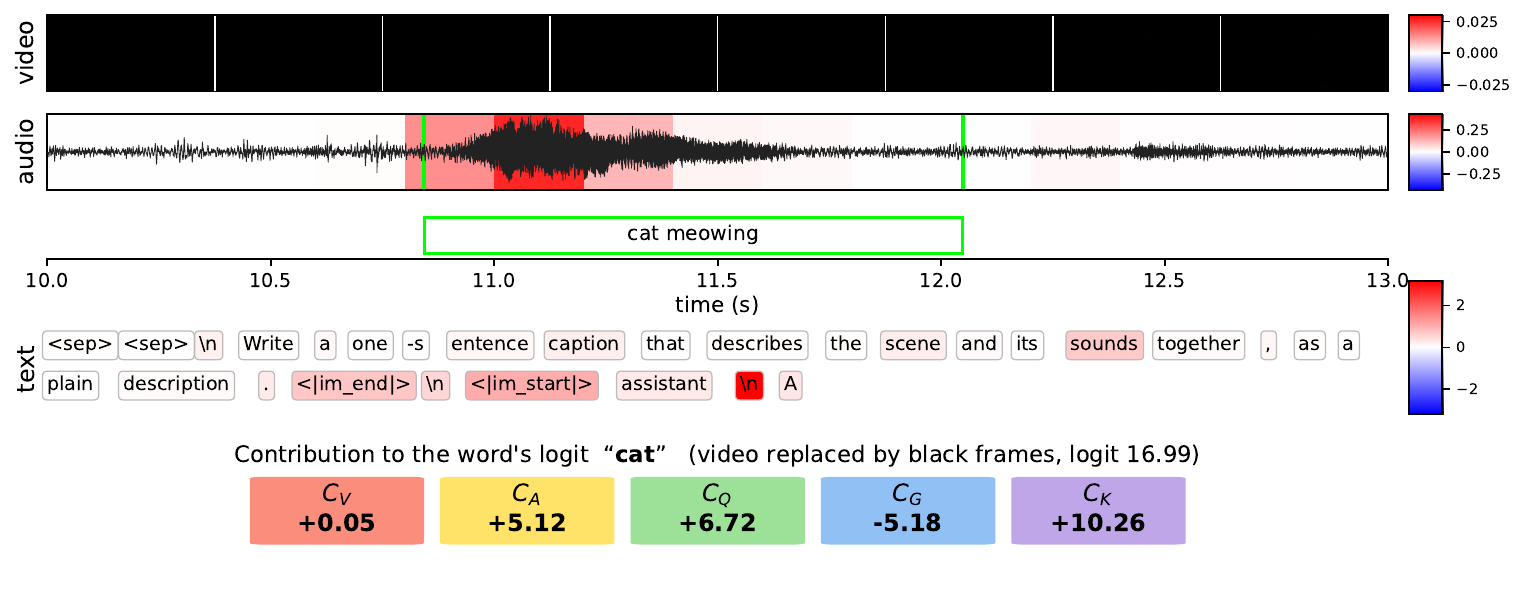}
        {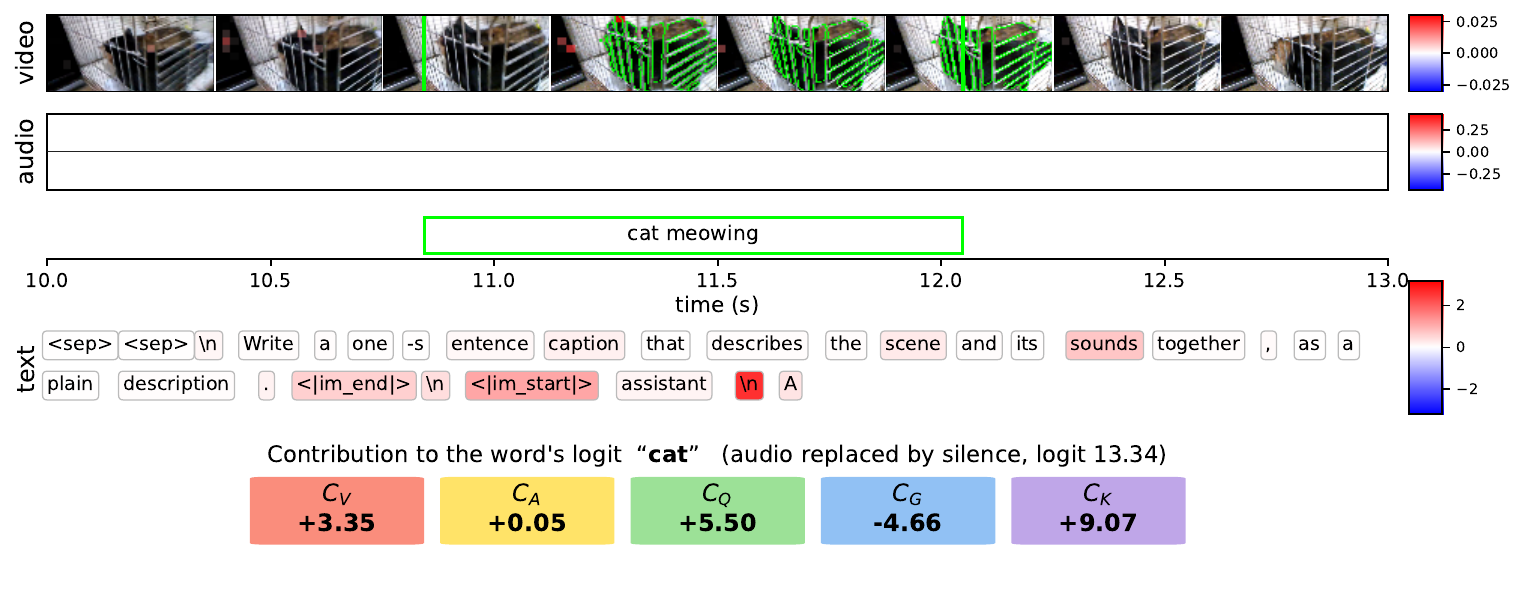}

    \cfwidegroup{AVF}
        {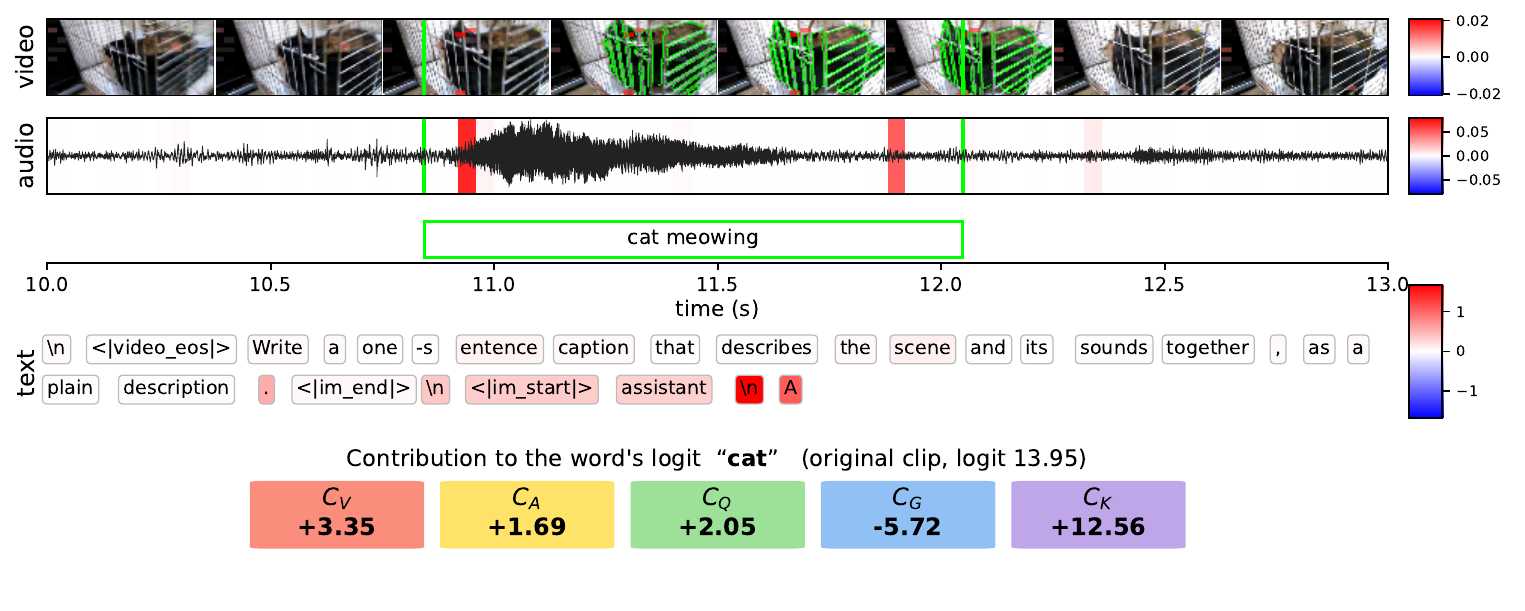}
        {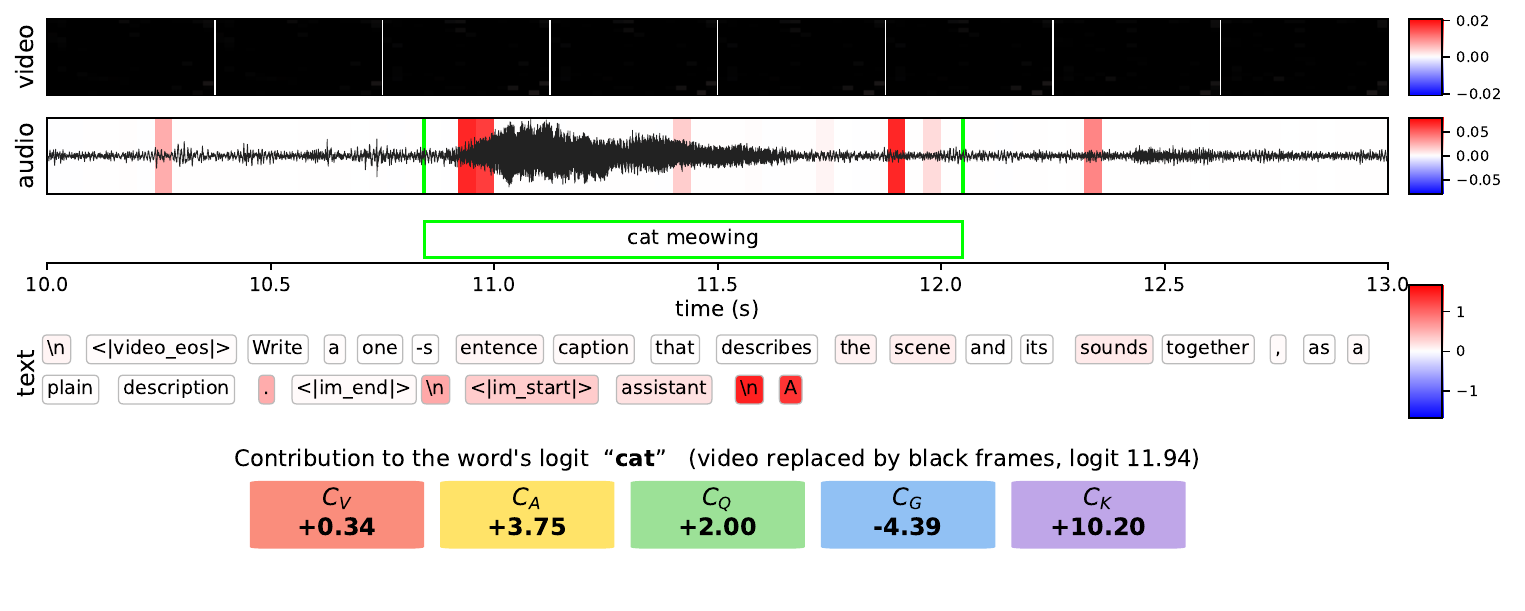}
        {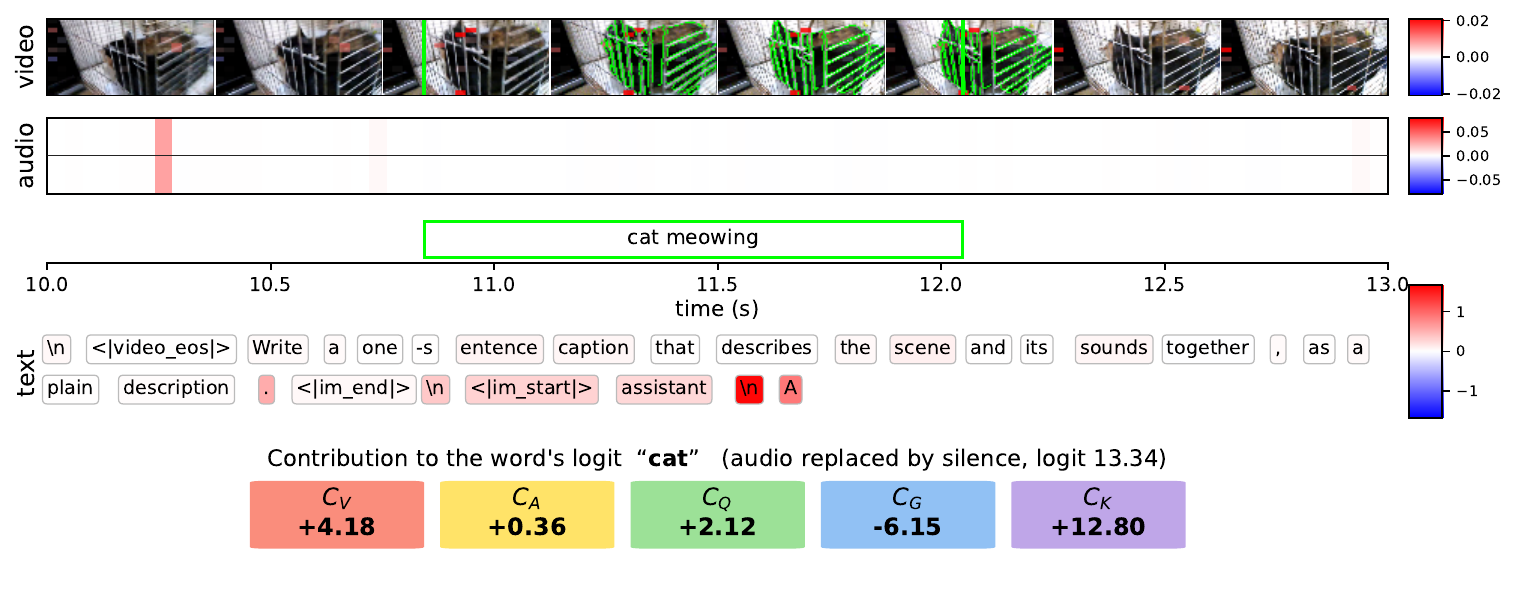}

    \caption{Qualitative MALT attributions under counterfactual modality obscuration for VL-AV, Ola-7B, and AVF. The three conditions use the same generated sequence and target token for each model. Heatmaps show signed per-position contributions, while bottom boxes report aggregated source and non-source contributions. Color conventions follow Fig.~\ref{fig:qualitative_counterfactual}.}
    \label{fig:qualitative_counterfactual_b}
\end{figure*}

\begin{figure*}[p]
    \centering

    \cfwidegroup{MCP 4.5}
        {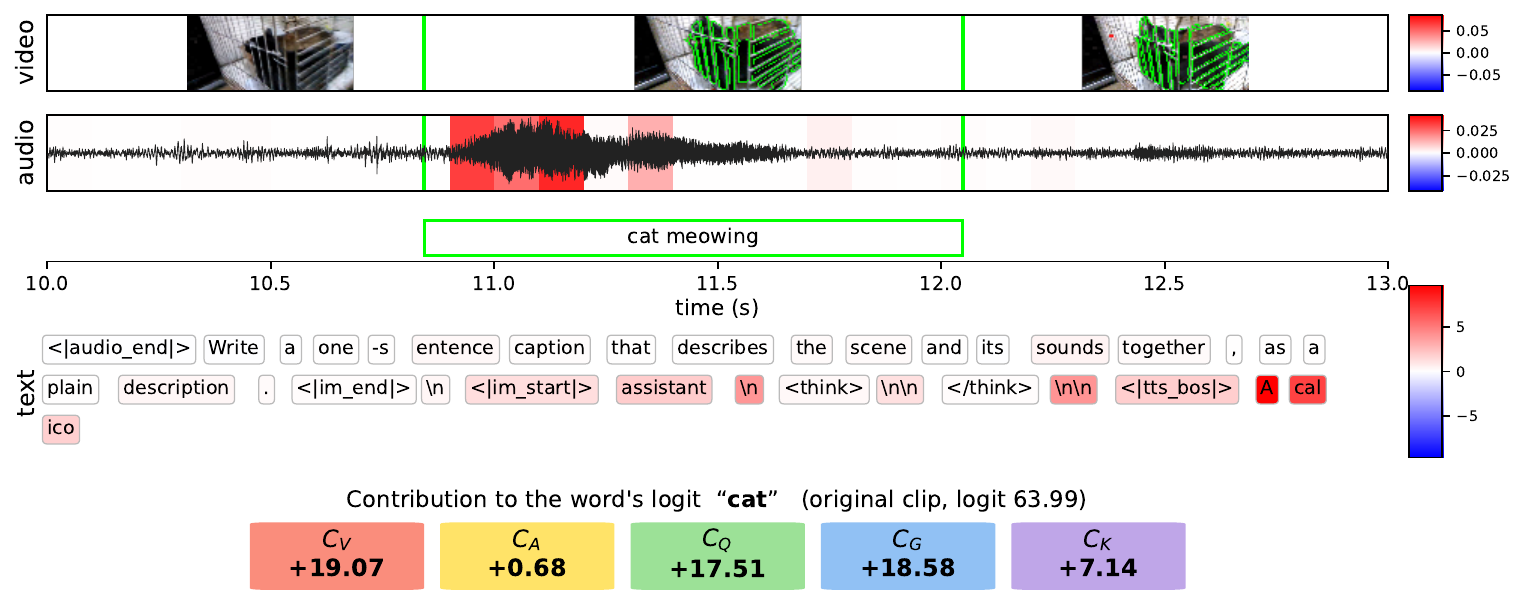}
        {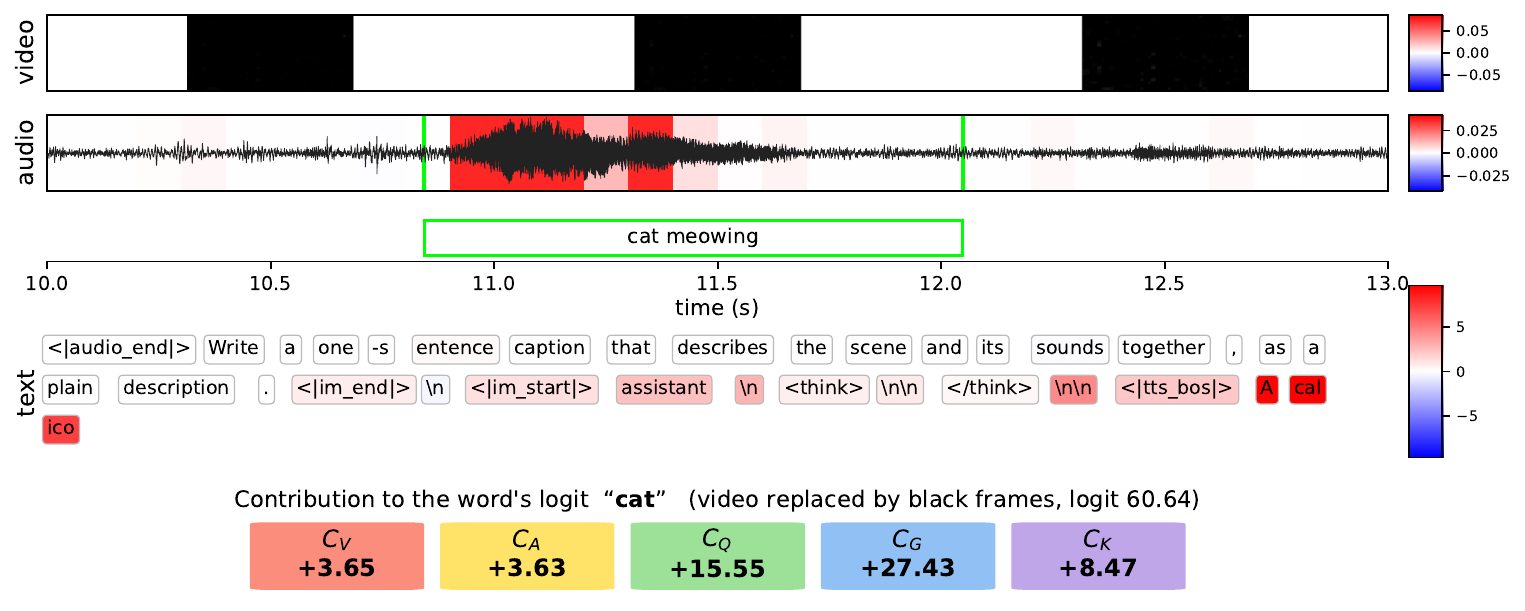}
        {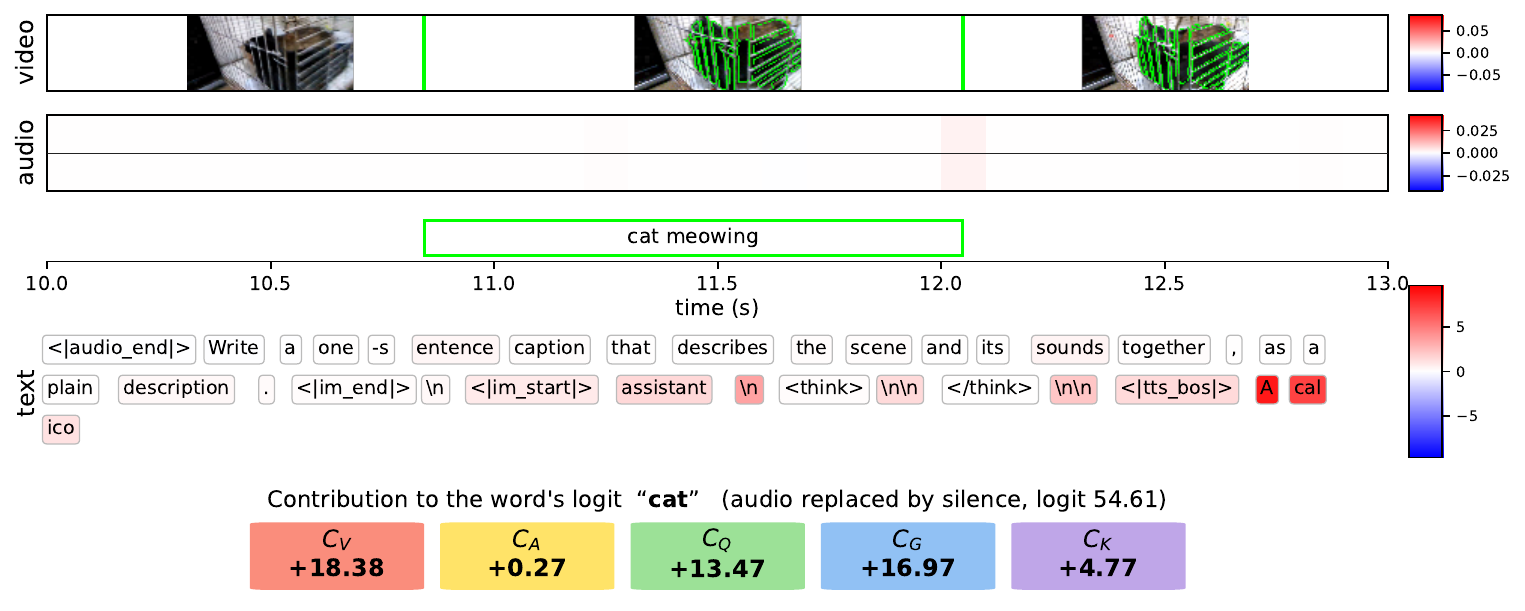}

    \cfwidegroup{vSAL 4B}
        {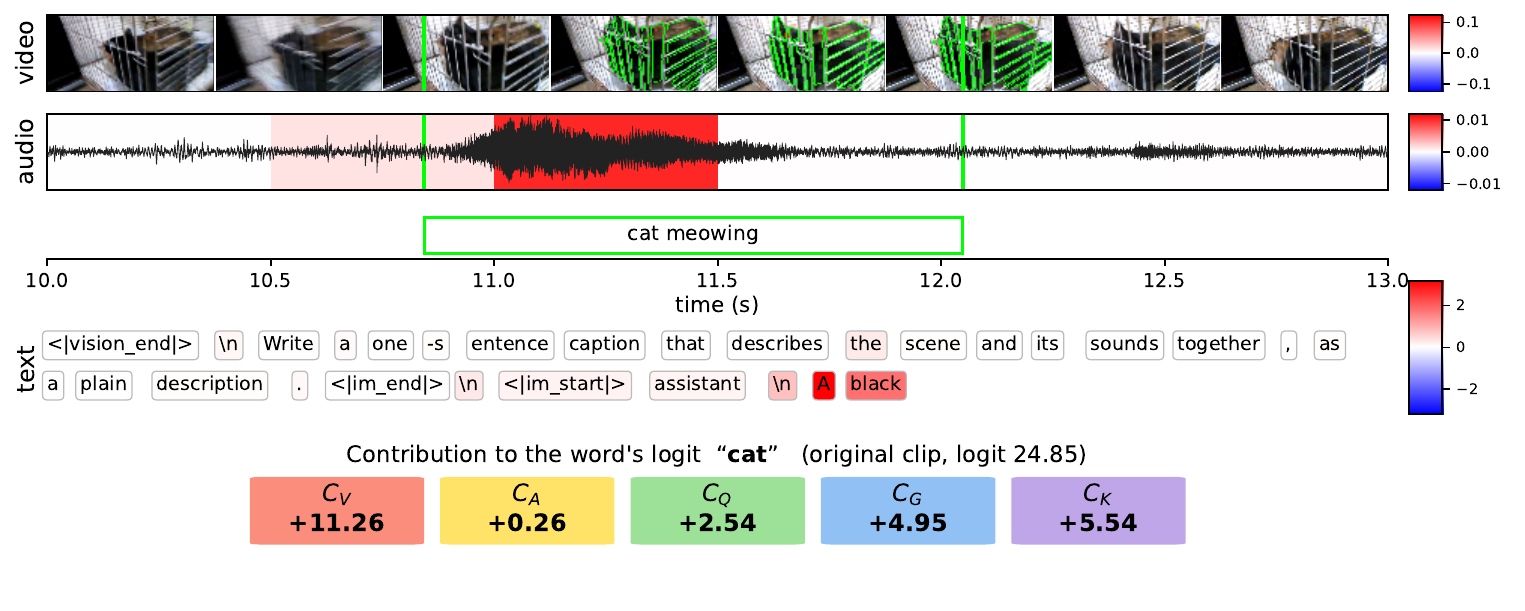}
        {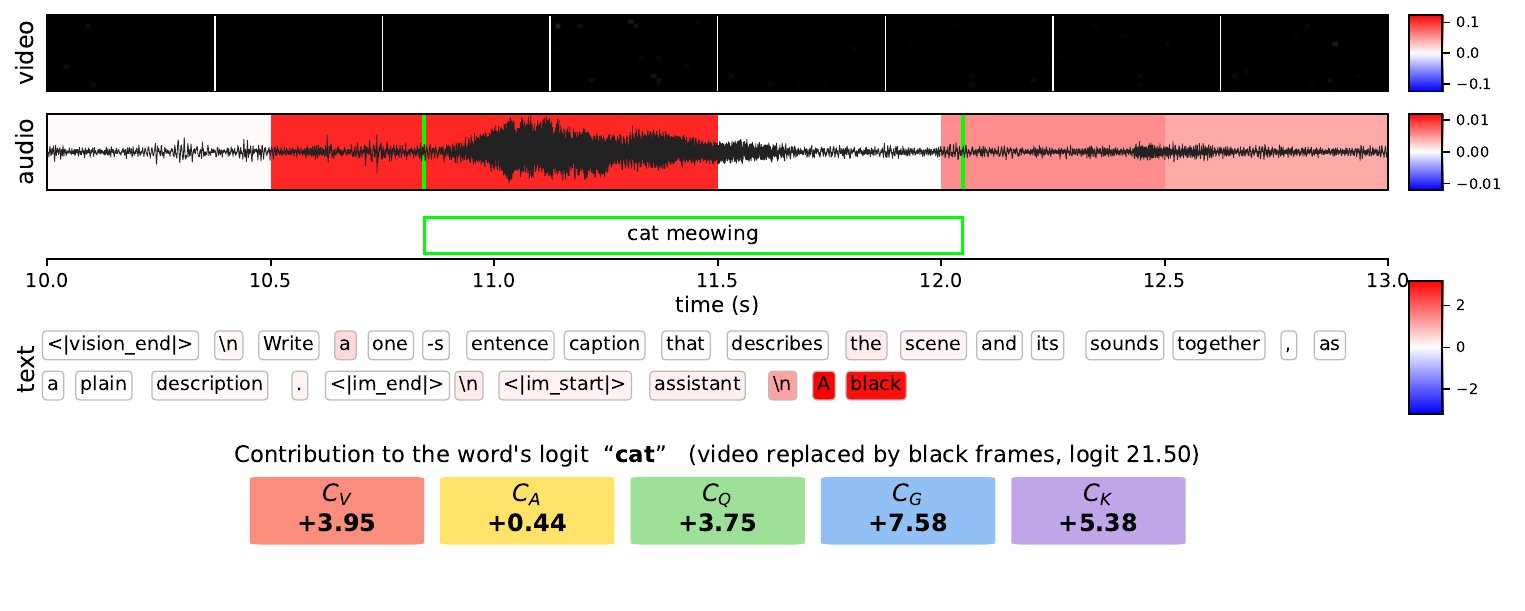}
        {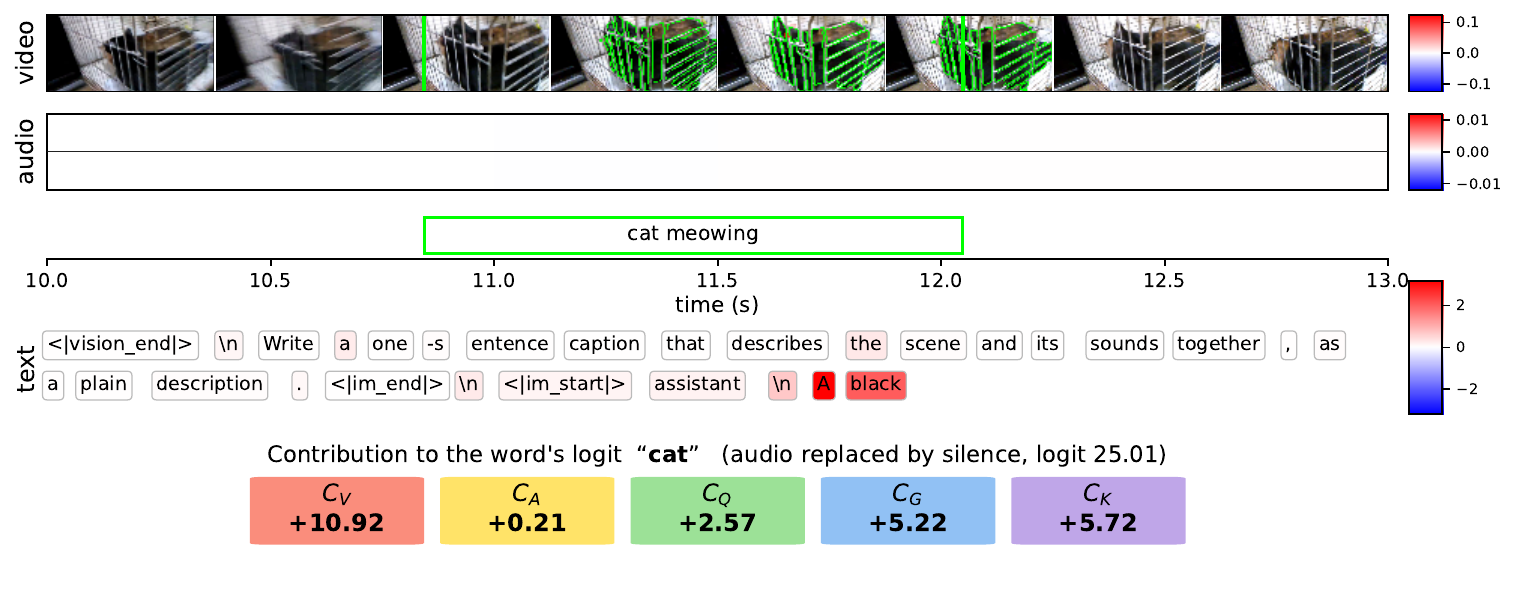}

    \cfwidegroup{vSAL 8B}
        {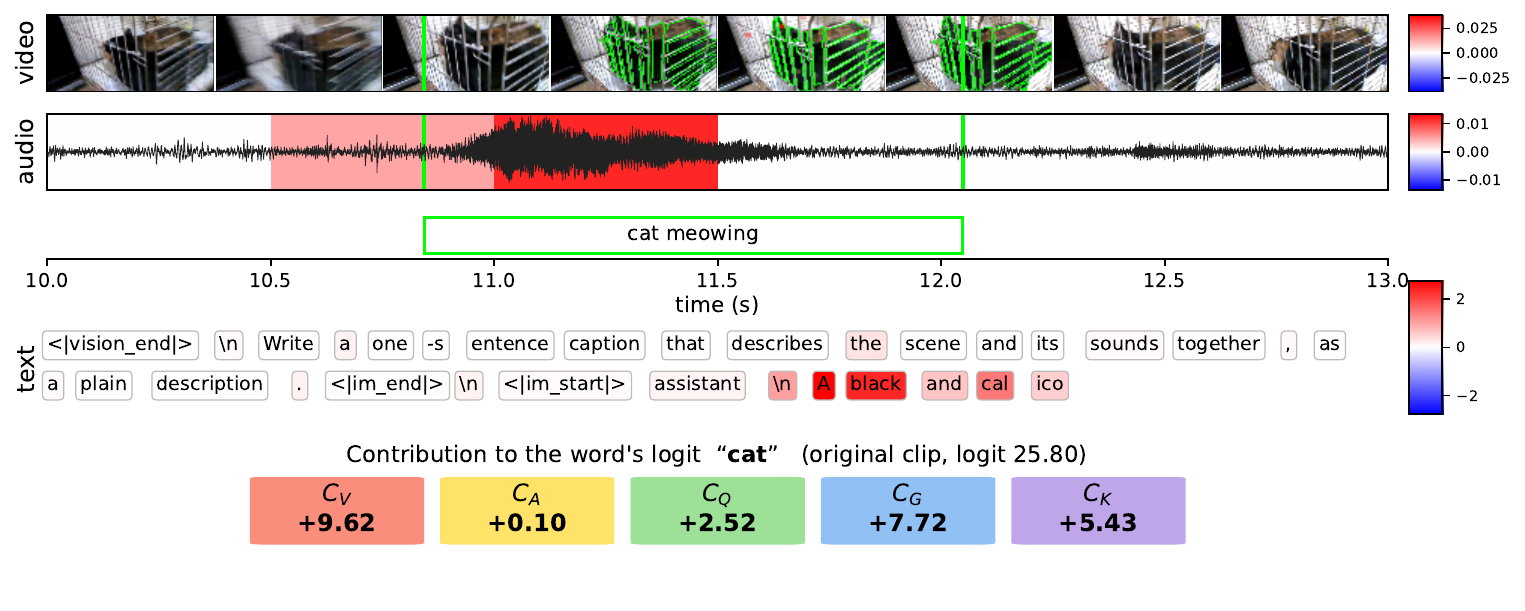}
        {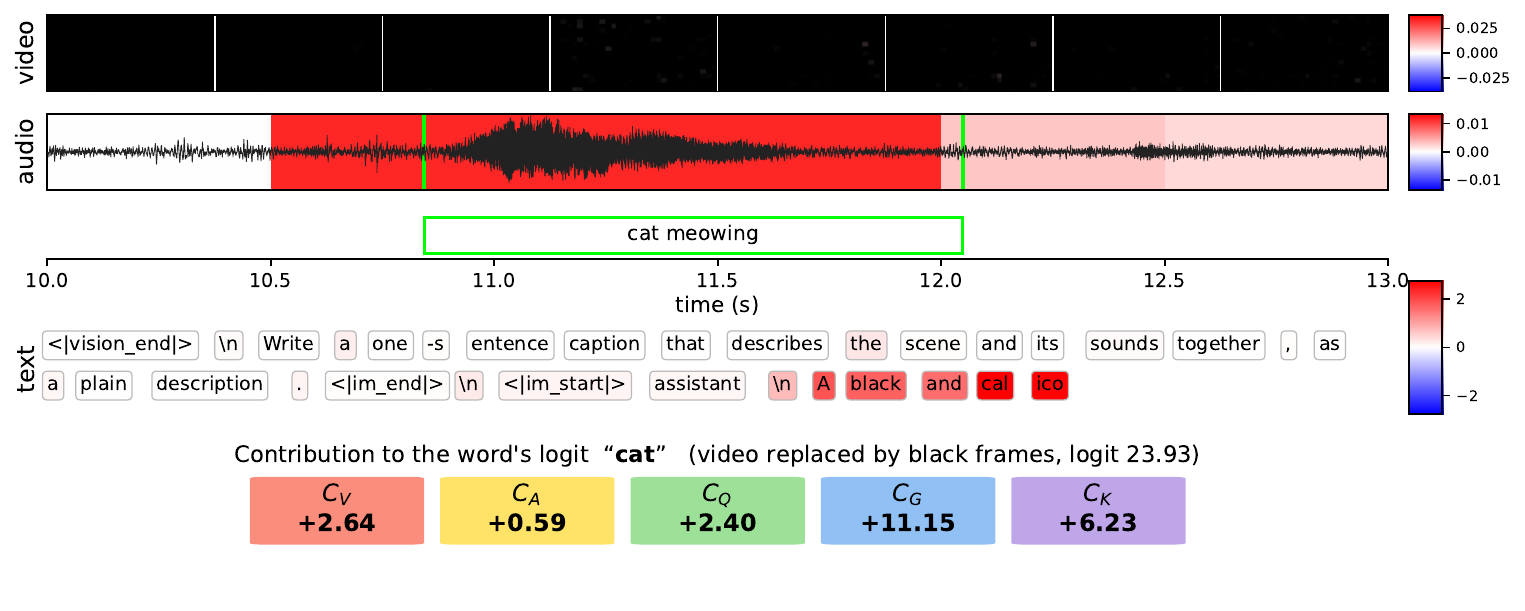}
        {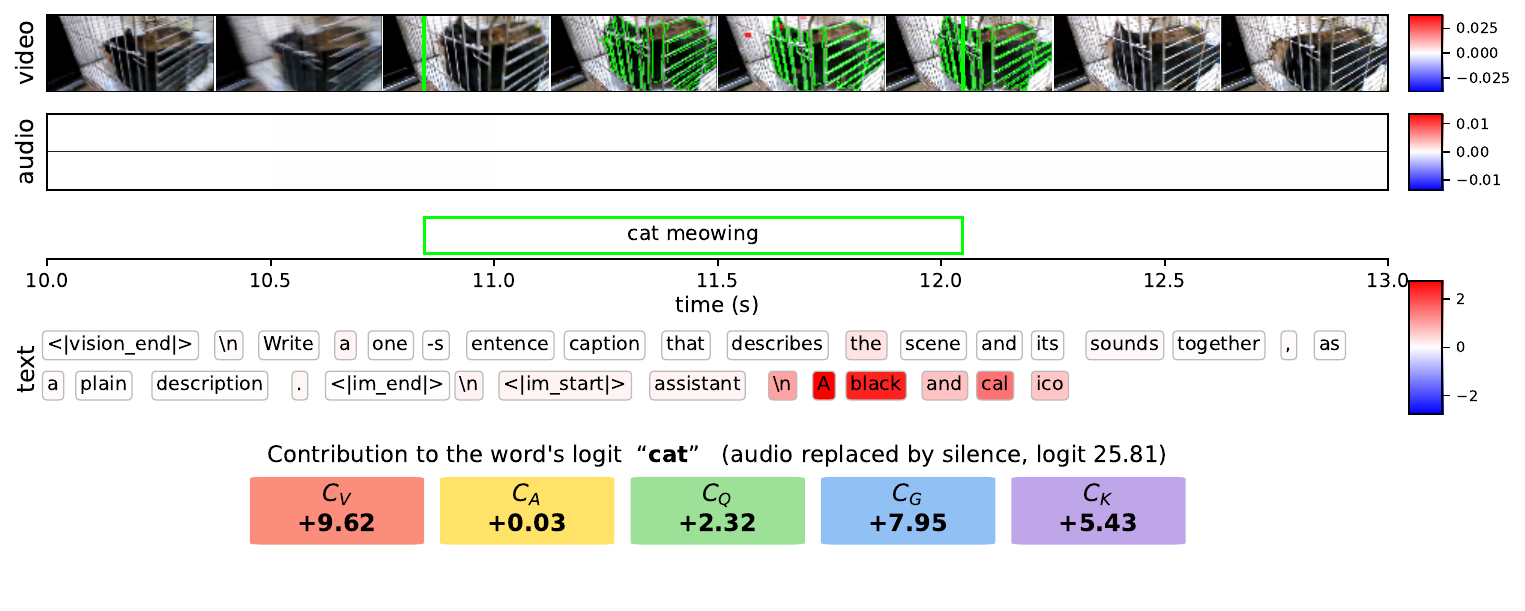}

    \caption{Qualitative MALT attributions under counterfactual modality obscuration for MCP 4.5 and video-SALMONN 2 Pro 4B/8B. The three conditions use the same generated sequence and target token for each model. Heatmaps show signed per-position contributions, while bottom boxes report aggregated source and non-source contributions. Color conventions follow Fig.~\ref{fig:qualitative_counterfactual}.}
    \label{fig:qualitative_counterfactual_c}
\end{figure*}

\subsection{Modality swap}
\Cref{fig:qualitative_swap_all_models,fig:qualitative_swap_all_models_b} present qualitative MALT attribution examples across all nine AVLLMs under conflicting audio-video evidence. For each model, the panels show generated words matched to the event carried by the video stream and by the audio
stream under swapped inputs. Panels are omitted when the selected
caption does not mention the corresponding event.

\colorlet{attrborder}{black!35}

\newcommand{\swapimg}[1]{%
    \adjustbox{valign=c}{%
        {%
            \setlength{\fboxsep}{0pt}%
            \setlength{\fboxrule}{0.15pt}%
            \color{attrborder}%
            \fbox{%
                \includegraphics[
                    width=\dimexpr0.47\textwidth-2\fboxrule\relax
                ]{#1}%
            }%
        }%
    }%
}
\newcommand{\swapmodel}[1]{%
    \adjustbox{valign=c}{%
        \rotatebox[origin=c]{90}{\textbf{#1}}%
    }%
}

\begin{figure*}[p]
    \centering

    \begin{tabular}{@{}c@{\hspace{4pt}}c@{\hspace{6pt}}c@{}}

        & \textbf{Video target token}
        & \textbf{Audio target token} \\[4pt]

        \swapmodel{Omni 3B}
        & \swapimg{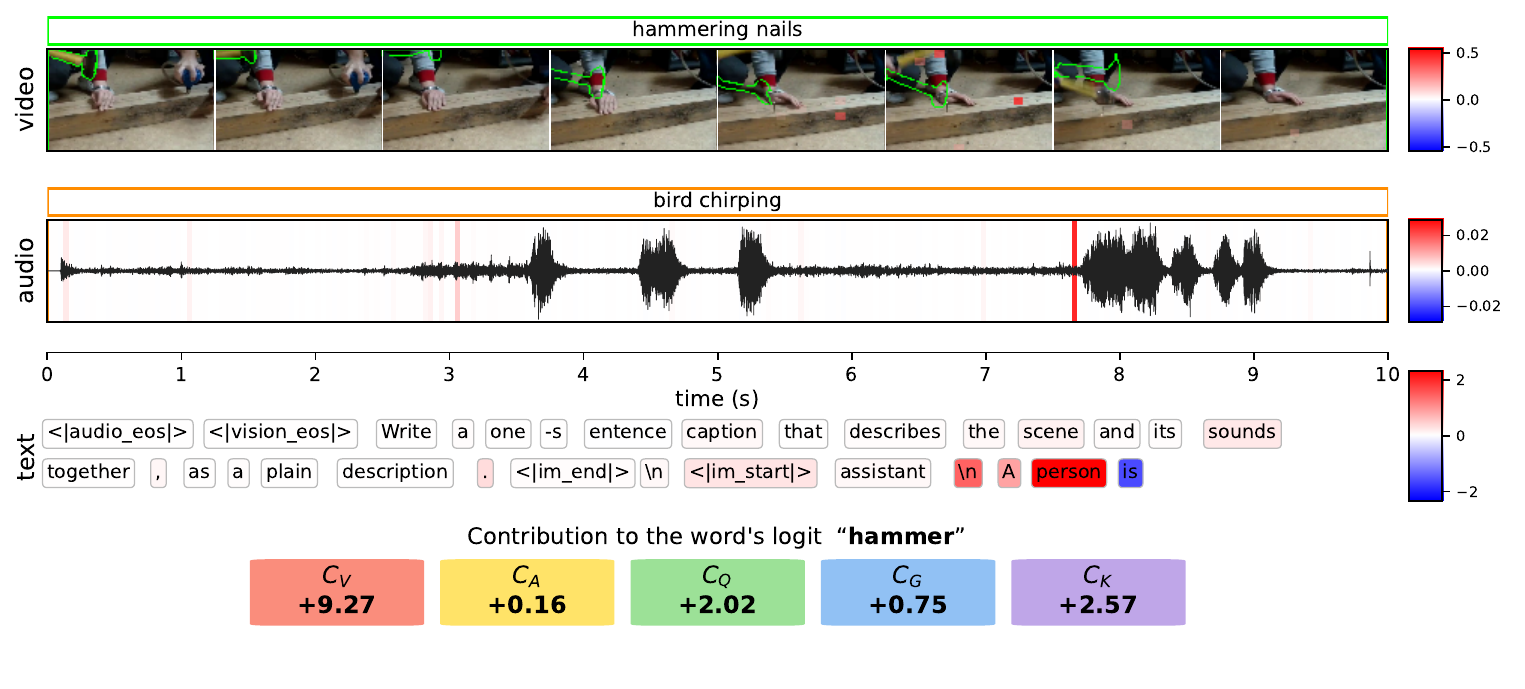}
        & \swapimg{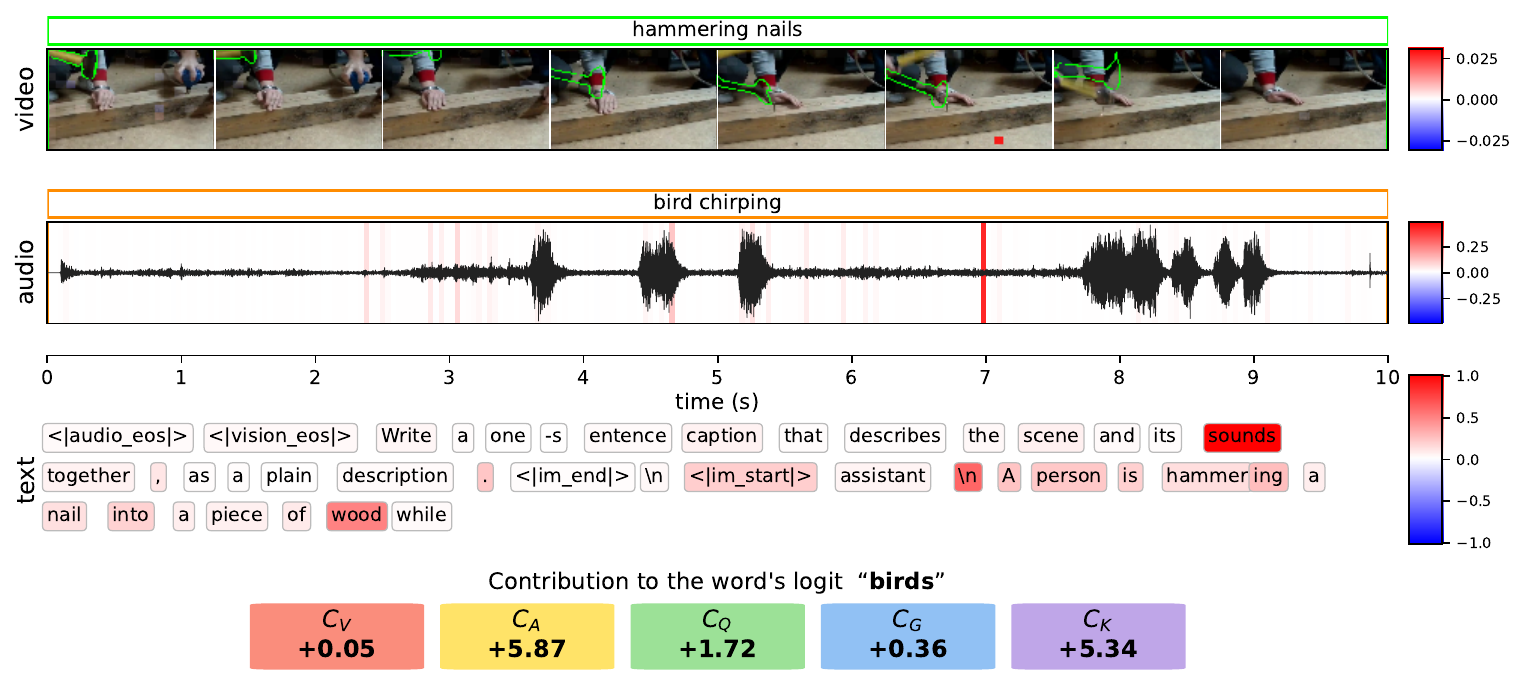}
        \\
        \noalign{\vskip 4pt}

        \swapmodel{Omni 7B}
        & \swapimg{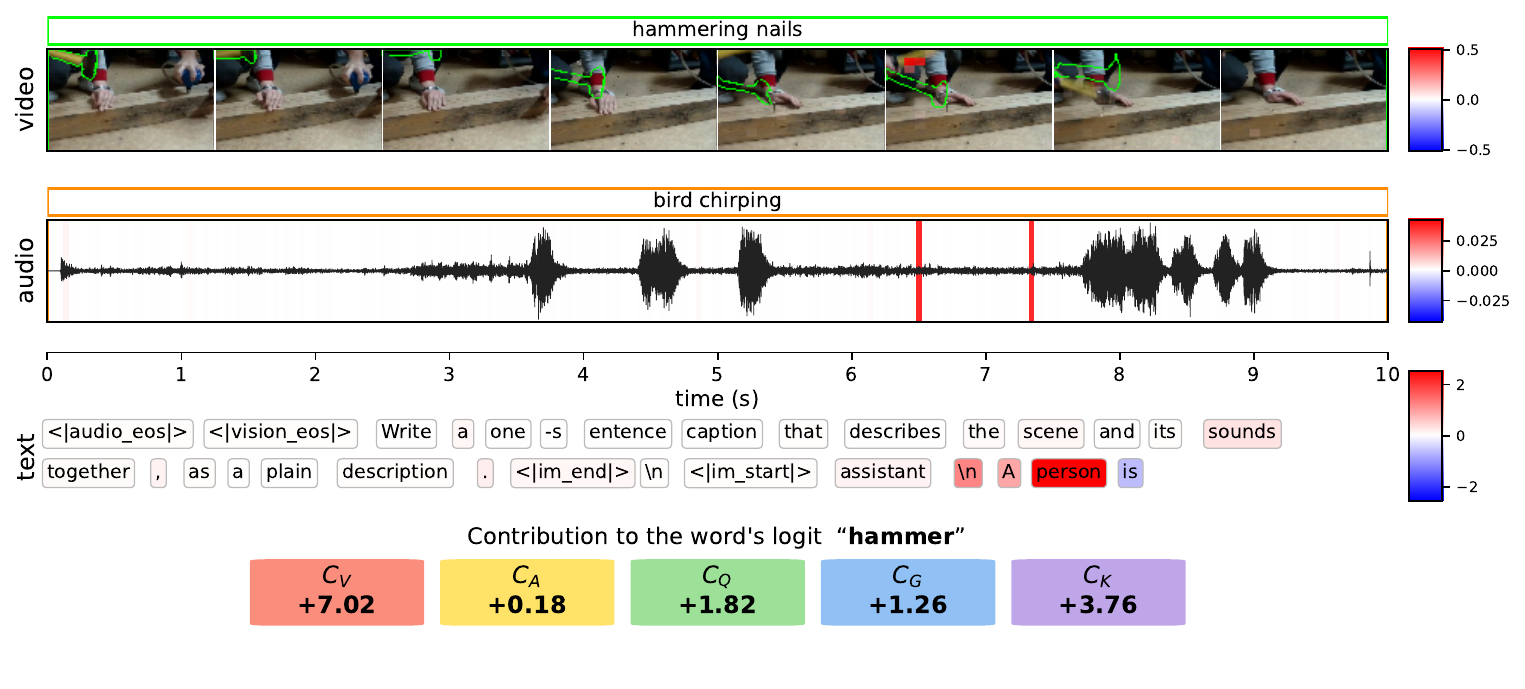}
        & \swapimg{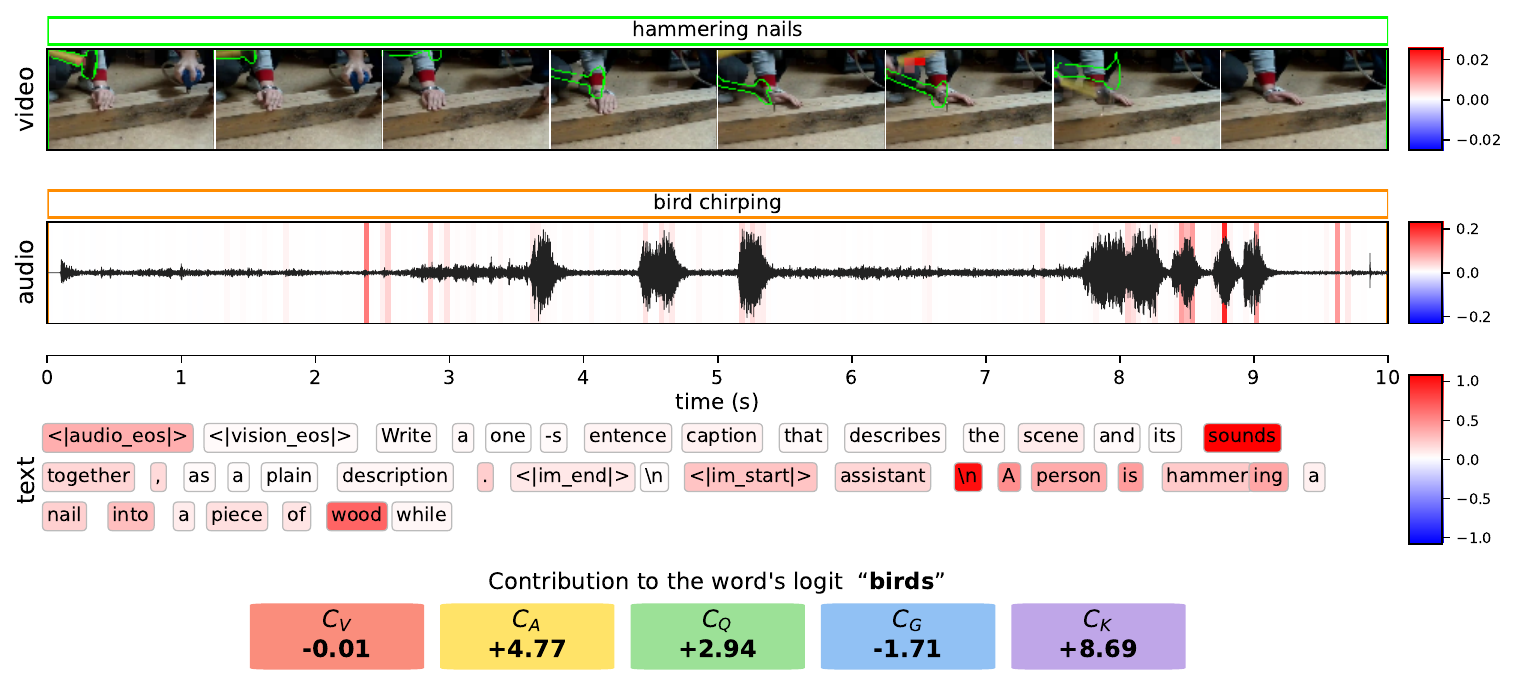}
        \\
        \noalign{\vskip 4pt}

        \swapmodel{MCP 2.6}
        & \swapimg{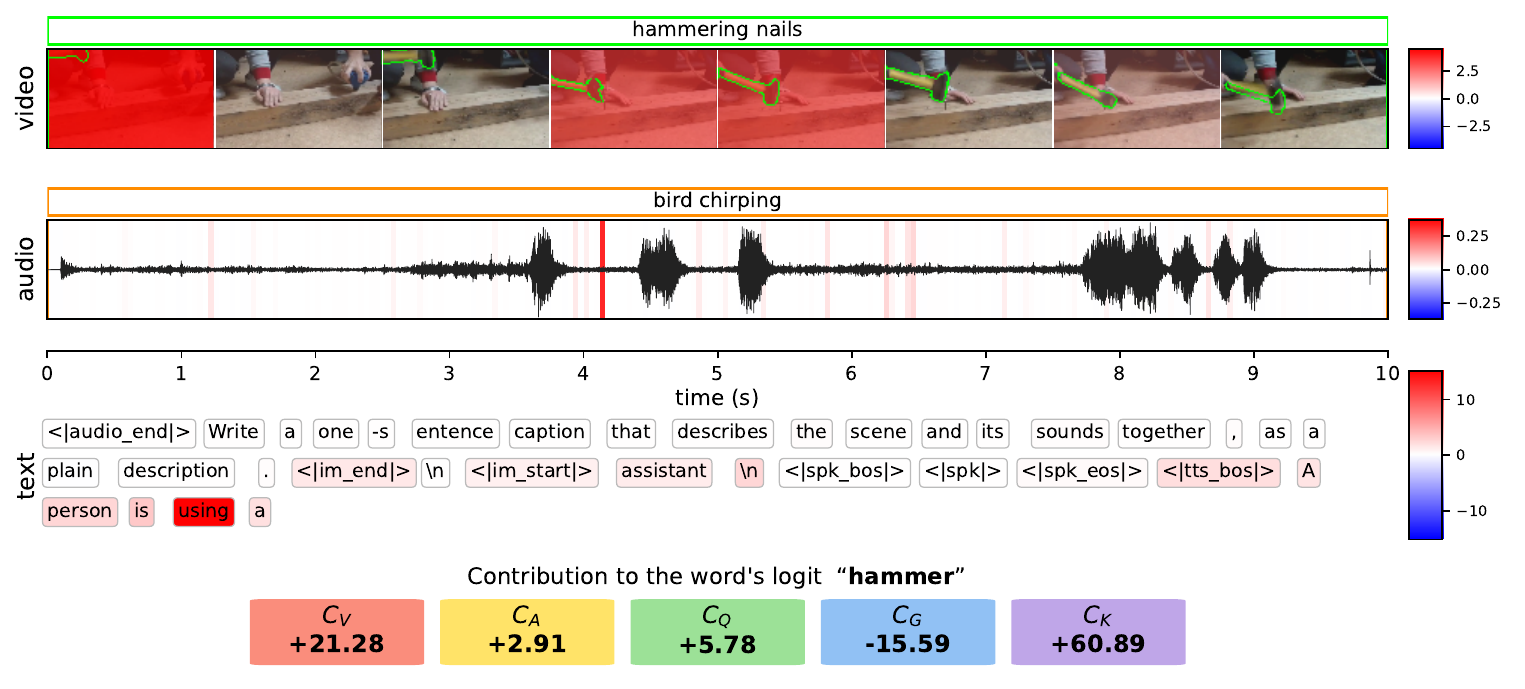}
        & \swapimg{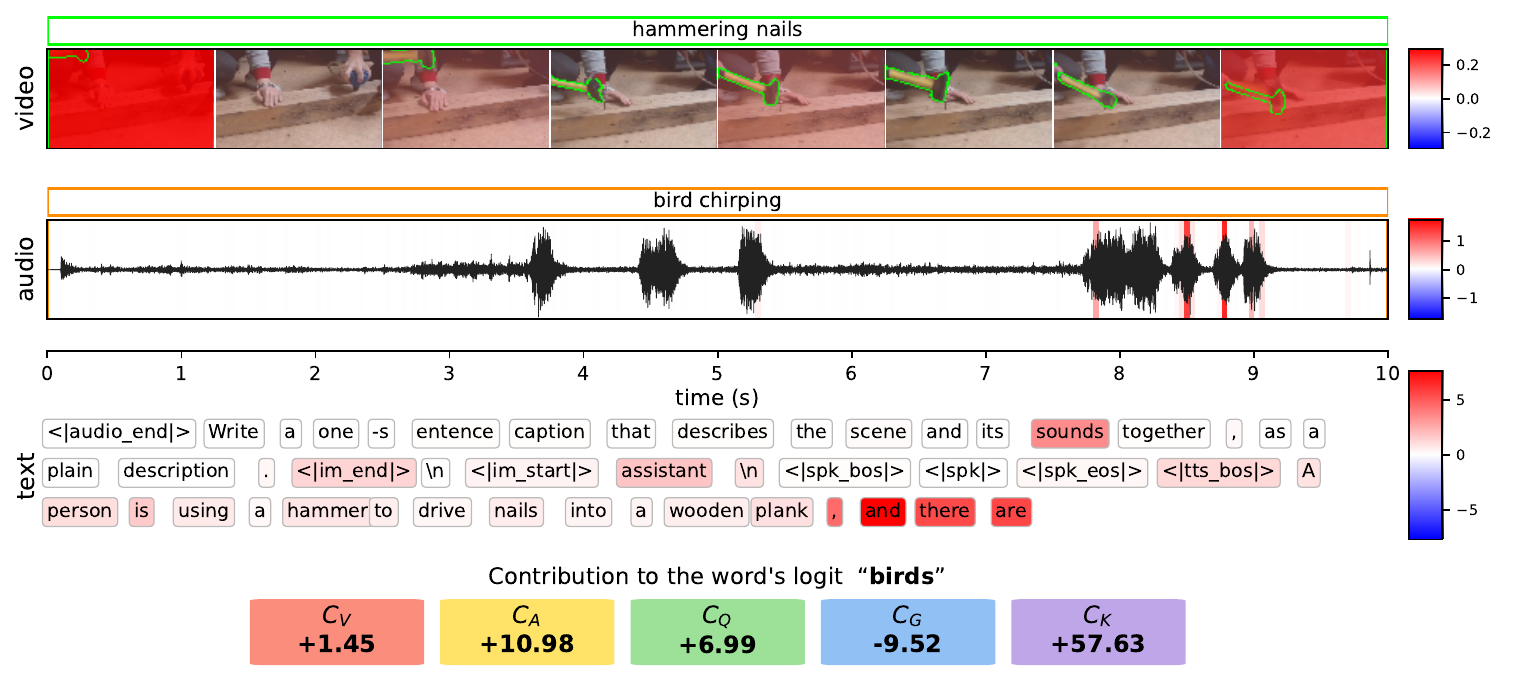}
        \\
        \noalign{\vskip 4pt}

        \swapmodel{VL-AV}
        & \swapimg{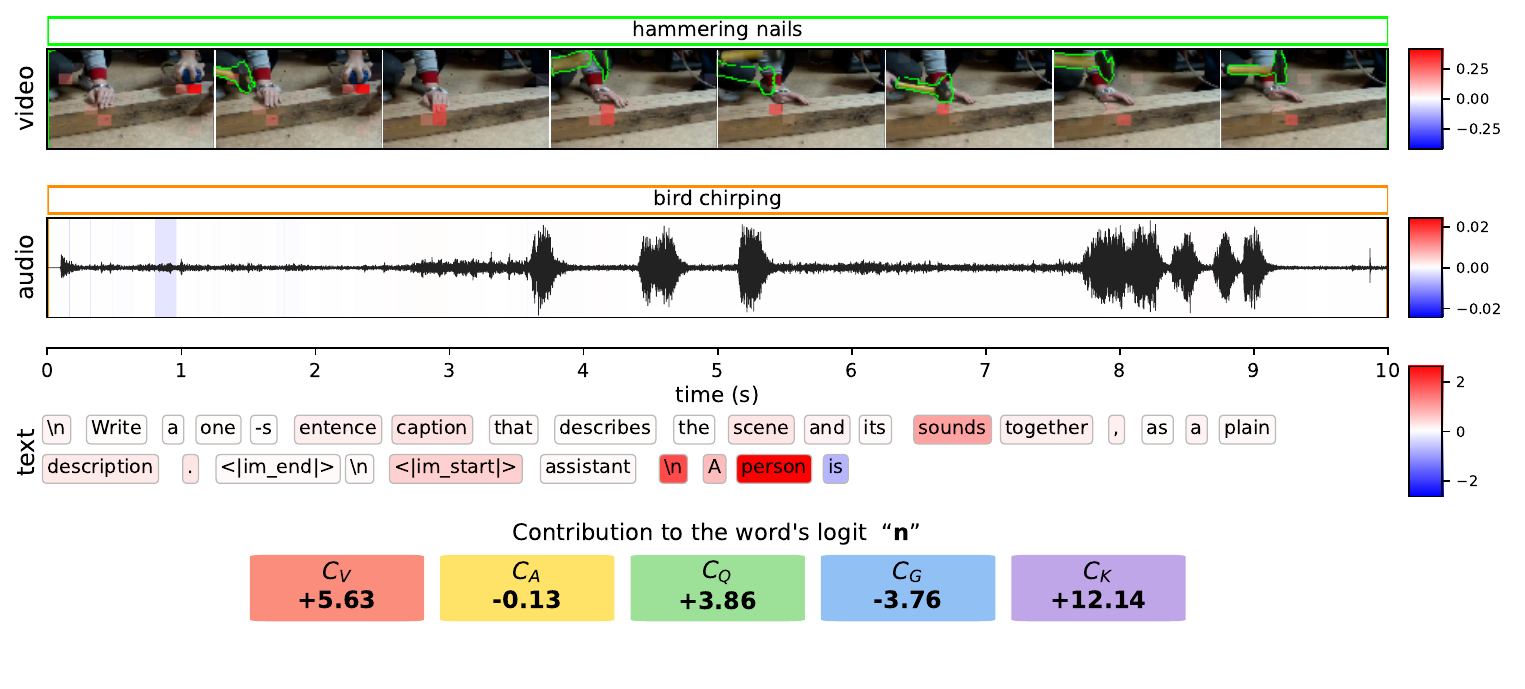}
        & \swapimg{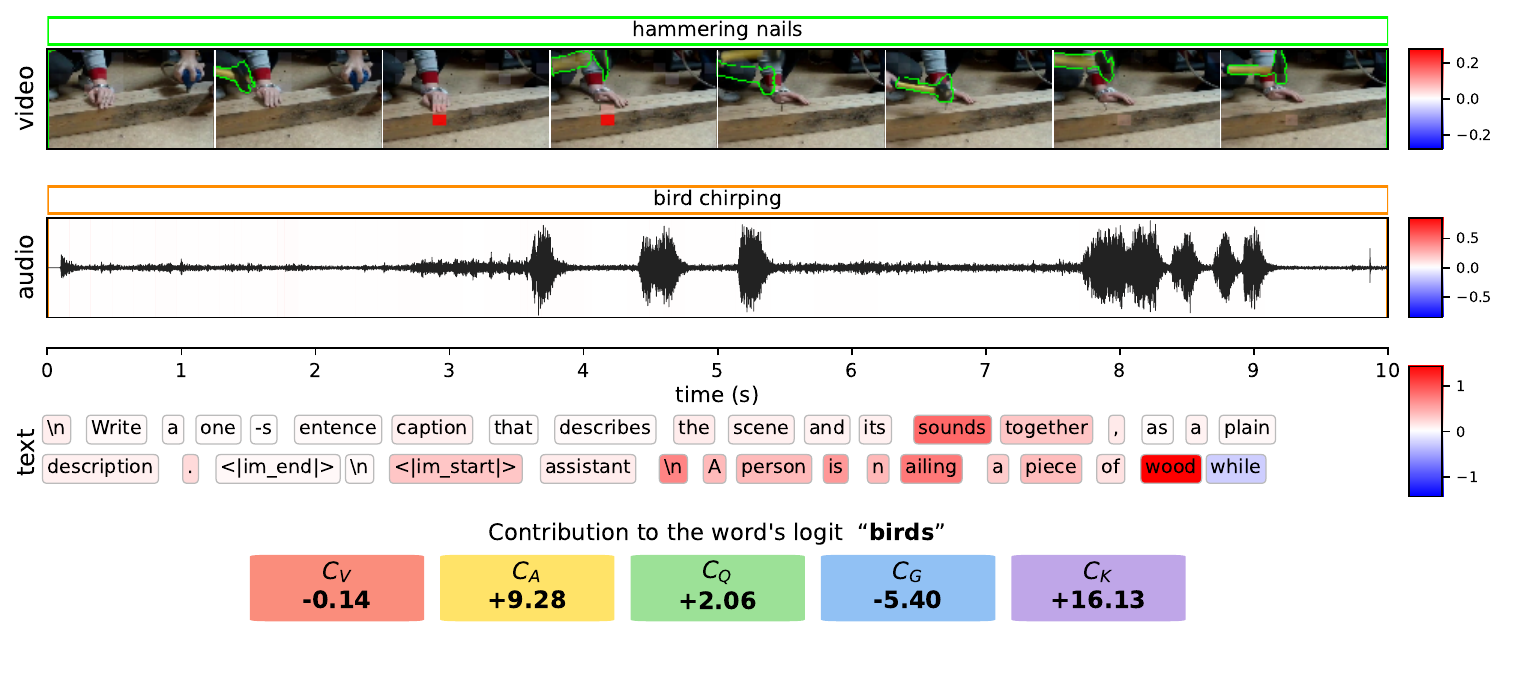}
        \\
        \noalign{\vskip 4pt}

        \swapmodel{Ola-7B}
        & \swapimg{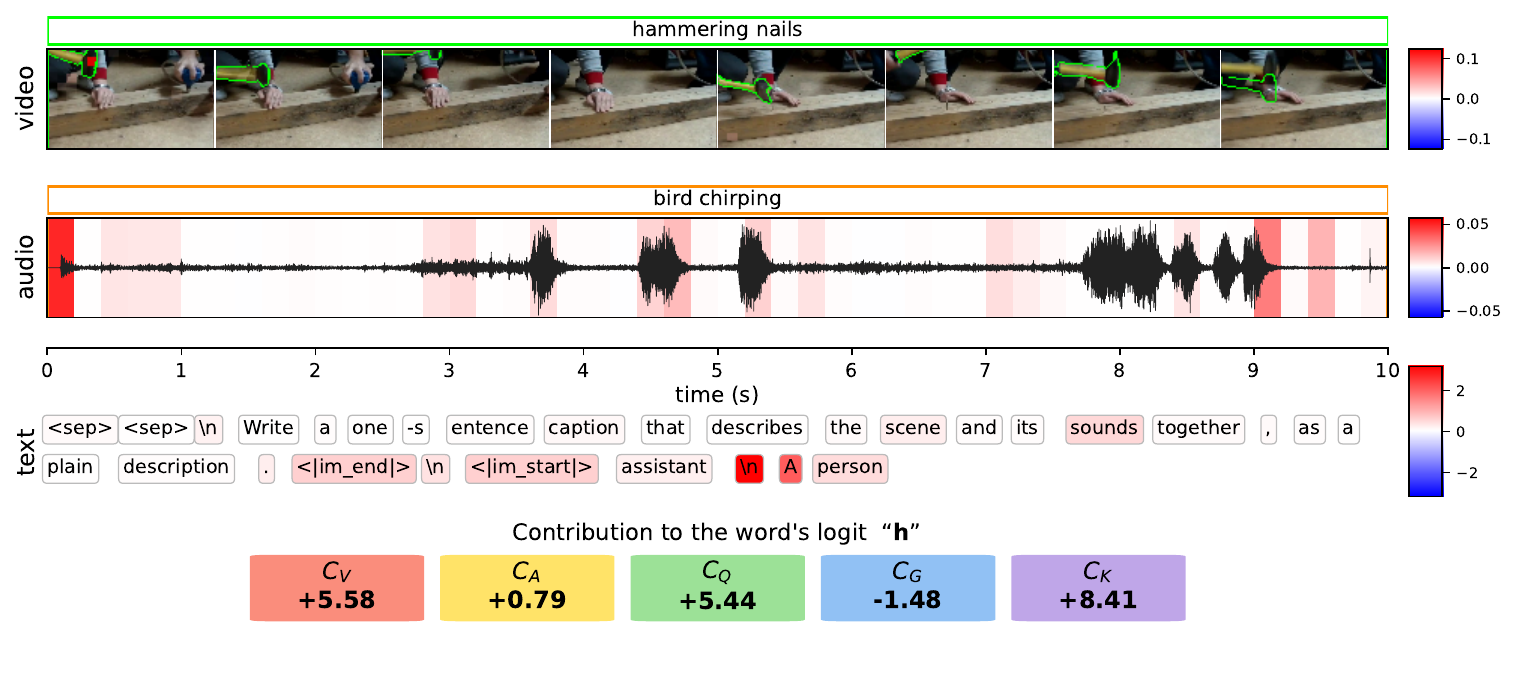}
        & \swapimg{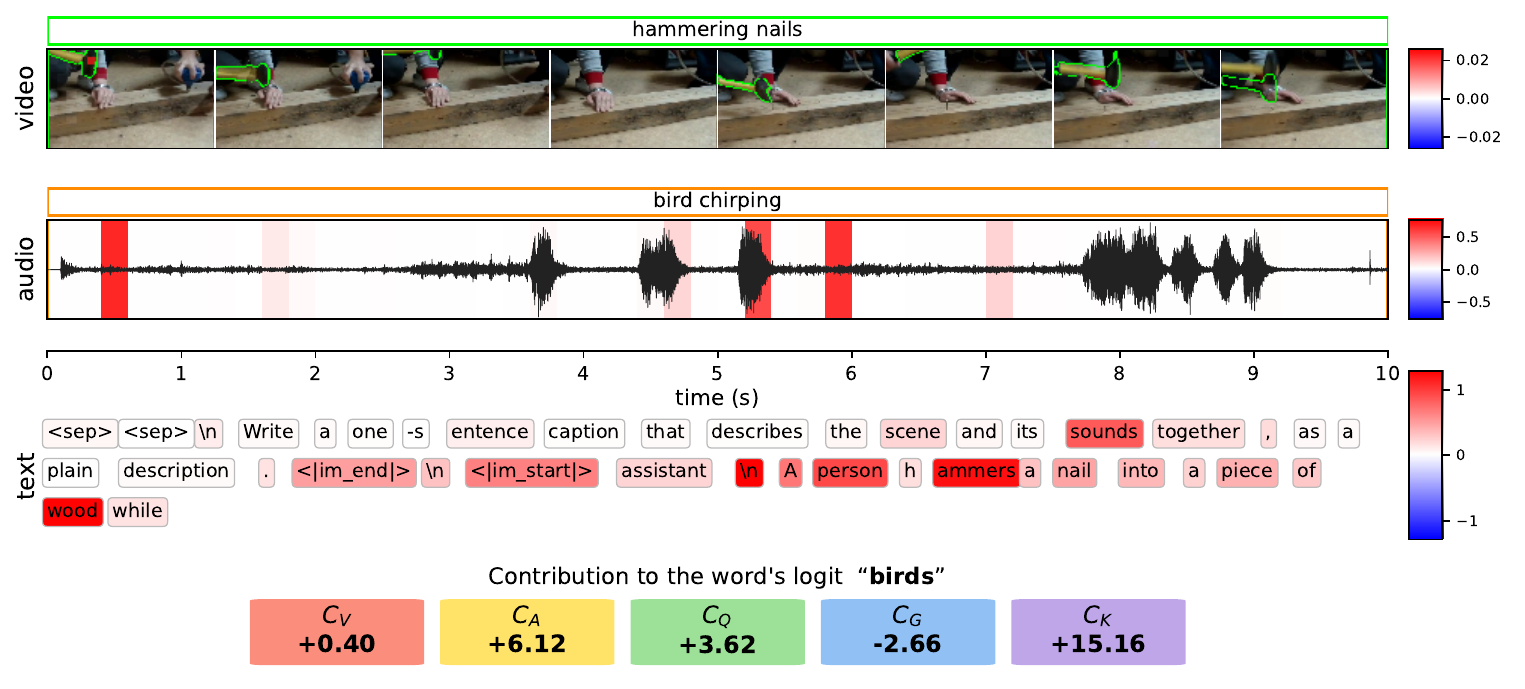}

    \end{tabular}

    \caption{Qualitative MALT attributions under controlled
    audio-video swaps (Omni 3B, Omni 7B, MCP 2.6, VL-AV,
    and Ola-7B). The video depicts hammering nails,
    while the audio contains bird chirping. Each row
    corresponds to a model, and the columns show
    attributions for the first token of a generated word
    referring to the video-carried event (left) or
    audio-carried event (right). Red and blue indicate
    positive and negative contributions, respectively;
    colorbars use independent scales across modalities
    and models. Green annotations mark the video event,
    while orange annotations indicate the audio event.
    Bottom boxes report aggregated contributions
    $C_V$, $C_A$, $C_Q$, $C_G$, and $C_K$ in logit
    units; $E$ and $\epsilon$ are omitted.}

    \label{fig:qualitative_swap_all_models}
\end{figure*}

\begin{figure*}[p]
    \centering

    \begin{tabular}{@{}c@{\hspace{4pt}}c@{\hspace{6pt}}c@{}}

        & \textbf{Video target token}
        & \textbf{Audio target token} \\[4pt]

        \swapmodel{AVF}
        & \swapimg{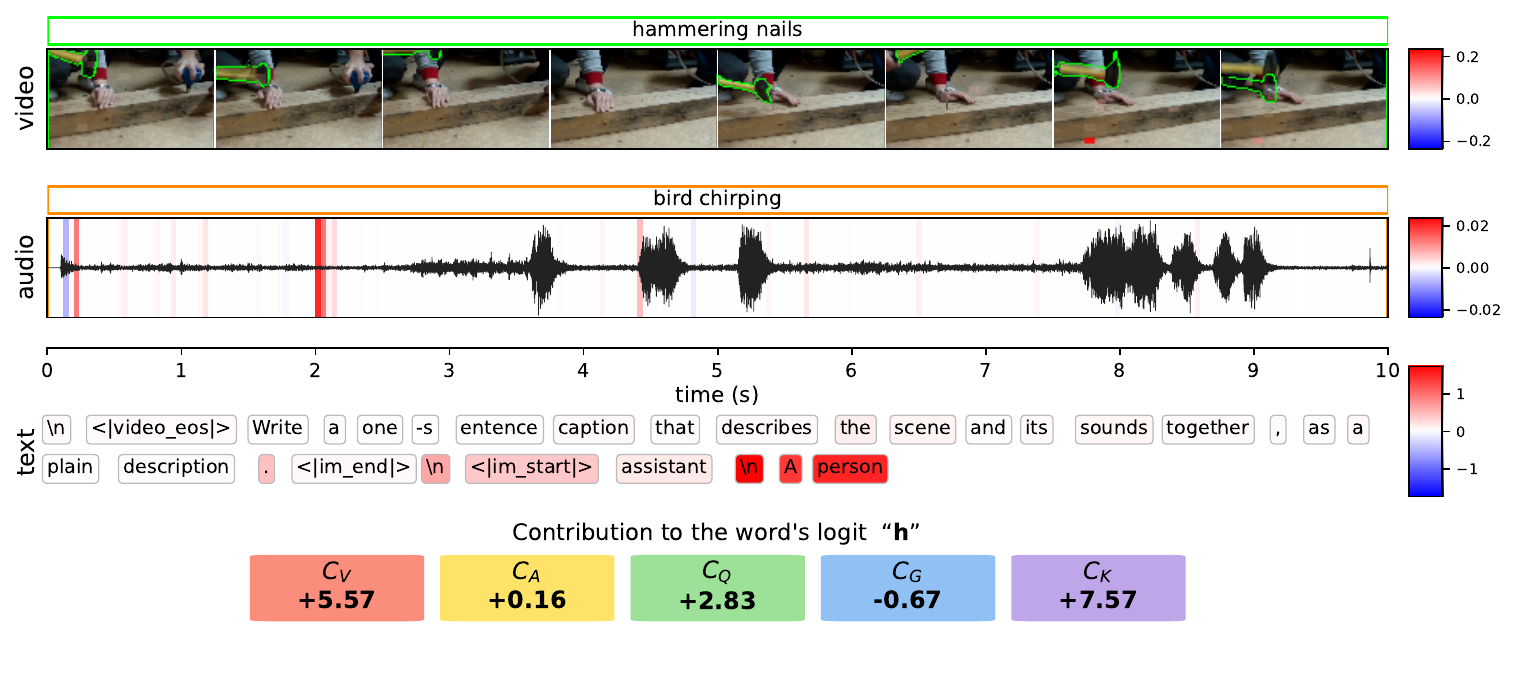}
        & \swapimg{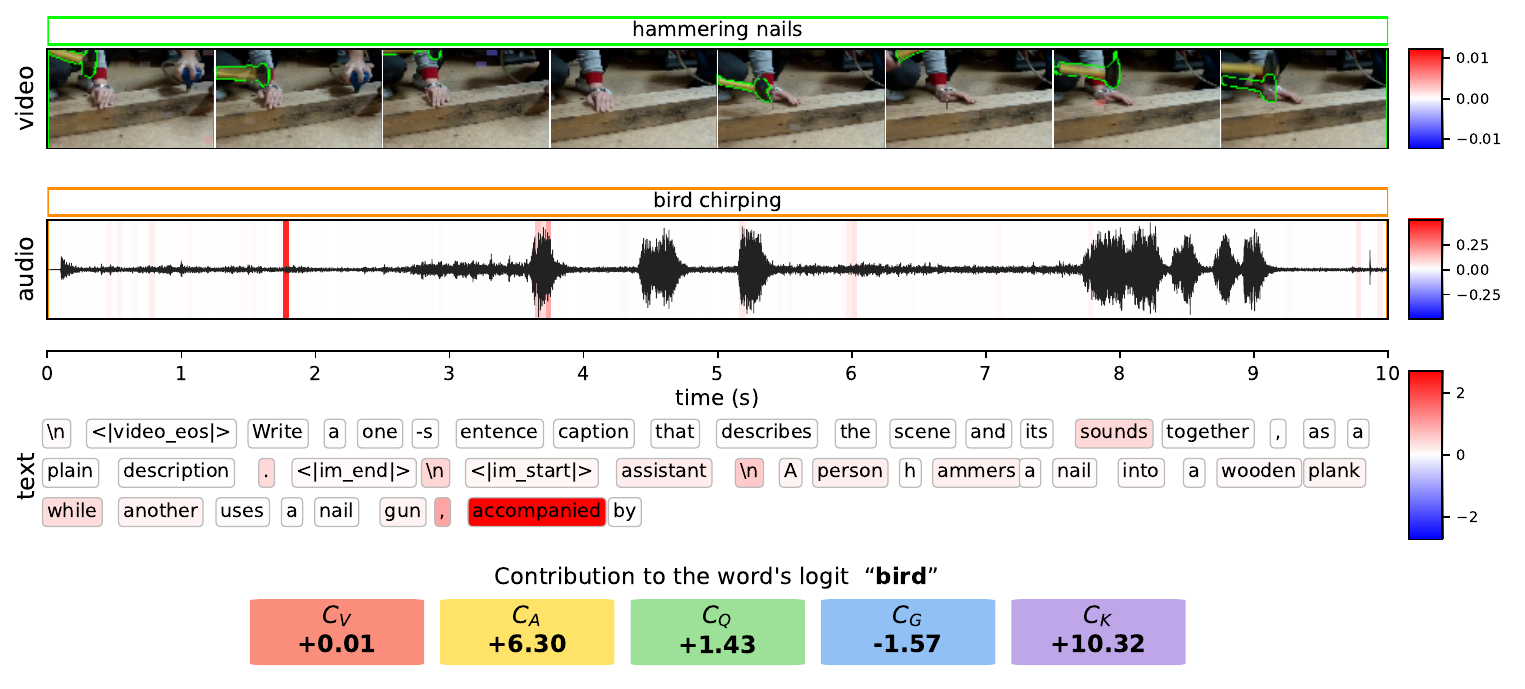}
        \\
        \noalign{\vskip 6pt}

        \swapmodel{MCP 4.5}
        & \swapimg{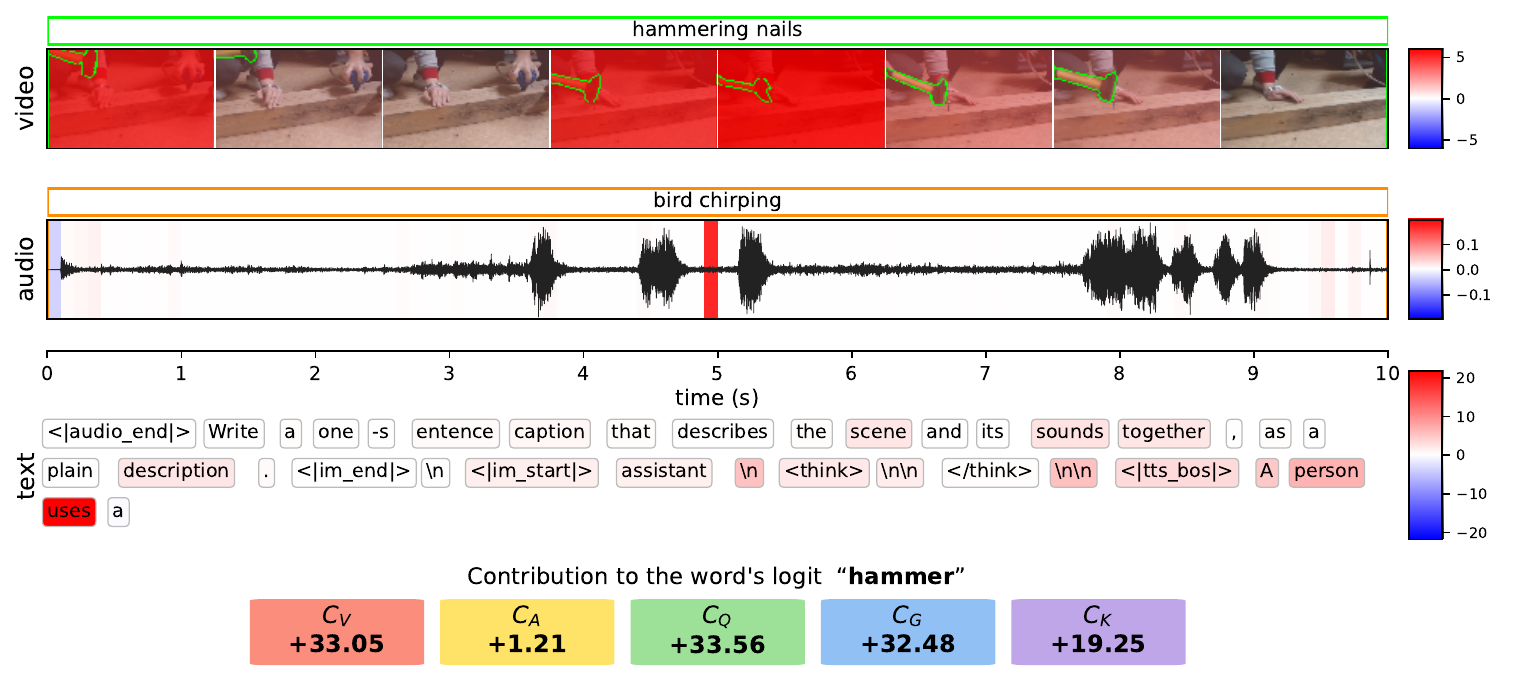}
        & \swapimg{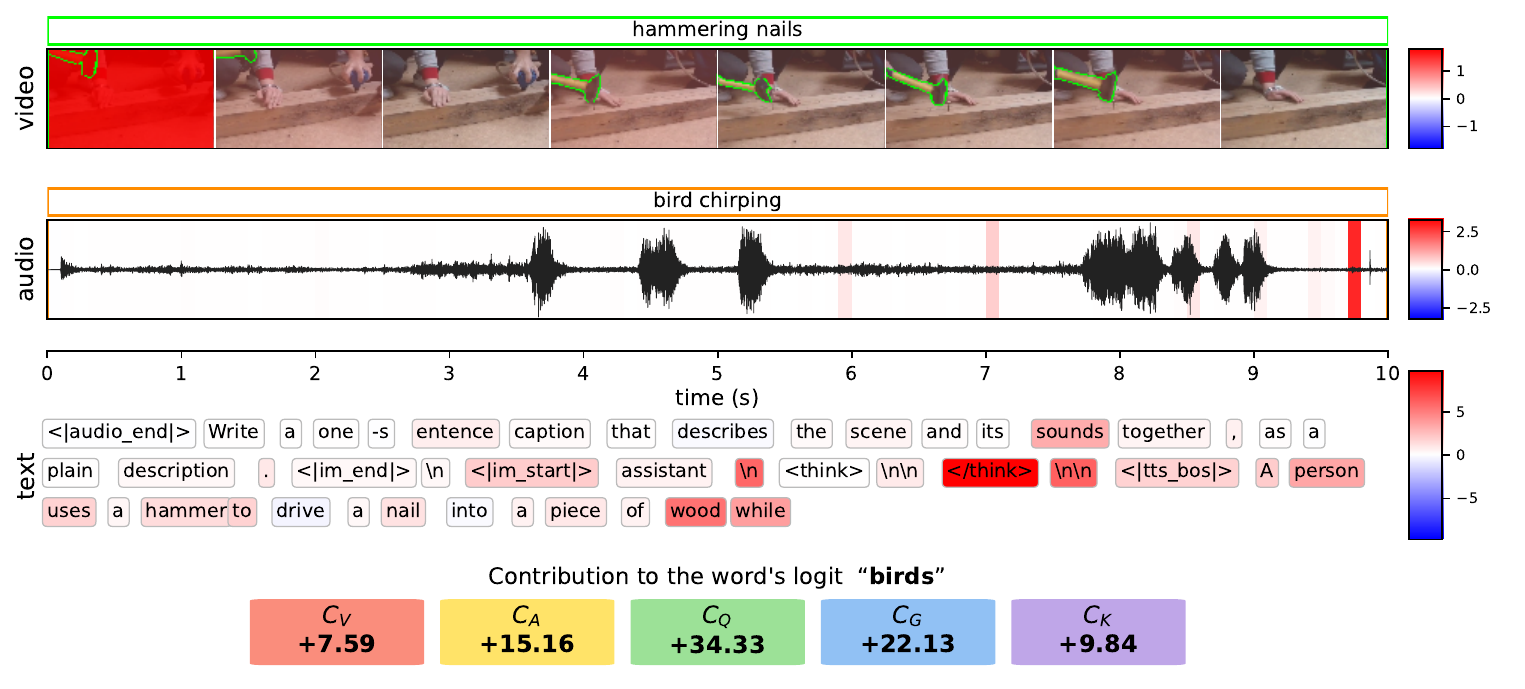}
        \\
        \noalign{\vskip 6pt}

        \swapmodel{vSAL 4B}
        & \swapimg{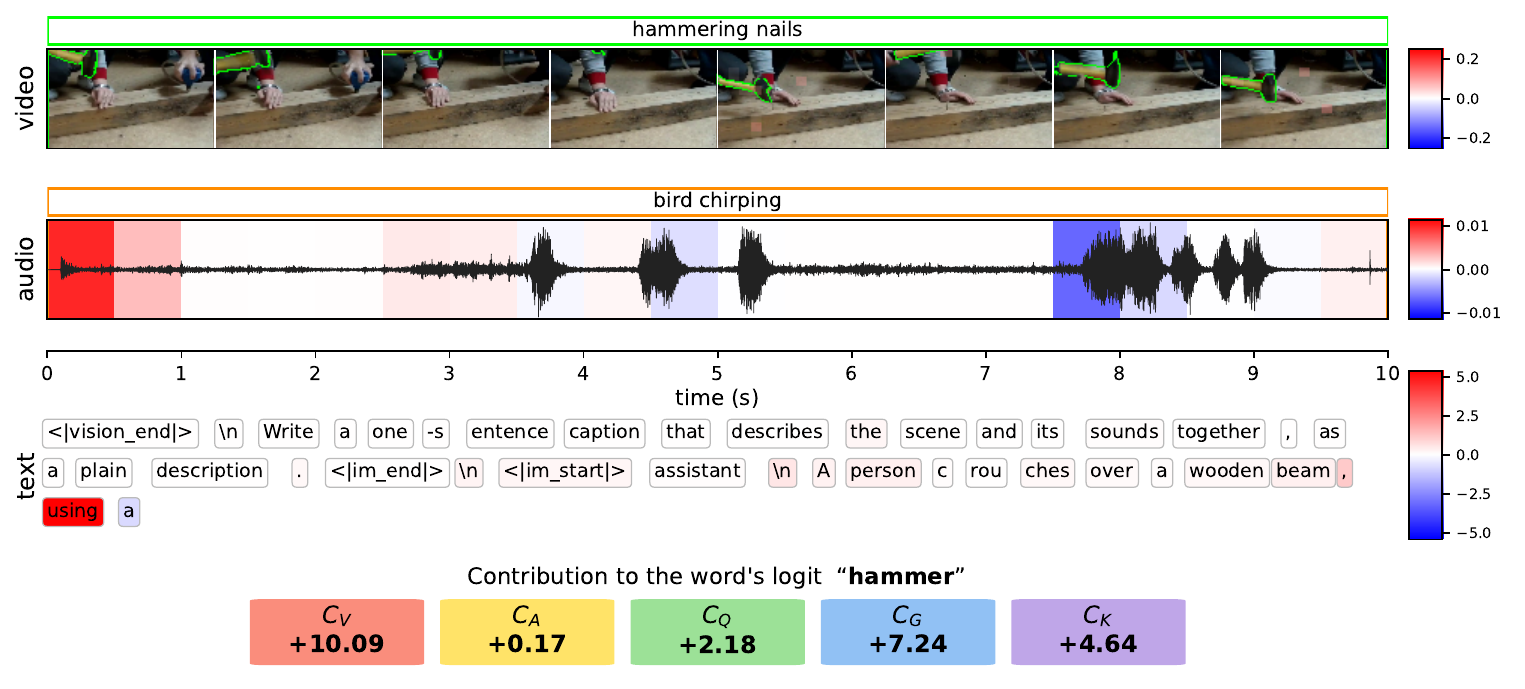}
        & \multicolumn{1}{c}{}
        \\
        \noalign{\vskip 6pt}

        \swapmodel{vSAL 8B}
        & \swapimg{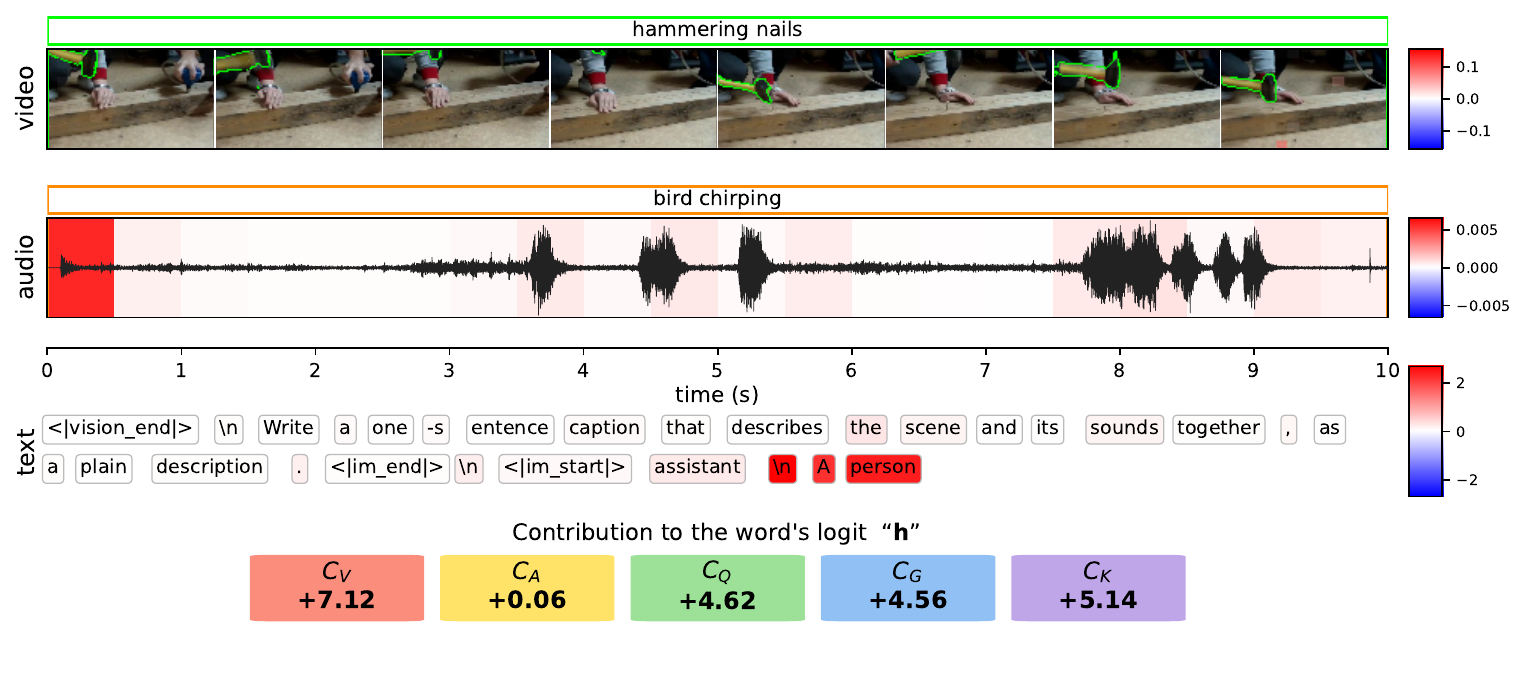}
        & \multicolumn{1}{c}{}
    \end{tabular}
    \caption{Qualitative MALT attributions under controlled
    audio-video swaps for AVF, MCP 4.5, and
    video-SALMONN 2 Pro 4B/8B, continuing
    Fig.~\ref{fig:qualitative_swap_all_models}.
    The columns show attributions for words referring
    to the video-carried and audio-carried events,
    respectively. Color conventions and component
    definitions follow
    Fig.~\ref{fig:qualitative_swap_all_models}.
    Audio-target panels are omitted for video-SALMONN
    because the selected captions do not mention
    the audio-carried event.}
    \label{fig:qualitative_swap_all_models_b}
\end{figure*}

\subsection{Baselines comparison}

\begin{figure*}[t]
    \centering
    \setlength{\tabcolsep}{2pt}
    \setlength{\fboxsep}{0pt}
    \setlength{\fboxrule}{0.15pt}
    \colorlet{attrborder}{black!35}

    \newcommand{\attrimg}[1]{%
        {\color{attrborder}\fbox{%
            \includegraphics[width=0.39\textwidth]{#1}%
        }}%
    }

    \newcommand{\attrmodel}[1]{%
        \adjustbox{valign=c}{%
            \rotatebox[origin=c]{90}{ #1}%
        }%
    }

    \newcommand{\attrblock}[4]{%
        \adjustbox{valign=c}{%
            \begin{tabular}{@{}c@{\hspace{4pt}}c@{}}
                \textbf{MALT}
                & \textbf{Gradient$\times$Input}
                \\[2pt]

                \attrimg{#1}
                & \attrimg{#2}
                \\[5pt]

                \textbf{AttnLRP}
                & \textbf{ALTI-Logit}
                \\[2pt]

                \attrimg{#3}
                & \attrimg{#4}
            \end{tabular}%
        }%
    }

    \begin{adjustbox}{
        max totalsize={\textwidth}{0.90\textheight},
        center
    }

    \begin{tabular}{@{}c@{\hspace{5pt}}c@{}}

        \attrmodel{\textbf{Omni 3B}}
        &
        \attrblock
        {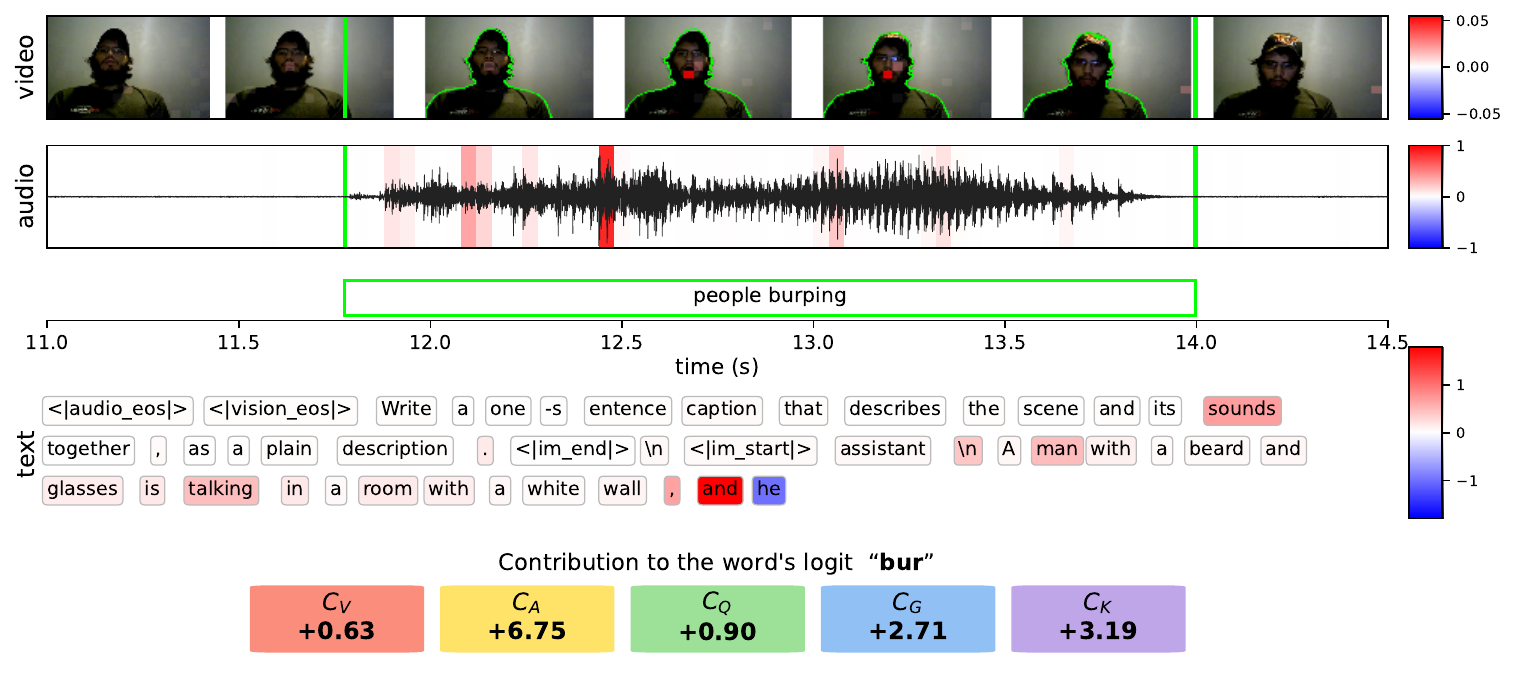}
        {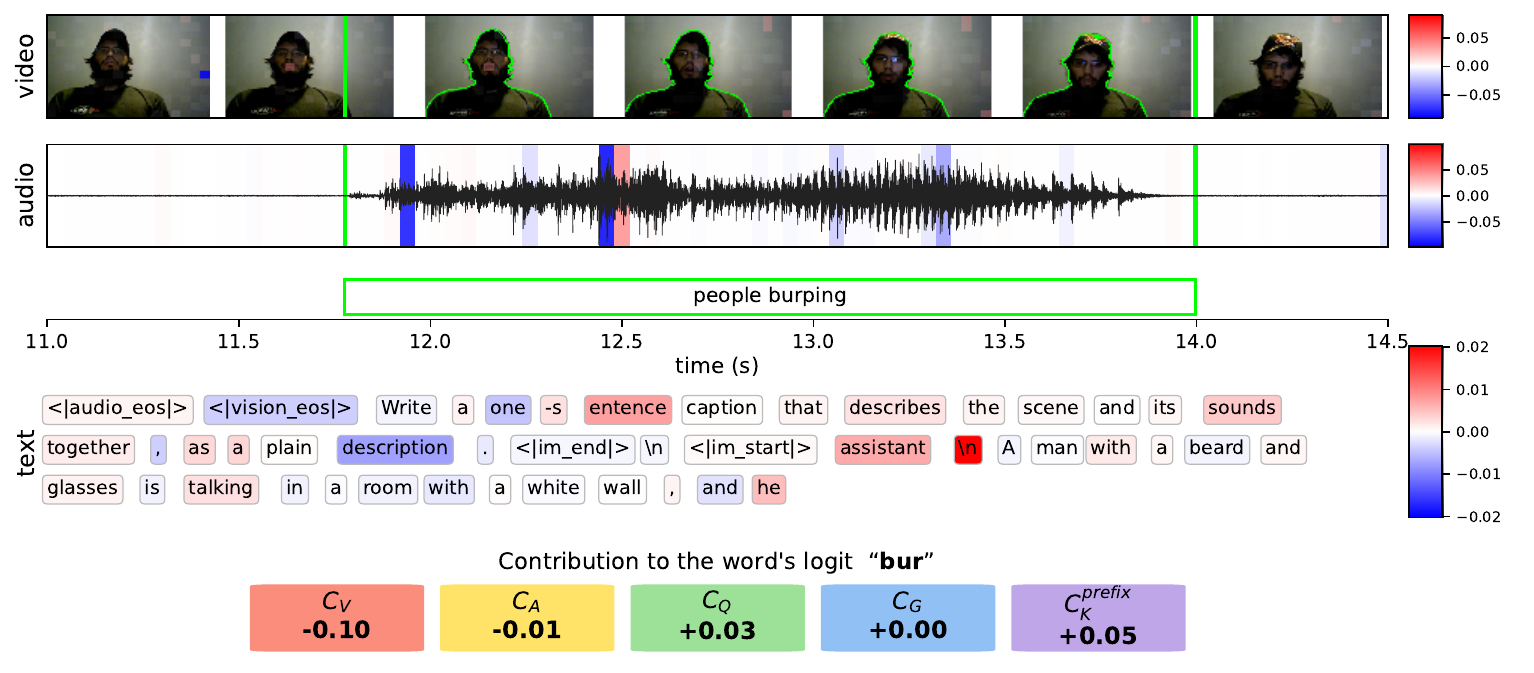}
        {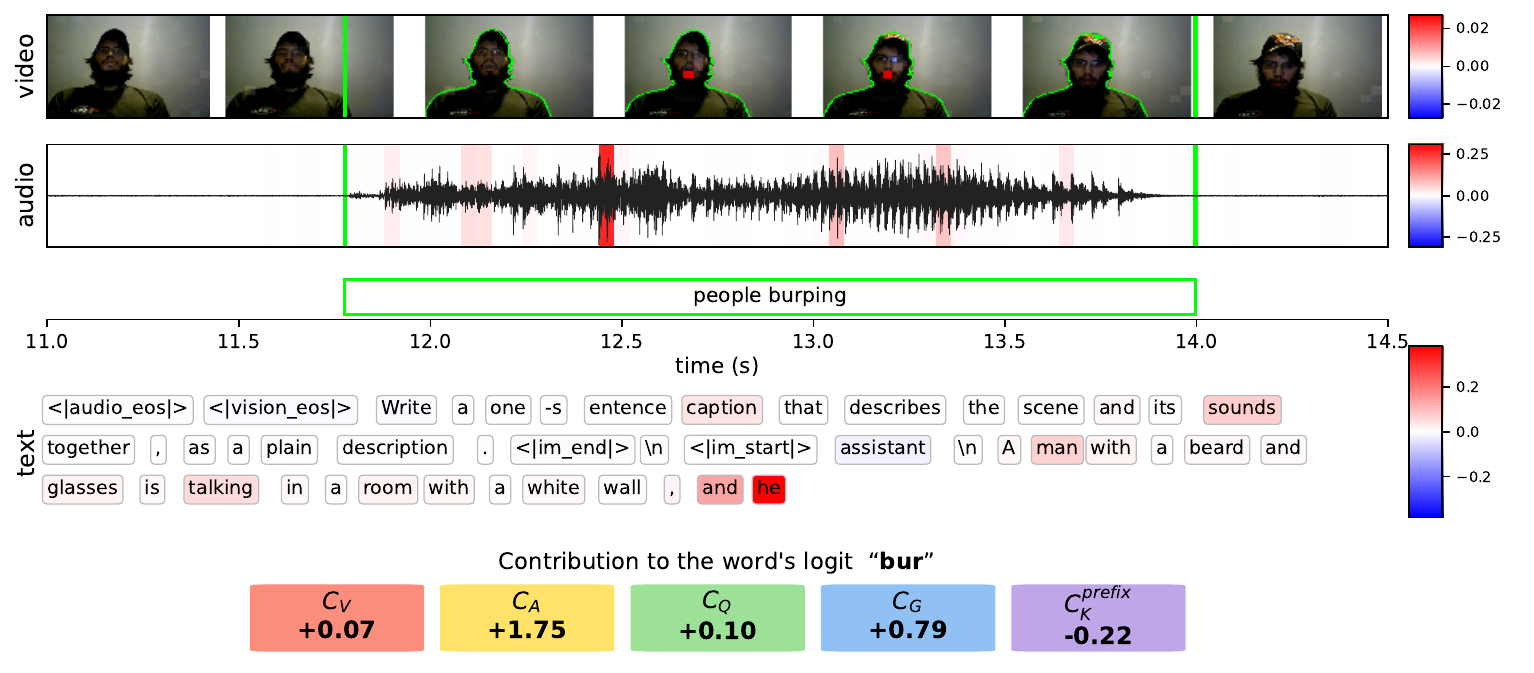}
        {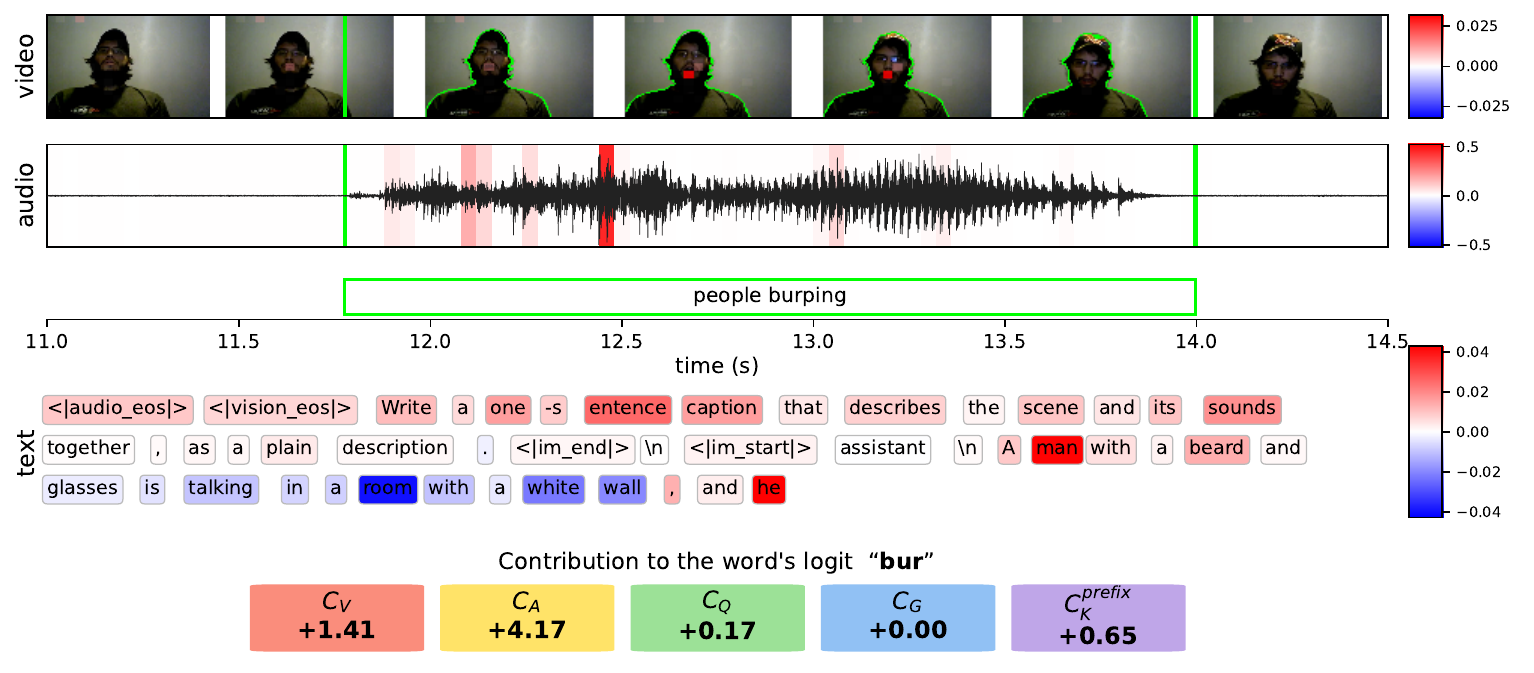}
        \\

        \noalign{\vskip 14pt}

        \attrmodel{\textbf{VL-AV}}
        &
        \attrblock
        {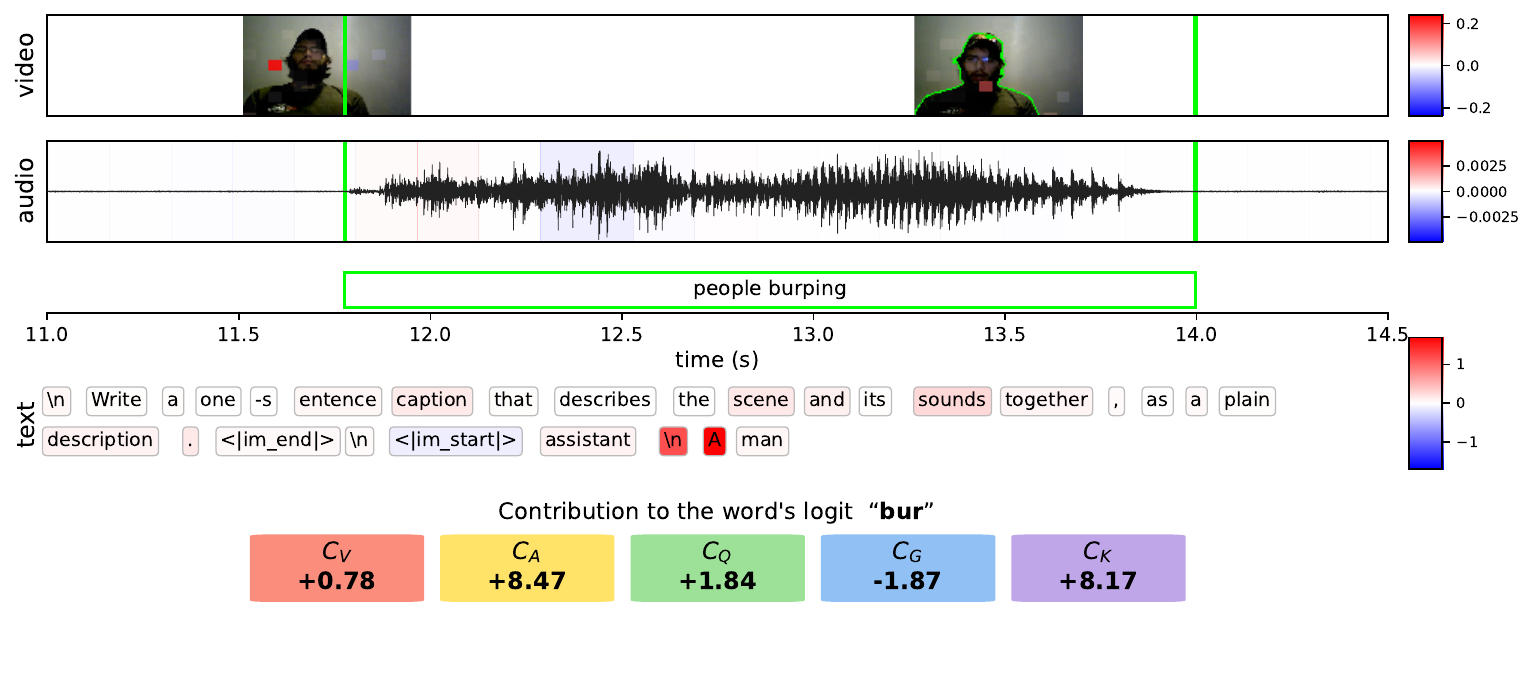}
        {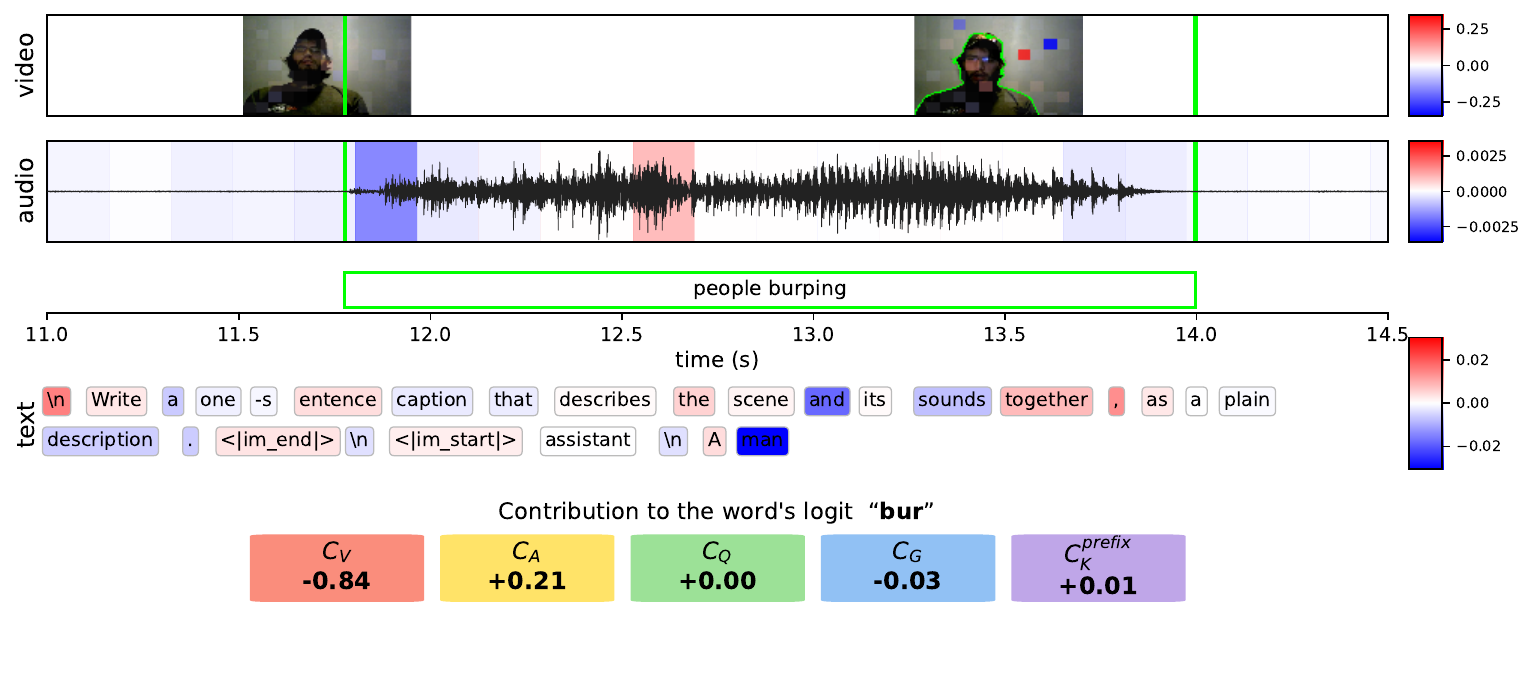}
        {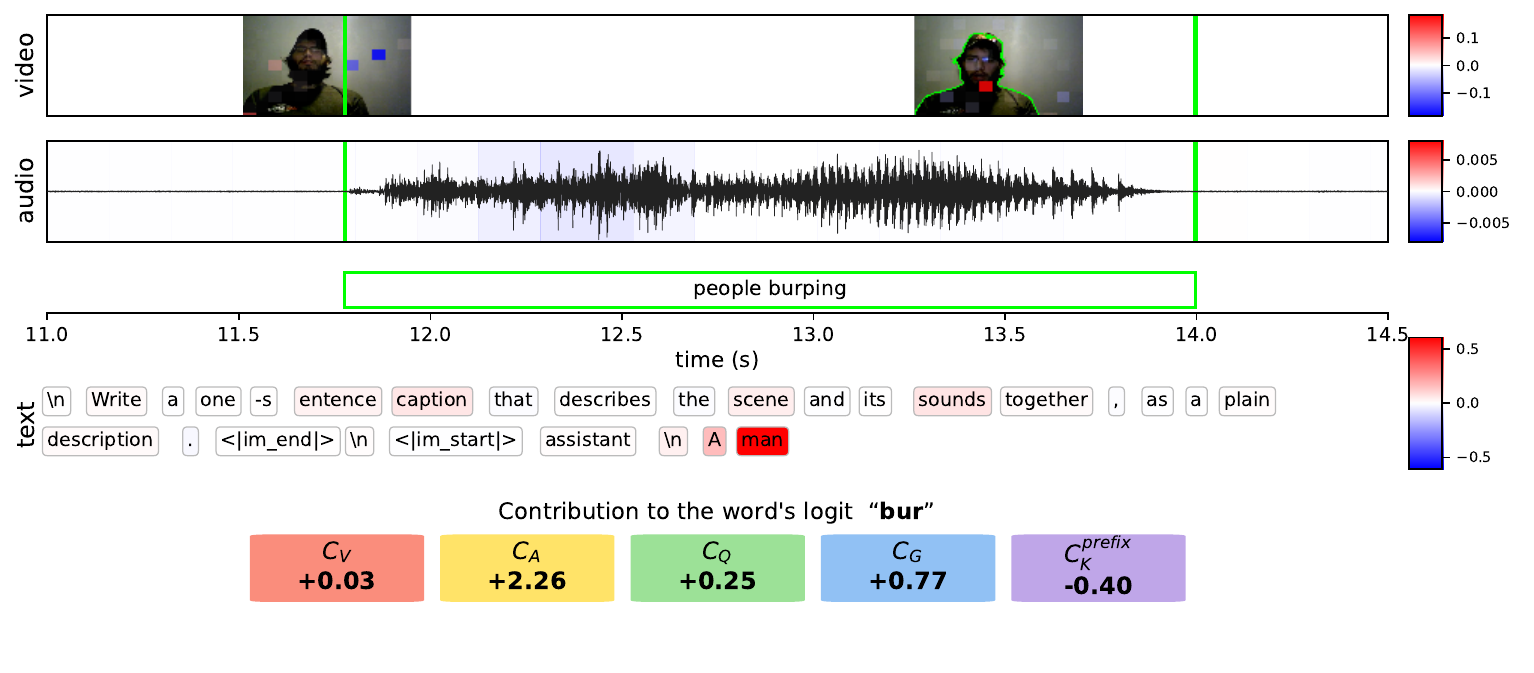}
        {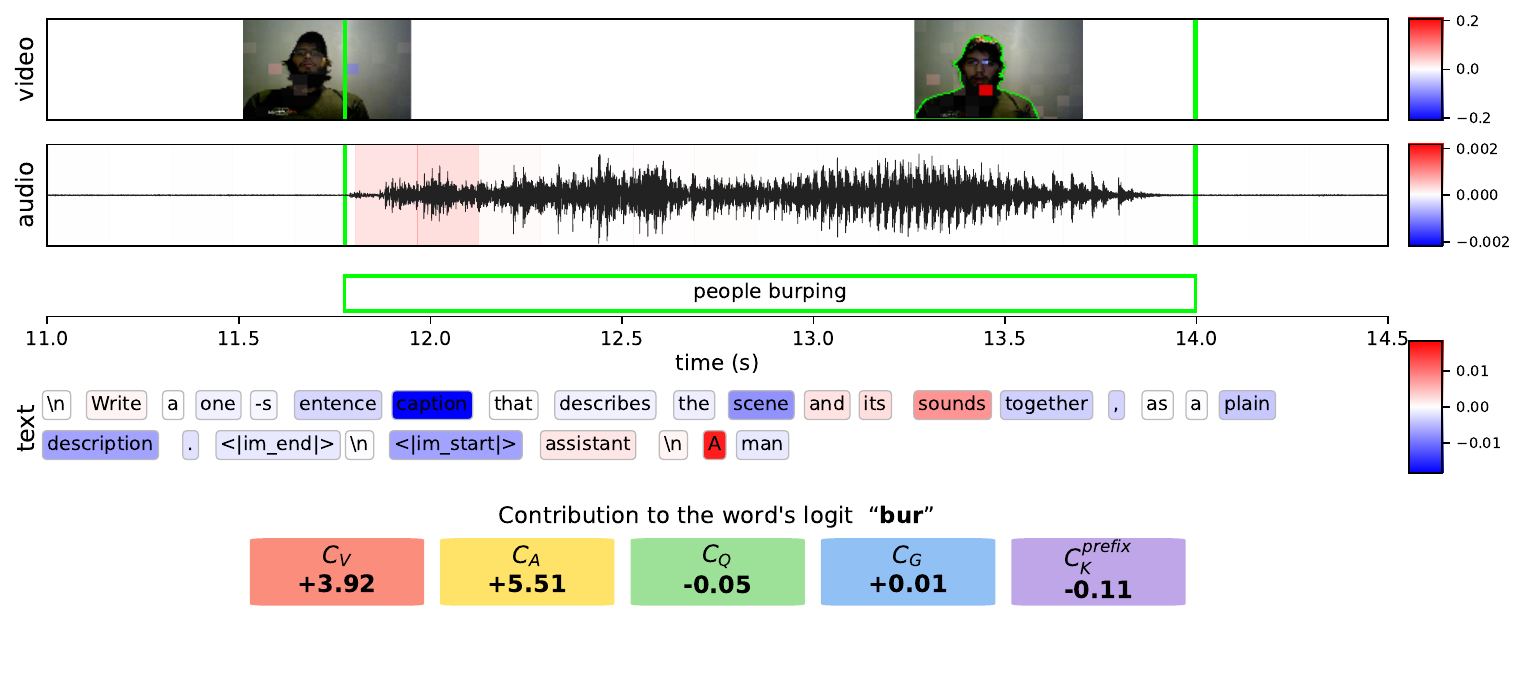}
        \\

        \noalign{\vskip 14pt}

        \attrmodel{\textbf{MCP 4.5}}
        &
        \attrblock
        {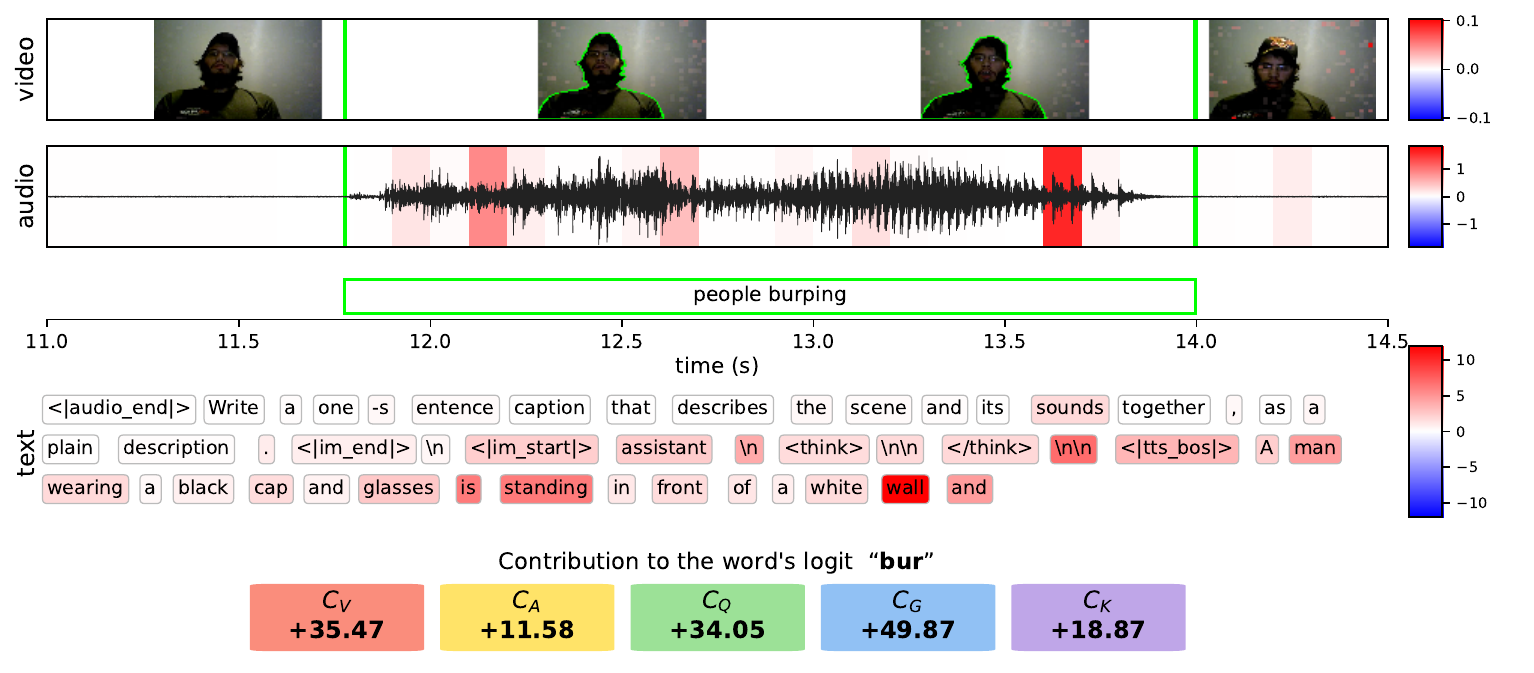}
        {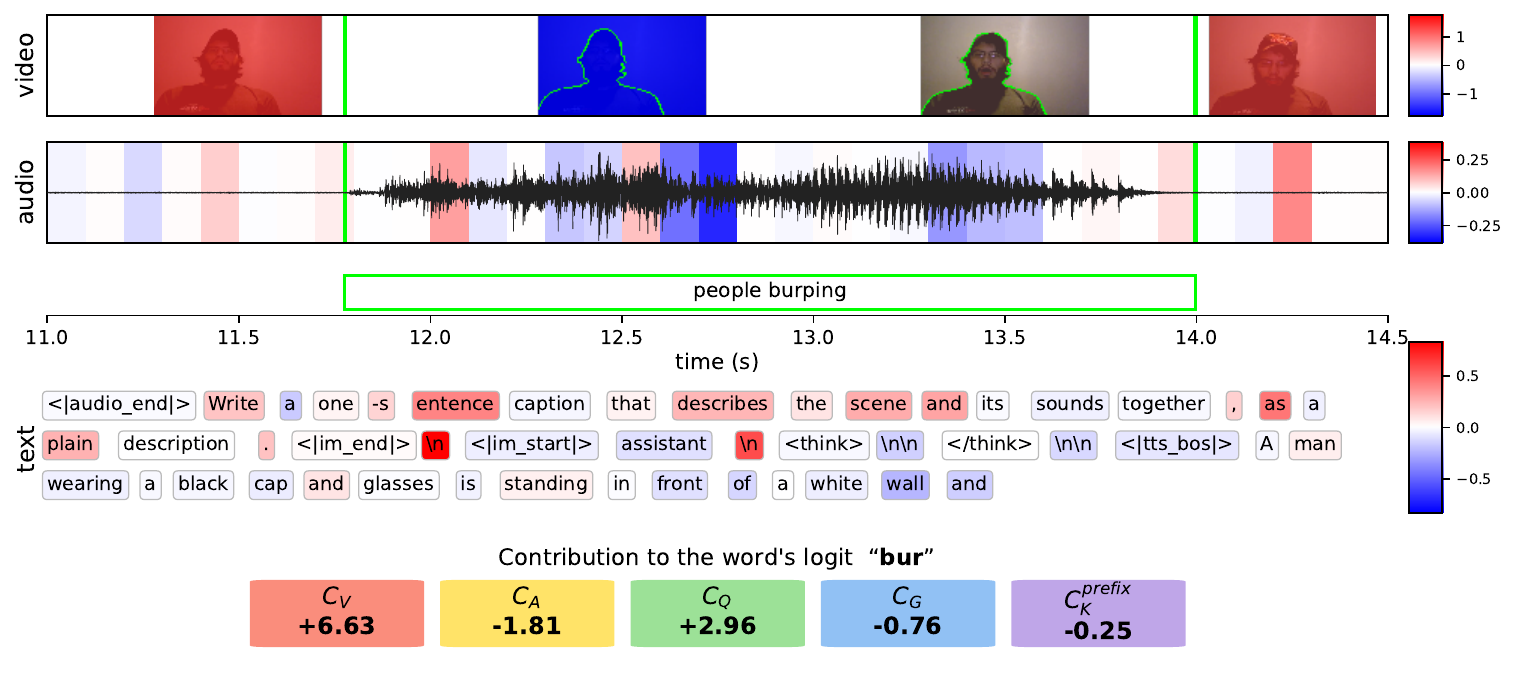}
        {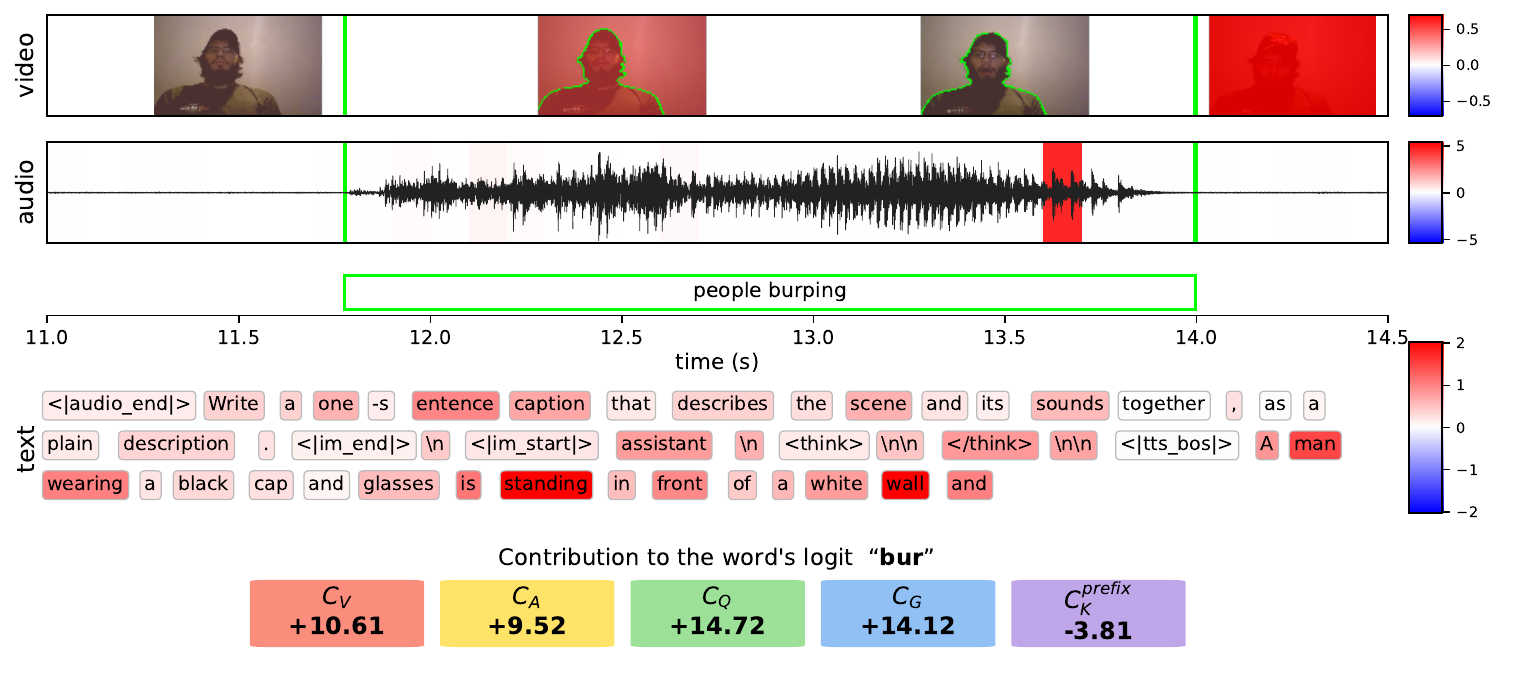}
        {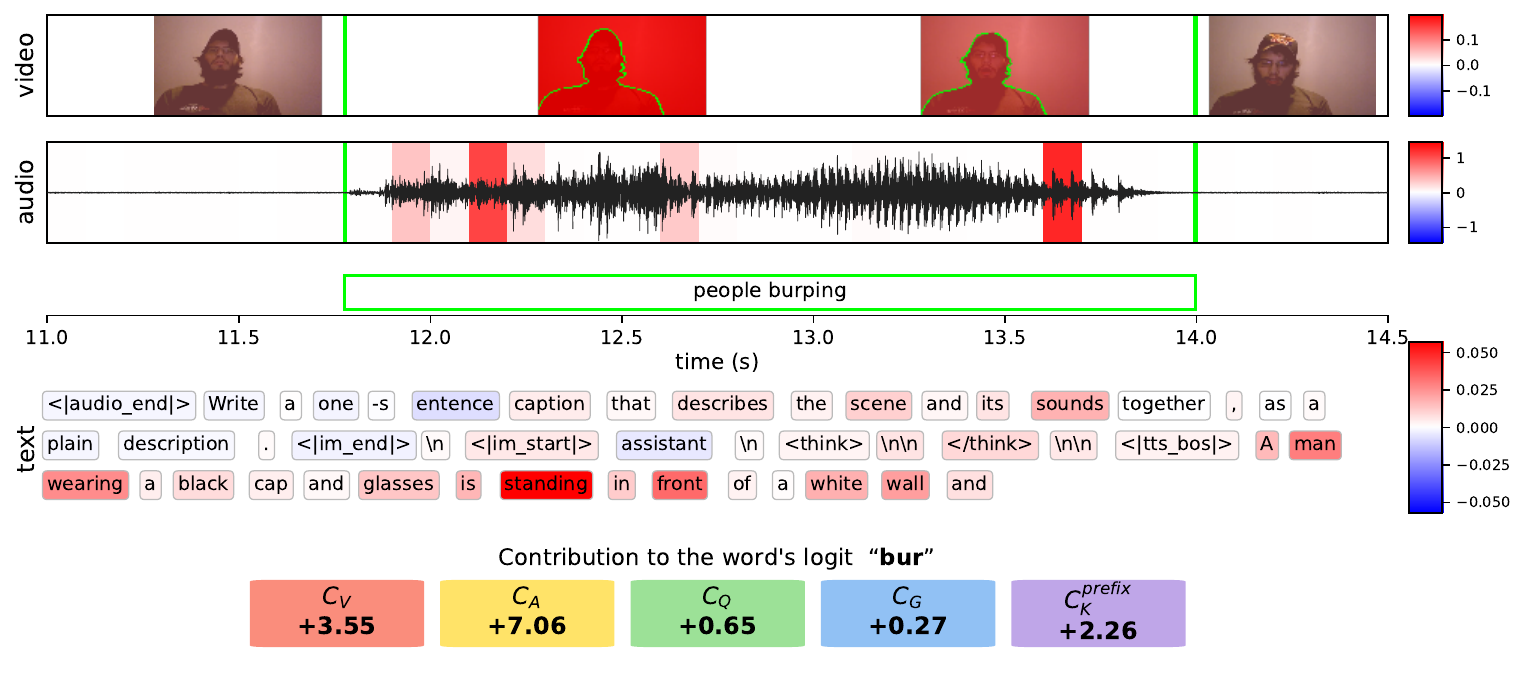}

    \end{tabular}

    \end{adjustbox}
\caption{Qualitative comparison of MALT with attribution baselines on Qwen2.5-Omni~3B, VideoLLaMA2.1-7B-AV, and MiniCPM-o~4.5. For each model, MALT, Gradient$\times$Input, AttnLRP, and ALTI-Logit are evaluated on the same generated event token and teacher-forced sequence. Heatmaps show signed position-level attribution scores for video, audio, and text: red indicates positive scores, blue negative scores, and white values near zero. Colorbars use independent scales for each method, modality, and model; color intensities are therefore not directly comparable across panels. Green annotations mark the spatial and temporal extent of the annotated sound event. Bottom boxes summarize attribution scores aggregated by component.}
\label{fig:qualitative_baseline_comparison}
\end{figure*}

\Cref{fig:qualitative_baseline_comparison} presents a qualitative
comparison of MALT, Gradient$\times$Input, AttnLRP, and ALTI-Logit
on Qwen2.5-Omni~3B, VideoLLaMA2.1-7B-AV, and MiniCPM-o~4.5. These three models were selected as representative cases of
different attribution-intervention consistency patterns, with
the comparison restricted to them due to space constraints.
For each model, the methods are applied to the same audio-video
input and generated target token, highlighting differences in
attribution patterns across modalities. The attention-only baselines are omitted due to their weaker attribution-intervention consistency in the quantitative evaluation.

%% file: main.bib
@String(CVPR= {IEEE Conf. Comput. Vis. Pattern Recog.})

@String(ICIP = {IEEE Int. Conf. Image Process.})

@String(AAAI = {AAAI})

@String(CVPR  = {CVPR})

@String(ICIP  = {ICIP})

@inproceedings{ishikawa2026listening,
  title={{Listening Without Looking: Modality Bias in Audio-Visual Captioning}},
  author={Ishikawa, Yuchi and Manabe, Toranosuke and Komatsu, Tatsuya and Aoki, Yoshimitsu},
  booktitle={2026 IEEE International Conference on Image Processing (ICIP)},
  pages={1--6},
  year={2026},
  organization={IEEE}
}

@inproceedings{ye2026eyes,
  title={{When Eyes and Ears Disagree: Can MLLMs Discern Audio-Visual Confusion?}},
  author={Ye, Qilang and Zeng, Wei and Liu, Meng and Zhang, Jie and Hu, Yupeng and Yu, Zitong and Zhou, Yu},
  booktitle={Proceedings of the AAAI Conference on Artificial Intelligence},
  volume={40},
  number={14},
  pages={11955--11963},
  year={2026}
}

@InProceedings{Selvakumar2026CVPR,
    author    = {Selvakumar, Ramaneswaran and Jayakumar, Kaousheik and Sakshi, S and Ghosh, Sreyan and Gao, Ruohan and Manocha, Dinesh},
    title     = {{Do Audio-Visual Large Language Models Really See and Hear?}},
    booktitle = {Proceedings of the IEEE/CVF Conference on Computer Vision and Pattern Recognition (CVPR) Findings},
    month     = {June},
    year      = {2026},
    pages     = {5892-5902}
}

@inproceedings{jung2026probing,
title={{Probing Cross-modal Information Hubs in Audio-Visual LLMs}},
author={Jihoo Jung and Chaeyoung Jung and Ji-Hoon Kim and Joon Son Chung},
booktitle={Forty-third International Conference on Machine Learning},
year={2026},
}

@inproceedings{liu2024benchmarking,
  title={{Benchmarking Audio Visual Segmentation for Long-Untrimmed Videos}},
  author={Liu, Chen and Li, Peike Patrick and Yu, Qingtao and Sheng, Hongwei and Wang, Dadong and Li, Lincheng and Yu, Xin},
  booktitle={2024 IEEE/CVF Conference on Computer Vision and Pattern Recognition (CVPR)},
  pages={22712--22722},
  year={2024},
  organization={IEEE}
}

@inproceedings{sung2025avhbench,
  title={{AVHBench: A Cross-Modal Hallucination Benchmark for Audio-Visual Large Language Models}},
  author={Sung-Bin, Kim and Hyun-Bin, Oh and Jung-Mok, Lee and Senocak, Arda and Chung, Joon Son and Oh, Tae-Hyun},
  booktitle={International Conference on Learning Representations},
  volume={2025},
  pages={24244--24271},
  year={2025}
}

@inproceedings{chen2026some,
  title={{Some Modalities are More Equal Than Others: Decoding and Architecting Multimodal Integration in MLLMs}},
  author={Chen, Tianle and Chakka, Chaitanya and Akula, Arjun Reddy and Thomas, Xavier and Ghadiyaram, Deepti},
  booktitle={Proceedings of the IEEE/CVF Conference on Computer Vision and Pattern Recognition},
  pages={2142--2151},
  year={2026}
}

@inproceedings{fang2026towards,
title={{Towards Understanding Modality Interaction in Multimodal Language Models via Partial Information Decomposition}},
author={Wanlong Fang and Tianle Zhang and Wen Tao and Alvin Chan},
booktitle={Forty-third International Conference on Machine Learning},
year={2026},
}

@inproceedings{Ancona2017TowardsBU,
  title={{Towards Better Understanding of Gradient-based Attribution Methods for Deep Neural Networks}},
  author={Marco Ancona and Enea Ceolini and Cengiz Oztireli and Markus H. Gross},
  booktitle={International Conference on Learning Representations},
  year={2017},
}

@article{yao2026survey,
  title={A survey on audio-visual large language models},
  author={Yao, Wenyi and Yahyaabadi, Roksana and Hassani, Hossein and Nikan, Soodeh},
  journal={Knowledge-Based Systems},
  pages={116569},
  year={2026},
  publisher={Elsevier}
}

@inproceedings{baid2026don,
  title={{Don’t Let the Video Speak: Audio-Contrastive Preference Optimization for Audio-visual Language Models}},
  author={Baid, Ami and Xue, Zihui and Grauman, Kristen},
  booktitle={European Conference on Computer Vision},
  pages={54--72},
  year={2026},
  organization={Springer}
}

@article{tang2025videosalmonn2captionenhancedaudiovisual,
      title={{video-SALMONN 2: Caption-Enhanced Audio-Visual Large Language Models}}, 
      author={Changli Tang and Yixuan Li and Yudong Yang and Jimin Zhuang and Guangzhi Sun and Wei Li and Zejun Ma and Chao Zhang},
      journal={ArXiv preprint arXiv:2506.15220},
      year={2025},
}

@article{Liu2025OlaPT ,
  title={{Ola: Pushing the Frontiers of Omni-Modal Language Model with Progressive Modality Alignment}},
  author={Zu-Yan Liu and Yuhao Dong and Jiahui Wang and Ziwei Liu and Winston Hu and Ji-Wen Lu and Yongming Rao},
  journal={arXiv preprint arXiv:2502.04328},
  year={2025},
}

@article{Cui2026MiniCPMo4T,
  title={{MiniCPM-o 4.5: Towards Real-Time Full-Duplex Omni-Modal Interaction}},
  author={Jun-Bo Cui and Bo-Kai Xu and Chong-Yi Wang and Tianyu Yu and Wei-Yu Sun and Yingjin Xu and Tian-Rui Wang and Zhi-Hui He and Wen-Shuo Ma and Tian-Chi Cai and Jian-Cheng Gui and Luo-Yuan Zhang and Xian Sun and Fu-Wei Huang and Moye Chen and Zhuohang Lin and Han-Yu Liu and Qi Gui and Qing-Yuan Han and Yuyang Wen and Hui-Ping Liu and Rongkang Wang and Ya-Qi Zhang and Hong-Liang Wei and Chi Chen and You Li and Kechen Fang and Jie Zhou and Yuxuan Li and Guo-Yang Zeng and Chao-Jun Xiao and Yankai Lin and Xu Han and Mao-Song Sun and Zhi-Yuan Liu and Yuan Yao},
  journal={ArXiv preprint arXiv:2604.27393},
  year={2026},
}

@misc{yao2024minicpm,
  author       = {{OpenBMB}},
  title        = {{MiniCPM-o 2.6}},
  howpublished = {Hugging Face model card},
  year         = {2025},
  note         = {\url{https://huggingface.co/openbmb/MiniCPM-o-2_6}}
}

@article{Qwen2.5-Omni,
  title={{Qwen2.5-Omni Technical Report}},
  author={Jin Xu and Zhifang Guo and Jinzheng He and Hangrui Hu and Ting He and Shuai Bai and Keqin Chen and Jialin Wang and Yang Fan and Kai Dang and Bin Zhang and Xiong Wang and Yunfei Chu and Junyang Lin},
  journal={arXiv preprint arXiv:2503.20215},
  year={2025}
}

@article{damonlpsg2024videollama2,
  title={{VideoLLaMA 2: Advancing Spatial-Temporal Modeling and Audio Understanding in Video-LLMs}},
  author={Cheng, Zesen and Leng, Sicong and Zhang, Hang and Xin, Yifei and Li, Xin and Chen, Guanzheng and Zhu, Yongxin and Zhang, Wenqi and Luo, Ziyang and Zhao, Deli and Bing, Lidong},
  journal={arXiv preprint arXiv:2406.07476},
  year={2024},
}

@article{Ghosh2026AudioVisualFO,
  title={{Audio-Visual Flamingo: Open Audio-Visual Intelligence for Long and Complex Videos}},
  author={Sreyan Ghosh and Arushi Goel and Kaousheik Jayakumar and Lasha Koroshinadze and Nishit Anand and Siddharth Gururani and Han-Rong Ye and Pritam Biswas and Yuanhang Su and Ehsan Hosseini-Asl and Sang-gil Lee and Zhi-Feng Kong and Jaehyeon Kim and Sungwon Kim and S. Sakshi and Ramani Duraiswami and Dinesh Manocha and Andrew Tao and Mohammad Shoeybi and Bryan Christopher Catanzaro and Ming-Yu Liu and Wei Ping},
  journal={ArXiv preprint arXiv:2607.16107},
  year={2026},
}

@inproceedings{3692070.3692076,
author = {Achtibat, Reduan and Hatefi, Sayed Mohammad Vakilzadeh and Dreyer, Maximilian and Jain, Aakriti and Wiegand, Thomas and Lapuschkin, Sebastian and Samek, Wojciech},
title     = {{AttnLRP: Attention-Aware Layer-Wise Relevance Propagation for Transformers}},
  booktitle = {Proceedings of the 41st International Conference on Machine Learning},
  series    = {Proceedings of Machine Learning Research},
  volume    = {235},
  pages     = {135--168},
  publisher = {PMLR},
  year      = {2024}
}

@inproceedings{ali2022xai,
  title={{XAI for Transformers: Better Explanations through Conservative Propagation}},
  author={Ali, Ameen and Schnake, Thomas and Eberle, Oliver and Montavon, Gr{\'e}goire and M{\"u}ller, Klaus-Robert and Wolf, Lior},
  booktitle={International conference on machine learning},
  pages={435--451},
  year={2022},
  organization={PMLR}
}

@article{Yang2024Qwen2TR,
  title={{Qwen2 Technical Report}},
  author={An Yang and Bao-Song Yang and Binyuan Hui and Bo Zheng and Bo-Wen Yu and Chang Zhou and Cheng-Peng Li and Cheng-Yuan Li and Dayiheng Liu and Fei Huang and Guanting Dong and Hao-Ran Wei and Huan Lin and Jia-Long Tang and Jia-Lin Wang and Jian Yang and Jian-Hong Tu and Jian-Wei Zhang and Jian-Xin Ma and Jian-Xin Yang and Jin Xu and Jing-Ren Zhou and Jin-Ze Bai and Jin-Zheng He and Jun-Yang Lin and Kai Dang and Ke-Ming Lu and Ke-Yang Chen and Kexin Yang and Mei Li and Min Xue and Na Ni and Pei Zhang and Peng Wang and Ru Peng and Rui Men and Rui-Ze Gao and Run-Ji Lin and Shi-Jie Wang and Shuai Bai and Si-Nan Tan and Tianhang Zhu and Tian-Hao Li and Tianyu Liu and Wen-Bin Ge and Xiaodong Deng and Xiao-Huan Zhou and Xing-Zhang Ren and Xinyu Zhang and Xipin Wei and Xuancheng Ren and Xue-Jing Liu and Yang Fan and Yang Yao and Yi-Chang Zhang and Yunyang Wan and Yun-Fei Chu and Yu-Qiong Liu and Zeyu Cui and Zhen-Ru Zhang and Zhi-Fang Guo and Zhi-Wei Fan},
  journal={ArXiv preprint ArXiv:2407.10671},
  year={2024},
}

@article{Yang2024Qwen25TR,
  title={{Qwen2.5 Technical Report}},
  author={An Yang and Bao-Song Yang and Bei-Chen Zhang and Bin-Yuan Hui and Bo Zheng and Bo-Wen Yu and Cheng-Yuan Li and Dayiheng Liu and Fei Huang and Guanting Dong and Hao-Ran Wei and Huan Lin and Jian Yang and Jian-Hong Tu and Jian-Wei Zhang and Jian-Xin Yang and Jia-Xin Yang and Jing-Ren Zhou and Jun-Yang Lin and Kai Dang and Ke-Ming Lu and Ke-Qin Bao and Kexin Yang and Le Yu and Mei Li and Mingfeng Xue and Pei Zhang and Qin Zhu and Rui Men and Run-Ji Lin and Tian-Hao Li and Tingyu Xia and Xing-Zhang Ren and Xuan-Cheng Ren and Yang Fan and Yang Su and Yi-Chao Zhang and Yunyang Wan and Yuqi Liu and Zeyu Cui and Zhen-Ru Zhang and Zi-Han Qiu and Shanghaoran Quan and Zekun Wang},
  journal={ArXiv preprint ArXiv:2412.15115},
  year={2024},
}

@article{yang2025qwen3technicalreport,
      title={{Qwen3 Technical Report}}, 
      author={An Yang and Anfeng Li and Baosong Yang and Beichen Zhang and Binyuan Hui and Bo Zheng and Bowen Yu and Chang Gao and Chengen Huang and Chenxu Lv and Chujie Zheng and Dayiheng Liu and Fan Zhou and Fei Huang and Feng Hu and Hao Ge and Haoran Wei and Huan Lin and Jialong Tang and Jian Yang and Jianhong Tu and Jianwei Zhang and Jianxin Yang and Jiaxi Yang and Jing Zhou and Jingren Zhou and Junyang Lin and Kai Dang and Keqin Bao and Kexin Yang and Le Yu and Lianghao Deng and Mei Li and Mingfeng Xue and Mingze Li and Pei Zhang and Peng Wang and Qin Zhu and Rui Men and Ruize Gao and Shixuan Liu and Shuang Luo and Tianhao Li and Tianyi Tang and Wenbiao Yin and Xingzhang Ren and Xinyu Wang and Xinyu Zhang and Xuancheng Ren and Yang Fan and Yang Su and Yichang Zhang and Yinger Zhang and Yu Wan and Yuqiong Liu and Zekun Wang and Zeyu Cui and Zhenru Zhang and Zhipeng Zhou and Zihan Qiu},
      journal={ArXiv preprint ArXiv:2505.09388},
      year={2025},
}

@article{Bai2025Qwen3VLTR,
  title={{Qwen3-VL Technical Report}},
  author={Shuai Bai and Yuxuan Cai and Ruizhe Chen and Keqin Chen and Xionghui Chen and Zesen Cheng and Lianghao Deng and Wei Ding and Chang Gao and Chunjiang Ge and Wenbin Ge and Zhifang Guo and Qidong Huang and Jie Huang and Fei Huang and Binyuan Hui and Shutong Jiang and Zhaohai Li and Mingsheng Li and Mei Li and Kaixin Li and Zicheng Lin and Junyang Lin and Xuejing Liu and Jiawei Liu and Chenglong Liu and Yang Liu and Dayiheng Liu and Shixuan Liu and Dunjie Lu and Ruilin Luo and Chenxu Lv and Rui Men and Lingchen Meng and Xuancheng Ren and Xingzhang Ren and Sibo Song and Yuchong Sun and Jun Tang and Jianhong Tu and Jianqiang Wan and Peng Wang and Pengfei Wang and Qiuyue Wang and Yuxuan Wang and Tianbao Xie and Yiheng Xu and Haiyang Xu and Jin Xu and Zhibo Yang and Mingkun Yang and Jianxin Yang and An Yang and Bowen Yu and Fei Zhang and Hang Zhang and Xi Zhang and Bo Zheng and Humen Zhong and Jingren Zhou and Fan Zhou and Jing Zhou and Yuanzhi Zhu and Ke Zhu},
  journal={ArXiv preprint ArXiv:2511.21631},
  year={2025},
}

@inproceedings{miller-1994-wordnet,
    title = {{WordNet: A Lexical Database for English}},
    author = "Miller, George A.",
    booktitle = "{H}uman {L}anguage {T}echnology: Proceedings of a Workshop held at {P}lainsboro, {N}ew {J}ersey, {M}arch 8-11, 1994",
    year = "1994",
}

@misc{honnibal2020spacy,
    title  = {{spaCy: Industrial-strength Natural Language Processing in Python}},
    author = {Honnibal, Matthew and Montani, Ines and Van Landeghem, Sofie and Boyd, Adriane},
    year   = {2020},
    doi    = {10.5281/zenodo.1212303},
    note         = {\url{https://github.com/explosion/spaCy}}
}

@inproceedings{ferrando2023explaining,
    title = {{Explaining How Transformers Use Context to Build Predictions}},
    author = "Ferrando, Javier  and
      G{\'a}llego, Gerard I.  and
      Tsiamas, Ioannis  and
      Costa-juss{\`a}, Marta R.",
    editor = "Rogers, Anna  and
      Boyd-Graber, Jordan  and
      Okazaki, Naoaki",
    booktitle = "Proceedings of the 61st Annual Meeting of the {A}ssociation for {C}omputational {L}inguistics (Volume 1: Long Papers)",
    month = jul,
    year = "2023",
    address = "Toronto, Canada",
    publisher = "Association for Computational Linguistics",
    doi = "10.18653/v1/2023.acl-long.301",
    pages = "5486--5513",
}

@article{arras2025closelook,
  title={{A Close Look at Decomposition-based XAI-Methods for Transformer Language Models}},
  author={Arras, Leila and Puri, Bruno and Kahardipraja, Patrick and Lapuschkin, Sebastian and Samek, Wojciech},
  journal={arXiv preprint arXiv:2502.15886},
  year={2025}
}

@inproceedings{abnar2020rollout,
  title={{Quantifying Attention Flow in Transformers}},
  author={Abnar, Samira and Zuidema, Willem},
  booktitle={Proceedings of the 58th annual meeting of the association for computational linguistics},
  pages={4190--4197},
  year={2020}
}

@inproceedings{kobayashi2020attention,
  title     = {{Attention is Not Only a Weight: Analyzing Transformers with Vector Norms}},
  author    = {Kobayashi, Goro and Kuribayashi, Tatsuki and Yokoi, Sho and Inui, Kentaro},
  booktitle = {Proceedings of the 2020 Conference on Empirical Methods in Natural Language Processing (EMNLP)},
  pages     = {7057--7075},
  year      = {2020}
}

@inproceedings{parcalabescu2023mmshap,
  title={{MM-SHAP: A Performance-agnostic Metric for Measuring Multimodal Contributions in Vision and Language Models \& Tasks}},
  author={Parcalabescu, Letitia and Frank, Anette},
  booktitle={Proceedings of the 61st Annual Meeting of the Association for Computational Linguistics (Volume 1: Long Papers)},
  pages={4032--4059},
  year={2023}
}

@article{sturmfels2020baselines,
  title={{Visualizing the Impact of Feature Attribution Baselines}},
  author={Sturmfels, Pascal and Lundberg, Scott and Lee, Su-In},
  journal={Distill},
  volume={5},
  number={1},
  pages={e22},
  year={2020}
}

@inproceedings{chang2018counterfactual,
  title     = {{Explaining Image Classifiers by Counterfactual Generation}},
  author    = {Chang, Chun-Hao and Creager, Elliot and Goldenberg, Anna and Duvenaud, David},
  booktitle = {International Conference on Learning Representations},
  year      = {2019}
}

@inproceedings{jantsch2026dpa,
title={{Dual Path Attribution: Efficient Attribution for SwiGLU-Transformers through Layer-Wise Target Propagation}},
author={Lasse M. Jantsch and Dong-Jae Koh and Seonghyeon Lee and Young-Kyoon Suh},
booktitle={Mechanistic Interpretability Workshop at ICML 2026},
year={2026},
}
